\documentclass[runningheads]{llncs}

\usepackage{eccv}

\usepackage{eccvabbrv}

\usepackage{graphicx}
\usepackage{booktabs}

\usepackage[accsupp]{axessibility}  %

\usepackage[pagebackref,breaklinks,colorlinks,citecolor=eccvblue]{hyperref}
\usepackage{orcidlink}

\usepackage[square,numbers,sort&compress]{natbib}
\usepackage[dvipsnames]{xcolor}

\renewcommand{\paragraph}[1]{\medskip\noindent\textbf{#1}~~}

\renewcommand{\colorbox}[2]{\protect\tikz[baseline]\protect\node[anchor=base,fill=#1,rounded corners=1pt,inner sep=1pt]{#2};}

\newcommand{\first}[1] {\colorbox{red!25}{#1}}
\newcommand{\second}[1] {\colorbox{orange!25}{#1}}
\newcommand{\third}[1] {\colorbox{yellow!25}{#1}}

\usepackage{tikz}
\usepackage{pgfplots}
\usetikzlibrary{hobby, fit, shapes.geometric, patterns, calc, math, patterns}
\usetikzlibrary{3d, shapes.arrows}
\usetikzlibrary{positioning}
\usetikzlibrary{spy}
\usepgfplotslibrary{groupplots}

\newlength{\figurewidth}
\newlength{\figureheight}

\usepackage{amsmath,amsfonts,bm}
\usepackage{caption}
\usepackage{subcaption}
\usepackage{comment}
\usepackage{enumitem}
\usepackage{colortbl} 
\usepackage{xcolor}

\usepackage{tabularx}
    \newcolumntype{L}{>{\raggedright\arraybackslash}X}

\newcommand{\vx}{\mathbf{x}} 
\newcommand{\vd}{\mathbf{d}} 
\newcommand{\vc}{\mathbf{c}} 
\newcommand{\vr}{\mathbf{r}} 
\newcommand{\tnear}{t_{n}} 
\newcommand{\tfar}{t_{f}} 
\newcommand{\vmu}{\bm{\mu}}

\newcommand{\vw}{\mathbf{w}} 

\newcommand{\mC}{\mathbf{C}} 
\newcommand{\mP}{\mathbf{P}} 
\newcommand{\mJ}{\mathbf{J}} 

\newcommand{\gL}{\mathcal{L}} 

\newcommand{\E}{\mathbb{E}} 
\newcommand{\Var}{\mathrm{Var}} 
\newcommand{\Cov}{\mathrm{Cov}} 

\newcommand{\gD}{\mathcal{D}} 
\newcommand{\R}{\mathbb{R}} 
\newcommand{\bS}{\mathbb{S}} 
\newcommand{\vtheta}{\bm{\theta}} 

\DeclareMathOperator*{\argmax}{arg\,max} 
\newcommand{\Hess}{\mathbf{H}} 

\newcommand{\mSigma}{\mathbf{\Sigma}} %
\newcommand{\mW}{\bm{W}} %

\newcommand{\dd}{\,\mathrm{d}}

\renewcommand{\mid}{\,|\,}

\usepackage{multirow}

\DeclareRobustCommand{\num}[1]{\protect\tikz[baseline=-.5ex]{\protect\node[circle,draw=black!80,thick,inner sep=1pt,font=\bf\scriptsize,xshift=1em](a){#1};}}

\usepackage{pifont}
\pgfplotsset{axis on top, scale only axis, y grid style={line width=.1pt, draw=gray!10,dashed},x grid style={line width=.1pt, draw=gray!10,dashed}}

\begin{document}

\title{Sources of Uncertainty \\ in 3D Scene Reconstruction} 

\titlerunning{Sources of Uncertainty in 3D Scene Reconstruction}

\author{Marcus Klasson$^{1}$\orcidlink{0000-0002-8633-281X} \and
Riccardo Mereu$^{1}$\orcidlink{0000-0002-8932-9341} \and
Juho Kannala$^{1,2}$\orcidlink{0000-0001-5088-4041} \and
Arno Solin$^{1}$\orcidlink{0000-0002-0958-7886}}

\authorrunning{M.~Klasson \etal}
\institute{$^1$~Aalto University, $^2$~University of Oulu \\ %
\email{\{firstname.lastname\}@aalto.fi}
}

\maketitle
\begin{abstract}
The process of 3D scene reconstruction can be affected by numerous uncertainty sources in real-world scenes. While Neural Radiance Fields (NeRFs) and 3D Gaussian Splatting (GS) achieve high-fidelity rendering, they lack built-in mechanisms to directly address or quantify uncertainties arising from the presence of noise, occlusions, confounding outliers, and imprecise camera pose inputs. In this paper, we introduce a taxonomy that categorizes different sources of uncertainty inherent in these methods. Moreover, we extend NeRF- and GS-based methods with uncertainty estimation techniques, including learning uncertainty outputs and ensembles, and perform an empirical study to assess their ability to capture the sensitivity of the reconstruction. Our study highlights the need for addressing various uncertainty aspects when designing NeRF/GS-based methods for uncertainty-aware 3D reconstruction. %

\end{abstract}

\crefname{table}{Table}{Tables}
\crefname{appendix}{App.}{Apps.}

\section{Introduction}
\label{sec:introduction}
The process of 3D scene reconstruction of real-world scenes can give rise to artefacts in regions that are affected by various sources of uncertainty. Despite that Neural Radiance Fields (NeRFs, \cite{mildenhall2020nerf}) and 3D Gaussian Splatting (GS, \cite{kerbl20233d}) achieve high-fidelity reconstruction of real-world scenes, they are sensitive when learning from sparse views, in the presence of occluders and moving objects. Furthermore, they lack built-in mechanisms to quantify uncertainties arising from such settings, which makes them unsuitable for deployment in robust, immersive reconstruction or structure-aware applications like navigation and planning (incl.\ safety-critical use cases). 

Recent works equipping NeRF and GS with uncertainty estimation have focused on recognizing unobserved scene parts~\cite{shen2022conditional,goli2024bayes,sunderhauf2023density} and next-best view selection~\cite{pan2022activenerf,jiang2023fisherrf,jin2023neu}. These methods aim to reduce the uncertainty arising from lack of information about the scene, often referred to as {\it epistemic} uncertainty. Other works have focused on detecting moving objects, or distractors, in the scene~\cite{sabour2023robustnerf,sabour2024spotlesssplats,ren2024nerf}, which is a type of {\it aleatoric} uncertainty that appears due to random effects in the scene. The approaches to capture the different aspects of uncertainty present in 3D scene reconstruction are tied to the downstream application task rather than framing uncertainty in a general framework. Hence, it is important to define what types of uncertainties NeRFs and GS need to quantify to enable robust 3D scene reconstruction.

\begin{figure}[t!]
  \centering\scriptsize
  \begin{subfigure}[c]{.47\textwidth}
  \centering
  \setlength{\figurewidth}{1.5\textwidth}
  \scalebox{.70}{%
  \begin{tikzpicture}[inner sep=0,outer sep=0]

    \tikzstyle{blob}=[inner sep=2pt,font=\bf,circle,draw=black,thick]
    \tikzstyle{label}=[inner sep=2pt,text width=7em,font=\scriptsize,align=center]  
    \tikzstyle{arr}=[draw=black!80,-latex]  

    \coordinate (center) at (.5\figurewidth,.475\figurewidth);
    \begin{scope}[shift={(center)},rotate=-55]
      \draw[fill=black!05,draw=none] (0,0) circle [x radius=2.3cm, y radius=3cm];
    \end{scope}    

    \node[anchor=south west] (bg) at (0,0) {\includegraphics[width=\figurewidth]{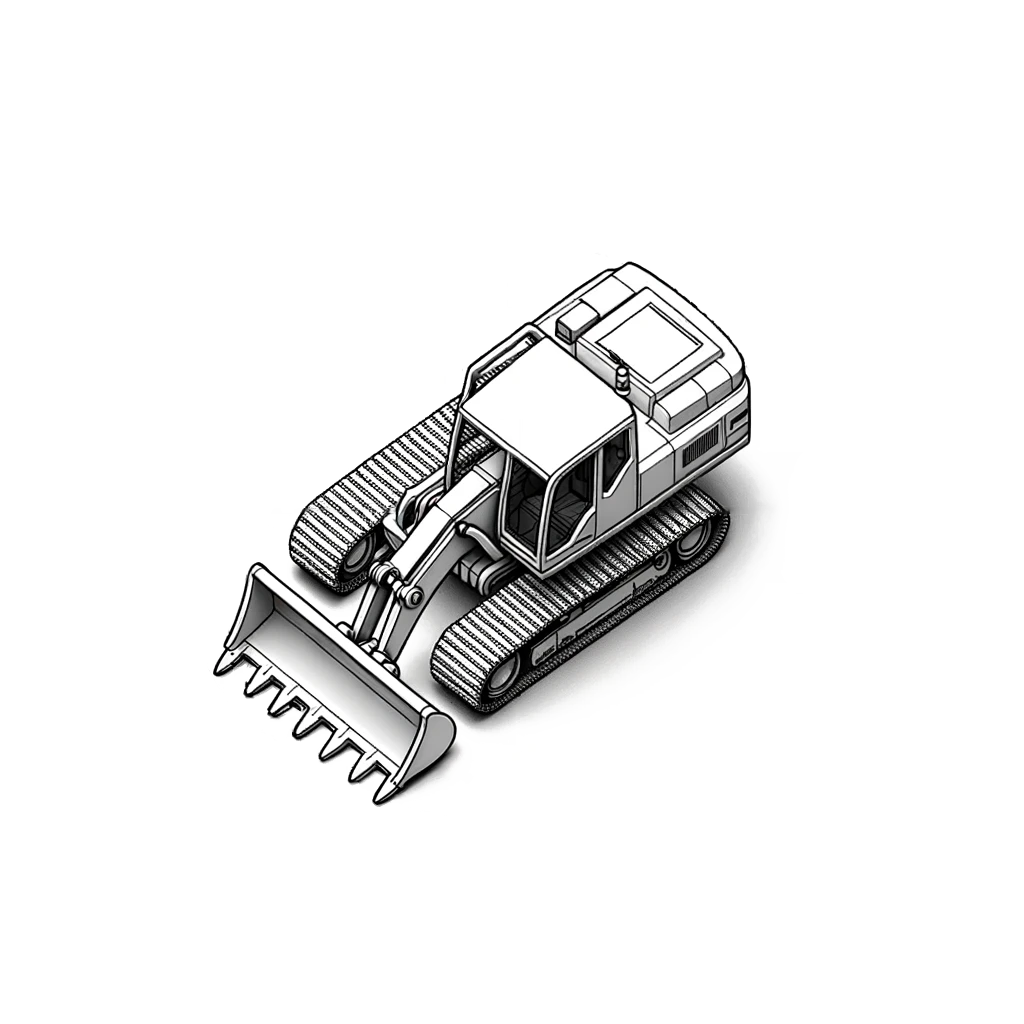}};  

    \begin{scope}
      \clip[rotate=-55] (center) ellipse (2.3cm and 3cm);
      \node[anchor=south west] (bg) at (0,0) {\includegraphics[width=\figurewidth]{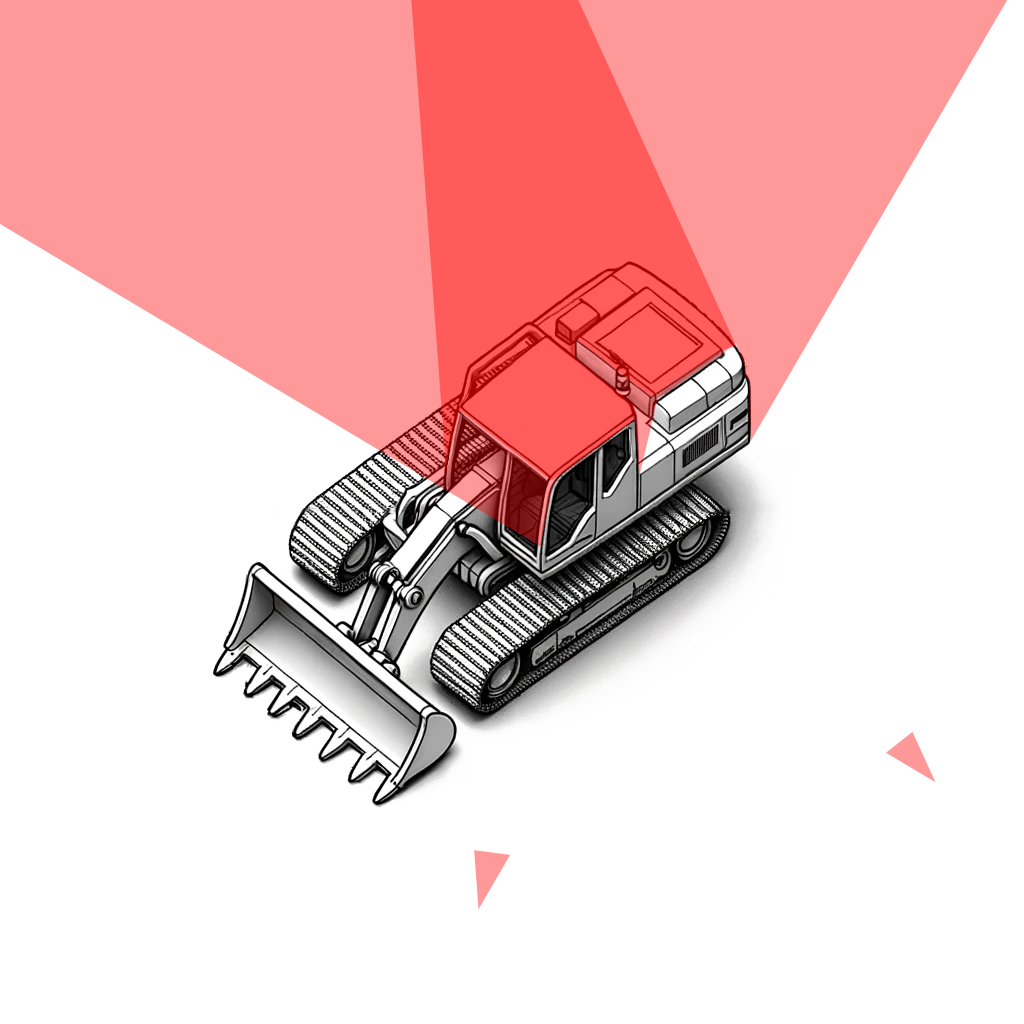}};
    \end{scope}

    \begin{scope}[x={(bg.south east)},y={(bg.north west)}]

      \node[draw=none, single arrow, minimum height=10mm, minimum width=2pt, single arrow head extend=4pt, fill=black!80, anchor=center, rotate=0, inner sep=5pt, rounded corners=1pt,fill=red!50,rotate=-130] at (.75,.31) {};
      \node at (.75,.35) {\includegraphics[width=8.5mm]{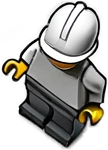}};

      \node[blob] at (.40,.10) {1};
      \node[blob] at (.25,.88) {2};
      \node[blob] at (.90,.54) {3};
      \node[blob] at (.90,.15) {4};

      \node[label,text width=5em] (L1) at (.22,.1) {Uncertainty due to variation};%
      \node[label,text width=5em] (L2) at (.90,.46) {Non-static confounders};%
      \node[label] (L3) at (.25,.80) {Uncertainty due to lack of information};%
      \node[label] (L4) at (.75,.15) {Sensitivity of camera poses};%

      \draw[arr] (L2.south) to[bend left=30] ++(-.7cm,-.5cm);    
      \draw[arr] (L3.south) to[bend right=30] ++(1cm,-1cm);    

      \draw[dashed] (0.4697,0.1162) -- (0.3994,1);  
      \draw[dashed] (0.4697,0.1162) -- (0.9844,1);  

      \draw[dashed] (0.9131,0.2354) -- (0,0.7842);  
      \draw[dashed] (0.9131,0.2354) -- (0.5645,1);        

      \draw[red!50,double, double distance=2pt] (L1.south west) to[out=140,in=230] (L1.north west);
      \draw[red!50,double, double distance=2pt] (L1.south east) to[out=40,in=-50] (L1.north east);

      \begin{scope}[shift={(0.9131,0.2354)}, rotate=-42]
        \draw[red,->,thick] (0,0) arc (0:20:1cm);
        \draw[red,->,thick] (0,0) arc (0:-20:1cm);        
      \end{scope}

      \begin{scope}[shift={(0.9131,0.2354)}, rotate=132]
        \tikzmath{\fov=40; \a=0.04;}
        \draw[blue,thick]  
          (0,0) -- ({\a*cos(\fov/2)},{\a*sin(\fov/2)}) --
                   ({\a*cos(-\fov/2)},{\a*sin(-\fov/2)}) -- cycle;        
      \end{scope}

      \begin{scope}[shift={(0.4697,0.1162)}, rotate=77]
        \tikzmath{\fov=40; \a=0.04;}
        \draw[blue,thick] 
          (0,0) -- ({\a*cos(\fov/2)},{\a*sin(\fov/2)}) --
                   ({\a*cos(-\fov/2)},{\a*sin(-\fov/2)}) -- cycle;        
      \end{scope}
          
    \end{scope}
             
  \end{tikzpicture}}
  \end{subfigure}
  \hfill
  \begin{subfigure}[c]{.51\textwidth}
  \textbf{Types of Uncertainty in NeRF and GS:}
  \begin{itemize}[itemsep=4pt]
    \item[\num{1}] \textbf{Irreducible uncertainty} (\emph{aleatoric}) \\ due to random effects in the observations, \eg, varying lighting and motion blur. %
    \item[\num{2}] \textbf{Reducible uncertainty} (\emph{epistemic}) \\ due to lack of information in the scene (\eg, occluded parts); can be reduced by observing more data included from new poses. %
    \item[\num{3}] \textbf{Confounding outliers} (\emph{aleatoric}/\emph{epistemic})\looseness-1 \\ due to non-static scenes (passers by, moving objects, \etc); often showing as blur or hallucinations in the reconstruction. %
    \item[\num{4}] \textbf{Pose uncertainty} (\emph{aleatoric}/\emph{epistemic}) \\ Sensitivity to the camera poses in the scene. Also viewed as the dual of uncertainty in the reconstruction.
  \end{itemize}
  \end{subfigure}
  \caption{We study four types \num{1}--\num{4} of uncertainty associated with NeRF/GS methods.}
  \label{fig:uncertainty}
\end{figure}

We introduce a taxonomy for different aspects of uncertainty inherent to NeRFs and GS in novel view synthesis and 3D reconstruction of real-world scenes. 
Moreover, we argue that developing frameworks that can quantify ambiguities present in real-world scenes is critical for enabling robust 3D reconstruction. Therefore, as an initial step, this work presents an empirical study of these uncertainty sources across several NeRF- and GS-based methods extended with uncertainty estimation techniques
~\cite{gal2016dropout,lakshminarayanan2017simple,kendall2017uncertainties,daxberger2021laplace}.\looseness-1

Our contributions are two-fold: {\em (i)}~We identify and categorize sources of uncertainties in 3D scene reconstruction and propose methods for systematically evaluating their impact. We introduce a taxonomy (\cref{fig:uncertainty}) that addresses various aspects of uncertainty intrinsic to NeRFs and GS, in the context of novel view synthesis and 3D reconstruction of real-world scenes. {\em (ii)}~We perform an empirical study using efficient NeRF and GS models from Nerfstudio~\cite{tancik2023nerfstudio} to compare the performance of various uncertainty estimation techniques on the sources of uncertainty.
Our study highlights the need to address multiple uncertainty aspects when designing NeRF/GS-based methods towards tackling the challenge of robust novel view synthesis and uncertainty-aware 3D reconstruction. 

\section{Uncertainty in Neural Scene Reconstruction}
\label{sec:related_work}
We identify four types of uncertainty: \num{1}~Irreducible uncertainty (aleatoric), stemming from random effects in observations (inherent noise); %
\num{2}~Reducible uncertainty (epistemic), due to insufficient information in parts of the scene (out-of-distribution, OOD), which can be reduced by capturing data from new poses; \num{3}~Confounding outliers, caused by non-static scenes, with elements such as moving objects or vegetation, which lead to ambiguities like blur or hallucinations; \num{4}~Input uncertainty, which relates to sensitivity to camera poses and can be seen as the dual of reconstruction uncertainty, focusing on how changing inputs can reduce uncertainty or enhance quality. 
See \cref{fig:uncertainty} for an illustration. 

In practice, these types of uncertainty can be deeply intertwined. In machine learning, the modeling assumptions/setups make up a surrogate for the original problem. This surrogate can exhibit aleatoric uncertainty that depends on or interacts with the epistemic uncertainty of the modeling parameters. 
Consequently, this uncertainty cannot solely be classified in any of the classes anymore but as a more general inferential uncertainty (see, \eg, \cite{ranftl2021bayesian}).

Especially in cases where the data is limited or noisy, a model might interpret noise in the data (aleatoric) as a feature it has not learned yet (epistemic). Thus, both uncertainty types can manifest as an imprecise rendered output, though their sources are different. In a scene that exhibits both uncertainty sources, 
distinguishing the cause of an uncertain reconstruction in the final output becomes complex. 
Similarly, non-static elements in a scene, such as moving people or vegetation, introduce variability that is often interpreted as aleatoric noise. However, these elements can also obscure parts of the scene, acting as a source of occlusion for parts of the scene.
In NeRF and GS, a confounding outlier can often satisfy the volumetric rendering constraints by either being considered arbitrarily close to the observing camera or as a thin 2D slice obstructing exactly one of the views (see \cref{sec:outliers}). 
Moreover, imprecise camera pose estimates may arise from noisy observations with varying lighting and motion blur (aleatoric) or challenging scenes with, \eg, low texture, repetitive patterns, and insufficient overlap in images (epistemic). 
In real-world applications, the limited availability of data and the presence of bias in data collection make the true nature of the uncertainties involved indistinguishable. Thus, it is reasonable to study one joint manifestation of sensitivity to perturbations as in the following techniques.

\paragraph{Related Work on Uncertainty Estimation in Deep Learning} 
Deep neural networks lack the ability to model predictive uncertainty as such. A common approach to model aleatoric uncertainty is to predict a probability distribution from the %
network~\cite{kendall2017uncertainties}---typically limited to capturing only the two first moments (mean and variance). 
However, this modification is insufficient to model epistemic uncertainty in the model parameters, which enables the network to reason between known observations and unseen phenomena. 
Bayesian deep learning methods~\cite{papamarkou2024position} give means to quantify epistemic uncertainty through posterior approximations~\cite{mackay1992bayesian,zhang2018advances} to obtain the predictive distribution. 
For large-scale problems, Ensembles~\cite{lakshminarayanan2017simple} and MC-Dropout~\cite{gal2016dropout} have become popular alternatives for estimating epistemic uncertainty in computer vision~\cite{gustafsson2020evaluating,ovadia2019can}. 
Ensembles estimate uncertainty by predictions from multiple networks trained with different weight initializations, while MC-Dropout performs predictions by masking weights in the network by enabling dropout~\cite{srivastava2014dropout} at test time.
More recently, the Laplace approximation~\cite{mackay1992bayesian} %
has been shown to be a scalable and fast option to obtain predictive uncertainties from already-trained networks in a post-hoc fashion~\cite{daxberger2021laplace,ritter2018scalable}. 
In our study, we extend NeRFs and GS with these various uncertainty estimation techniques where applicable (see \cref{sec:method}). %

\paragraph{Related Work on Uncertainty Estimation in NeRFs and GS} 
Various NeRF-based methods have explored different facets of uncertainty estimation. 
ActiveNeRF~\cite{pan2022activenerf} models a Gaussian distribution over rendered RGB pixels with the goal of next-best view selection, which spurred interest in this application~\cite{ran2023neurar,jin2023neu} as well as exploring more flexible probability distributions~\cite{seo2023mixnerf,seo2023flipnerf}.  
Estimating epistemic uncertainty in few-view settings was studied in \cite{shen2021stochastic,shen2022conditional}, which require significant modifications to the NeRF architecture as they use variational inference~\cite{zhang2018advances} for optimization. 
Later works have focused on architecture-agnostic approaches, \eg, \cite{sunderhauf2023density} which used ensembles of efficient NeRF backbones~\cite{muller2022instant} to estimate uncertainties. 
Additionally, \cite{amini2023calibrated} proposes a calibration method for correcting uncalibrated predictions on novel scenes of already-trained NeRFs. 
Bayes' Rays~\cite{goli2024bayes} uses perturbations in a spatial grid to define a NeRF architecture-agnostic spatial uncertainty estimated using the Laplace approximation, while FisherRF~\cite{jiang2023fisherrf} computes the Fisher information over the parameters in both NeRF- and GS-based methods to quantify the uncertainty of novel views.\looseness-1 

Recently, robustness to confounding outliers and removing distractors has been studied for both NeRF~\cite{sabour2023robustnerf,ren2024nerf,martin2021nerf} and GS~\cite{sabour2024spotlesssplats,kulhanek2024wildgaussians}. These works have studied using view-specific embeddings~\cite{martin2021nerf}, leveraging pre-trained networks~\cite{ren2024nerf,kulhanek2024wildgaussians}, and using robust optimization~\cite{sabour2023robustnerf,sabour2024spotlesssplats} to learn what scene parts are static and dynamic. 
Other works have aimed to consider aleatoric noise, \eg, motion blur and rolling shutter effects, by explicitly modeling for these in the pipeline~\cite{ma2022deblur,seiskari2024gaussian}. 
Furthermore, camera pose optimizers have been proposed to correct inaccurate camera parameters alongside optimizing the scene~\cite{lin2021barf,jeong2021self,matsuki2024gaussian}. 
Our study is complementary to the works mentioned above, because our aim is to provide an in-depth analysis of all the different sources of uncertainty and assess their influence on the reconstructed scenes through our experiments.

\section{Neural Radiance Fields and Gaussian Splatting}
\label{sec:background}
\paragraph{Neural Radiance Fields} 
NeRFs~\cite{mildenhall2020nerf} learn a continuous volumetric scene representation with a neural network $f_{\vtheta}$ that maps spatial coordinates $\vx \in \R^3$ and viewing directions $\vd \in \bS^2$ into volumetric density $\sigma \in \R^{+}$ and RGB color $\vc \in \R^3$ at the coordinate $\vx$. 
The pixel colors are rendered by approximating the volume rendering integral using quadrature rule as
\begin{equation}\label{eq:nerf-rendering}
  \vc_{\text{NeRF}} = \textstyle\sum_{i=1}^{N_s} T_i \alpha_i \vc_i, \text{~with~}  T_i = \textstyle\prod_{i=1}^{j-1} (1 - \alpha_i), \quad \alpha_j = (1 - \exp{(-\sigma_j \delta_j)}),
\end{equation}
where $T_i$ is the transmittance, $\alpha_i$ is the alpha value for $\vx_i$, and $\delta_i = \|\vx_{i+1} - \vx_{i}\|_2$ is the distance between adjacent coordinates. The network is optimized with gradient descent by minimizing the reconstruction loss %
    $\gL(\vtheta) =\sum_{\vr \sim \mathcal{R}} \| \vc_{\text{NeRF}}(\vr) - \vc_{\text{GT}}(\vr)\|_{2}^{2}$, 
where $\vr$ is a batch of rays drawn from the set of all rays $\mathcal{R}$ going from the camera origin through the pixels in the view. Many efforts have been made to improve the training speed~\cite{muller2022instant} and rendering quality~\cite{barron2022mip,barron2023zip} in NeRFs. 

\paragraph{Gaussian Splatting} While a NeRF is an implicit scene representation, 3D Gaussian Splatting (GS, \cite{kerbl20233d}) is an explicit point-based representation. More specifically, GS learns a set of 3D Gaussians, each one parameterized using its center position $\vmu \in \R^3$, a 3D covariance matrix $\mSigma \in \R^{3 \times 3}$, opacity $o \in [0, 1]$, and color $\vc$. The color $\vc$ is predicted using spherical harmonics for modeling view-dependent appearance. 
Rendering pixels %
begins by sorting the Gaussians that overlap with the target pixel according to their depth, which can be computed via the viewing transformation $\mW$. The pixel color is then rendered using alpha blending of the sorted Gaussians as
\begin{equation}\label{eq:gs-rendering}
    \vc_{\text{GS}} = \textstyle\sum_{i=1}^{N_p} T_i \alpha_i  \vc_i, \quad \text{~with~} \quad T_i = \textstyle\prod_{j=1}^{i-1} (1 - \alpha_j),
\end{equation}
where $N_p$ is the number of overlapping Gaussians. The blending weight $\alpha_i$ is obtained by 
    $\alpha_i = o_i \cdot \exp\!\left( - \frac{1}{2} (\vx' - \vmu_i')^{\top} \mSigma_i' (\vx' - \vmu_i') \right)$,
where $\vx'$ and $\vmu_i'$ are coordinates in the projected space, and the 2D covariance is $\mSigma' = \mJ \mW \mSigma \mW^{\top} \mJ^{\top}$ where $\mJ$ is the Jacobian of the affine approximation of the projective  transformation \cite{kerbl20233d}. The scene is optimized by minimizing the reconstruction loss
    $\gL = (1 - \lambda) \gL_1 + \lambda \gL_{\text{SSIM}}$,
where $\gL_1$ is an $L_1$ loss, $\gL_{\text{SSIM}}$ is a D-SSIM loss, and $\lambda$ is the weighting factor. The Gaussians are typically initialized from a sparse point cloud obtained from a Structure-from-Motion approach (\eg, COLMAP \cite{schoenberger2016sfm}). A density control procedure performs duplication and splitting of Gaussians in over- and under-reconstructed areas during the optimization, as well as pruning of highly transparent and overly large Gaussians. 
GS has become immensely popular due to its efficient training and high-quality rendering.

\section{Methods}
\label{sec:method}
We take approaches designed for aleatoric and epistemic uncertainty estimation and apply these to NeRF and GS. In particular, for the aleatoric ones, we adapt the approach proposed in Active-NeRF~\cite{pan2022activenerf}, while, for the epistemic approaches, we use MC-Dropout~\cite{gal2016dropout}, the Laplace approximations~\cite{daxberger2021laplace} and ensembles~\cite{lakshminarayanan2017simple}. We limit the use of MC-Dropout and LA to NeRFs since both are Bayesian deep learning methods and, thus, non-trivial to extend to GS. See \cref{app:methods} for details. 

\paragraph{Active-NeRF/GS} Both NeRF and GS can be modified with a learned uncertainty parameter to enable uncertainty estimation for the rendered pixels \cite{martin2021nerf,pan2022activenerf,jiang2023fisherrf}. The idea is to treat the color $\vc$ as a Gaussian random variable, \ie, $\vc \sim \mathcal{N}(\vc; \bar{\vc}, \beta)$, where the mean $\bar{\vc} \in \mathbb{R}^3$ and variance $\beta \in \mathbb{R}^{+}$ are learned. For NeRF, the variance is predicted with an additional network output. We render the pixel color and its uncertainty using
\begin{align}\label{eq:active-nerf-color-and-unc}
  \vc_{\text{Active-NeRF}} = \textstyle\sum_{i=1}^{N_s} T_i \alpha_i \bar{\vc}_i \text{~~and~~} \text{Var}(\vc_{\text{Active-NeRF}}) = \sum_{i=1}^{N_s} T_i^2 \alpha_i^2 \beta_i ,
\end{align} 
where the transmittance $T_i$ and alpha value $\alpha_i$ are squared~\cite{pan2022activenerf}. We use the Gaussian log-likelihood loss from Active-NeRF\cite{pan2022activenerf} during training with an added $L_1$ regularizer to enforce sparser density values.  
For GS, we add an additional learnable parameter to every 3D Gaussian. We render the pixel color and its uncertainty as in \cite{martin2021nerf} using 
\begin{align}\label{eq:active-gs-color-and-unc}
  \vc_{\text{Active-GS}} = \textstyle\sum_{i=1}^{N_p} T_i \alpha_i \bar{\vc}_i \text{~~and~~} \text{Var}(\vc_{\text{Active-GS}}) = \sum_{i=1}^{N_p} T_i \alpha_i \beta_i ,
\end{align} 
which allows rendering the uncertainty similarly as the color is rasterized in GS. 
The $\gL_1$-loss is replaced with the Gaussian log-likelihood to learn the per-Gaussian uncertainties during training. 
We noticed that the number of Gaussians increases fast when learning the uncertainty. Hence, we apply $L_1$ regularization to sparsify the opacity values to reduce the memory footprint of the method.\looseness-1

\paragraph{MC-Dropout NeRF} As the NeRF architecture mainly consists of MLPs, MC-Dropout can be used for uncertainty estimation by adding dropout layers into the network. The uncertainty is estimated by applying dropout $M$ times during inference to obtain $M$ rendered RGB predictions $\{\vc_{\text{NeRF}}^{(m)}\}_{m=1}^{M}$ from a parameter set $\{\hat{\vtheta}^{(m)}\}_{m=1}^{M}$ of masked weights $\hat{\vtheta}^{(m)} \subset \vtheta$. We compute the mean prediction and its variance across the rendered RGB images as
\begin{align}
  \begin{split}\label{eq:mcdropout-nerf-color-and-unc}
    \vc_{\text{MC-Dropout}} & = \textstyle\frac{1}{M} \sum_{m=1}^{M} \vc_{\text{NeRF}}^{(m)} = \frac{1}{M} \sum_{m=1}^{M} \sum_{i=1}^{N_s} T_i^{(m)} \alpha_i^{(m)} \vc_i^{(m)}, \\ %
    \text{Var}(\vc_{\text{MC-Dropout}}) & \textstyle\approx \frac{1}{M} \sum_{m=1}^{M} \vc_{\text{MC-Dropout}}^2 - \left( \frac{1}{M} \sum_{m=1}^{M} \vc_{\text{MC-Dropout}} \right)^2 . %
  \end{split}
\end{align}

\paragraph{Laplace NeRF} The Laplace approximation~\cite{mackay1992bayesian, daxberger2021laplace, ritter2018scalable} can be applied post-hoc to any network architecture for uncertainty estimation.  %
The idea is to approximate the intractable posterior distribution over the weights with a Gaussian distribution centered around the mode of $p(\vtheta \mid \gD)$, where $\gD$ is the training data set. 
We set the mean to be a local maximum (mode) of the posterior $\vtheta^{*} = \argmax_{\vtheta} \log{p(\vtheta \mid \gD)}$ obtained by training the network until convergence. %
The covariance matrix is obtained by Taylor expanding $\log{p(\vtheta \mid \gD)}$ around $\vtheta^{*}$ up to a second order as
\begin{equation}\label{eq:laplace_logposterior_taylor_expansion}
        \log{h(\vtheta)} \approx \log{h(\vtheta^{*})} - \frac{1}{2}(\vtheta - \vtheta^{*})^\top \Hess (\vtheta - \vtheta^{*}), 
\end{equation}
where the first-order term is zero at $\vtheta^{*}$ and $\Hess = - \nabla_{\vtheta}^2 \log{h(\vtheta)}|_{\vtheta^{*}}$ is the Hessian matrix of the unnormalized log-posterior at $\vtheta^{*}$. This yields the Laplace posterior approximation as the Gaussian distribution 
    $p(\vtheta \mid \gD) \approx q(\vtheta) = \mathcal{N}(\vtheta \mid \vtheta^{*}, \Hess^{-1})$,
with $\vtheta^{*}$ and $\Hess^{-1}$ as the mean and covariance respectively. The Hessian can be efficiently approximated using the generalised Gauss--Newton matrix~\cite{botev2017practical,immer2021improving} and assume a diagonal form to trade-off between scalability and approximation quality. For NeRF prediction, we compute an empirical color mean $\hat{\vc}$ and variance $\hat{\beta}$ for every input coordinate $\vx$ along the ray via Monte Carlo sampling from the approximate posterior. The pixel and its uncertainty are then rendered using: %
\begin{equation}\label{eq:laplace-nerf-color-and-unc}
  \vc_{\text{Laplace}} = \textstyle\sum_{i=1}^{N_s} T_i \alpha_i \hat{\vc}_i \text{~~and~~}
  \text{Var}(\vc_{\text{Laplace}}) = \textstyle\sum_{i=1}^{N_s} T_i^2 \alpha_i^2 \hat{\beta}_i .
\end{equation}

\paragraph{Ensemble NeRF/GS} Epistemic uncertainty can be measured by training an ensemble of models and averaging their predictions together \cite{lakshminarayanan2017simple}. This involves training $M$ models with different weight initializations to obtain solutions from different local minima. The prediction from each trained model are combined to compute the prediction and predictive variance as
\begin{equation}\label{eq:ensemble-color-and-unc}
  \!\!\vc_{\text{ens}} = \textstyle\frac{1}{M} \sum_{m=1}^{M} \vc_{\text{NeRF/GS}}^{(m)} \text{~and~}
  \text{Var}(\vc_{\text{ens}}) \approx \frac{1}{M} \sum_{m=1}^{M} \vc_{\text{ens}}^2 - \left( \frac{1}{M} \sum_{m=1}^{M} \vc_{\text{ens}} \right)^2 .
\end{equation}
Analogously, the ensemble approach can be applied to both NeRFs and GS by training multiple models on the scene. Ensembling can be viewed as a approximate Bayesian method, where the model parameters have been sampled from the posterior distribution to capture its multi-modality \cite{gustafsson2020evaluating}.

\section{Experiments}
\label{sec:experiments}
We present four-fold experiments covering the different uncertainty aspects \num{1}--\num{4} (see \cref{fig:uncertainty}).
We implement all methods in Nerfstudio~\cite{tancik2023nerfstudio} with their default settings unless stated otherwise. All experiments are run on a Nvidia A100 GPU on a cluster. 
The code is publicly available under the MIT license\footnote{Code: \url{https://github.com/AaltoML/uncertainty-nerf-gs}}. 

\paragraph{Methods} We base each NeRF and GS method on Nerfacto and Splatfacto respectively from Nerfstudio~\cite{tancik2023nerfstudio}. 
\begin{itemize}[leftmargin=*,itemsep=3pt]
  \item \textbf{Active-Nerfacto:} Re-implementation of Active-NeRF~\cite{pan2022activenerf} using Nerfacto architecture with an additional uncertainty output at the density MLP. The uncertainty is rendered using \cref{eq:active-nerf-color-and-unc}. %
  \item \textbf{Active-Splatfacto:} Splatfacto with an uncertainty parameter added to each Gaussian primitive that is rendered using \cref{eq:active-gs-color-and-unc}. %
  \item \textbf{MC-Dropout-Nerfacto:} Nerfacto with dropout layers added in the color and density MLPs. Uncertainty is estimated by performing $M$ forward passes with %
  masked weights sampled using dropout at test time~\cite{gal2016dropout} (see \cref{eq:mcdropout-nerf-color-and-unc}). 
  \item \textbf{Laplace-Nerfacto:} Nerfacto with the Laplace approximation applied to the last-layers of the color and density MLPs to estimate uncertainty (see \cref{eq:laplace-nerf-color-and-unc}).  %
  \item \textbf{Ensemble-Nerfacto:} Ensemble of $M$ Nerfacto models trained with different initializations \cite{lakshminarayanan2017simple}. The uncertainty is estimated by computing the variance on the rendered RGB images from each model (see \cref{eq:ensemble-color-and-unc}). 
  \item \textbf{Ensemble-Splatfacto:} Ensemble of $M$ Splatfacto models trained with different initializations. Same uncertainty estimation as for Ensemble-Nerfacto. 
\end{itemize}
For MC-Dropout and Ensemble, we set $M=5$ to estimate the uncertainty maps. See \cref{app:methods} for more details on the methods.

\paragraph{Data Sets} 
We use the Mip-NeRF 360~\cite{barron2022mip} and Blender~\cite{mildenhall2020nerf} data sets to assess the uncertainty aspects \num{1} and \num{2}. For \num{2}, we also use the Light Field (LF) data set~\cite{yucer2016efficient,zhang2020nerf++} and follow the few-shot setting from \cite{shen2021stochastic, shen2022conditional}. 
For \num{3}, we use scenes from the RobustNeRF~\cite{sabour2023robustnerf} and On-the-go~\cite{ren2024nerf} data sets to evaluate robustness against confounding objects in the training views. 
Finally, we use scenes from Mip-NeRF 360 data set to assess the input pose uncertainty aspect~\num{4}. 

\paragraph{Metrics} 
We focus on assessing novel view synthesis and corresponding per-pixel uncertainty estimates. Image quality is evaluated using PSNR, SSIM \cite{zhang2018unreasonable}, and LPIPS~\cite{wang2003multiscale} for generated test views. For uncertainty estimation, we use the (test) negative log likelihood (NLL, \cite{lakshminarayanan2017simple,loquercio2020general}) that captures both the reconstruction error and corresponding uncertainty (see \cite{shen2021stochastic,shen2022conditional}). 
Furthermore, we compare the methods using sparsification curves \cite{qu2021bayesian,bae2021estimating,poggi2020uncertainty,ilg2018uncertainty} and the related Area Under Sparsification Error (AUSE) metric that summarizes how well the uncertainty is correlated with the prediction error. 
As a perfect AUSE can be achieved despite over/under-estimating the uncertainty, we use the Area Under Calibration Error (AUCE) metric proposed in \cite{gustafsson2020evaluating} to assess the calibration of each method.

\begin{figure}[t]
  \centering
  \setlength{\figurewidth}{0.2\textwidth}
  \setlength{\figureheight}{.08\textheight}
  \begin{subfigure}[b]{.9\linewidth}
    \centering
    \pgfplotsset{every axis title/.append style={at={(0.5,0.80)}}} 
\pgfplotsset{every axis x label/.append style={at={(0.5,0.05)}}}
\pgfplotsset{every axis y label/.append style={at={(0.15,0.5)}}}

\begin{tikzpicture}
\tikzstyle{every node}=[font=\scriptsize]
\definecolor{crimson2143940}{RGB}{214,39,40}
\definecolor{darkgray176}{RGB}{176,176,176}
\definecolor{darkorange25512714}{RGB}{255,127,14}
\definecolor{forestgreen4416044}{RGB}{44,160,44}
\definecolor{mediumpurple148103189}{RGB}{148,103,189}
\definecolor{sienna1408675}{RGB}{140,86,75}
\definecolor{steelblue31119180}{RGB}{31,119,180}

\definecolor{lightgray204}{RGB}{204,204,204}

\begin{groupplot}[
  group style={group size= 3 by 2, horizontal sep=1.25cm, vertical sep=.5cm},
  tick align=outside,
  tick pos=left,
  grid=both,
  xlabel={Noise scale},
  xmin=-0.1, xmax=2.1,
  xtick={0,1,2},
  xticklabels={0,0.1,0.2},
  xtick style={color=black},
  x grid style={darkgray176,solid},
  y grid style={darkgray176,solid},
  ytick style={color=black},
  tick label style={font=\tiny}
]

\nextgroupplot[
height=\figureheight,
width=\figurewidth,
legend cell align={left},
legend columns=3,
legend style={
  nodes={scale=0.8},
  fill opacity=0.8, 
  draw opacity=1, 
  text opacity=1, 
  at={(0.03,1.1)}, 
  anchor=south west,
  draw=lightgray204},
xlabel=\empty,
xticklabel=\empty,
ylabel={PSNR~$\rightarrow$},
ymin=20.3102368195852, ymax=27.3380916436513,
ytick style={color=black}
]
\addplot [thick, steelblue31119180, mark=*, mark size=1, mark options={solid}]
table {%
0 23.8740194108751
1 23.5029703776042
2 21.9756047990587
};
\addlegendentry{Active-Nerfacto}
\addplot [thick, darkorange25512714, mark=*, mark size=1, mark options={solid}]
table {%
0 22.4727408091227
1 22.569727367825
2 21.0084487067329
};
\addlegendentry{Active-Splatfacto}
\addplot [thick, forestgreen4416044, mark=*, mark size=1, mark options={solid}]
table {%
0 22.6171943876478
1 22.334359910753
2 21.5347341961331
};
\addlegendentry{MC-Dropout-Nerfacto}
\addplot [thick, crimson2143940, mark=*, mark size=1, mark options={solid}]
table {%
0 22.4005655712552
1 21.9973420037164
2 20.6296847661336
};
\addlegendentry{Laplace-Nerfacto}
\addplot [thick, mediumpurple148103189, mark=*, mark size=1, mark options={solid}]
table {%
0 25.2237396240234
1 24.8094796074761
2 23.4711473253038
};
\addlegendentry{Ensemble-Nerfacto}
\addplot [thick, sienna1408675, mark=*, mark size=1, mark options={solid}]
table {%
0 27.0186436971029
1 25.0420648786757
2 22.38482454088
};
\addlegendentry{Ensemble-Splatfacto}

\nextgroupplot[
height=\figureheight,
width=\figurewidth,
xlabel=\empty,
xticklabel=\empty,
ylabel={SSIM~$\rightarrow$},
ymin=0.480235027604633, ymax=0.820075001319249,
ytick style={color=black}
]
\addplot [thick, steelblue31119180, mark=*, mark size=1, mark options={solid}]
table {%
0 0.733699613147312
1 0.66665733522839
2 0.589328295654721
};
\addplot [thick, darkorange25512714, mark=*, mark size=1, mark options={solid}]
table {%
0 0.695707185400857
1 0.642685595485899
2 0.563246157434252
};
\addplot [thick, forestgreen4416044, mark=*, mark size=1, mark options={solid}]
table {%
0 0.67000350356102
1 0.63779252105289
2 0.592538959450192
};
\addplot [thick, crimson2143940, mark=*, mark size=1, mark options={solid}]
table {%
0 0.665156006813049
1 0.621716744369931
2 0.562686724795236
};
\addplot [thick, mediumpurple148103189, mark=*, mark size=1, mark options={solid}]
table {%
0 0.771125919289059
1 0.730321890778012
2 0.674042562643687
};
\addplot [thick, sienna1408675, mark=*, mark size=1, mark options={solid}]
table {%
0 0.804627729786767
1 0.668944845596949
2 0.495682299137115
};

\nextgroupplot[
height=\figureheight,
width=\figurewidth,
xlabel=\empty,
xticklabel=\empty,
ylabel={$\leftarrow$~LPIPS},
ymin=0.162717424747017, ymax=0.467847581124968,
ytick style={color=black}
]
\addplot [thick, steelblue31119180, mark=*, mark size=1, mark options={solid}]
table {%
0 0.247570802768072
1 0.315553506215413
2 0.405008210076226
};
\addplot [thick, darkorange25512714, mark=*, mark size=1, mark options={solid}]
table {%
0 0.259776590598954
1 0.292903875311216
2 0.367652136418555
};
\addplot [thick, forestgreen4416044, mark=*, mark size=1, mark options={solid}]
table {%
0 0.32970024810897
1 0.361631848745876
2 0.414765338102976
};
\addplot [thick, crimson2143940, mark=*, mark size=1, mark options={solid}]
table {%
0 0.326107394364145
1 0.373305201530457
2 0.453978028562334
};
\addplot [thick, mediumpurple148103189, mark=*, mark size=1, mark options={solid}]
table {%
0 0.228619457946883
1 0.261565988262494
2 0.311135315232807
};
\addplot [thick, sienna1408675, mark=*, mark size=1, mark options={solid}]
table {%
0 0.176586977309651
1 0.270581858025657
2 0.424149973524941
};

\nextgroupplot[
height=\figureheight,
width=\figurewidth,
ylabel={$\leftarrow$~NLL},
ymin=-2.05502232602901, ymax=9.65894613191485,
ytick style={color=black}
]
\addplot [thick, steelblue31119180, mark=*, mark size=1, mark options={solid}]
table {%
0 -1.19427672359678
1 -1.27899430195491
2 -0.910379767417908
};
\addplot [thick, darkorange25512714, mark=*, mark size=1, mark options={solid}]
table {%
0 2.13446897102727
1 -0.0908242712418238
2 -0.32290573252572
};
\addplot [thick, forestgreen4416044, mark=*, mark size=1, mark options={solid}]
table {%
0 6.26601327790154
1 6.26904302173191
2 6.15238029095862
};
\addplot [thick, crimson2143940, mark=*, mark size=1, mark options={solid}]
table {%
0 7.4083449906773
1 7.45401774512397
2 9.12649302019013
};
\addplot [thick, mediumpurple148103189, mark=*, mark size=1, mark options={solid}]
table {%
0 -0.929207103119956
1 -0.82457605501016
2 -0.303033505876859
};
\addplot [thick, sienna1408675, mark=*, mark size=1, mark options={solid}]
table {%
0 -1.52256921430429
1 -1.37401976188024
2 -0.955274631579717
};

\nextgroupplot[
height=\figureheight,
width=\figurewidth,
ylabel={$\leftarrow$~AUSE},
ymin=0.270509494592746, ymax=0.541784371021721,
ytick style={color=black}
]
\addplot [thick, steelblue31119180, mark=*, mark size=1, mark options={solid}]
table {%
0 0.303053165475527
1 0.37176791495747
2 0.510363572173648
};
\addplot [thick, darkorange25512714, mark=*, mark size=1, mark options={solid}]
table {%
0 0.30360756152206
1 0.417276392380397
2 0.47402357061704
};
\addplot [thick, forestgreen4416044, mark=*, mark size=1, mark options={solid}]
table {%
0 0.475279129213757
1 0.521630687846078
2 0.526338133547041
};
\addplot [thick, crimson2143940, mark=*, mark size=1, mark options={solid}]
table {%
0 0.438191360897488
1 0.496116555399365
2 0.529453694820404
};
\addplot [thick, mediumpurple148103189, mark=*, mark size=1, mark options={solid}]
table {%
0 0.338944264584117
1 0.390462767746713
2 0.46210468477673
};
\addplot [thick, sienna1408675, mark=*, mark size=1, mark options={solid}]
table {%
0 0.282840170794063
1 0.45313345723682
2 0.500664876566993
};

\nextgroupplot[
height=\figureheight,
width=\figurewidth,
ylabel={$\leftarrow$~AUCE},
ymin=0.0491828801206215, ymax=0.497074754758009,
ytick style={color=black}
]
\addplot [thick, steelblue31119180, mark=*, mark size=1, mark options={solid}]
table {%
0 0.130791312755635
1 0.109509591124696
2 0.163397019990711
};
\addplot [thick, darkorange25512714, mark=*, mark size=1, mark options={solid}]
table {%
0 0.131509277600232
1 0.163285878768019
2 0.172597514065176
};
\addplot [thick, forestgreen4416044, mark=*, mark size=1, mark options={solid}]
table {%
0 0.383694742426325
1 0.402421212990735
2 0.417002459227045
};
\addplot [thick, crimson2143940, mark=*, mark size=1, mark options={solid}]
table {%
0 0.466542990198246
1 0.473790424474117
2 0.476716033183582
};
\addplot [thick, mediumpurple148103189, mark=*, mark size=1, mark options={solid}]
table {%
0 0.257554197665349
1 0.284176840336243
2 0.316285162012345
};
\addplot [thick, sienna1408675, mark=*, mark size=1, mark options={solid}]
table {%
0 0.186099422591707
1 0.104804854070817
2 0.0695416016950482
};

\end{groupplot}

\end{tikzpicture}%
    \caption{Gaussian noise applied to input images with varying noise scale (0 means no noise). }
    \label{fig:noisyview-mipnerf360-noise}
  \end{subfigure}
  \hfill
  \begin{subfigure}[b]{.09\linewidth}
    \centering\tiny
    \tikz\node[minimum width=\textwidth,minimum height=\textwidth,inner sep=0pt]{\includegraphics[width=\textwidth,height=\textwidth,trim={6.25cm 5.0cm 17cm 10.0cm},clip]{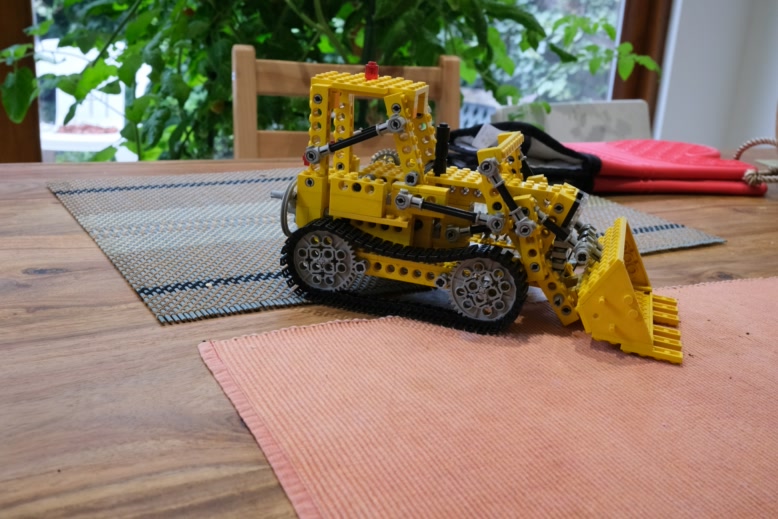}};\\
    No noise\\[4pt]
    \tikz\node[minimum width=\textwidth,minimum height=\textwidth,inner sep=0pt]{\includegraphics[width=\textwidth,height=\textwidth,trim={6.25cm 5.0cm 17cm 10.0cm},clip]{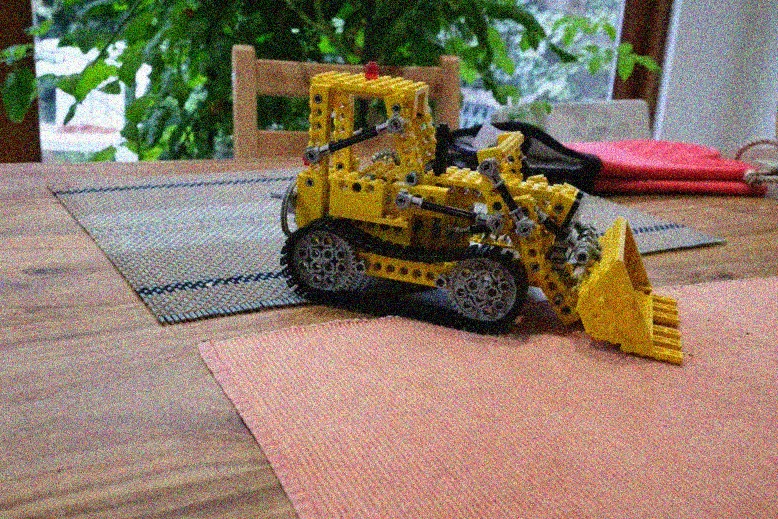}};\\
    $\nu = 0.1$ \\[4pt]
    \tikz\node[minimum width=\textwidth,minimum height=\textwidth,inner sep=0pt]{\includegraphics[width=\textwidth,height=\textwidth,trim={6.25cm 5.0cm 17cm 10.0cm},clip]{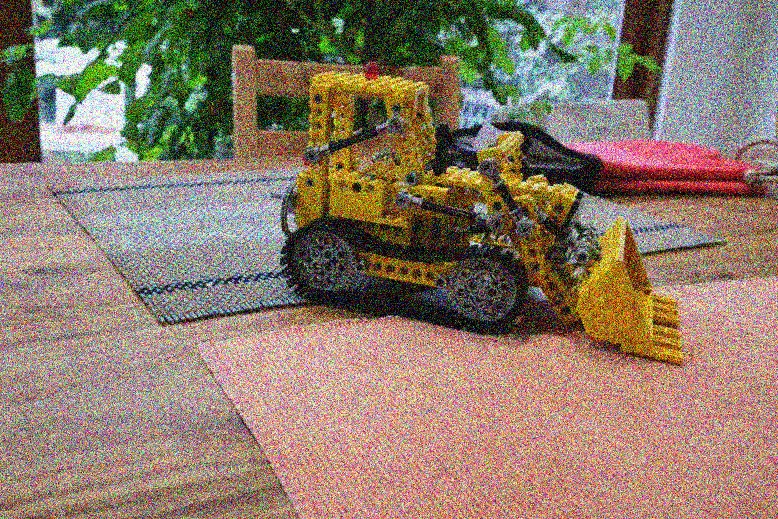}};\\
    $\nu = 0.2$\\[1.3em]~
  \end{subfigure}\\
  \begin{subfigure}[b]{.9\linewidth}
    \centering
    \pgfplotsset{every axis title/.append style={at={(0.5,0.80)}}} 
\pgfplotsset{every axis x label/.append style={at={(0.5,0.05)}}}
\pgfplotsset{every axis y label/.append style={at={(0.15,0.5)}}}

\begin{tikzpicture}
\tikzstyle{every node}=[font=\scriptsize]
\definecolor{crimson2143940}{RGB}{214,39,40}
\definecolor{darkgray176}{RGB}{176,176,176}
\definecolor{darkorange25512714}{RGB}{255,127,14}
\definecolor{forestgreen4416044}{RGB}{44,160,44}
\definecolor{gray127}{RGB}{127,127,127}
\definecolor{mediumpurple148103189}{RGB}{148,103,189}
\definecolor{orchid227119194}{RGB}{227,119,194}
\definecolor{sienna1408675}{RGB}{140,86,75}
\definecolor{steelblue31119180}{RGB}{31,119,180}

\definecolor{lightgray204}{RGB}{204,204,204}

\begin{groupplot}[
  group style={group size= 3 by 2, horizontal sep=1.25cm, vertical sep=.5cm},
  tick align=outside,
  tick pos=left,
  grid=both,
  xlabel={Kernel size},
  xmin=-0.1, xmax=2.1,
  xtick={0,1,2},
  xticklabels={0,7,15},
  xtick style={color=black},
  x grid style={darkgray176,solid},
  y grid style={darkgray176,solid},
  ytick style={color=black},
  tick label style={font=\tiny}
]

\nextgroupplot[
height=\figureheight,
width=\figurewidth,
legend cell align={left},
legend columns=3,
legend style={
  nodes={scale=0.8},
  fill opacity=0.8, 
  draw opacity=1, 
  text opacity=1, 
  at={(0.03,1.1)}, 
  anchor=south west,
  draw=lightgray204},
xlabel=\empty,
xticklabel=\empty,
ylabel={PSNR~$\rightarrow$},
ymin=19.9025772624546, ymax=27.4044806480408,
ytick style={color=black}
]
\addplot [thick, steelblue31119180, mark=*, mark size=1, mark options={solid}]
table {%
0 23.8740194108751
1 23.2362321217855
2 22.298807144165
};
\addplot [thick, darkorange25512714, mark=*, mark size=1, mark options={solid}]
table {%
0 22.4727408091227
1 20.9945499632094
2 20.854869418674
};
\addplot [thick, forestgreen4416044, mark=*, mark size=1, mark options={solid}]
table {%
0 22.6171943876478
1 21.8395608266195
2 20.9517285029093
};
\addplot [thick, crimson2143940, mark=*, mark size=1, mark options={solid}]
table {%
0 22.4005655712552
1 21.2367777294583
2 20.2435728708903
};
\addplot [thick, mediumpurple148103189, mark=*, mark size=1, mark options={solid}]
table {%
0 25.2404975891113
1 24.2640410529243
2 23.1080934736464
};
\addplot [thick, sienna1408675, mark=*, mark size=1, mark options={solid}]
table {%
0 27.063485039605
1 25.242483774821
2 23.6934388478597
};

\nextgroupplot[
height=\figureheight,
width=\figurewidth,
xlabel=\empty,
xticklabel=\empty,
ylabel={SSIM~$\rightarrow$},
ymin=0.526846115291119, ymax=0.819184302124712,
ytick style={color=black}
]
\addplot [thick, steelblue31119180, mark=*, mark size=1, mark options={solid}]
table {%
0 0.733699613147312
1 0.680241902669271
2 0.603958007362154
};
\addplot [thick, darkorange25512714, mark=*, mark size=1, mark options={solid}]
table {%
0 0.695707185400857
1 0.617217739423116
2 0.56066706445482
};
\addplot [thick, forestgreen4416044, mark=*, mark size=1, mark options={solid}]
table {%
0 0.67000350356102
1 0.617804033888711
2 0.558407425880432
};
\addplot [thick, crimson2143940, mark=*, mark size=1, mark options={solid}]
table {%
0 0.665156006813049
1 0.597902841038174
2 0.540134214692646
};
\addplot [thick, mediumpurple148103189, mark=*, mark size=1, mark options={solid}]
table {%
0 0.771827816963196
1 0.704601917001936
2 0.623098297251595
};
\addplot [thick, sienna1408675, mark=*, mark size=1, mark options={solid}]
table {%
0 0.805896202723185
1 0.707998282379574
2 0.617722041077084
};

\nextgroupplot[
height=\figureheight,
width=\figurewidth,
xlabel=\empty,
xticklabel=\empty,
ylabel={$\leftarrow$~LPIPS},
ymin=0.157294780471259, ymax=0.599369759733478,
ytick style={color=black}
]
\addplot [thick, steelblue31119180, mark=*, mark size=1, mark options={solid}]
table {%
0 0.247570802768072
1 0.362205944127507
2 0.512475050157971
};
\addplot [thick, darkorange25512714, mark=*, mark size=1, mark options={solid}]
table {%
0 0.259776590598954
1 0.422379675838682
2 0.531003167231878
};
\addplot [thick, forestgreen4416044, mark=*, mark size=1, mark options={solid}]
table {%
0 0.32970024810897
1 0.435532035099136
2 0.557575484116872
};
\addplot [thick, crimson2143940, mark=*, mark size=1, mark options={solid}]
table {%
0 0.326107394364145
1 0.455104281504949
2 0.579275442494286
};
\addplot [thick, mediumpurple148103189, mark=*, mark size=1, mark options={solid}]
table {%
0 0.229607306420803
1 0.364799439907074
2 0.522457980447345
};
\addplot [thick, sienna1408675, mark=*, mark size=1, mark options={solid}]
table {%
0 0.17738909771045
1 0.378424887855848
2 0.549604455629985
};

\nextgroupplot[
height=\figureheight,
width=\figurewidth,
ylabel={$\leftarrow$~NLL},
ymin=-2.10400164292918, ymax=10.1865787622001,
ytick style={color=black}
]
\addplot [thick, steelblue31119180, mark=*, mark size=1, mark options={solid}]
table {%
0 -1.19427672359678
1 -0.411547591081924
2 0.682429732961787
};
\addplot [thick, darkorange25512714, mark=*, mark size=1, mark options={solid}]
table {%
0 2.13446897102727
1 3.87357145547867
2 3.01651658448908
};
\addplot [thick, forestgreen4416044, mark=*, mark size=1, mark options={solid}]
table {%
0 6.26601327790154
1 6.89622917440203
2 7.26019139422311
};
\addplot [thick, crimson2143940, mark=*, mark size=1, mark options={solid}]
table {%
0 7.4083449906773
1 8.80896974934472
2 9.62791601651245
};
\addplot [thick, mediumpurple148103189, mark=*, mark size=1, mark options={solid}]
table {%
0 -0.943986160887612
1 -0.508686893930038
2 0.207264713115162
};
\addplot [thick, sienna1408675, mark=*, mark size=1, mark options={solid}]
table {%
0 -1.54533889724149
1 -1.11273472031785
2 -0.42817593117555
};

\nextgroupplot[
height=\figureheight,
width=\figurewidth,
ylabel={$\leftarrow$~AUSE},
ymin=0.260180508262581, ymax=0.485521920687623,
ytick style={color=black}
]
\addplot [thick, steelblue31119180, mark=*, mark size=1, mark options={solid}]
table {%
0 0.303053165475527
1 0.328577780061298
2 0.368505669964684
};
\addplot [thick, darkorange25512714, mark=*, mark size=1, mark options={solid}]
table {%
0 0.30360756152206
1 0.31485566827986
2 0.342828899621964
};
\addplot [thick, forestgreen4416044, mark=*, mark size=1, mark options={solid}]
table {%
0 0.475279129213757
1 0.459633360306422
2 0.474806272321277
};
\addplot [thick, crimson2143940, mark=*, mark size=1, mark options={solid}]
table {%
0 0.438191360897488
1 0.459630661540561
2 0.470328562789493
};
\addplot [thick, mediumpurple148103189, mark=*, mark size=1, mark options={solid}]
table {%
0 0.326354955633481
1 0.36131669415368
2 0.410263564851549
};
\addplot [thick, sienna1408675, mark=*, mark size=1, mark options={solid}]
table {%
0 0.270423299736447
1 0.336167529225349
2 0.393045152227084
};

\nextgroupplot[
height=\figureheight,
width=\figurewidth,
ylabel={$\leftarrow$~AUCE},
ymin=0.113682488049693, ymax=0.490076631580419,
ytick style={color=black}
]
\addplot [thick, steelblue31119180, mark=*, mark size=1, mark options={solid}]
table {%
0 0.130791312755635
1 0.261239725825672
2 0.325477725036402
};
\addplot [thick, darkorange25512714, mark=*, mark size=1, mark options={solid}]
table {%
0 0.131509277600232
1 0.230692127135098
2 0.265450894950962
};
\addplot [thick, forestgreen4416044, mark=*, mark size=1, mark options={solid}]
table {%
0 0.383694742426325
1 0.406136661661046
2 0.419075004056557
};
\addplot [thick, crimson2143940, mark=*, mark size=1, mark options={solid}]
table {%
0 0.466542990198246
1 0.470870092450427
2 0.472967806874477
};
\addplot [thick, mediumpurple148103189, mark=*, mark size=1, mark options={solid}]
table {%
0 0.253620567664045
1 0.299821538505289
2 0.324747325459715
};
\addplot [thick, sienna1408675, mark=*, mark size=1, mark options={solid}]
table {%
0 0.179853590133493
1 0.272116135268962
2 0.317297902786112
};

\end{groupplot}

\end{tikzpicture}%
    \caption{Gaussian blur applied to input images with varying kernel size (0 means no blurring). }
    \label{fig:noisyview-mipnerf360-blur}
  \end{subfigure}
  \hfill
  \begin{subfigure}[b]{.09\linewidth}
    \centering\tiny
    \tikz\node[minimum width=\textwidth,minimum height=\textwidth,inner sep=0pt]{\includegraphics[width=\textwidth,height=\textwidth,trim={6.25cm 5.0cm 17cm 10.0cm},clip]{figures-main/noisyview-mipnerf360-examples/kitchen/frame_00001.JPG}};\\
    No blur\\[4pt]
    \tikz\node[minimum width=\textwidth,minimum height=\textwidth,inner sep=0pt]{\includegraphics[width=\textwidth,height=\textwidth,trim={6.25cm 5.0cm 17cm 10.0cm},clip]{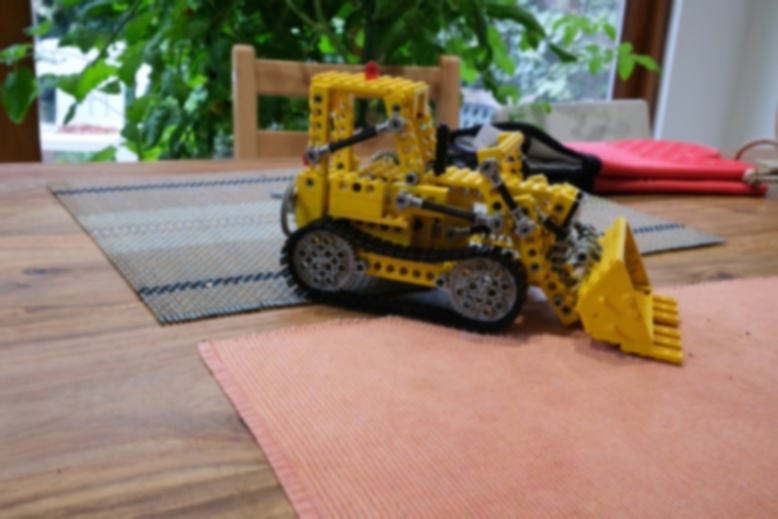}};\\
    $k = 7$ \\[4pt]
    \tikz\node[minimum width=\textwidth,minimum height=\textwidth,inner sep=0pt]{\includegraphics[width=\textwidth,height=\textwidth,trim={6.25cm 5.0cm 17cm 10.0cm},clip]{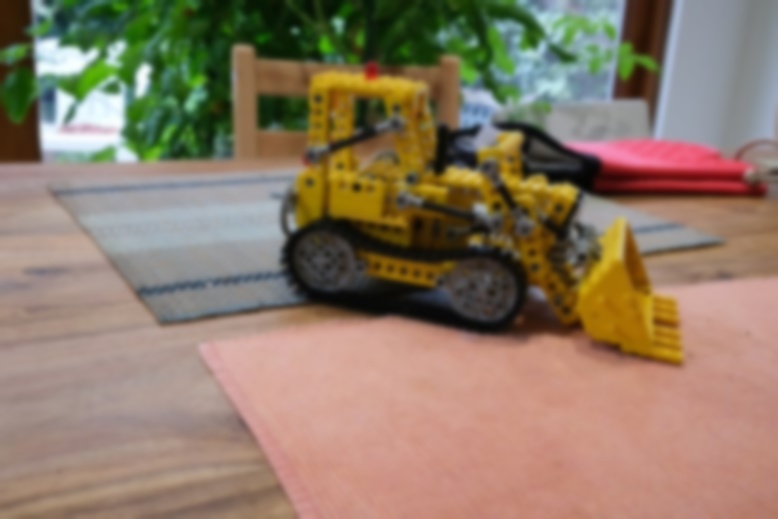}};\\
    $k = 15$\\[1.3em]~
  \end{subfigure}\\
  \caption{Confounding effects in input images \num{1}: %
  Performance metrics over different magnitudes of noise and blur applied to the training views averaged across scenes from the Mip-NeRF 360 data set. 
  }
  \label{fig:noisyview-mipnerf360}
\end{figure}

\subsection{Experiments on Aleatoric Uncertainty (\num{1})}
\label{sec:aleatoric}
We assess the sensitivity under different confounding effects applied to the training images. %
Following \cite{barron2022mip}, we downsize the images from the Mip-NeRF 360 scenes $\times4$ to a resolution of $1.0-1.6$ megapixels, and apply Gaussian noise with scales $\nu = 0.1$ and $0.2$, as well as Gaussian blur with a kernel size of $k= 7 $ and $15$ pixels (see examples in \cref{fig:noisyview-mipnerf360-examples-appendix}). We evaluate the performance on test images without noise 
that are regularly subsampled to contain 10\% of the original input images.\looseness-1

\paragraph{Mip-NeRF 360 Results} In \cref{fig:noisyview-mipnerf360}, we show the performance metrics when training with noisy and blurred input images. The image quality performance decreases for all methods for both confounding effects, where Active-Nerfacto and the Ensemble methods perform best. Active-Splatfacto performs similarly to MC-Dropout- and Laplace-Nerfacto, possibly due to the opacity regularization necessary to avoid high memory footprints. 
In \cref{fig:noisyview-mipnerf360-noise} (Gaussian noise), 
we note that the NLL is similar for the Active and Ensemble methods but for different reasons. For Ensemble-Nerfacto, the average variance is stable across the noise scales, while it increases for other methods (see \cref{fig:noisyview-mipnerf360-average-variance-appendix}). 
As its pixel error decreases more slowly, Ensemble-Nerfacto achieves a similar NLL as the other best methods. %
For AUCE, we note that the other methods are better calibrated than Ensemble-Nerfacto, although their AUSE scores are similar. 
In \cref{fig:noisyview-mipnerf360-blur} (Gaussian blur), Ensemble-Splatfacto achieves the best PSNR  but manages to capture structural and perceptual aspects similarly well as Ensemble-Nerfacto according to the SSIM and LPIPS scores. %
Their average variances decrease slightly, but the NLL still increases since their pixel errors become larger with blurrier training images. %
The AUSE and AUCE scores increase for all methods, which shows that all methods underfit the uncertainty with more blurred training views.

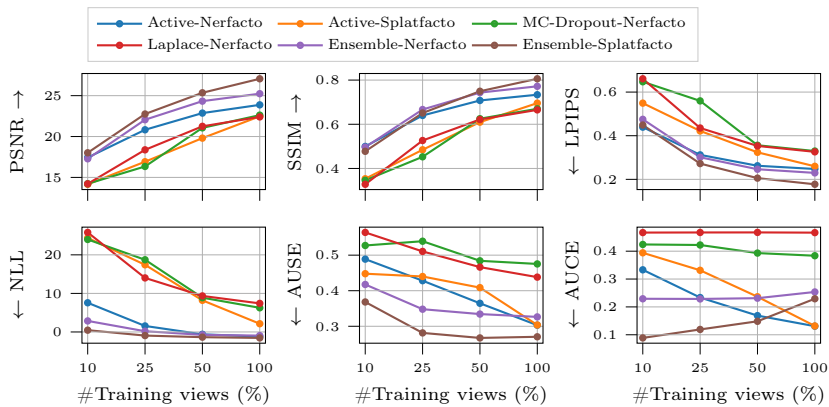
\begin{figure}[t]
  \centering
  \setlength{\figurewidth}{0.2\textwidth}
  \setlength{\figureheight}{.08\textheight}
  \pgfplotsset{every axis title/.append style={at={(0.5,0.80)}}} 
\pgfplotsset{every axis x label/.append style={at={(0.5,0.05)}}}
\pgfplotsset{every axis y label/.append style={at={(0.15,0.5)}}}

\begin{tikzpicture}
\tikzstyle{every node}=[font=\scriptsize]
\definecolor{crimson2143940}{RGB}{214,39,40}
\definecolor{darkgray176}{RGB}{176,176,176}
\definecolor{darkorange25512714}{RGB}{255,127,14}
\definecolor{forestgreen4416044}{RGB}{44,160,44}
\definecolor{lightgray204}{RGB}{204,204,204}
\definecolor{mediumpurple148103189}{RGB}{148,103,189}
\definecolor{sienna1408675}{RGB}{140,86,75}
\definecolor{steelblue31119180}{RGB}{31,119,180}

\begin{groupplot}[
  group style={group size= 3 by 2, horizontal sep=1.25cm, vertical sep=.5cm},
  tick align=outside,
  tick pos=left,
  xlabel={\#Training views (\%)},
  grid=both,
  xmin=-0.1, xmax=3.1,
  xtick={0,1,2,3},
  xticklabels={10,25,50,100},
  xtick style={color=black},
  x grid style={darkgray176,solid},
  y grid style={darkgray176,solid},
  ytick style={color=black},
  tick label style={font=\tiny}
]

\nextgroupplot[
height=\figureheight,
width=\figurewidth,
legend cell align={left},
legend columns=3,
legend style={
  nodes={scale=0.8},
  fill opacity=0.8, 
  draw opacity=1, 
  text opacity=1, 
  at={(0.03,1.1)}, 
  anchor=south west,
  draw=lightgray204},
ylabel={PSNR~$\rightarrow$},
xlabel=\empty,
xticklabel=\empty,
ymin=13.5135778956943, ymax=27.7087187131246,
]
\addplot [thick, steelblue31119180, mark=*, mark size=1, mark options={solid}]
table {%
0 17.4356079101562
1 20.8256729973687
2 22.8675140804715
3 23.8740194108751
};
\addlegendentry{Active-Nerfacto}
\addplot [thick, darkorange25512714, mark=*, mark size=1, mark options={solid}]
table {%
0 14.1588115692139
1 16.9187837176853
2 19.8043461905585
3 22.4727408091227
};
\addlegendentry{Active-Splatfacto}
\addplot [thick, forestgreen4416044, mark=*, mark size=1, mark options={solid}]
table {%
0 14.2221079932319
1 16.3394226498074
2 21.0598461363051
3 22.6171943876478
};
\addlegendentry{MC-Dropout-Nerfacto}
\addplot [thick, crimson2143940, mark=*, mark size=1, mark options={solid}]
table {%
0 14.2141388787164
1 18.3723644680447
2 21.247065226237
3 22.4005655712552
};
\addlegendentry{Laplace-Nerfacto}
\addplot [thick, mediumpurple148103189, mark=*, mark size=1, mark options={solid}]
table {%
0 17.2762996885512
1 22.0559352238973
2 24.3284687466092
3 25.2404975891113
};
\addlegendentry{Ensemble-Nerfacto}
\addplot [thick, sienna1408675, mark=*, mark size=1, mark options={solid}]
table {%
0 18.0120443767971
1 22.7542751100328
2 25.3511227501763
3 27.063485039605
};
\addlegendentry{Ensemble-Splatfacto}

\nextgroupplot[
height=\figureheight,
width=\figurewidth,
ylabel={SSIM~$\rightarrow$},
xlabel=\empty,
xticklabel=\empty,
ymin=0.304434649646282, ymax=0.829775324298276,
]
\addplot [thick, steelblue31119180, mark=*, mark size=1, mark options={solid}]
table {%
0 0.499653100967407
1 0.639761146571901
2 0.707741452587975
3 0.733699613147312
};
\addplot [thick, darkorange25512714, mark=*, mark size=1, mark options={solid}]
table {%
0 0.353367500834995
1 0.484487858083513
2 0.609612762928009
3 0.695707185400857
};
\addplot [thick, forestgreen4416044, mark=*, mark size=1, mark options={solid}]
table {%
0 0.346251899997393
1 0.452688641018338
2 0.625136809216605
3 0.67000350356102
};
\addplot [thick, crimson2143940, mark=*, mark size=1, mark options={solid}]
table {%
0 0.328313771221373
1 0.526490926742554
2 0.621714686354001
3 0.665156006813049
};
\addplot [thick, mediumpurple148103189, mark=*, mark size=1, mark options={solid}]
table {%
0 0.496396349536048
1 0.666767239570618
2 0.74329588148329
3 0.771827816963196
};
\addplot [thick, sienna1408675, mark=*, mark size=1, mark options={solid}]
table {%
0 0.477582212951448
1 0.651018758614858
2 0.749498334195879
3 0.805896202723185
};

\nextgroupplot[
height=\figureheight,
width=\figurewidth,
ylabel={$\leftarrow$~LPIPS},
xlabel=\empty,
xticklabel=\empty,
ymin=0.153212623091208, ymax=0.685095064714551,
]
\addplot [thick, steelblue31119180, mark=*, mark size=1, mark options={solid}]
table {%
0 0.439236382643382
1 0.3122345606486
2 0.262287500831816
3 0.247570802768072
};
\addplot [thick, darkorange25512714, mark=*, mark size=1, mark options={solid}]
table {%
0 0.549288577503628
1 0.421846785479122
2 0.323909988005956
3 0.259776590598954
};
\addplot [thick, forestgreen4416044, mark=*, mark size=1, mark options={solid}]
table {%
0 0.647234216332436
1 0.559585594468647
2 0.356297806733184
3 0.32970024810897
};
\addplot [thick, crimson2143940, mark=*, mark size=1, mark options={solid}]
table {%
0 0.660918590095308
1 0.435292897125085
2 0.353157036834293
3 0.326107394364145
};
\addplot [thick, mediumpurple148103189, mark=*, mark size=1, mark options={solid}]
table {%
0 0.475113835599687
1 0.301306924886174
2 0.246356144547462
3 0.229607306420803
};
\addplot [thick, sienna1408675, mark=*, mark size=1, mark options={solid}]
table {%
0 0.449032379521264
1 0.272289199961556
2 0.205615468323231
3 0.17738909771045
};

\nextgroupplot[
height=\figureheight,
width=\figurewidth,
ylabel={$\leftarrow$~NLL},
ymin=-2.91347389262584, ymax=27.1854960058298,
ytick style={color=black}
]
\addplot [thick, steelblue31119180, mark=*, mark size=1, mark options={solid}]
table {%
0 7.57043382856581
1 1.55749535891745
2 -0.609880194895797
3 -1.19427672359678
};
\addplot [thick, darkorange25512714, mark=*, mark size=1, mark options={solid}]
table {%
0 24.3851063516405
1 17.4176178508335
2 8.20463991165161
3 2.13446897102727
};
\addplot [thick, forestgreen4416044, mark=*, mark size=1, mark options={solid}]
table {%
0 23.9882398777538
1 18.7254042441232
2 8.89109775672356
3 6.26601327790154
};
\addplot [thick, crimson2143940, mark=*, mark size=1, mark options={solid}]
table {%
0 25.8173610104455
1 14.0164702335993
2 9.34446213642756
3 7.4083449906773
};
\addplot [thick, mediumpurple148103189, mark=*, mark size=1, mark options={solid}]
table {%
0 2.8634731852346
1 0.190963864326477
2 -0.75601135691007
3 -0.943986160887612
};
\addplot [thick, sienna1408675, mark=*, mark size=1, mark options={solid}]
table {%
0 0.46204592121972
1 -0.93886677424113
2 -1.35191483381722
3 -1.54533889724149
};

\nextgroupplot[
height=\figureheight,
width=\figurewidth,
ylabel={$\leftarrow$~AUSE},
ymin=0.25238764633735, ymax=0.57876041183869,
]
\addplot [thick, steelblue31119180, mark=*, mark size=1, mark options={solid}]
table {%
0 0.488798065318002
1 0.428185823890898
2 0.364608135488298
3 0.303053165475527
};
\addplot [thick, darkorange25512714, mark=*, mark size=1, mark options={solid}]
table {%
0 0.447831047905816
1 0.439996325307422
2 0.408674902386136
3 0.30360756152206
};
\addplot [thick, forestgreen4416044, mark=*, mark size=1, mark options={solid}]
table {%
0 0.527173091967901
1 0.539166725344128
2 0.484074162112342
3 0.475279129213757
};
\addplot [thick, crimson2143940, mark=*, mark size=1, mark options={solid}]
table {%
0 0.563925286134084
1 0.510478764772415
2 0.466351028945711
3 0.438191360897488
};
\addplot [thick, mediumpurple148103189, mark=*, mark size=1, mark options={solid}]
table {%
0 0.417526715331607
1 0.347996960083644
2 0.334294656912486
3 0.326354955633481
};
\addplot [thick, sienna1408675, mark=*, mark size=1, mark options={solid}]
table {%
0 0.368320521381166
1 0.281325161457062
2 0.267222772041957
3 0.270423299736447
};

\nextgroupplot[
height=\figureheight,
width=\figurewidth,
ylabel={$\leftarrow$~AUCE},
ymin=0.069787126277569, ymax=0.486297864132142,
]
\addplot [thick, steelblue31119180, mark=*, mark size=1, mark options={solid}]
table {%
0 0.333140188775636
1 0.233625767444806
2 0.168758575147007
3 0.130791312755635
};
\addplot [thick, darkorange25512714, mark=*, mark size=1, mark options={solid}]
table {%
0 0.394154505805122
1 0.33140388635745
2 0.236495438402949
3 0.131509277600232
};
\addplot [thick, forestgreen4416044, mark=*, mark size=1, mark options={solid}]
table {%
0 0.424128762473426
1 0.422220757127767
2 0.393183555533309
3 0.383694742426325
};
\addplot [thick, crimson2143940, mark=*, mark size=1, mark options={solid}]
table {%
0 0.466766662027516
1 0.467180615937727
2 0.467365557866025
3 0.466542990198246
};
\addplot [thick, mediumpurple148103189, mark=*, mark size=1, mark options={solid}]
table {%
0 0.229173440587098
1 0.228595404339919
2 0.231307218017495
3 0.253620567664045
};
\addplot [thick, sienna1408675, mark=*, mark size=1, mark options={solid}]
table {%
0 0.0887194325436859
1 0.119076366322484
2 0.148341871185838
3 0.229276956821967
};

\end{groupplot}

\end{tikzpicture}%
  \caption{Varying training views \num{2}: Including more training views (sampled uniformly) improves test metrics. Performance metrics over varying number of training views (in \%) averaged across scenes from the Mip-NeRF 360 data set.
  }
  \label{fig:varyingview-mipnerf360}
\end{figure}

\subsection{Experiments on Epistemic Uncertainty (\num{2})}
\label{sec:reducible}
We assess the sensitivity to epistemic uncertainty using three  settings: \emph{(i)} sensitivity to the number of training views; \emph{(ii)}  sensitivity to views that are different from the training views (out-of-distribution, OOD); \emph{(iii)}  sensitivity to few views with limited scene coverage. 

\paragraph{Varying Training Views} We randomly sample proportions of 10\%, 25\%, 50\%, and 100\% of the views for training. We regularly subsample 10\% of the input images as the test set before the training set sampling. We present results for the Mip-NeRF 360 data set and show results on Blender scenes in \cref{app:epistemic}. 

In \cref{fig:varyingview-mipnerf360}, we observe that Active-Nerfacto and the Ensemble methods achieve the highest image quality where Ensemble-Splatfacto surpasses the rest when more training views are used. 
Active-splatfacto achieves similar image quality as MC-Dropout- and Laplace-Nerfacto, possibly due to the opacity regularization necessary to avoid high memory footprints. 
The NLL and AUSE metrics decrease with more training views for all methods which indicates that the uncertainties correlate better with the prediction errors for each method. On the AUCE metric, we observe that most methods achieve lower scores when adding more views, while it increases for Ensemble-Nerfacto and -Splatfacto. The Active methods underestimate the variance when trained on fewer views, such that fewer pixels than expected are covered by the wider prediction intervals (higher $p$-value), but the estimates become better calibrated when adding more views. The opposite is observed for the Ensemble methods where the under-estimation of the uncertainty becomes worse when using more training views. See \cref{fig:varyingview-mipnerf360-auce-coverage} for coverage curves of the AUCE for each method.

\begin{table}[t]
  \centering
  \caption{OOD setting \num{2}: Metrics for RGB predictions on test views in the occluded hemisphere  (OOD) of scenes of Mip-NeRF 360 data set. The \first{first}, \second{second}, and \third{third} values are highlighted.
  }
  \begin{minipage}{.75\textwidth}
  \resizebox{\textwidth}{!}{
  \setlength{\tabcolsep}{6pt}
  \begin{tabular}{lcccccc}
  \toprule
  Method & PSNR $\uparrow$ & SSIM $\uparrow$ & LPIPS $\downarrow$ & NLL $\downarrow$ & AUSE $\downarrow$ & AUCE $\downarrow$ \\
  \midrule
  Active-Nerfacto & \cellcolor{orange!25}16.28 & \cellcolor{yellow!25}0.43 & \cellcolor{yellow!25}0.58 & \cellcolor{orange!25}7.23 & 0.45 & 0.31 \\  
  Active-Splatfacto & 13.63 & 0.37 & \cellcolor{yellow!25}0.58 & 14.30 & 0.44 & \cellcolor{orange!25}0.29 \\  
  MC-Dropout-Nerfacto & 16.07 & 0.41 & 0.63 & 11.14 & \cellcolor{orange!25}0.37 & 0.42 \\  
  Laplace-Nerfacto & \cellcolor{yellow!25}16.34 & 0.42 & 0.60 & 13.61 & \cellcolor{yellow!25}0.38 & 0.47 \\  
  Ensemble-Nerfacto & \cellcolor{red!25}16.84 & \cellcolor{orange!25}0.49 & \cellcolor{orange!25}0.50 & \cellcolor{red!25}3.54 & \cellcolor{red!25}0.33 & \cellcolor{yellow!25}0.30 \\  
  Ensemble-Splatfacto & 15.46 & \cellcolor{red!25}0.51 & \cellcolor{red!25}0.49 & \cellcolor{yellow!25}9.02 & 0.48 & \cellcolor{red!25}0.22 \\  
  \bottomrule
  \end{tabular}
  }
  \end{minipage}
  \hfill
  \begin{minipage}{.2\textwidth}
  \newlength{\fw}\newlength{\fh}
  \setlength{\fw}{5cm}
  \setlength{\fh}{.7\fw}
  
  \resizebox{\textwidth}{!}{%
  \begin{tikzpicture}[use Hobby shortcut]

    \tikzstyle{blob}=[draw=black!60, closed hobby, fill=none, opacity=0.75, line width=.75pt]

    \path[blob, opacity=.75] ([closed]0.57\fw,0.75\fh).. (0.80\fw,0.90\fh).. (0.93\fw,0.83\fh).. (0.98\fw,0.52\fh).. (0.93\fw,0.34\fh).. (0.71\fw,0.08\fh).. (0.53\fw,0.06\fh).. (0.41\fw,0.17\fh).. (0.39\fw,0.41\fh);

    \draw[black,dashed] (1.5,-.5) -- (5.5,3.5);

    \begin{scope}
      \clip (1.5,3.5) -- (1.5,-.5) -- (5.5,3.5) -- cycle;

      \path[blob,pattern=north west lines, pattern color=black!20, opacity=.75] ([closed]0.57\fw,0.75\fh).. (0.80\fw,0.90\fh).. (0.93\fw,0.83\fh).. (0.98\fw,0.52\fh).. (0.93\fw,0.34\fh).. (0.71\fw,0.08\fh).. (0.53\fw,0.06\fh).. (0.41\fw,0.17\fh).. (0.39\fw,0.41\fh);
    \end{scope}

    \newcommand{\tri}[3]{
      \begin{scope}[shift={(#1,#2)}, rotate=#3, scale=.33]
          \draw[thick] 
              (1.5,0) -- (0,0) -- (0.75,1.299);    
          \draw[thick, fill=lightgray] 
              (1,0) -- (0,0) -- (0.5,0.866); %
      \end{scope}
    }

    \tri{2}{-1}{30}
    \tri{4}{-.75}{90}
    \tri{5.5}{0}{100}  
    \tri{6.5}{1.5}{140}

    \tri{.5}{2}{-50}
    \tri{1.5}{3.5}{-70}  
    \tri{4}{4.5}{-110}

    \node[rotate=45] at (2,2.75) {\small Test views};
    \node[rotate=45] at (5,.75) {\small Training views};  
  
  \end{tikzpicture}%
  }
  \end{minipage}
  \hfill
  \label{tab:ood-mipnerf360}
  \end{table}

\paragraph{OOD Setting} We separate the training and test sets by splitting all input images based on their camera pose positions (see illustration in \cref{tab:ood-mipnerf360}). We make this split for the Mip-NeRF 360 scenes where the views are captured roughly in a hemisphere. Results for a similar setup on Blender are in \cref{app:epistemic}. 

In \cref{tab:ood-mipnerf360}, based on the PSNR, we see that the Nerfacto-based methods are better at obtaining pixel-wise accuracy on unseen camera views compared to the Splatfacto methods.  
However, the Ensemble-Nerfacto and -Splatfacto perform similarly on SSIM and LPIPS, which indicates that both methods manage to capture structural and perceptual aspects equally well on the OOD images. For the uncertainty metrics, Ensemble-Nerfacto obtains the best NLL and AUSE scores while Ensemble-Splatfacto achieves the best AUCE score.

\begin{figure}[t]
  \centering
  \setlength{\figurewidth}{0.136\textwidth}
  \begin{tikzpicture}[image/.style = {inner sep=0, outer sep=0, minimum width=\figurewidth, anchor=north west, text width=\figurewidth}, node distance = 1pt and 1pt, every node/.style={font= {\tiny}}, label/.style = {scale=0.75,font={\tiny},anchor=south,inner sep=0pt,outer sep=2pt,rotate=0}] 

    \node [image] (bicycle-1) {\includegraphics[width=\figurewidth]{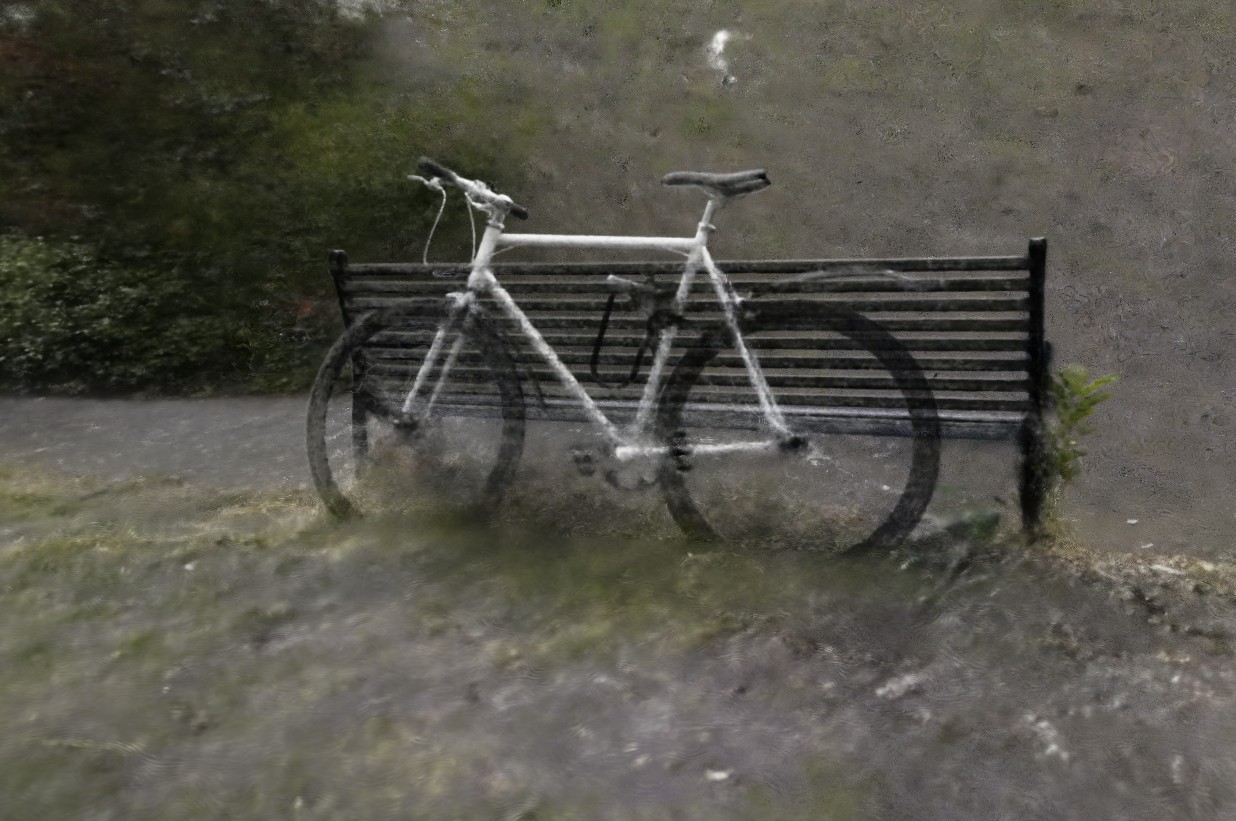}};
    \node [image,right=of bicycle-1] (bicycle-2) {\includegraphics[width=\figurewidth]{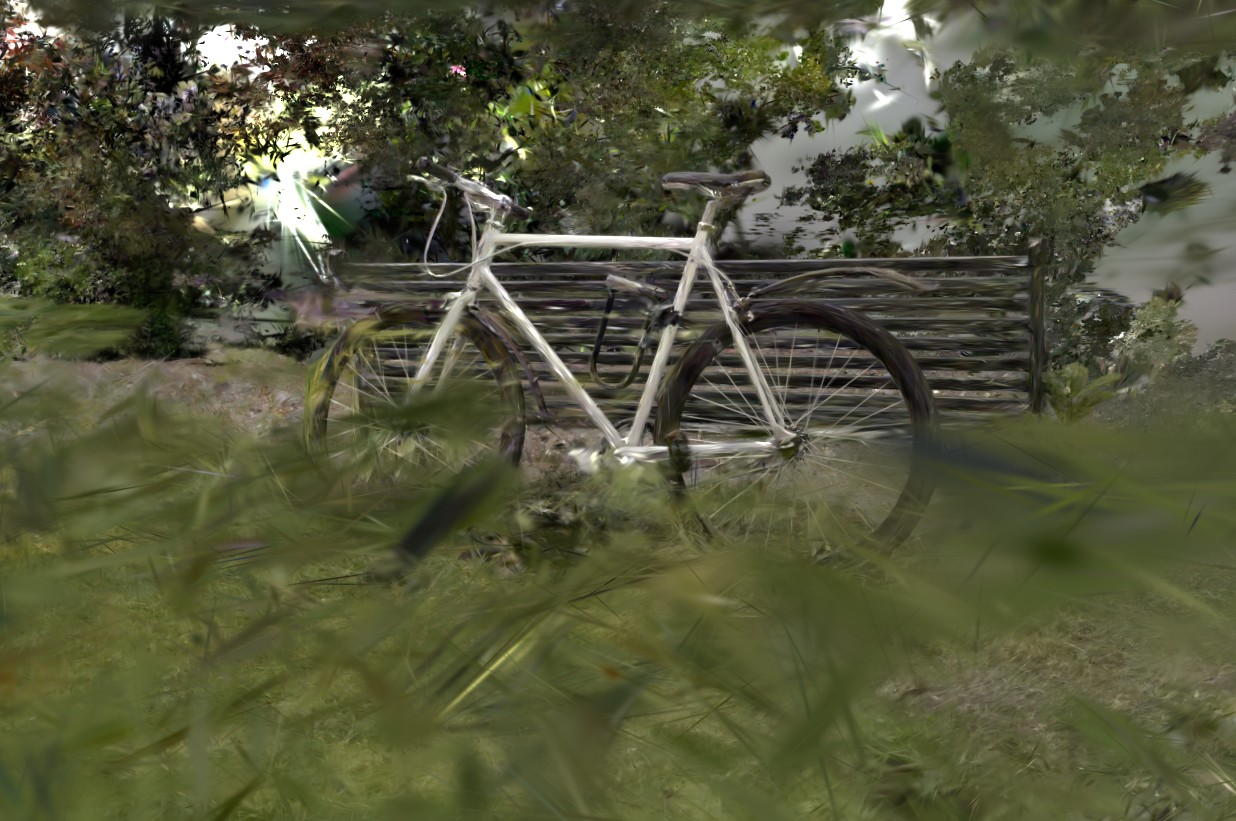}};
    \node [image,right=of bicycle-2] (bicycle-3) {\includegraphics[width=\figurewidth]{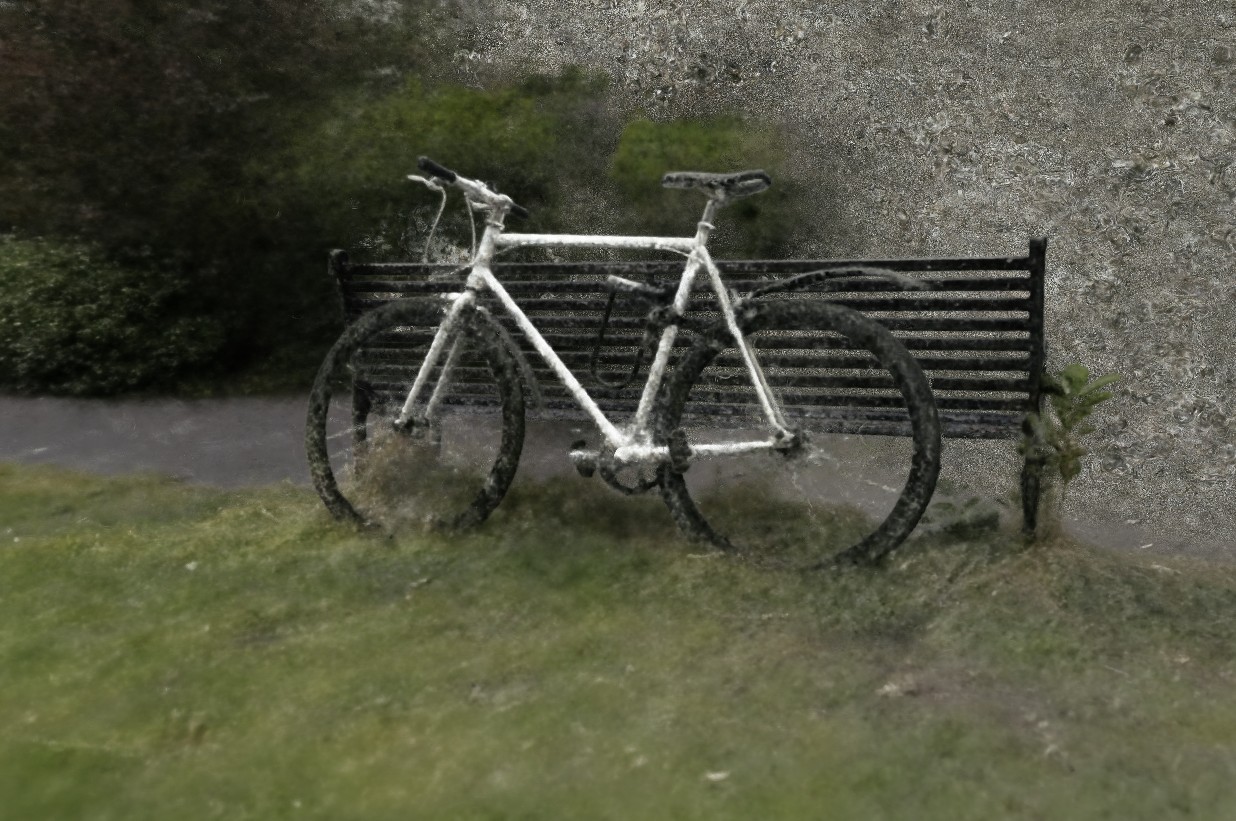}};
    \node [image,right=of bicycle-3] (bicycle-4) {\includegraphics[width=\figurewidth]{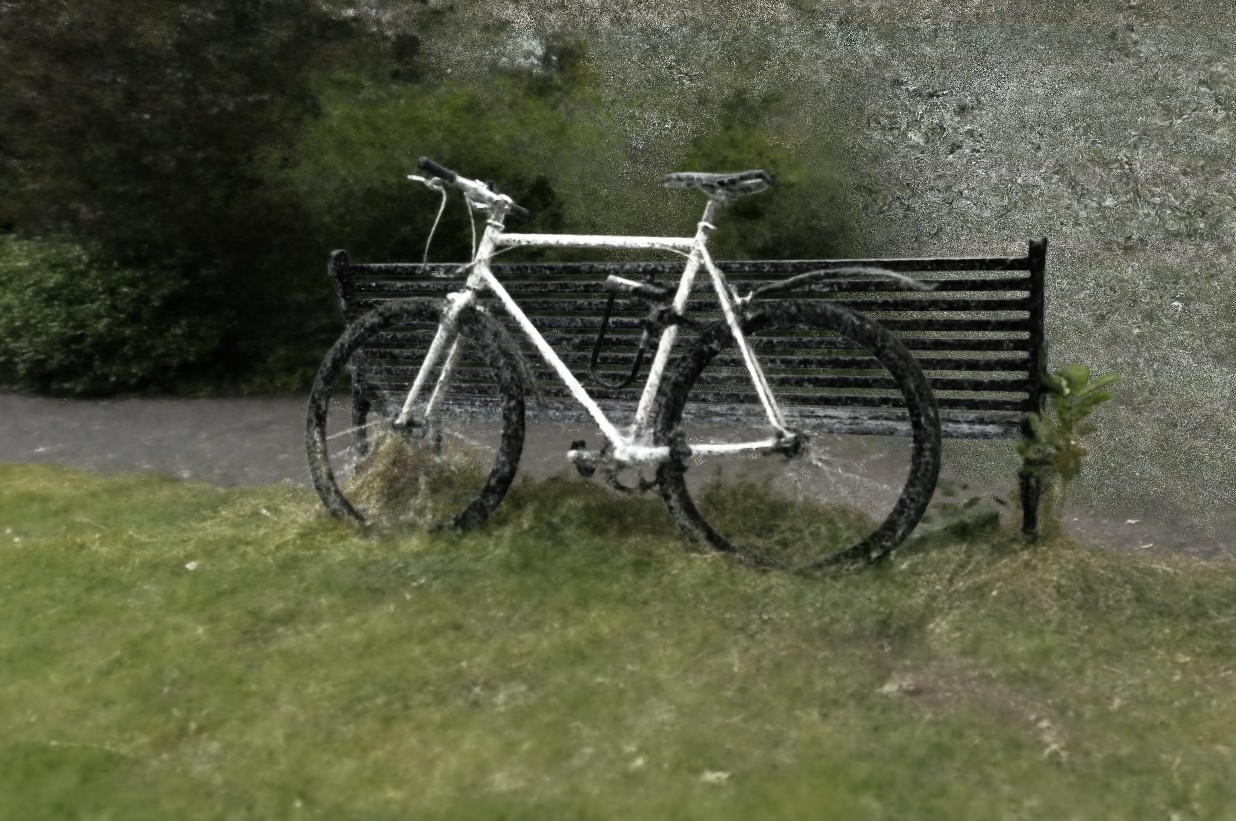}};
    \node [image,right=of bicycle-4] (bicycle-5) {\includegraphics[width=\figurewidth]{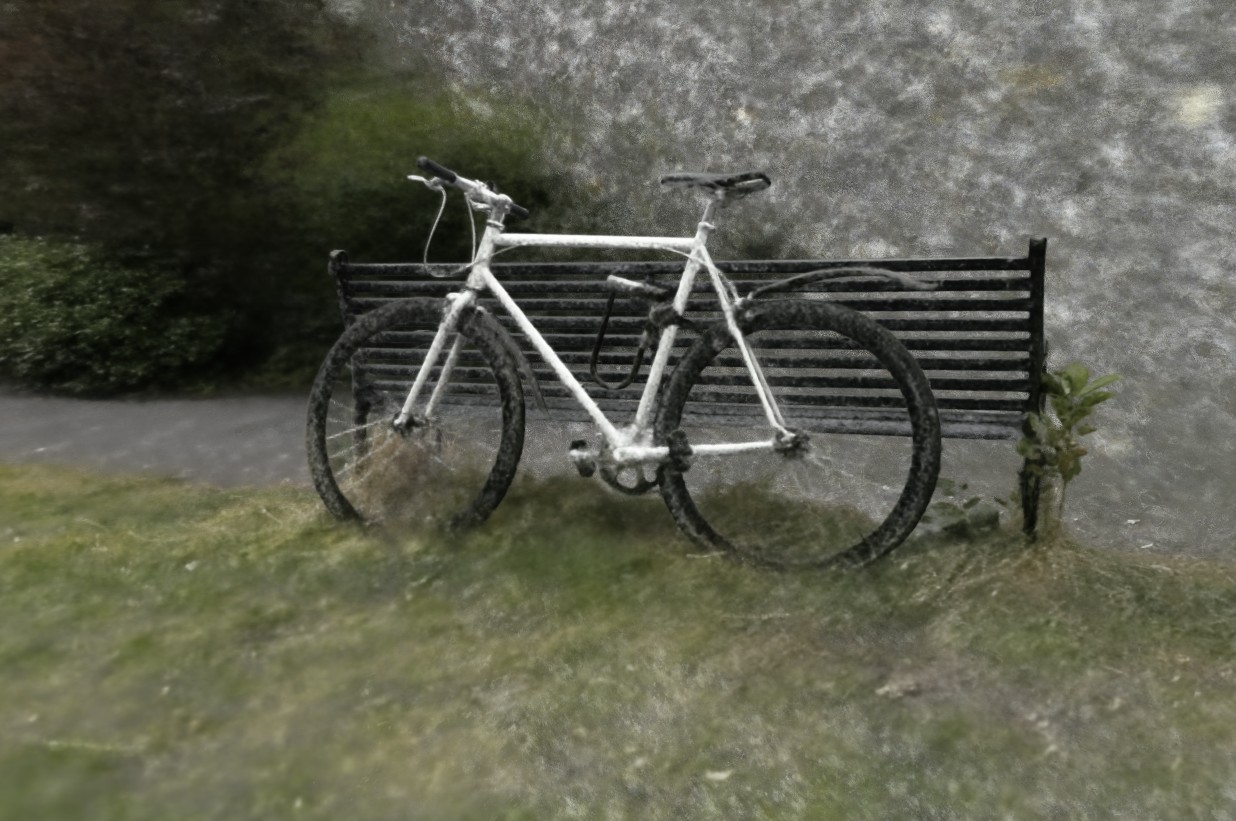}};
    \node [image,right=of bicycle-5] (bicycle-6) {\includegraphics[width=\figurewidth]{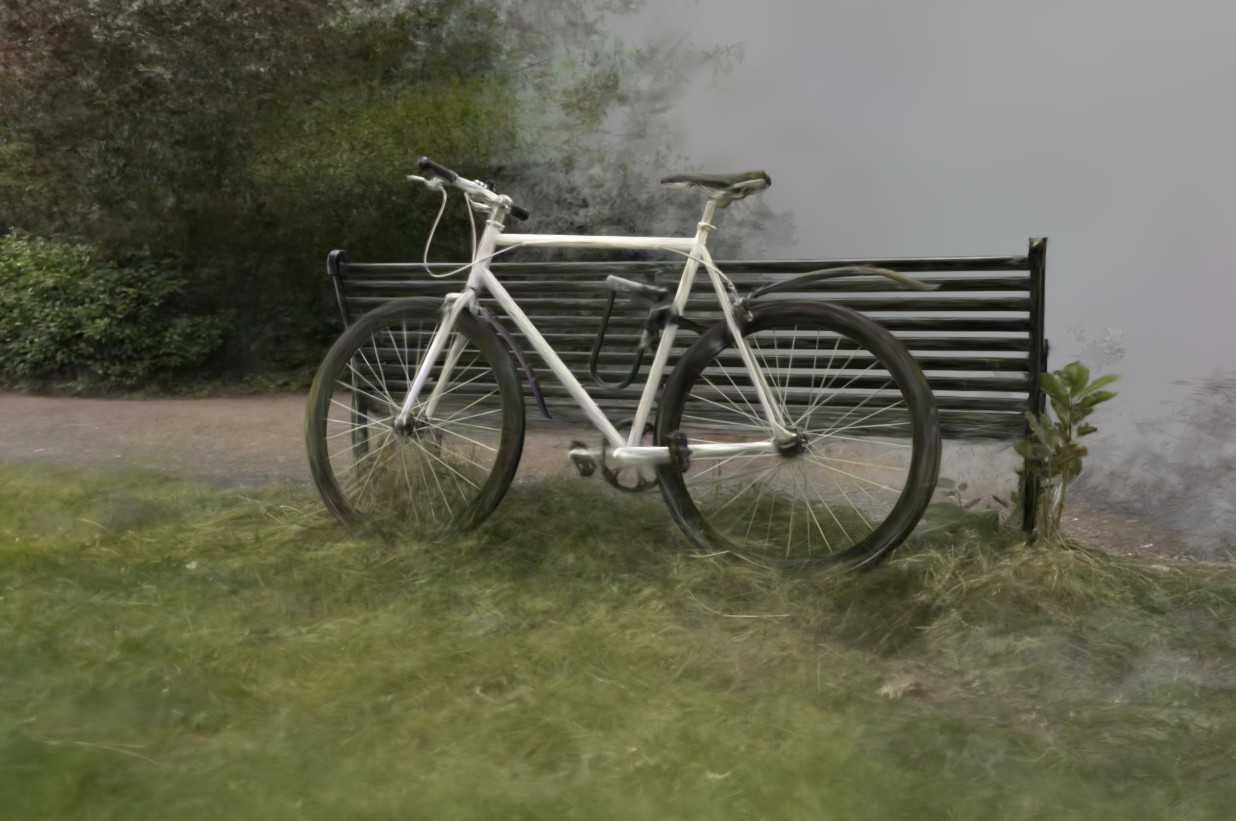}};
    \node [image,right=of bicycle-6] (bicycle-gt) {\includegraphics[width=\figurewidth]{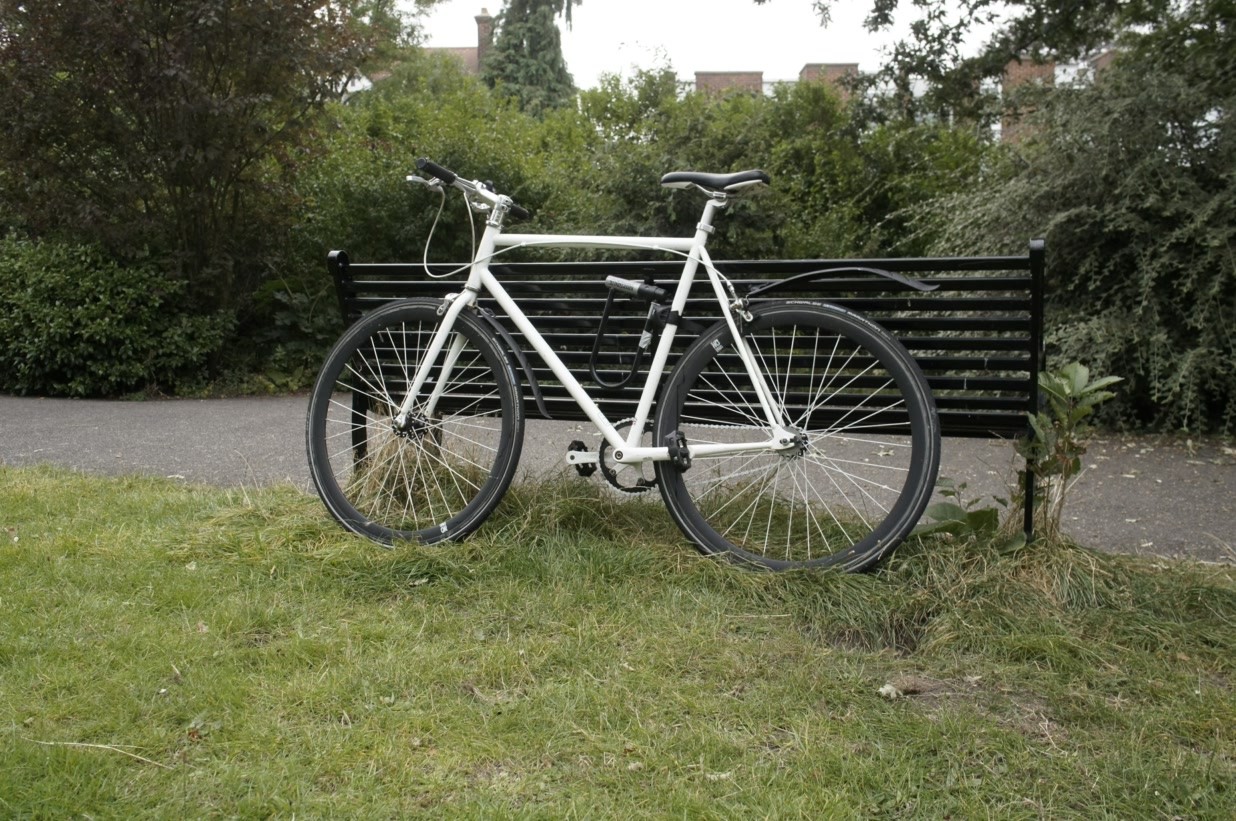}};

    \node[image,below=of bicycle-1] (bicycle-unc-1) {\includegraphics[width=\figurewidth]{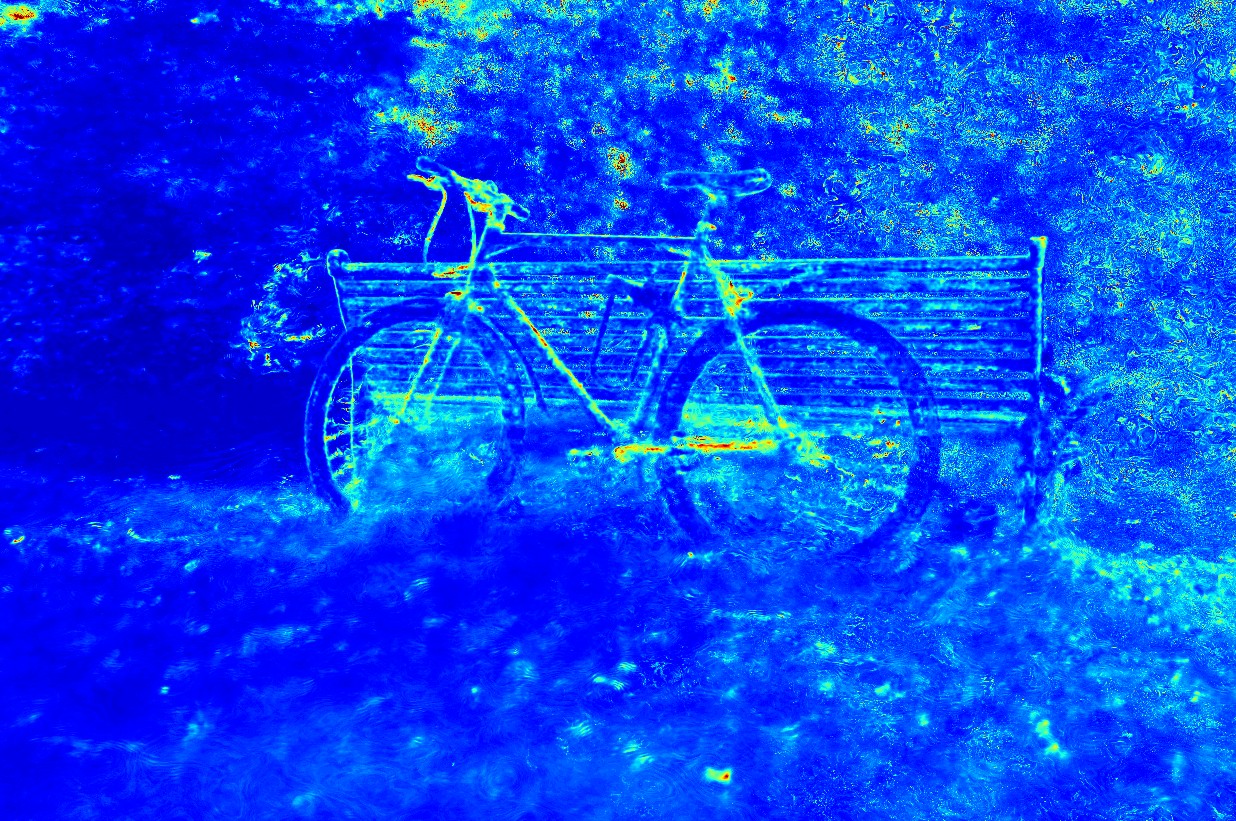}};
    \node[image,right=of bicycle-unc-1] (bicycle-unc-2) {\includegraphics[width=\figurewidth]{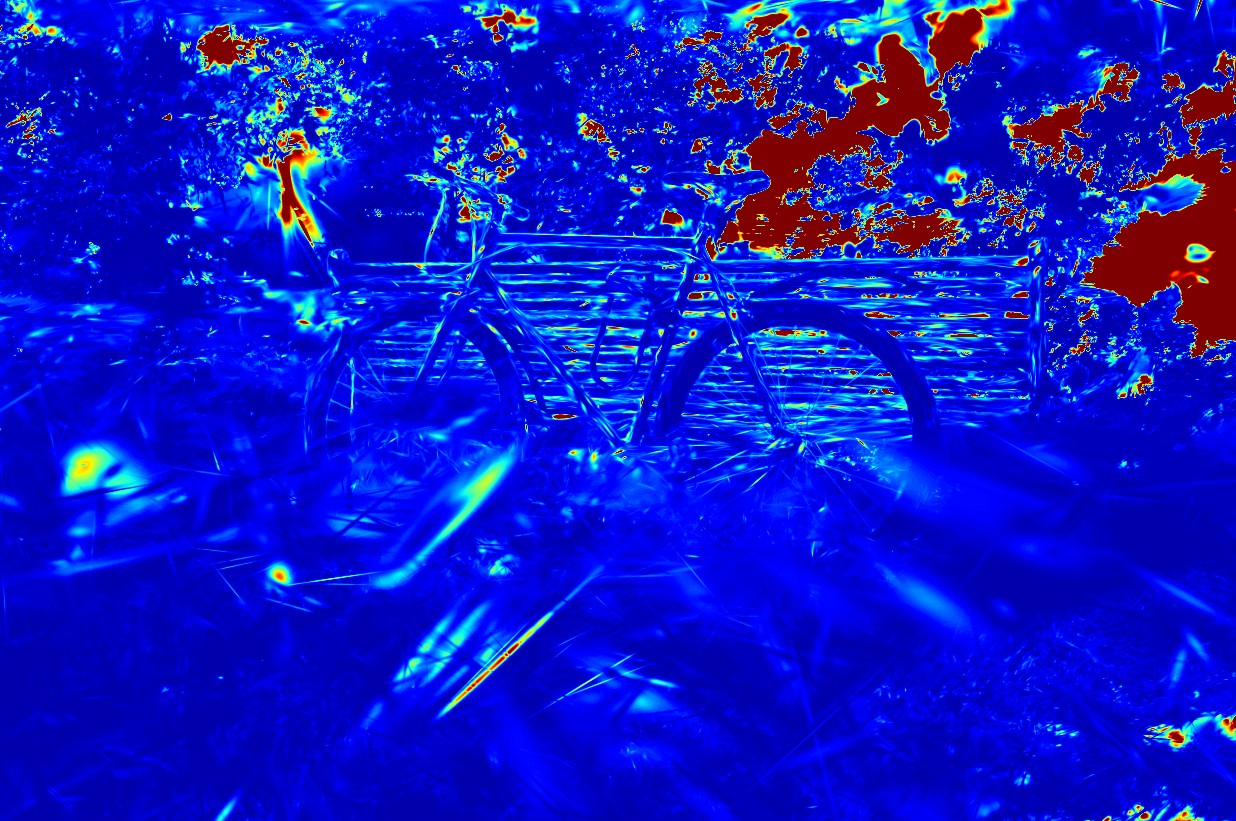}};
    \node[image,right=of bicycle-unc-2] (bicycle-unc-3) {\includegraphics[width=\figurewidth]{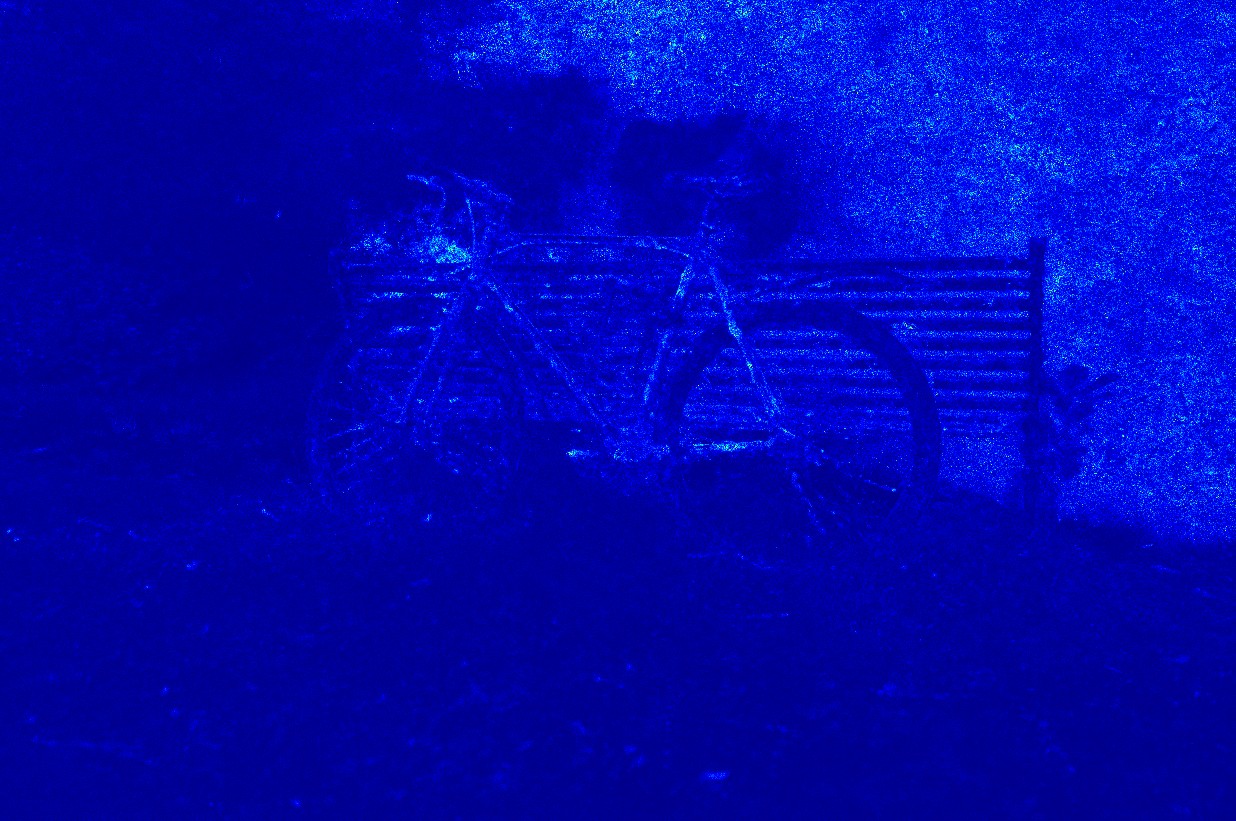}};
    \node[image,right=of bicycle-unc-3] (bicycle-unc-4) {\includegraphics[width=\figurewidth]{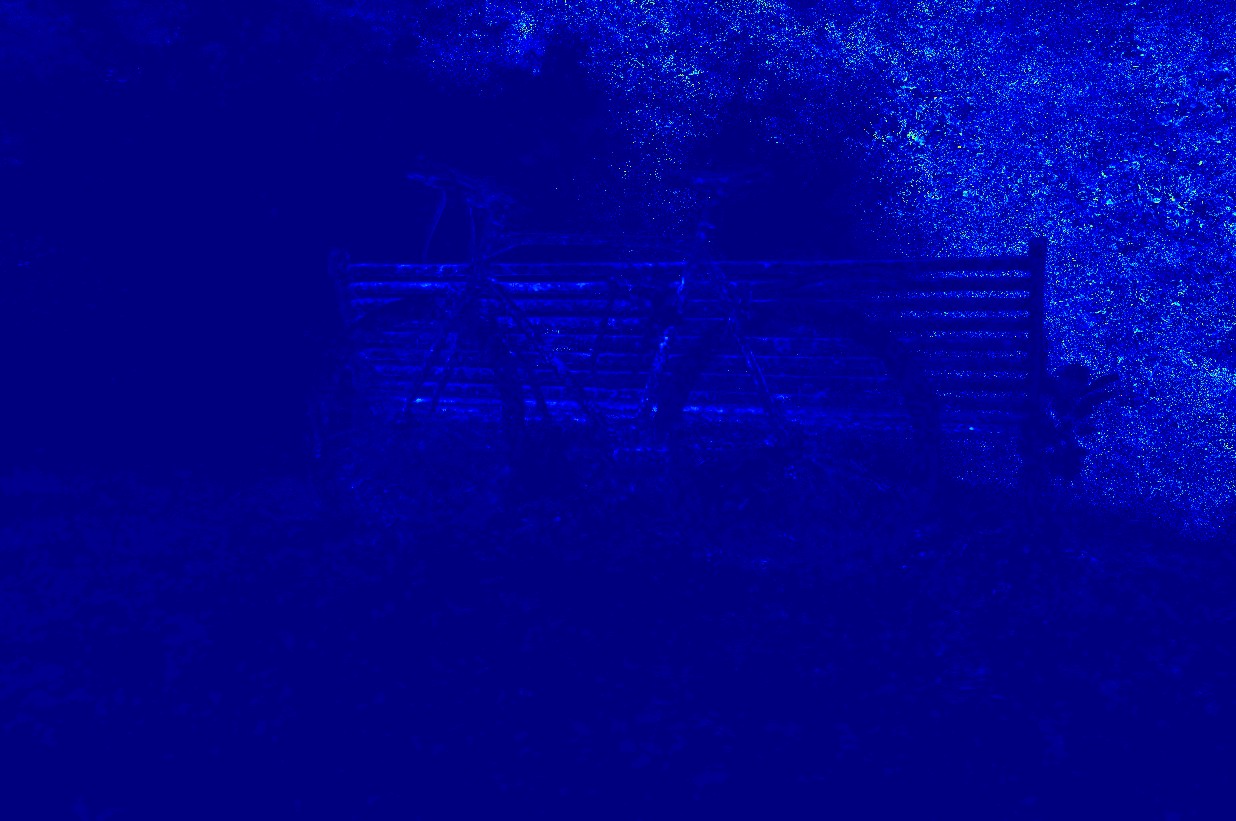}};
    \node[image,right=of bicycle-unc-4] (bicycle-unc-5) {\includegraphics[width=\figurewidth]{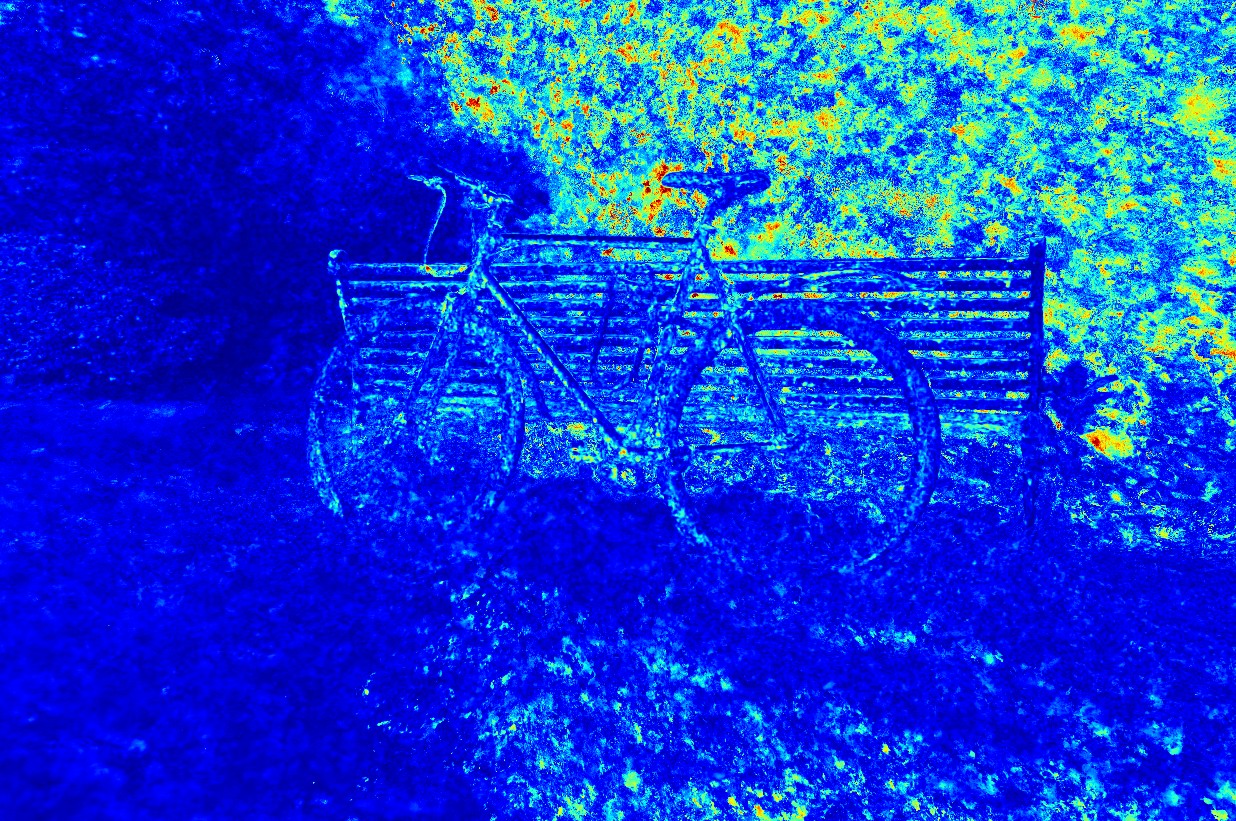}};
    \node[image,right=of bicycle-unc-5] (bicycle-unc-6) {\includegraphics[width=\figurewidth]{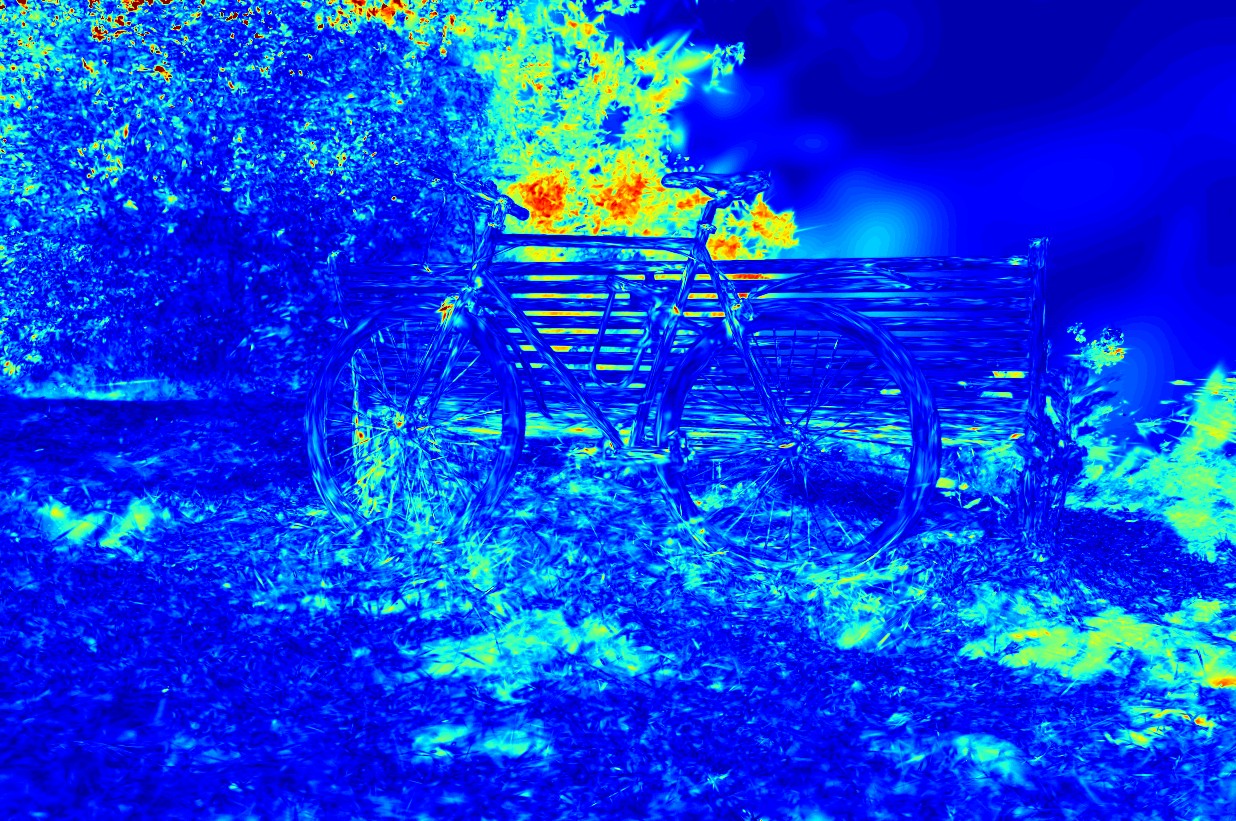}};
    \node [image,below=of bicycle-unc-1] (garden-1) {\includegraphics[width=\figurewidth]{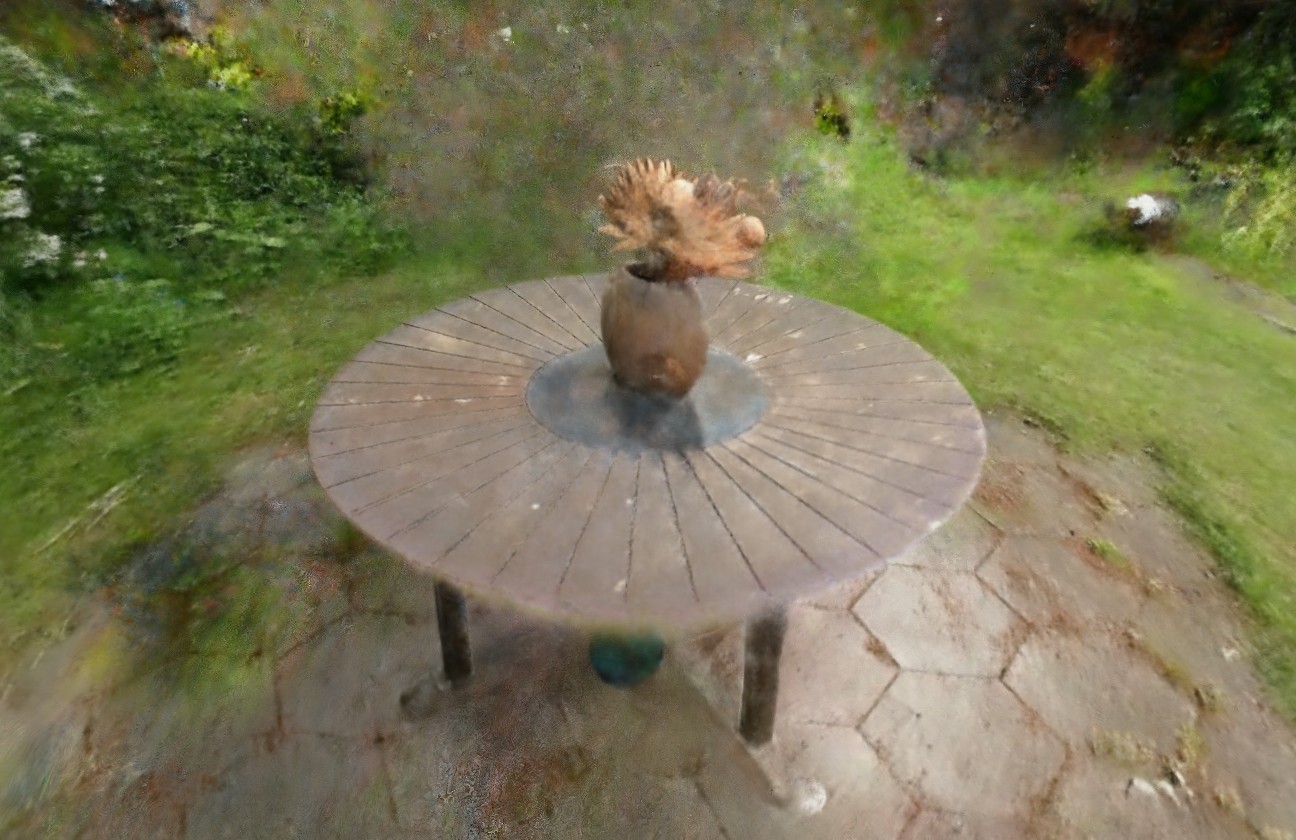}};
    \node [image,right=of garden-1] (garden-2) {\includegraphics[width=\figurewidth]{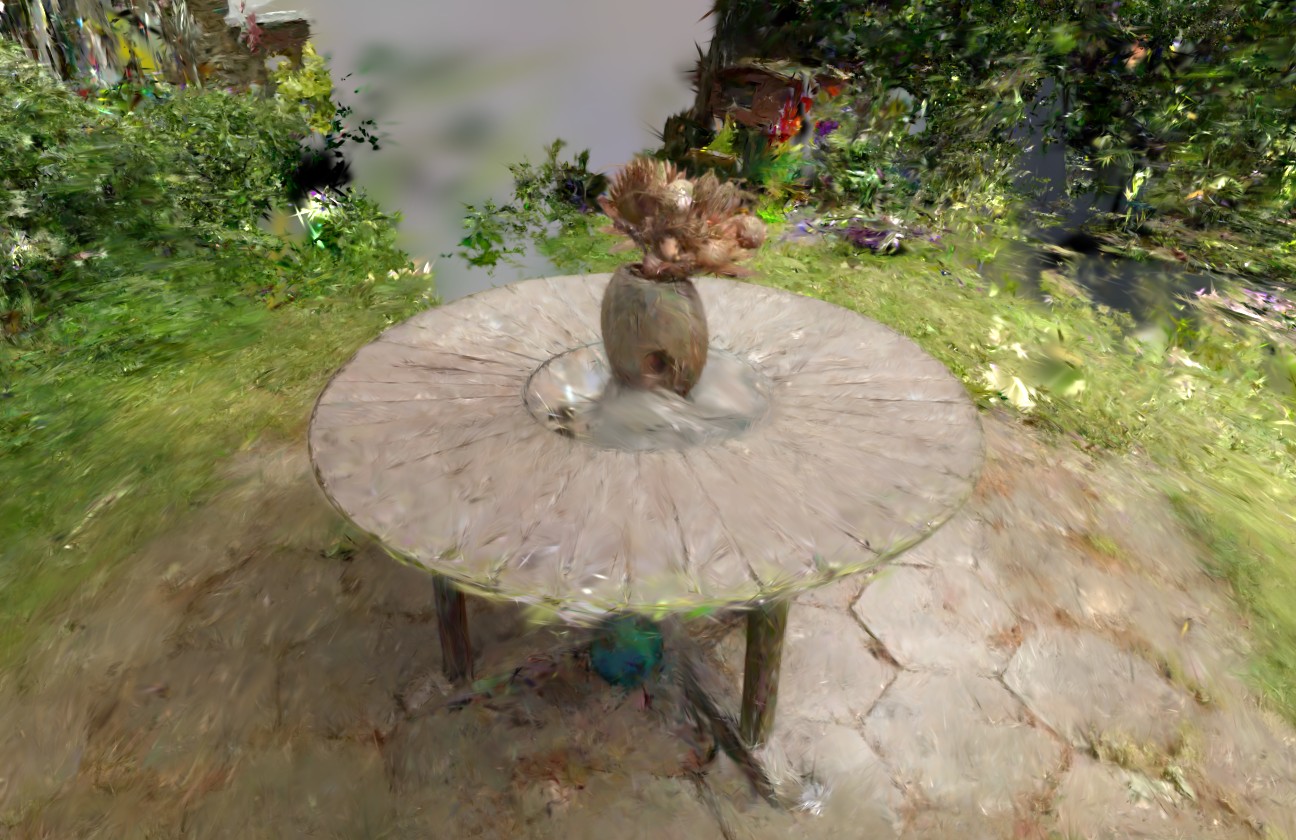}};
    \node [image,right=of garden-2] (garden-3) {\includegraphics[width=\figurewidth]{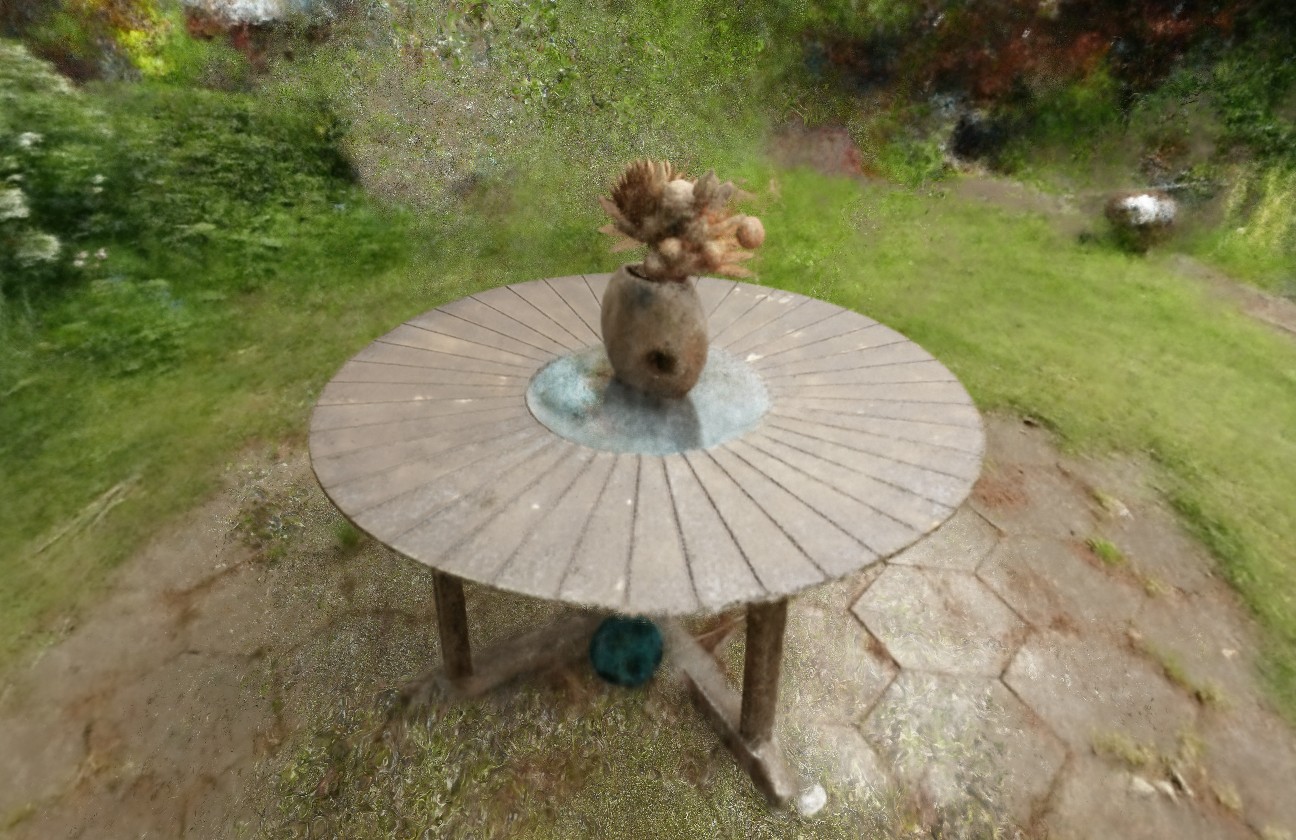}};
    \node [image,right=of garden-3] (garden-4) {\includegraphics[width=\figurewidth]{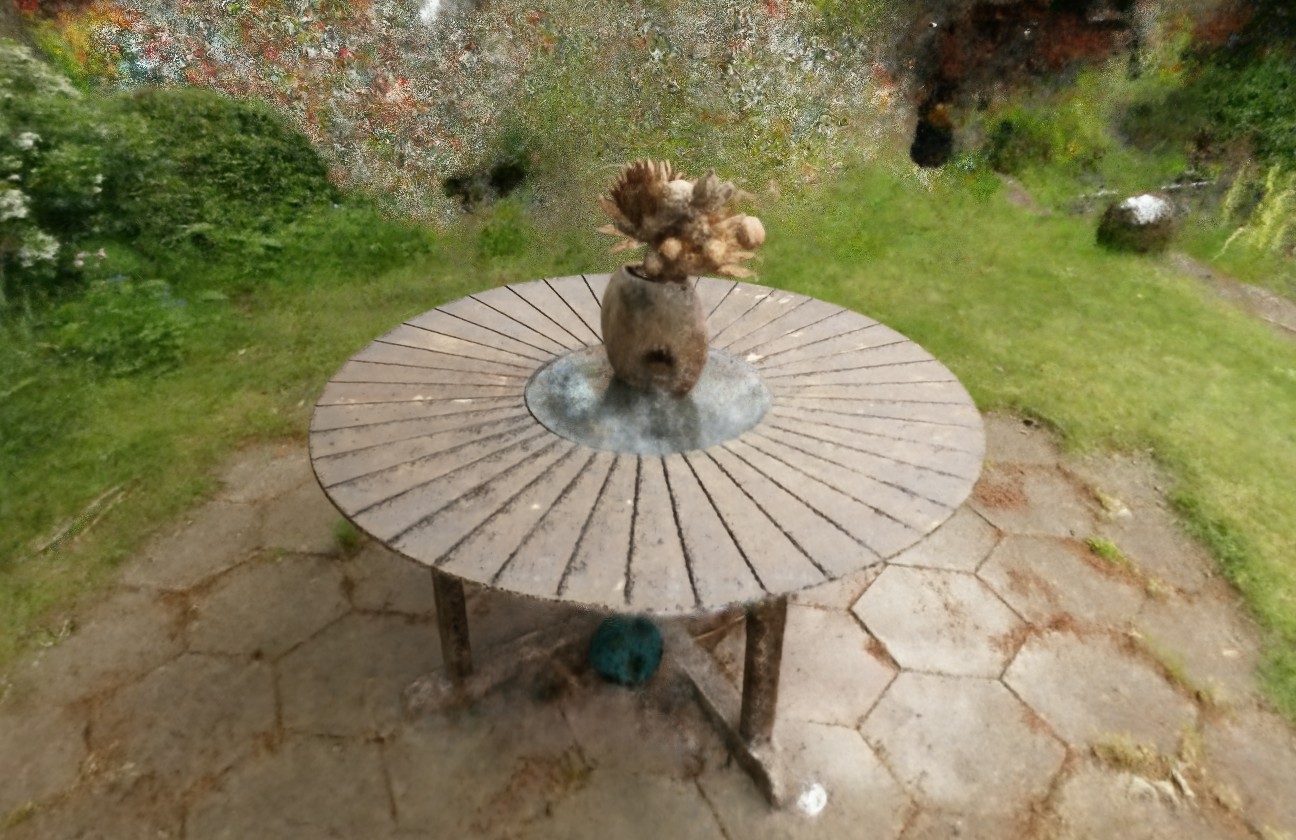}};
    \node [image,right=of garden-4] (garden-5) {\includegraphics[width=\figurewidth]{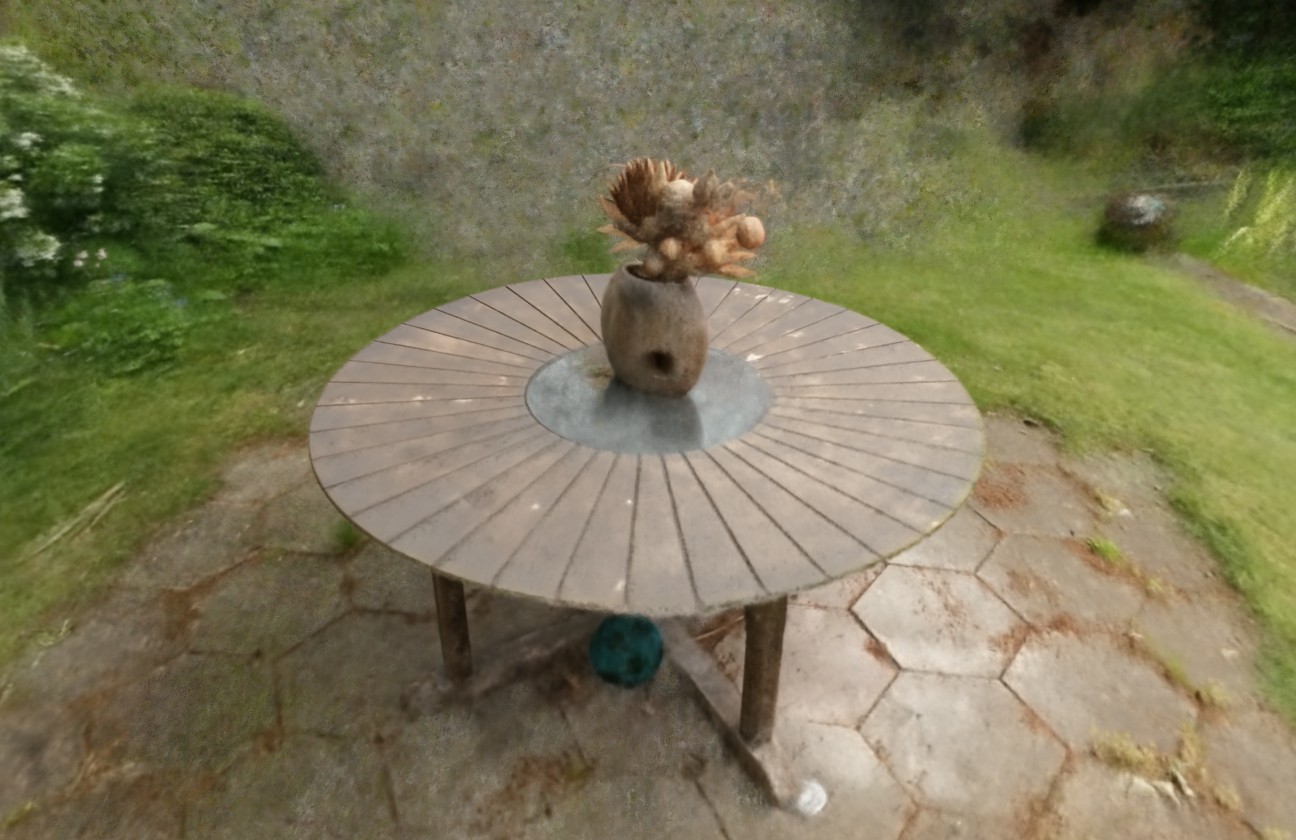}};
    \node [image,right=of garden-5] (garden-6) {\includegraphics[width=\figurewidth]{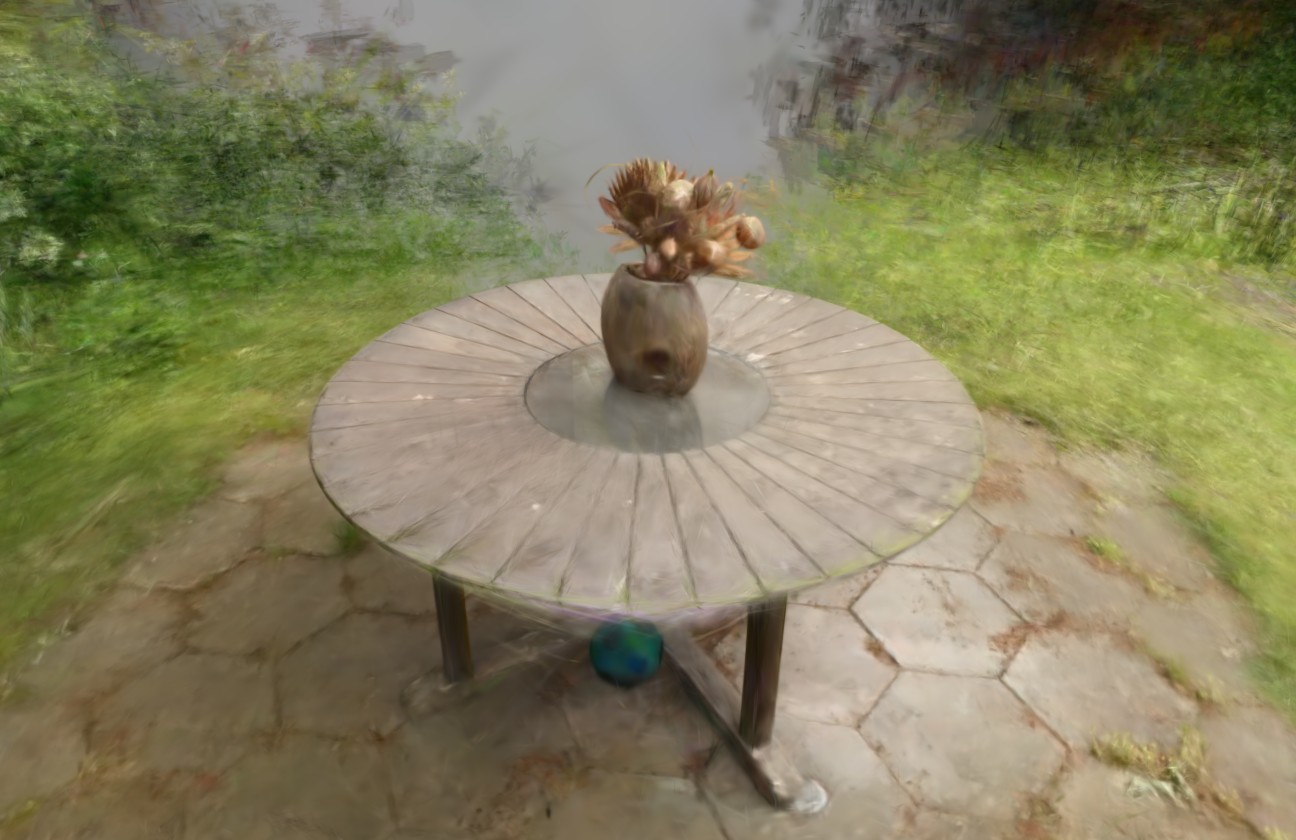}};
    \node [image,right=of garden-6] (garden-gt) {\includegraphics[width=\figurewidth]{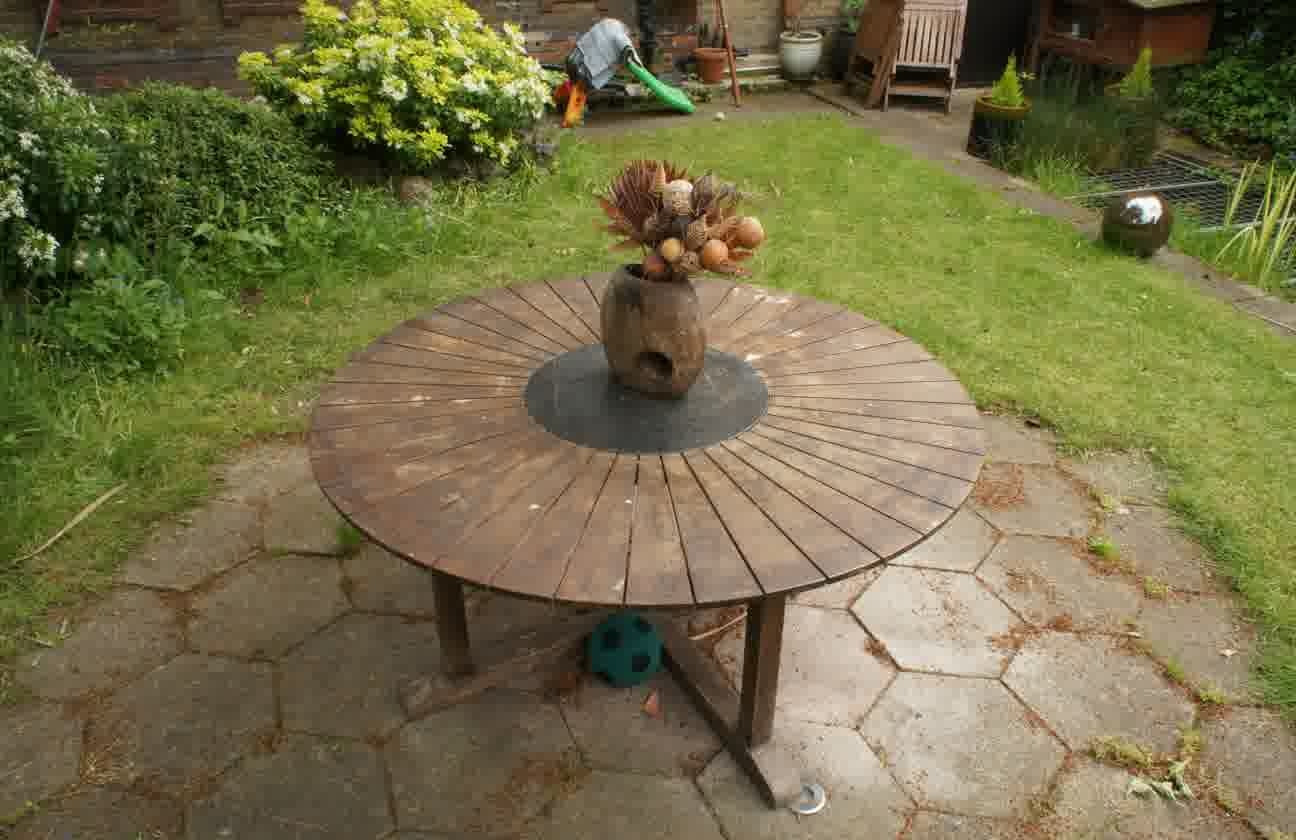}};

    \node[image,below=of garden-1] (garden-unc-1) {\includegraphics[width=\figurewidth]{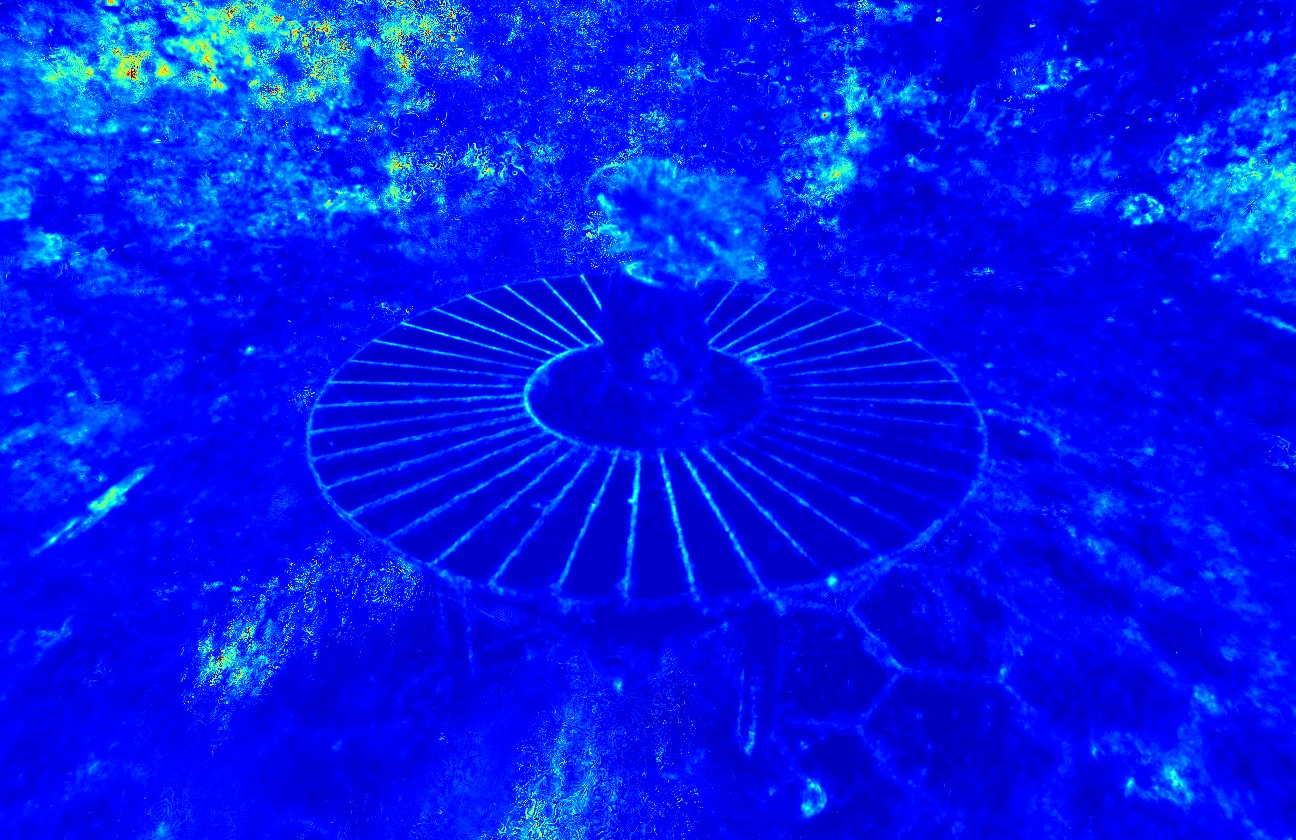}};
    \node[image,right=of garden-unc-1] (garden-unc-2) {\includegraphics[width=\figurewidth]{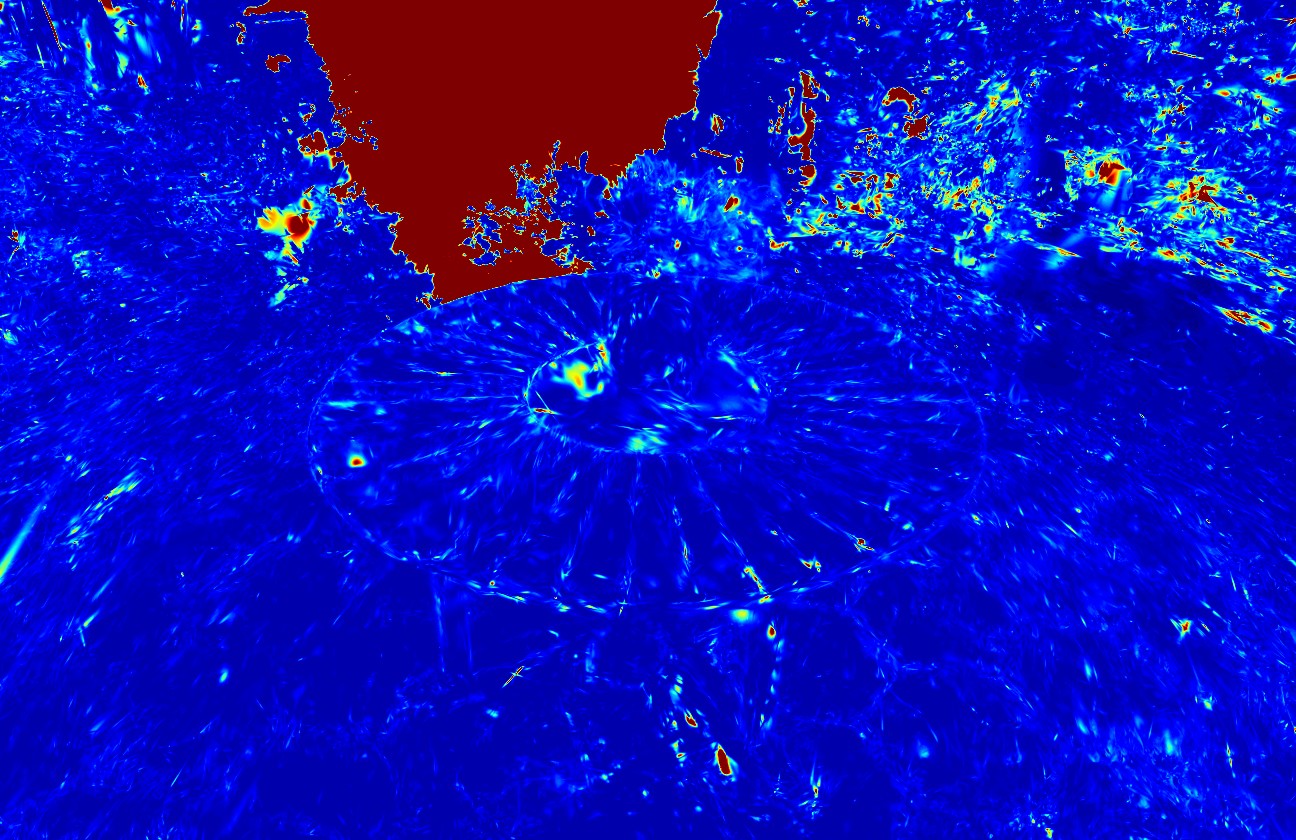}};
    \node[image,right=of garden-unc-2] (garden-unc-3) {\includegraphics[width=\figurewidth]{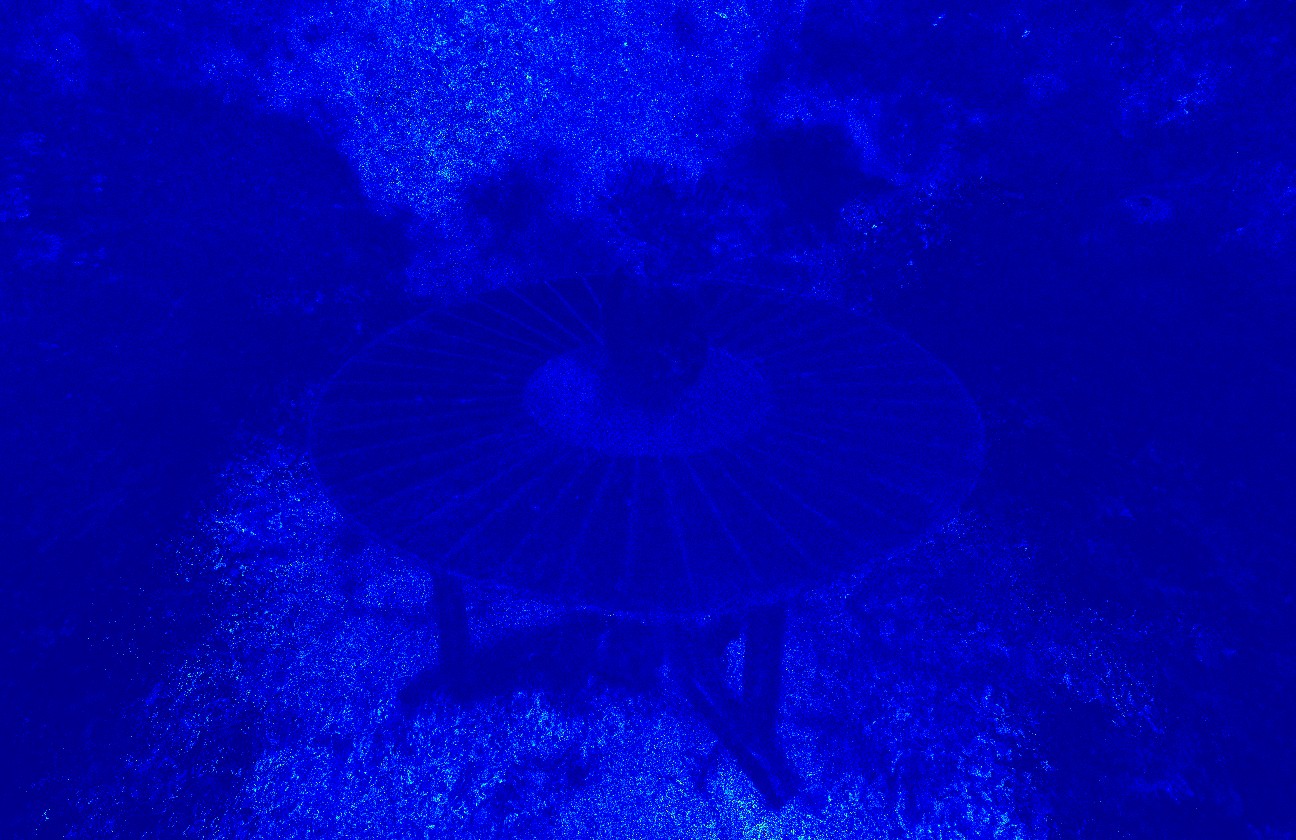}};
    \node[image,right=of garden-unc-3] (garden-unc-4) {\includegraphics[width=\figurewidth]{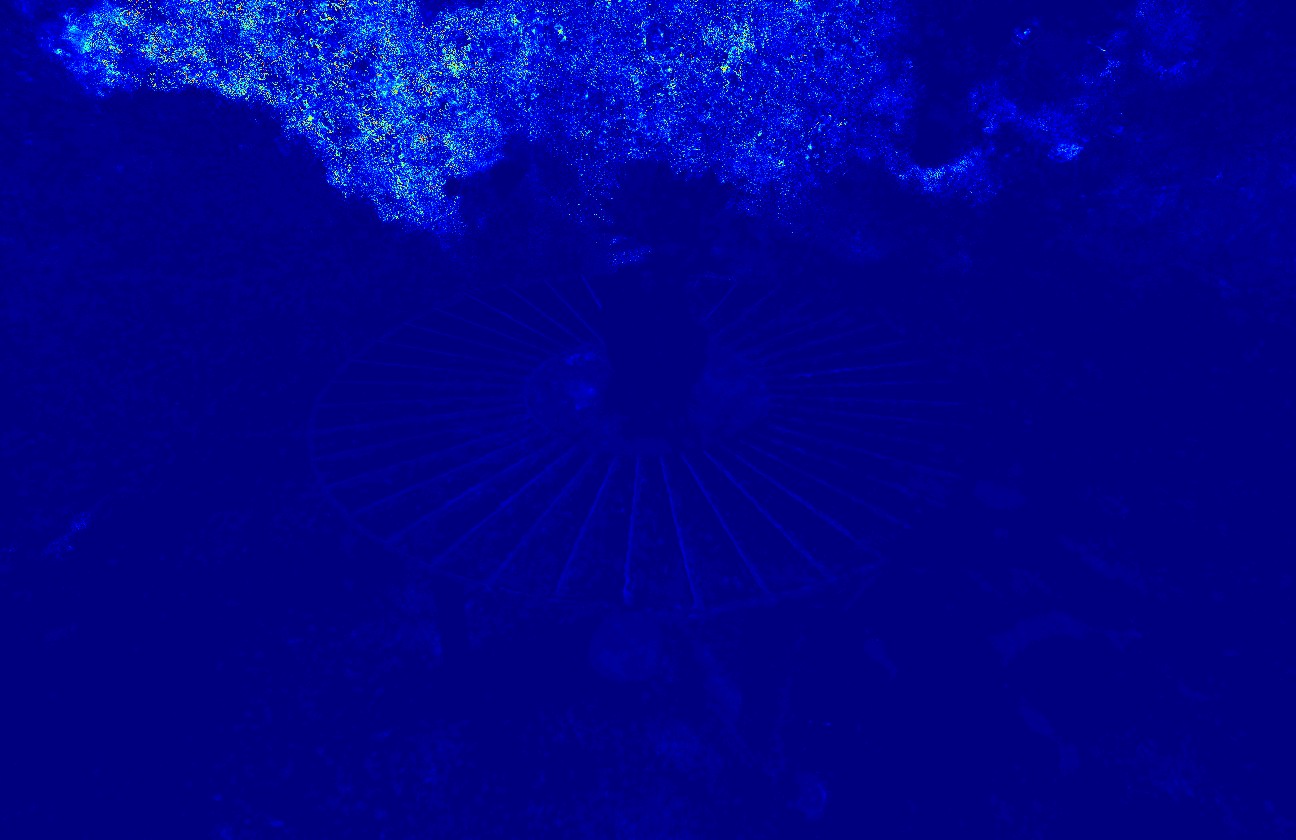}};
    \node[image,right=of garden-unc-4] (garden-unc-5) {\includegraphics[width=\figurewidth]{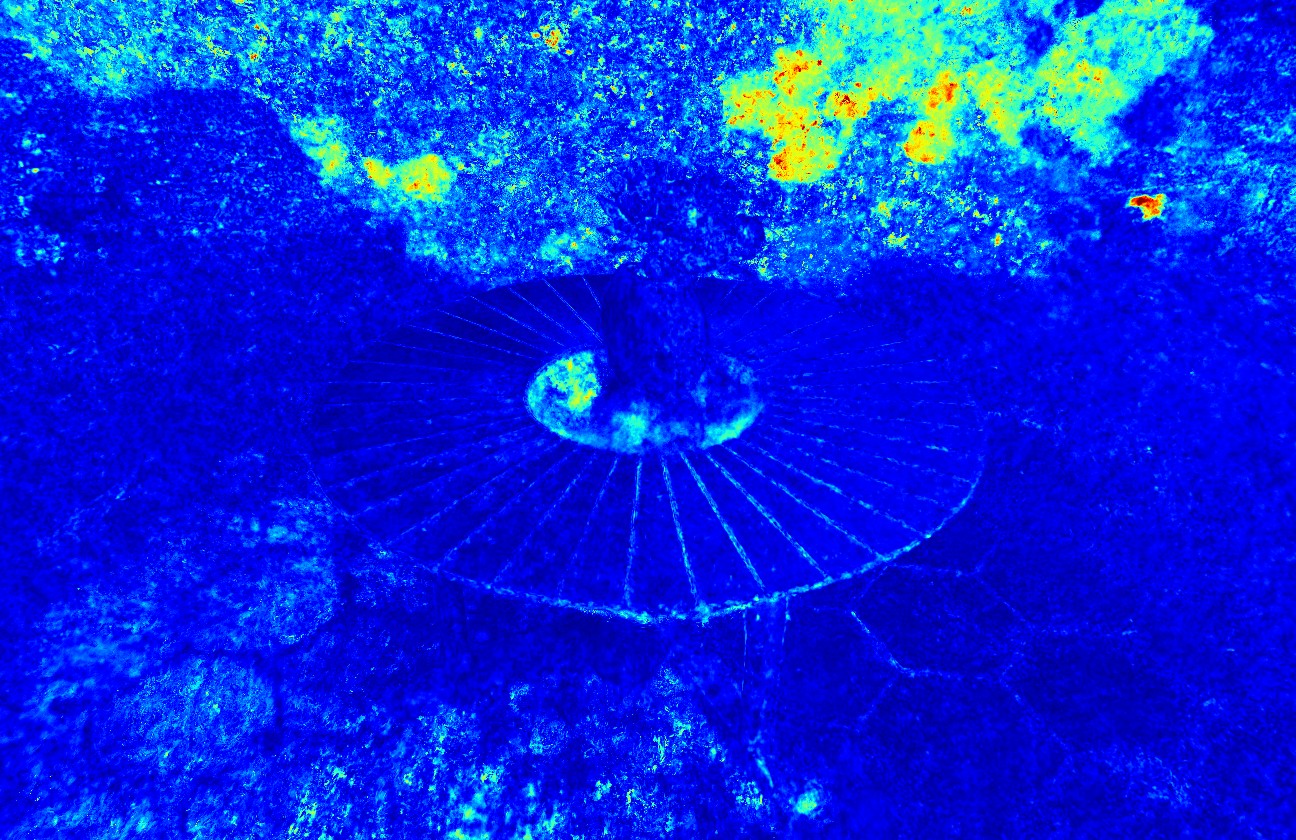}};
    \node[image,right=of garden-unc-5] (garden-unc-6) {\includegraphics[width=\figurewidth]{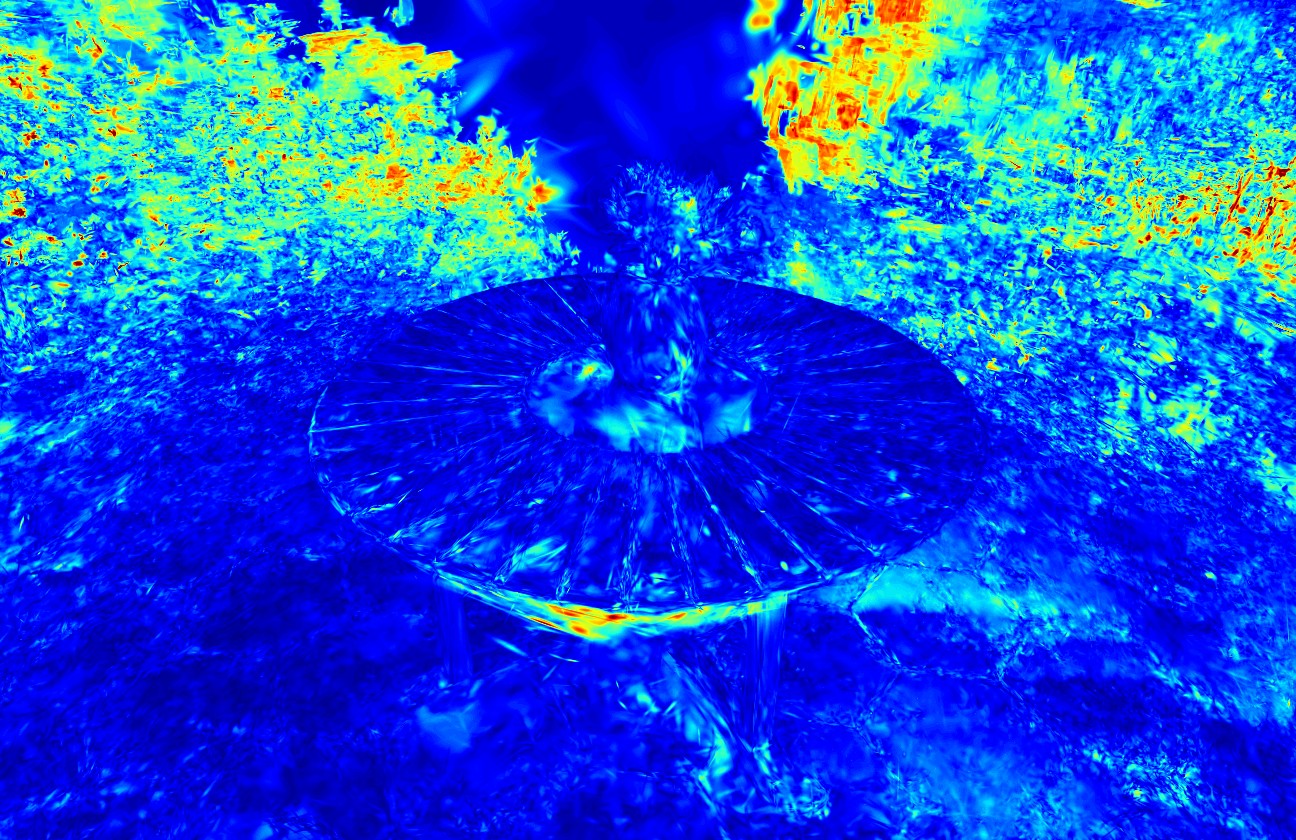}};
    \node[label] (label1) at (bicycle-1.north) {Active-Nerfacto\vphantom{p}};
    \node[label] (label2) at (bicycle-2.north) {Active-Splatfacto\vphantom{p}};
    \node[label] (label3) at (bicycle-3.north) {MC-Dropout-Nerfacto\vphantom{p}};
    \node[label] (label4) at (bicycle-4.north) {Laplace-Nerfacto\vphantom{p}};
    \node[label] (label5) at (bicycle-5.north) {Ensemble-Nerfacto\vphantom{p}};
    \node[label] (label6) at (bicycle-6.north) {Ensemble-Splatfacto\vphantom{p}};
    \node[label] (label-gt) at (bicycle-gt.north) {Ground Truth\vphantom{p}};

    \node[anchor=south,inner sep=1pt,rotate=90] (scene1) at (bicycle-1.south west) {Bicycle\vphantom{p}};
    \node[anchor=south,inner sep=1pt,rotate=90] (scene2) at (garden-1.south west) {Garden\vphantom{p}};
    
    \end{tikzpicture}
  \caption{OOD setting \num{2}: Rendered RGB and uncertainty for test views from Mip-NeRF 360 scenes next to the ground truth RGB. The uncertainty is visualized by the standard deviation (0.0~\protect\includegraphics[width=3em,height=.7em]{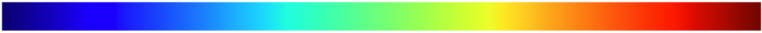}~0.3). See \cref{fig:ood-mipnerf360-qualitative-appendix} for more scenes.}
  \label{fig:ood-mipnerf360-qualitative}
\end{figure}
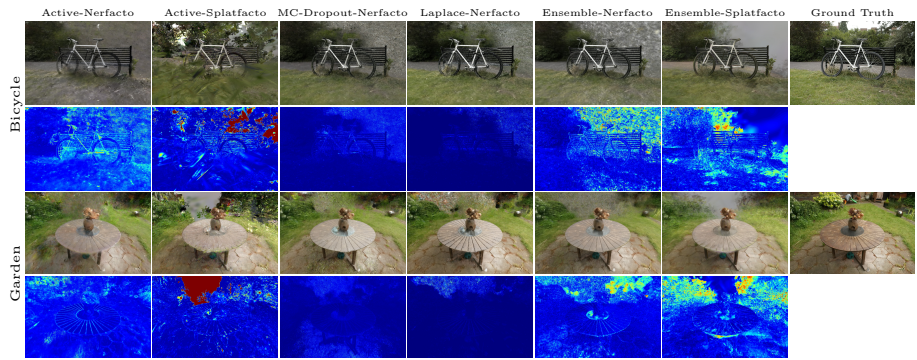

In \cref{fig:ood-mipnerf360-qualitative}, we qualitatively assess the rendered RGB and uncertainty %
on test views from two scenes (more examples in \cref{app:epistemic}). %
Active-Nerfacto focuses its uncertainties on known parts that can be reconstructed sufficiently while estimating sparse uncertainty values on unseen regions. Active-Splatfacto achieves poor reconstruction quality but yields high uncertainty on the unseen unknown regions. MC-Dropout- and Laplace-Nerfacto place most of their uncertainty on the unseen background regions. Ensemble-Nerfacto estimates uncertainty on both unseen regions and poorly reconstructed parts, while Ensemble-Splatfacto focuses on poorly reconstructed parts rather than the unseen regions.

\paragraph{Few-view Setting} We take the few-view setting with the LF data set from \cite{shen2021stochastic, shen2022conditional}, where we evaluate the uncertainty estimation on both RGB and depth (see details in \cref{app:methods}). To summarize, the Nerfacto-based methods yield plasuible reconstruction and uncertainty maps. 
Although the test views are nearby the training views, Active-Splatfacto and Ensemble-Splatfacto need larger number of views of the scenes to achieve a lower reconstruction error. See \cref{fig:fewview-lf-qualitative-appendix} and \cref{tab:ood-lfdataset-rgb-depth-appendix} in \cref{app:epistemic} for qualitative and quantitative results respectively.

\subsection{Experiments on Sensitivity to Cluttered Inputs (\num{3})}
\label{sec:outliers}
In this experiment, we assess the sensitivity of each method to learning from cluttered training views. 
We run experiments with scenes from the RobustNeRF \cite{sabour2023robustnerf} and the On-the-go \cite{ren2024nerf} data set. 
Both data sets consist of real-world scenes where various distractor items, \eg, toys or people, appear in different camera views, which results in floater artefacts in the rendered RGB. This setting allows us to study implicitly whether the methods are capable of recognizing static and non-static parts in the scene, where the uncertainties should capture which pixels are disturbed by confounding objects from the training views.

\begin{table}[t]
  \centering
  \caption{Cluttered RobustNeRF data set scenes \num{3}: Performance metrics averaged across the scenes in the data set. The \first{first}, \second{second}, and \third{third} values are highlighted.
  }
  \setlength{\tabcolsep}{4pt}
  \resizebox{0.7\textwidth}{!}{
  \begin{tabular}{lcccccc}
  \toprule
  Method & PSNR $\uparrow$ & SSIM $\uparrow$ & LPIPS $\downarrow$ & NLL $\downarrow$ & AUSE $\downarrow$ & AUCE $\downarrow$ \\
  \midrule
  Active-Nerfacto & 23.05 & 0.77 & \cellcolor{yellow!25}0.21 & \cellcolor{yellow!25}0.06 & \cellcolor{yellow!25}0.32 & \cellcolor{red!25}0.17 \\ 
  Active-Splatfacto & 21.29 & 0.73 & 0.28 & 3.48 & \cellcolor{orange!25}0.31 & \cellcolor{orange!25}0.22 \\
  MC-Dropout-Nerfacto & 22.67 & 0.75 & \cellcolor{orange!25}0.20 & 2.50 & 0.37 & 0.35 \\ 
  Laplace-Nerfacto & \cellcolor{yellow!25}23.14 & \cellcolor{yellow!25}0.76 & \cellcolor{orange!25}0.20 & 1.75 & 0.36 & 0.46 \\ 
  Ensemble-Nerfacto & \cellcolor{orange!25}23.51 & \cellcolor{orange!25}0.78 & \cellcolor{orange!25}0.20 & \cellcolor{orange!25}-0.35 & \cellcolor{red!25}0.28 & \cellcolor{yellow!25}0.23 \\ 
  Ensemble-Splatfacto & \cellcolor{red!25}26.66 & \cellcolor{red!25}0.86 & \cellcolor{red!25}0.14 & \cellcolor{red!25}-1.03 & \cellcolor{yellow!25}0.32 & \cellcolor{yellow!25}0.23 \\ 
  \bottomrule
  \end{tabular}
  }
  \label{tab:outliers-robustnerf}
\end{table}

  \begin{figure*}[t!]
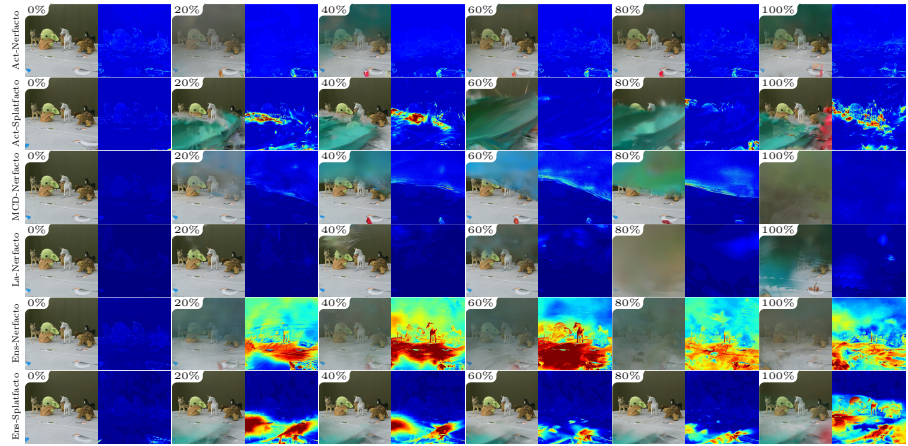

    \centering\scriptsize
    \begin{tikzpicture}[inner sep=0]
      \setlength{\figurewidth}{.16\textwidth}
      \setlength{\figureheight}{.5\figurewidth}
        
      \foreach \method [count=\j] in {ensemble-splatfacto,ensemble-nerfacto,nerfacto-laplace,nerfacto-mcdropout,active-splatfacto,active-nerfacto} {  

      \foreach \file / \label [count=\i] in {0.0/0\%,0.2/20\%,0.4/40\%,0.6/60\%,0.8/80\%,1.0/100\%} {
        \begin{scope}
          \node[anchor=north west, inner sep=1pt, font=\tiny\sc,scale=.75] (label-\i) at (\figurewidth*\i,\figureheight*\j) {\label};

          \clip[rounded corners=2pt] (label-\i.south west) -- (label-\i.south east) -- (label-\i.north east) -- (\figurewidth*\i+1.5\figurewidth,\figureheight*\j) -- (\figurewidth*\i+1.5\figurewidth,\figureheight*\j-1.5\figureheight) -- (\figurewidth*\i,\figureheight*\j-1.5\figureheight) -- cycle;

          \node [inner sep=0,minimum width=\figurewidth,minimum height=\figureheight,fill=white,anchor=north west,text width=\figurewidth] (node-\j-\i) at (\i*\figurewidth,\figureheight*\j) {%
            \includegraphics[width=.495\figurewidth]{figures-main/outliers-robustnerf-visuals/yoda/clutter_level_\file/\method/plots/119_rgb_pred}%
            \includegraphics[width=.495\figurewidth]{figures-main/outliers-robustnerf-visuals/yoda/clutter_level_\file/\method/plots/119_rgb_unc}
          };    
          \draw[white,line width=1pt] (\figurewidth*\i,\figureheight*\j) rectangle ++(\figurewidth,\figureheight);
          
        \end{scope}
      }
      }

      \node[anchor=south,inner sep=3pt,rotate=90,scale=0.5] at (node-1-1.west) {Ens-Splatfacto\vphantom{p}};
      \node[anchor=south,inner sep=3pt,rotate=90,scale=0.5] at (node-2-1.west) {Ens-Nerfacto\vphantom{p} };
      \node[anchor=south,inner sep=3pt,rotate=90,scale=0.5] at (node-3-1.west) {La-Nerfacto\vphantom{p}  };
      \node[anchor=south,inner sep=3pt,rotate=90,scale=0.5] at (node-4-1.west) {MCD-Nerfacto\vphantom{p}  };
      \node[anchor=south,inner sep=3pt,rotate=90,scale=0.5] at (node-5-1.west) {Act-Splatfacto\vphantom{p} };
      \node[anchor=south,inner sep=3pt,rotate=90,scale=0.5] at (node-6-1.west) {Act-Nerfacto\vphantom{p} };
      
    \end{tikzpicture}%
    \caption{Yoda scene \num{3}: Rendered RGB and uncertainty for test view for different proportions of cluttered training views (0--100\%, from only clean images to fully cluttered). The uncertainty is visualized by the standard deviation (0.0~\protect\includegraphics[width=3em,height=.7em]{figures-main/jet.png}~0.2). }
    \label{fig:outliers-yoda-clutter-visual}
  \end{figure*}

\paragraph{RobustNeRF Results} \cref{tab:outliers-robustnerf} shows the performance metrics averaged across scenes. We observe that Ensemble-Splatfacto achieves significantly higher image quality than the other methods while achieving uncertainty metrics on par with the Ensemble-Nerfacto and Active-Nerfacto. 
In \cref{fig:outliers-yoda-clutter-visual}, we visualize the rendered RGB and uncertainty for one test view in the Yoda scene to inspect how the uncertainty maps progress for different clutter proportions. 
This view is challenging as all methods render blurry RGB images occasionally with floaters already with 20\% cluttered views. 
Active-Nerfacto places uncertainty across the foggy pixels and detects the red floater in the lower right corner. 
Active-Splatfacto renders a wide floater in cyan but detects it poorly.  
MC-Dropout-Nerfacto renders a blurry floater that is accurately detected, especially with high uncertainty on its edge. 
Laplace-Nerfacto also manages to detect the floaters accurately, while having lower maximum uncertainty than the other methods. 
Ensemble-Nerfacto produces foggy images potentially due to averaging of rendered RGB and places most of its uncertainty on the table. 
Ensemble-Splatfacto manages to render the toys with high quality while detecting the floaters on the table.

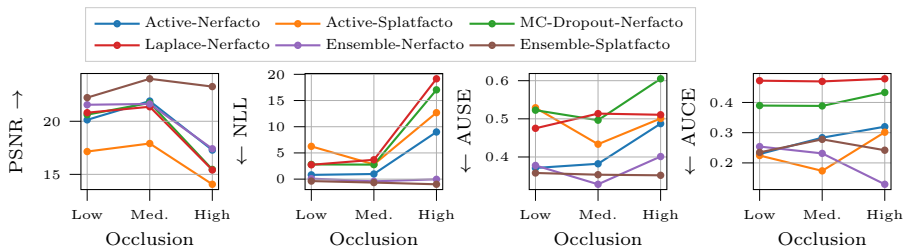
\begin{figure}[t]
  \centering
  \setlength{\figurewidth}{0.15\textwidth}
  \setlength{\figureheight}{.08\textheight}
  \pgfplotsset{every axis title/.append style={at={(0.5,0.80)}}} 
\pgfplotsset{every axis x label/.append style={at={(0.5,0.05)}}}
\pgfplotsset{every axis y label/.append style={at={(0.20,0.5)}}}

\begin{tikzpicture}
\tikzstyle{every node}=[font=\scriptsize]
\definecolor{crimson2143940}{RGB}{214,39,40}
\definecolor{darkgray176}{RGB}{176,176,176}
\definecolor{darkorange25512714}{RGB}{255,127,14}
\definecolor{forestgreen4416044}{RGB}{44,160,44}
\definecolor{lightgray204}{RGB}{204,204,204}
\definecolor{mediumpurple148103189}{RGB}{148,103,189}
\definecolor{sienna1408675}{RGB}{140,86,75}
\definecolor{steelblue31119180}{RGB}{31,119,180}

\begin{groupplot}[
  group style={group size= 4 by 1, 
    horizontal sep=1.15cm, 
    vertical sep=.5cm},
  tick align=outside,
  tick pos=left,
  grid=both,
  xlabel={Occlusion},
  xmin=-0.1, xmax=2.1,
  xtick={0,1,2},
  xticklabels={Low,Med.,High},
  xtick style={color=black},
  x grid style={darkgray176,solid},
  y grid style={darkgray176,solid},
  ytick style={color=black},
  tick label style={font=\tiny}
]

\nextgroupplot[
height=\figureheight,
width=\figurewidth,
legend cell align={left},
legend columns=3,
legend style={
  nodes={scale=0.8},
  fill opacity=0.8, 
  draw opacity=1, 
  text opacity=1, 
  at={(0.03,1.1)}, 
  anchor=south west,
  draw=lightgray204},
ylabel={PSNR~$\rightarrow$},
ymin=13.5683740615845, ymax=24.4979330062866,
ytick style={color=black}
]
\addplot [thick, steelblue31119180, mark=*, mark size=1, mark options={solid}]
table {%
0 20.1330404281616
1 21.9035453796387
2 17.2878751754761
};
\addlegendentry{Active-Nerfacto}
\addplot [thick, darkorange25512714, mark=*, mark size=1, mark options={solid}]
table {%
0 17.1491088867188
1 17.9064741134644
2 14.0651721954346
};
\addlegendentry{Active-Splatfacto}
\addplot [thick, forestgreen4416044, mark=*, mark size=1, mark options={solid}]
table {%
0 20.623607635498
1 21.7319622039795
2 15.46622133255
};
\addlegendentry{MC-Dropout-Nerfacto}
\addplot [thick, crimson2143940, mark=*, mark size=1, mark options={solid}]
table {%
0 20.8078327178955
1 21.3666868209839
2 15.4145460128784
};
\addlegendentry{Laplace-Nerfacto}
\addplot [thick, mediumpurple148103189, mark=*, mark size=1, mark options={solid}]
table {%
0 21.5546407699585
1 21.6325874328613
2 17.4112930297852
};
\addlegendentry{Ensemble-Nerfacto}
\addplot [thick, sienna1408675, mark=*, mark size=1, mark options={solid}]
table {%
0 22.2263355255127
1 24.0011348724365
2 23.2667007446289
};
\addlegendentry{Ensemble-Splatfacto}

\nextgroupplot[
height=\figureheight,
width=\figurewidth,
ylabel={$\leftarrow$~NLL},
ymin=-1.99267206043005, ymax=20.1557464644313,
ytick style={color=black}
]
\addplot [thick, steelblue31119180, mark=*, mark size=1, mark options={solid}]
table {%
0 0.810395002365112
1 0.977057844400406
2 8.99483752250671
};
\addplot [thick, darkorange25512714, mark=*, mark size=1, mark options={solid}]
table {%
0 6.25316321849823
1 2.80561584234238
2 12.6826195716858
};
\addplot [thick, forestgreen4416044, mark=*, mark size=1, mark options={solid}]
table {%
0 2.80239868164062
1 2.76995050907135
2 17.0389976501465
};
\addplot [thick, crimson2143940, mark=*, mark size=1, mark options={solid}]
table {%
0 2.73574614524841
1 3.69063830375671
2 19.1490001678467
};
\addplot [thick, mediumpurple148103189, mark=*, mark size=1, mark options={solid}]
table {%
0 0.0939229000359774
1 -0.423177175223827
2 -0.0353035777807236
};
\addplot [thick, sienna1408675, mark=*, mark size=1, mark options={solid}]
table {%
0 -0.372241258621216
1 -0.660018116235733
2 -0.985925763845444
};

\nextgroupplot[
height=\figureheight,
width=\figurewidth,
ylabel={$\leftarrow$~AUSE},
ymin=0.314597538858652, ymax=0.618801140040159,
ytick style={color=black}
]
\addplot [thick, steelblue31119180, mark=*, mark size=1, mark options={solid}]
table {%
0 0.371211677789688
1 0.382167533040047
2 0.487401783466339
};
\addplot [thick, darkorange25512714, mark=*, mark size=1, mark options={solid}]
table {%
0 0.529008969664574
1 0.433491975069046
2 0.500921934843063
};
\addplot [thick, forestgreen4416044, mark=*, mark size=1, mark options={solid}]
table {%
0 0.522689074277878
1 0.496458038687706
2 0.604973703622818
};
\addplot [thick, crimson2143940, mark=*, mark size=1, mark options={solid}]
table {%
0 0.475179240107536
1 0.513704940676689
2 0.510635197162628
};
\addplot [thick, mediumpurple148103189, mark=*, mark size=1, mark options={solid}]
table {%
0 0.377806484699249
1 0.328424975275993
2 0.401163682341576
};
\addplot [thick, sienna1408675, mark=*, mark size=1, mark options={solid}]
table {%
0 0.358079716563225
1 0.353514716029167
2 0.351965233683586
};

\nextgroupplot[
height=\figureheight,
width=\figurewidth,
ylabel={$\leftarrow$~AUCE},
ymin=0.111784739004289, ymax=0.495794014638915,
ytick style={color=black}
]
\addplot [thick, steelblue31119180, mark=*, mark size=1, mark options={solid}]
table {%
0 0.229265464055116
1 0.283444595452834
2 0.319819997486306
};
\addplot [thick, darkorange25512714, mark=*, mark size=1, mark options={solid}]
table {%
0 0.224433300454487
1 0.173645399689291
2 0.30157093039562
};
\addplot [thick, forestgreen4416044, mark=*, mark size=1, mark options={solid}]
table {%
0 0.389730845515816
1 0.388515340473203
2 0.433132076787619
};
\addplot [thick, crimson2143940, mark=*, mark size=1, mark options={solid}]
table {%
0 0.472032289836471
1 0.469605414552257
2 0.478339047564613
};
\addplot [thick, mediumpurple148103189, mark=*, mark size=1, mark options={solid}]
table {%
0 0.25406270536267
1 0.231455467930683
2 0.129239706078591
};
\addplot [thick, sienna1408675, mark=*, mark size=1, mark options={solid}]
table {%
0 0.23483982477912
1 0.277944951839921
2 0.24198258395496
};

\end{groupplot}

\end{tikzpicture}%
  \caption{Confounding outliers \num{3}: Performance metrics over occlusion levels in the training views averaged across the corresponding scenes from the On-the-go data set. 
  }
  \label{fig:outliers-nerfonthego-lightversion}
\end{figure}

\begin{figure}[t]
  \centering
  \setlength{\figurewidth}{0.136\textwidth}
  \begin{tikzpicture}[image/.style = {inner sep=0, outer sep=0, minimum width=\figurewidth, anchor=north west, text width=\figurewidth}, node distance = 1pt and 1pt, every node/.style={font= {\tiny}}, label/.style = {scale=0.75,font={\tiny},anchor=south,inner sep=0pt,outer sep=2pt,rotate=0}]

    \node [image] (mountain-1) {\includegraphics[width=\figurewidth]{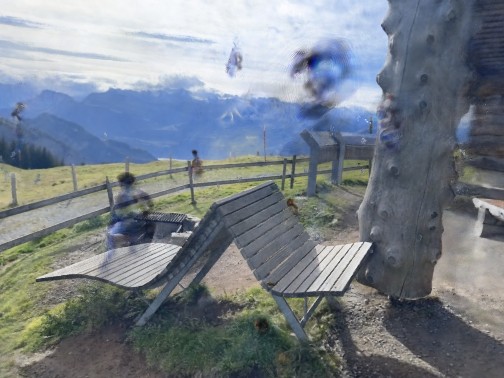}}; 
    \node [image,right=of mountain-1] (mountain-2) {\includegraphics[width=\figurewidth]{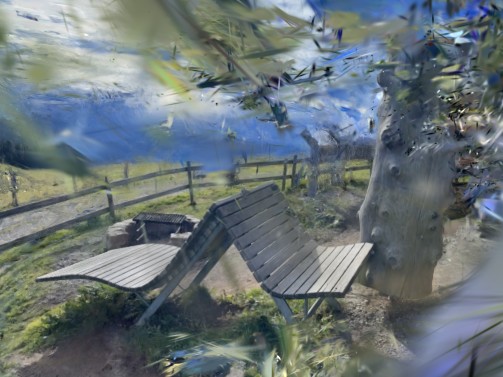}};
    \node [image,right=of mountain-2] (mountain-3) {\includegraphics[width=\figurewidth]{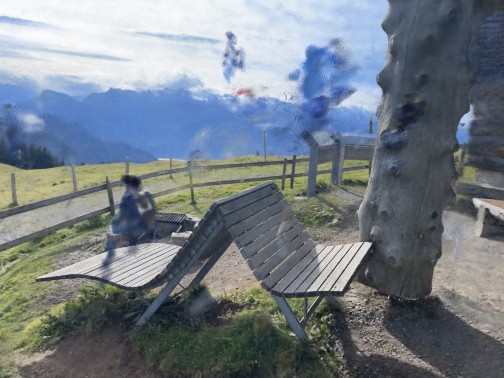}};
    \node [image,right=of mountain-3] (mountain-4) {\includegraphics[width=\figurewidth]{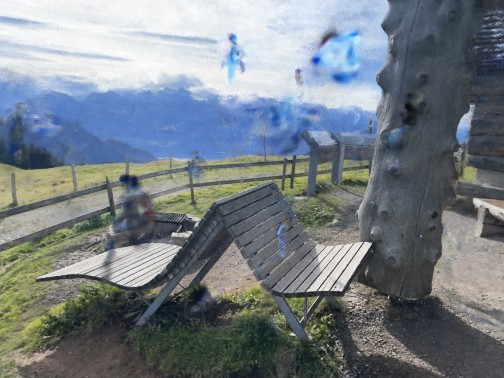}};
    \node [image,right=of mountain-4] (mountain-5) {\includegraphics[width=\figurewidth]{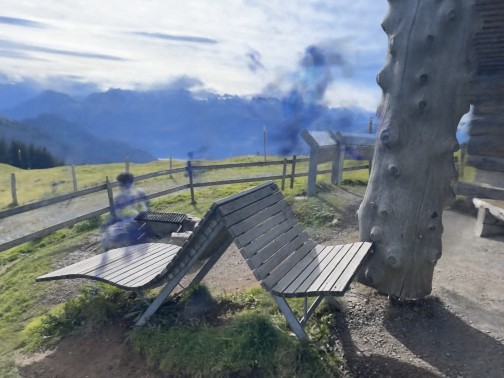}};
    \node [image,right=of mountain-5] (mountain-6) {\includegraphics[width=\figurewidth]{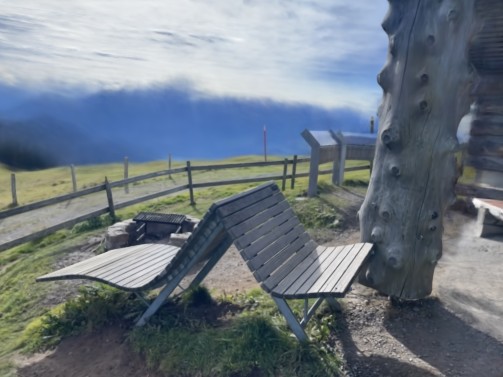}};
    \node [image,right=of mountain-6] (mountain-gt) {\includegraphics[width=\figurewidth]{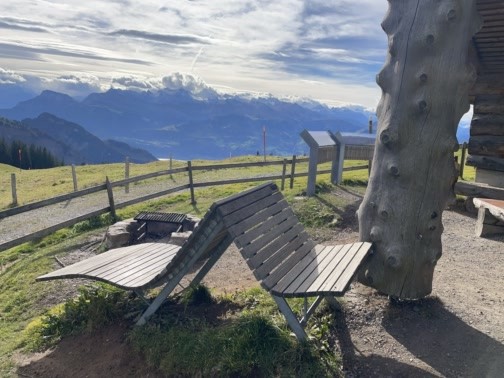}};
    
    \node[image,below=of mountain-1] (mountain-unc-1) {\includegraphics[width=\figurewidth]{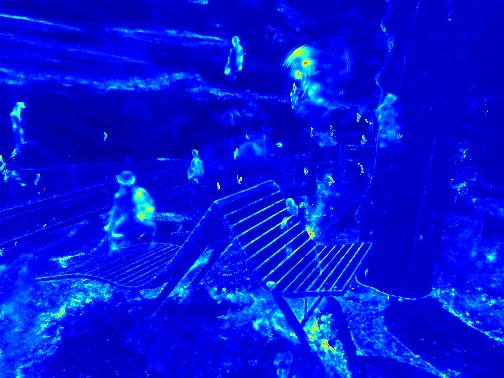}};
    \node[image,right=of mountain-unc-1] (mountain-unc-2) {\includegraphics[width=\figurewidth]{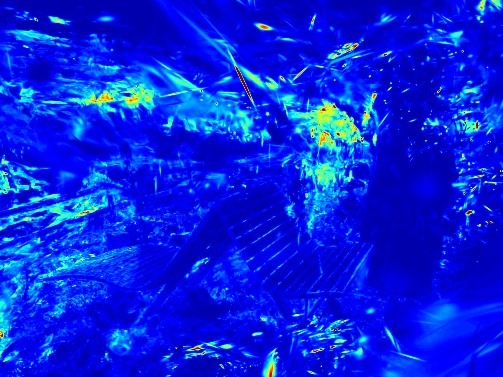}};
    \node[image,right=of mountain-unc-2] (mountain-unc-3) {\includegraphics[width=\figurewidth]{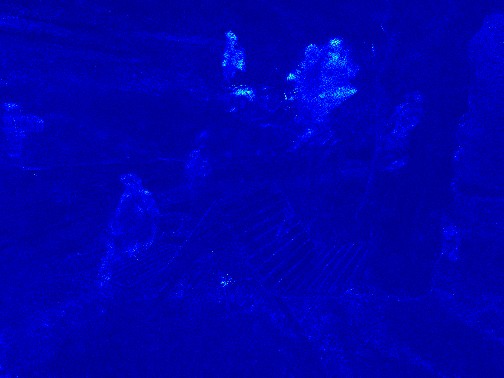}};
    \node[image,right=of mountain-unc-3] (mountain-unc-4) {\includegraphics[width=\figurewidth]{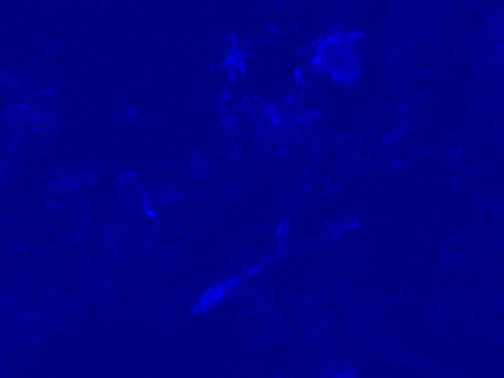}};
    \node[image,right=of mountain-unc-4] (mountain-unc-5) {\includegraphics[width=\figurewidth]{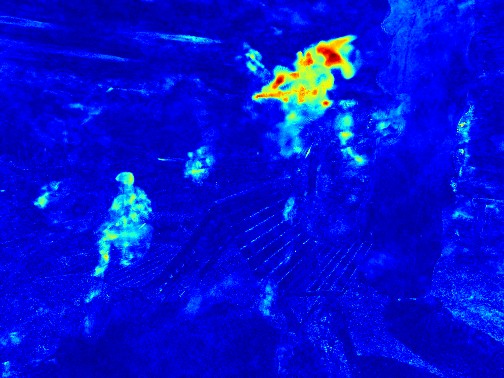}};
    \node[image,right=of mountain-unc-5] (mountain-unc-6) {\includegraphics[width=\figurewidth]{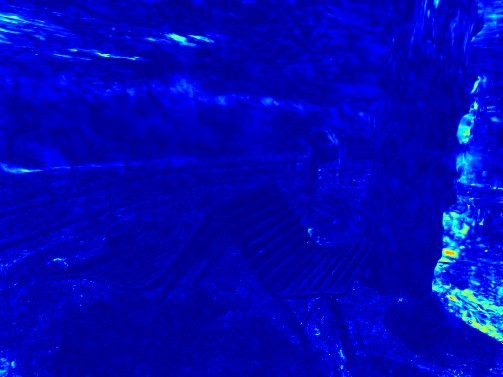}};

    \node [image,below=of mountain-unc-1] (patio-1) {\includegraphics[width=\figurewidth]{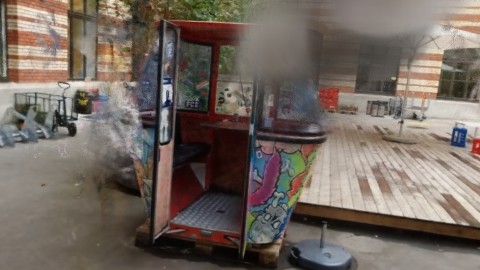}};
    \node [image,right=of patio-1] (patio-2) {\includegraphics[width=\figurewidth]{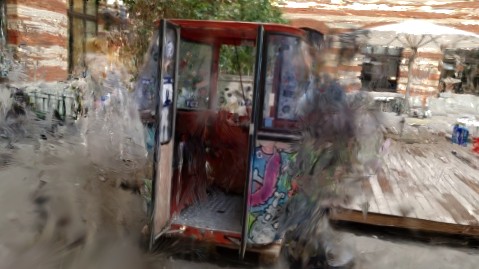}};
    \node [image,right=of patio-2] (patio-3) {\includegraphics[width=\figurewidth]{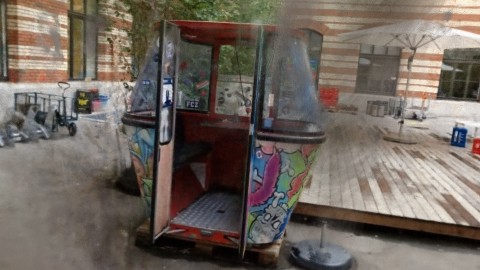}};
    \node [image,right=of patio-3] (patio-4) {\includegraphics[width=\figurewidth]{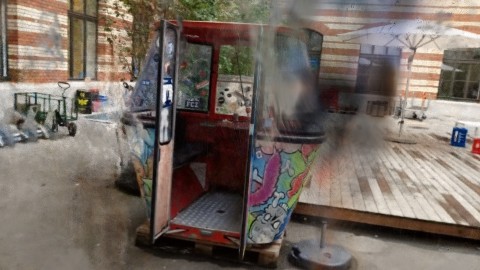}};
    \node [image,right=of patio-4] (patio-5) {\includegraphics[width=\figurewidth]{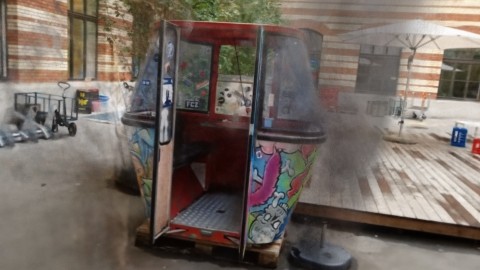}};
    \node [image,right=of patio-5] (patio-6) {\includegraphics[width=\figurewidth]{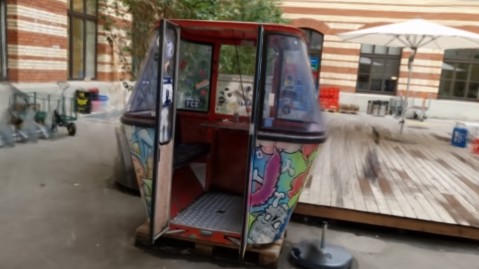}};
    \node [image,right=of patio-6] (patio-gt) {\includegraphics[width=\figurewidth]{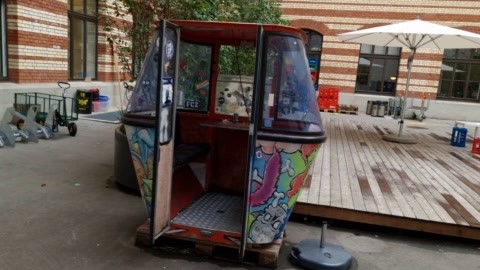}};

    \node[image,below=of patio-1] (patio-unc-1) {\includegraphics[width=\figurewidth]{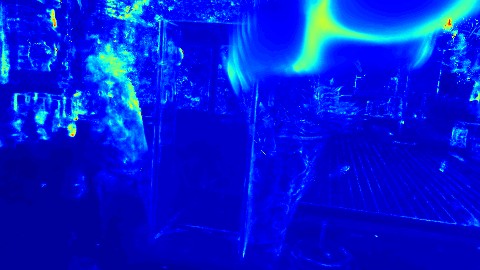}};
    \node[image,right=of patio-unc-1] (patio-unc-2) {\includegraphics[width=\figurewidth]{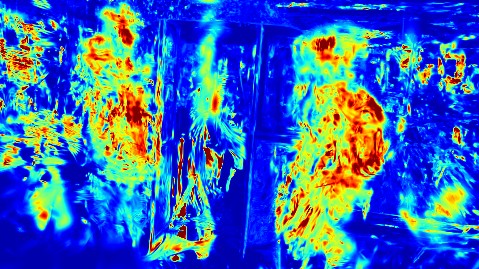}};
    \node[image,right=of patio-unc-2] (patio-unc-3) {\includegraphics[width=\figurewidth]{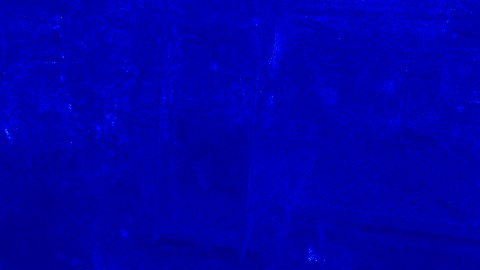}};
    \node[image,right=of patio-unc-3] (patio-unc-4) {\includegraphics[width=\figurewidth]{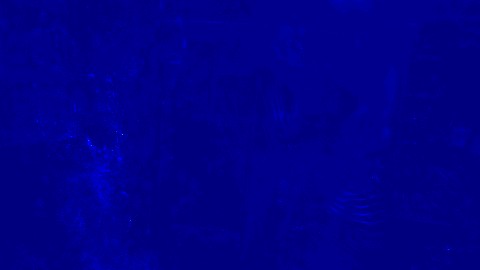}};
    \node[image,right=of patio-unc-4] (patio-unc-5) {\includegraphics[width=\figurewidth]{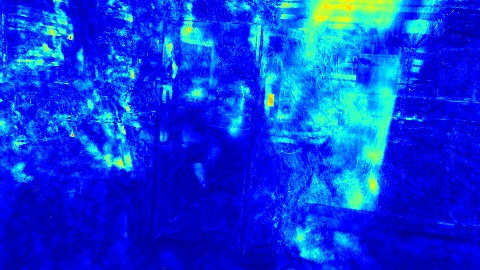}};
    \node[image,right=of patio-unc-5] (patio-unc-6) {\includegraphics[width=\figurewidth]{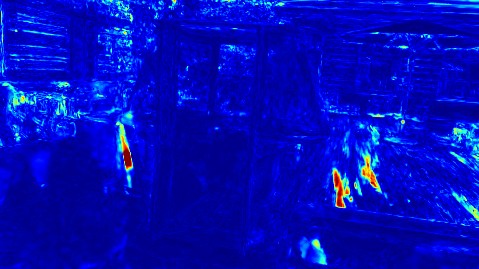}};
    \node[label] (label1) at (mountain-1.north) {Active-Nerfacto\vphantom{p}};
    \node[label] (label2) at (mountain-2.north) {Active-Splatfacto\vphantom{p}};
    \node[label] (label3) at (mountain-3.north) {MC-Dropout-Nerfacto\vphantom{p}};
    \node[label] (label4) at (mountain-4.north) {Laplace-Nerfacto\vphantom{p}};
    \node[label] (label5) at (mountain-5.north) {Ensemble-Nerfacto\vphantom{p}};
    \node[label] (label6) at (mountain-6.north) {Ensemble-Splatfacto\vphantom{p}};
    \node[label] (label-gt) at (mountain-gt.north) {Ground Truth\vphantom{p}};

    \node[anchor=south,inner sep=1pt,rotate=90] (scene2) at (mountain-1.south west) {Mountain\vphantom{p}};
    \node[anchor=south,inner sep=1pt,rotate=90] (scene3) at (patio-1.south west) {Patio\vphantom{p}};

  \end{tikzpicture}
  \caption{Confounding outliers \num{3}: Rendered RGB and uncertainty for test views from On-the-go scenes~\cite{ren2024nerf} next to the ground truth RGB. The uncertainty is visualized by the standard deviation (0.0~\protect\includegraphics[width=3em,height=.7em]{figures-main/jet.png}~0.3). See \cref{fig:outliers-nerfonthego-qualitative-appendix} for more scenes.}
  \label{fig:outliers-nerfonthego-qualitative}
\end{figure}

\paragraph{On-the-go Results} \cref{fig:outliers-nerfonthego-lightversion} shows the performance metrics averaged across the scenes by their different occlusion levels (see \cref{fig:outliers-nerfonthego-appendix} for all metrics). 
Ensemble-Splatfacto performs best on image quality, while the NeRF methods perform on par with each other. On the uncertainty metrics, the Ensemble methods perform consistently across the different occlusion levels, while the other methods have difficulties to correlate the uncertainty with the RGB error accurately on the high occlusion level scenes. 
In \cref{fig:outliers-nerfonthego-qualitative}, we visualize the rendered RGB and uncertainty of each method on the Mountain and Patio scenes, where Ensemble-Splatfacto effectively removes the floaters that appear in the other methods. However, Ensemble-Splatfacto produces blurry backgrounds especially in the Mountain scene. The other methods output higher uncertainty values on the floaters in general, which shows their capability of detecting such artefacts.

\subsection{Experiments on Sensitivity in Input Poses (\num{4})}
\label{sec:inputs}
We %
demonstrate the sensitivity (uncertainty) aspect of imprecise camera poses in 3D scene reconstruction. Our approach is to use the gradient of the rendered RGB $\vc$ w.r.t.\ the camera pose parameters $\mP$, \ie, $\partial \vc / \partial \mP \in \mathbb{R}^{3 \times 4}$, to quantify the sensitivity of the camera pose. 
We then perturb the camera pose and compute pixel-wise gradient norm maps to evaluate the degree of sensitivity in the reconstruction. The perturbations are simulated by shifting the camera position along the $z$-axis (zoom-out) with various magnitudes. The perturbation is small enough to enable pixel-wise comparisons between the perturbed vs.\ non-perturbed gradient norm maps. See \cref{app:pose} for more details.\looseness-1 

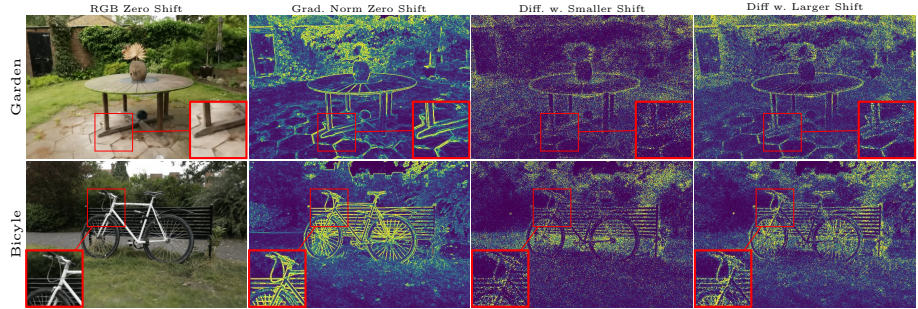
\begin{figure}[t]
    \centering
    \setlength{\figurewidth}{0.24\textwidth}
    \begin{tikzpicture}[image/.style = {inner sep=0, outer sep=0, minimum width=\figurewidth, anchor=north west, text width=\figurewidth}, node distance = 1pt and 1pt, every node/.style={font= {\tiny}}, label/.style = {scale=0.75,font={\tiny},anchor=south,inner sep=0pt,outer sep=2pt,rotate=0}, spy using outlines={rectangle, red, magnification=1.5, size=0.75cm, connect spies}] 

    \node [image] (img-0) {\includegraphics[width=\figurewidth]{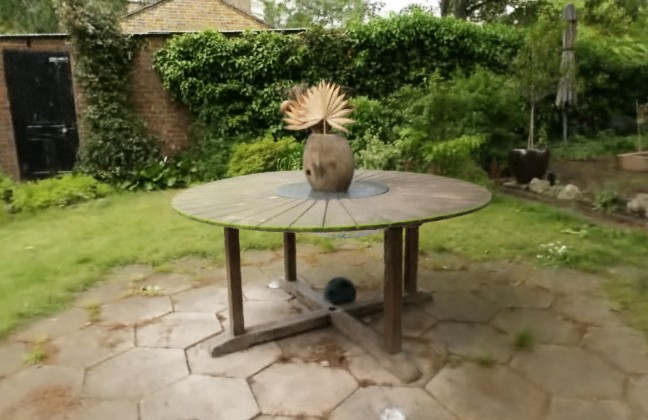}};
    \node [image,right=of img-0] (img-1) {\includegraphics[width=\figurewidth]{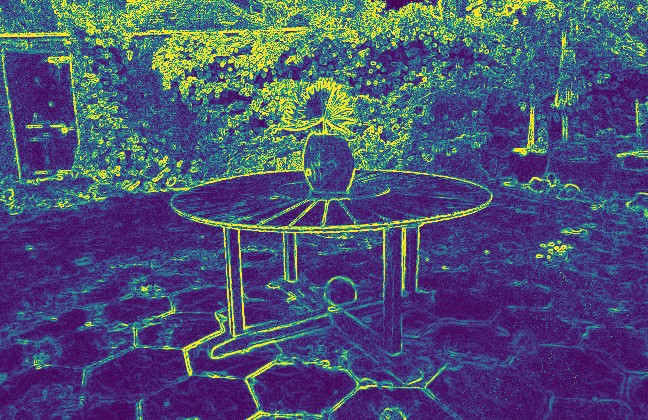}};
    \node [image,right=of img-1] (img-2) {\includegraphics[width=\figurewidth]{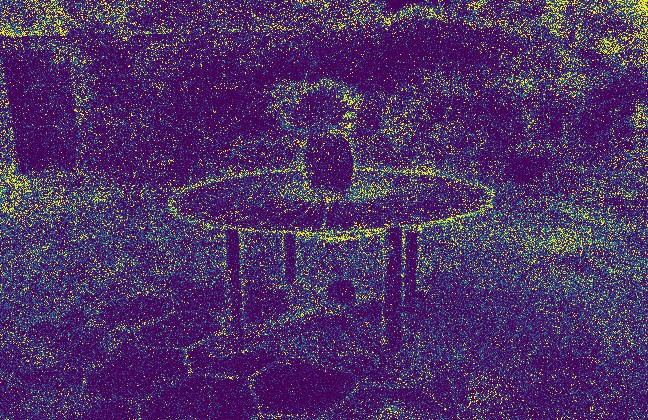}};
    \node [image,right=of img-2] (img-4) {\includegraphics[width=\figurewidth]{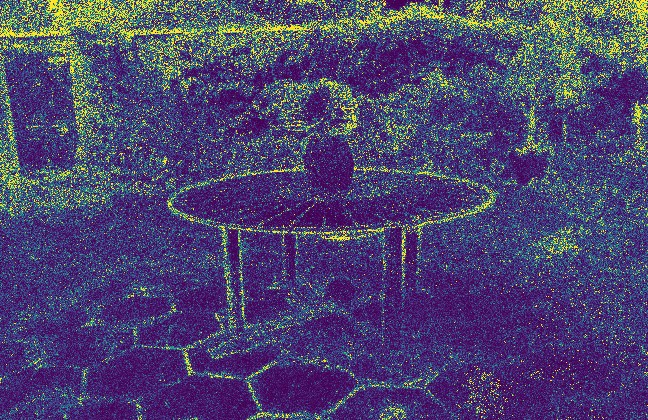}};

    \spy on ($(img-0) + (-0.30,-0.60)$) in node [left] at ($(img-0.south east) + (0,0.38)$);
    \spy on ($(img-1) + (-0.30,-0.60)$) in node [left] at ($(img-1.south east) + (0,0.38)$);
    \spy on ($(img-2) + (-0.30,-0.60)$) in node [left] at ($(img-2.south east) + (0,0.38)$);
    \spy on ($(img-4) + (-0.30,-0.60)$) in node [left] at ($(img-4.south east) + (0,0.38)$);
    \node[label] (label1) at (img-0.north) {RGB Zero Shift};
    \node[label] (label1) at (img-1.north) {Grad. Norm Zero Shift};
    \node[label] (label2) at (img-2.north) {Diff.\ w.\ Smaller  Shift};%
    \node[label] (label3) at (img-4.north) {Diff w.\ Larger Shift};%

    \node [image,below=of img-0] (img2-0) {\includegraphics[width=\figurewidth]{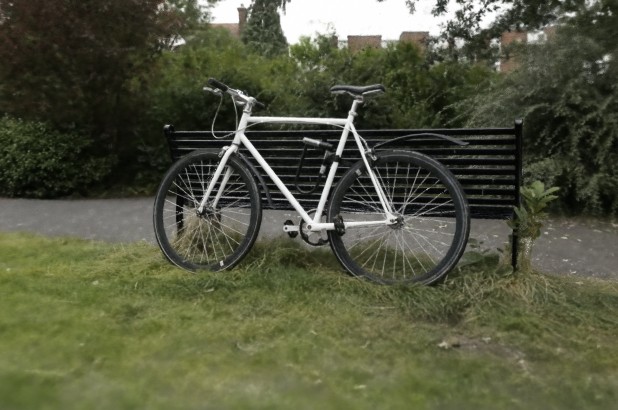}};
    \node [image,right=of img2-0] (img2-1) {\includegraphics[width=\figurewidth]{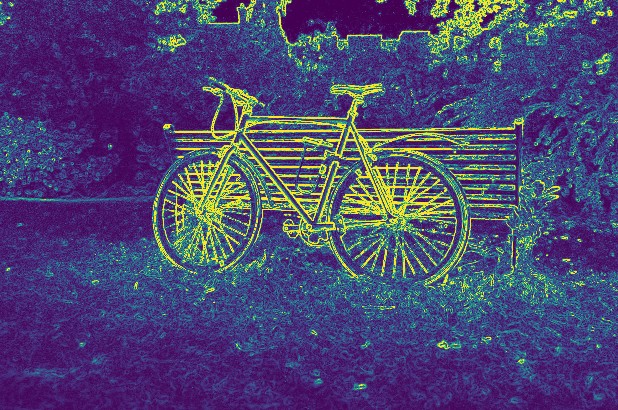}};
    \node [image,right=of img2-1] (img2-2) {\includegraphics[width=\figurewidth]{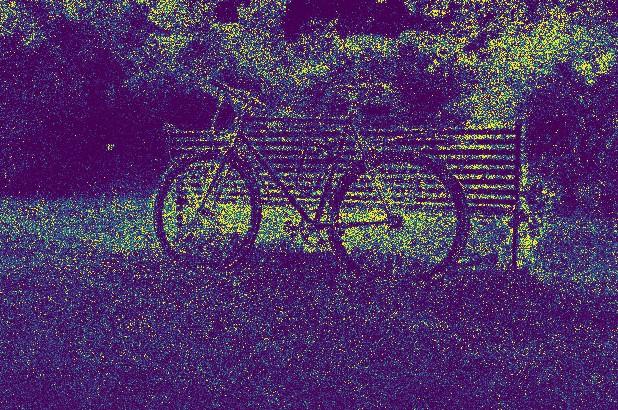}};
    \node [image,right=of img2-2] (img2-4) {\includegraphics[width=\figurewidth]{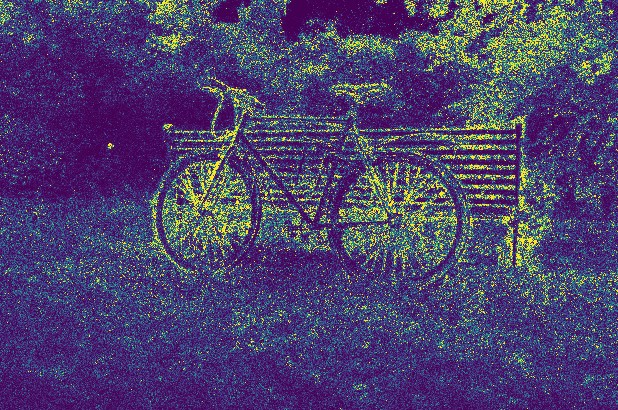}};

    \spy on ($(img2-0) + (-0.40,0.35)$) in node [right] at ($(img2-0.south west) + (0,0.38)$);
    \spy on ($(img2-1) + (-0.40,0.35)$) in node [right] at ($(img2-1.south west) + (0,0.38)$);
    \spy on ($(img2-2) + (-0.40,0.35)$) in node [right] at ($(img2-2.south west) + (0,0.38)$);
    \spy on ($(img2-4) + (-0.40,0.35)$) in node [right] at ($(img2-4.south west) + (0,0.38)$);

    \node[anchor=south,inner sep=1pt,rotate=90] (scene1) at (img-0.west) {Garden\vphantom{p}};
    \node[anchor=south,inner sep=1pt,rotate=90] (scene2) at (img2-0.west) {Bicyle\vphantom{p}};

    \end{tikzpicture}
    \caption{Rendered RGB, gradient norm map, and difference between gradient norms from perturbed \num{4} camera poses vs.\ non-shifted pose for perturbation shifts. We threshold the difference values by the 95\textsuperscript{th} percentile of the gradient norm maps individually. A larger shift results in larger gradient differences around edges with high-contrast. 
    }
    \label{fig:posegrad-mipnerf360-qualitative}
\end{figure}

\paragraph{Results} In \cref{fig:posegrad-mipnerf360-qualitative}, we show the rendered RGB from a Nerfacto model and its gradient norm from the original camera pose, and the gradient norm differences between the perturbed vs non-perturbed camera poses for two different degrees of shift. 
The gradient norm difference maps are thresholded by their own 95\textsuperscript{th} percentile value to enhance how the maps differ for increasing shifts. We observe that the high gradient norms dominate around edges and high-contrasts by inspecting the gradient norm maps with zero shift.  
In the difference maps, the gradient norms change the most in the fine-grained details and structural parts. Moreover, the differences become more concentrated in these parts when the shift increases, which means that the top-5\% highest gradient norms have become larger. For instance, the table edge and paving stone in the Garden scene, or the bench and bicycle edges in Bicycle, are more highlighted in the difference map with larger shift. This result demonstrates that the sensitivity aspect of camera poses exists in NeRFs and presumably GS-based methods.

\section{Conclusions}
\label{sec:conclusions}
In this paper, we covered various sources of uncertainty that arise in 3D reconstruction tasks for NeRF- and GS-based methods. 
We categorized four types of uncertainty sources that can appear in these scenarios (see \cref{fig:uncertainty}) and discussed how these can be intertwined as aleatoric uncertainty phenomenons can interact with the epistemic uncertainty of the modeling parameters (see \cref{sec:related_work}). 
To this end, we proposed different experimental setups to systematically evaluate their impact (see \cref{sec:experiments}) and performed an empirical study across several NeRF- and GS-based methods extended with common uncertainty estimation techniques in deep learning. 
Exploring new techniques that incorporate these uncertainties directly in the pipeline, via explicit modeling or regularization, is an important direction toward robust and uncertainty-aware 3D reconstruction.

\section*{Acknowledgments}
We acknowledge support from the Finnish Center for Artificial Intelligence (FCAI) and funding from the Research Council of Finland (grant 339730). We acknowledge CSC--IT Center for Science, Finland, for computational resources and acknowledge the computational resources provided by the Aalto Science-IT project.
We thank Martin Trapp and Matias Turkulainen for discussions on the idea. 
Finally, we thank the anonymous reviewers for their helpful suggestions and feedback.

\bibliographystyle{splncs04}

\begin{thebibliography}{10}
\providecommand{\url}[1]{\texttt{#1}}
\providecommand{\urlprefix}{URL }
\providecommand{\doi}[1]{https://doi.org/#1}

\bibitem{amini2023calibrated}
Amini-Naieni, N., Jakab, T., Vedaldi, A., Clark, R.: Calibrated uncertainties
  for neural radiance fields. arXiv preprint arXiv:2312.02350  (2023)

\bibitem{bae2021estimating}
Bae, G., Budvytis, I., Cipolla, R.: Estimating and exploiting the aleatoric
  uncertainty in surface normal estimation. In: Proceedings of the IEEE/CVF
  International Conference on Computer Vision (ICCV). pp. 13137--13146 (2021)

\bibitem{barron2022mip}
Barron, J.T., Mildenhall, B., Verbin, D., Srinivasan, P.P., Hedman, P.:
  {Mip-NeRF} 360: Unbounded anti-aliased neural radiance fields. In:
  Proceedings of the IEEE/CVF Conference on Computer Vision and Pattern
  Recognition (CVPR). pp. 5470--5479 (2022)

\bibitem{barron2023zip}
Barron, J.T., Mildenhall, B., Verbin, D., Srinivasan, P.P., Hedman, P.:
  Zip-nerf: Anti-aliased grid-based neural radiance fields. In: Proceedings of
  the IEEE/CVF International Conference on Computer Vision (ICCV). pp.
  19697--19705 (2023)

\bibitem{botev2017practical}
Botev, A., Ritter, H., Barber, D.: Practical {G}auss-{N}ewton optimisation for
  deep learning. In: International Conference on Machine Learning (ICML). pp.
  557--565. PMLR (2017)

\bibitem{daxberger2021laplace}
Daxberger, E., Kristiadi, A., Immer, A., Eschenhagen, R., Bauer, M., Hennig,
  P.: Laplace redux-effortless {B}ayesian deep learning. In: Advances in Neural
  Information Processing Systems 34 (NeurIPS). pp. 20089--20103 (2021)

\bibitem{gal2016dropout}
Gal, Y., Ghahramani, Z.: Dropout as a {B}ayesian approximation: Representing
  model uncertainty in deep learning. In: International Conference on Machine
  Learning (ICML). pp. 1050--1059. PMLR (2016)

\bibitem{goli2024bayes}
Goli, L., Reading, C., Sell{\'a}n, S., Jacobson, A., Tagliasacchi, A.: Bayes'
  rays: Uncertainty quantification for neural radiance fields. In: Proceedings
  of the IEEE/CVF Conference on Computer Vision and Pattern Recognition (CVPR).
  pp. 20061--20070 (2024)

\bibitem{gustafsson2020evaluating}
Gustafsson, F.K., Danelljan, M., Sch\"on, T.B.: Evaluating scalable {B}ayesian
  deep learning methods for robust computer vision. In: Proceedings of the
  IEEE/CVF Conference on Computer Vision and Pattern Recognition Workshops. pp.
  318--319 (2020)

\bibitem{he2024nerfs}
He, S., Osman, Z., Chaudhari, P.: From {NeRF}s to {G}aussian splats, and back.
  arXiv preprint arXiv:2405.09717  (2024)

\bibitem{ilg2018uncertainty}
Ilg, E., Cicek, O., Galesso, S., Klein, A., Makansi, O., Hutter, F., Brox, T.:
  Uncertainty estimates and multi-hypotheses networks for optical flow. In:
  European Conference on Computer Vision (ECCV). pp. 652--667 (2018)

\bibitem{immer2021improving}
Immer, A., Korzepa, M., Bauer, M.: Improving predictions of {B}ayesian neural
  nets via local linearization. In: International Conference on Artificial
  Intelligence and Statistics (AISTATS). pp. 703--711. PMLR (2021)

\bibitem{jeong2021self}
Jeong, Y., Ahn, S., Choy, C., Anandkumar, A., Cho, M., Park, J.:
  Self-calibrating neural radiance fields. In: Proceedings of the IEEE/CVF
  International Conference on Computer Vision (ICCV). pp. 5846--5854 (2021)

\bibitem{jiang2023fisherrf}
Jiang, W., Lei, B., Daniilidis, K.: {FisherRF}: Active view selection and
  uncertainty quantification for radiance fields using {F}isher information.
  arXiv preprint arXiv:2311.17874  (2023)

\bibitem{jin2023neu}
Jin, L., Chen, X., R{\"u}ckin, J., Popovi{\'c}, M.: {NeU-NBV}: Next best view
  planning using uncertainty estimation in image-based neural rendering. In:
  2023 IEEE/RSJ International Conference on Intelligent Robots and Systems
  (IROS). pp. 11305--11312. IEEE (2023)

\bibitem{kendall2017uncertainties}
Kendall, A., Gal, Y.: What uncertainties do we need in {B}ayesian deep learning
  for computer vision? In: Advances in Neural Information Processing Systems 30
  (NeurIPS) (2017)

\bibitem{kerbl20233d}
Kerbl, B., Kopanas, G., Leimk{\"u}hler, T., Drettakis, G.: {3D} {G}aussian
  splatting for real-time radiance field rendering. ACM Transactions on
  Graphics (TOG)  \textbf{42}(4),  1--14 (2023)

\bibitem{kulhanek2024wildgaussians}
Kulhanek, J., Peng, S., Kukelova, Z., Pollefeys, M., Sattler, T.:
  {WildGaussians}: {3D} {G}aussian splatting in the wild. arXiv preprint
  arXiv:2407.08447  (2024)

\bibitem{lakshminarayanan2017simple}
Lakshminarayanan, B., Pritzel, A., Blundell, C.: Simple and scalable predictive
  uncertainty estimation using deep ensembles. In: Advances in Neural
  Information Processing Systems 30 (NeurIPS) (2017)

\bibitem{lee2024bayesian}
Lee, S., Kang, K., Yu, H.: Bayesian {NeRF}: Quantifying uncertainty with volume
  density in neural radiance fields. arXiv preprint arXiv:2404.06727  (2024)

\bibitem{lin2021barf}
Lin, C.H., Ma, W.C., Torralba, A., Lucey, S.: Barf: Bundle-adjusting neural
  radiance fields. In: Proceedings of the IEEE/CVF International Conference on
  Computer Vision (ICCV). pp. 5741--5751 (2021)

\bibitem{loquercio2020general}
Loquercio, A., Segu, M., Scaramuzza, D.: A general framework for uncertainty
  estimation in deep learning. IEEE Robotics and Automation Letters
  \textbf{5}(2),  3153--3160 (2020)

\bibitem{ma2022deblur}
Ma, L., Li, X., Liao, J., Zhang, Q., Wang, X., Wang, J., Sander, P.V.:
  Deblur-{NeRF}: Neural radiance fields from blurry images. In: Proceedings of
  the IEEE/CVF Conference on Computer Vision and Pattern Recognition (CVPR).
  pp. 12861--12870 (2022)

\bibitem{mackay1992bayesian}
MacKay, D.J.: Bayesian interpolation. Neural Computation  \textbf{4}(3),
  415--447 (1992)

\bibitem{martin2021nerf}
Martin-Brualla, R., Radwan, N., Sajjadi, M.S., Barron, J.T., Dosovitskiy, A.,
  Duckworth, D.: {NeRF} in the wild: {N}eural radiance fields for unconstrained
  photo collections. In: Proceedings of the IEEE/CVF Conference on Computer
  Vision and Pattern Recognition (CVPR). pp. 7210--7219 (2021)

\bibitem{matsuki2024gaussian}
Matsuki, H., Murai, R., Kelly, P.H., Davison, A.J.: Gaussian splatting {SLAM}.
  In: Proceedings of the IEEE/CVF Conference on Computer Vision and Pattern
  Recognition (CVPR). pp. 18039--18048 (2024)

\bibitem{mildenhall2020nerf}
Mildenhall, B., Srinivasan, P., Tancik, M., Barron, J., Ramamoorthi, R., Ng,
  R.: {NeRF}: Representing scenes as neural radiance fields for view synthesis.
  In: European Conference on Computer Vision (ECCV) (2020)

\bibitem{muller2022instant}
M{\"u}ller, T., Evans, A., Schied, C., Keller, A.: Instant neural graphics
  primitives with a multiresolution hash encoding. ACM Transactions on Graphics
  (TOG)  \textbf{41}(4),  1--15 (2022)

\bibitem{ovadia2019can}
Ovadia, Y., Fertig, E., Ren, J., Nado, Z., Sculley, D., Nowozin, S., Dillon,
  J., Lakshminarayanan, B., Snoek, J.: Can you trust your model's uncertainty?
  evaluating predictive uncertainty under dataset shift. Advances in neural
  information processing systems  \textbf{32} (2019)

\bibitem{pan2022activenerf}
Pan, X., Lai, Z., Song, S., Huang, G.: {ActiveNeRF}: Learning where to see with
  uncertainty estimation. In: European Conference on Computer Vision (ECCV).
  pp. 230--246. Springer (2022)

\bibitem{papamarkou2024position}
Papamarkou, T., Skoularidou, M., Palla, K., Aitchison, L., Arbel, J., Dunson,
  D., Filippone, M., Fortuin, V., Hennig, P., Hern\'{a}ndez-Lobato, J.M.,
  Hubin, A., Immer, A., Karaletsos, T., Khan, M.E., Kristiadi, A., Li, Y.,
  Mandt, S., Nemeth, C., Osborne, M.A., Rudner, T.G.J., R\"{u}gamer, D., Teh,
  Y.W., Welling, M., Wilson, A.G., Zhang, R.: Position: {B}ayesian deep
  learning is needed in the age of large-scale {AI}. In: Proceedings of the
  41st International Conference on Machine Learning. vol.~235, pp. 39556--39586
  (2024)

\bibitem{poggi2020uncertainty}
Poggi, M., Aleotti, F., Tosi, F., Mattoccia, S.: On the uncertainty of
  self-supervised monocular depth estimation. In: Proceedings of the IEEE/CVF
  Conference on Computer Vision and Pattern Recognition (CVPR). pp. 3227--3237
  (2020)

\bibitem{qu2021bayesian}
Qu, C., Liu, W., Taylor, C.J.: Bayesian deep basis fitting for depth completion
  with uncertainty. In: Proceedings of the IEEE/CVF International Conference on
  Computer Vision (ICCV). pp. 16147--16157 (2021)

\bibitem{ran2023neurar}
Ran, Y., Zeng, J., He, S., Chen, J., Li, L., Chen, Y., Lee, G., Ye, Q.:
  {NeurAR}: Neural uncertainty for autonomous {3D} reconstruction with implicit
  neural representations. IEEE Robotics and Automation Letters  \textbf{8}(2),
  1125--1132 (2023)

\bibitem{ranftl2021bayesian}
Ranftl, S., von~der Linden, W.: Bayesian surrogate analysis and uncertainty
  propagation. In: Physical Sciences Forum. vol.~3, p.~6. MDPI (2021)

\bibitem{ren2024nerf}
Ren, W., Zhu, Z., Sun, B., Chen, J., Pollefeys, M., Peng, S.: {NeRF} on-the-go:
  Exploiting uncertainty for distractor-free {NeRF}s in the wild. In:
  Proceedings of the IEEE/CVF Conference on Computer Vision and Pattern
  Recognition (CVPR). pp. 8931--8940 (2024)

\bibitem{ritter2018scalable}
Ritter, H., Botev, A., Barber, D.: A scalable laplace approximation for neural
  networks. In: 6th International Conference on Learning Representations, ICLR
  2018-Conference Track Proceedings. vol.~6. International Conference on
  Representation Learning (2018)

\bibitem{roessle2022dense}
Roessle, B., Barron, J.T., Mildenhall, B., Srinivasan, P.P., Nie{\ss}ner, M.:
  Dense depth priors for neural radiance fields from sparse input views. In:
  Proceedings of the IEEE/CVF Conference on Computer Vision and Pattern
  Recognition (CVPR). pp. 12892--12901 (2022)

\bibitem{sabour2024spotlesssplats}
Sabour, S., Goli, L., Kopanas, G., Matthews, M., Lagun, D., Guibas, L.,
  Jacobson, A., Fleet, D.J., Tagliasacchi, A.: {SpotlessSplats}: Ignoring
  distractors in {3D} {G}aussian splatting. arXiv preprint arXiv:2406.20055
  (2024)

\bibitem{sabour2023robustnerf}
Sabour, S., Vora, S., Duckworth, D., Krasin, I., Fleet, D.J., Tagliasacchi, A.:
  {RobustNeRF}: {I}gnoring distractors with robust losses. In: Proceedings of
  the IEEE/CVF Conference on Computer Vision and Pattern Recognition (CVPR).
  pp. 20626--20636 (2023)

\bibitem{schoenberger2016sfm}
Sch\"{o}nberger, J.L., Frahm, J.M.: Structure-from-motion revisited. In:
  Proceedings of the IEEE/CVF Conference on Computer Vision and Pattern
  Recognition (CVPR) (2016)

\bibitem{seiskari2024gaussian}
Seiskari, O., Ylilammi, J., Kaatrasalo, V., Rantalankila, P., Turkulainen, M.,
  Kannala, J., Rahtu, E., Solin, A.: Gaussian splatting on the move: Blur and
  rolling shutter compensation for natural camera motion. arXiv preprint
  arXiv:2403.13327  (2024)

\bibitem{seo2023flipnerf}
Seo, S., Chang, Y., Kwak, N.: {FlipNeRF}: Flipped reflection rays for few-shot
  novel view synthesis. In: Proceedings of the IEEE/CVF International
  Conference on Computer Vision (ICCV). pp. 22883--22893 (2023)

\bibitem{seo2023mixnerf}
Seo, S., Han, D., Chang, Y., Kwak, N.: {MixNeRF}: Modeling a ray with mixture
  density for novel view synthesis from sparse inputs. In: Proceedings of the
  IEEE/CVF Conference on Computer Vision and Pattern Recognition (CVPR). pp.
  20659--20668 (2023)

\bibitem{shen2022conditional}
Shen, J., Agudo, A., Moreno-Noguer, F., Ruiz, A.: Conditional-flow {NeRF}:
  {A}ccurate {3D} modelling with reliable uncertainty quantification. In:
  European Conference on Computer Vision (ECCV). pp. 540--557. Springer (2022)

\bibitem{shen2021stochastic}
Shen, J., Ruiz, A., Agudo, A., Moreno-Noguer, F.: Stochastic neural radiance
  fields: Quantifying uncertainty in implicit {3D} representations. In:
  International Conference on 3D Vision (3DV). pp. 972--981. IEEE (2021)

\bibitem{srivastava2014dropout}
Srivastava, N., Hinton, G., Krizhevsky, A., Sutskever, I., Salakhutdinov, R.:
  Dropout: A simple way to prevent neural networks from overfitting. Journal of
  Machine Learning Research  \textbf{15}(1),  1929--1958 (2014)

\bibitem{sunderhauf2023density}
S{\"u}nderhauf, N., Abou-Chakra, J., Miller, D.: Density-aware {NeRF}
  ensembles: Quantifying predictive uncertainty in neural radiance fields. In:
  2023 IEEE International Conference on Robotics and Automation (ICRA). pp.
  9370--9376. IEEE (2023)

\bibitem{tancik2023nerfstudio}
Tancik, M., Weber, E., Ng, E., Li, R., Yi, B., Wang, T., Kristoffersen, A.,
  Austin, J., Salahi, K., Ahuja, A., et~al.: Nerfstudio: A modular framework
  for neural radiance field development. In: ACM SIGGRAPH 2023 Conference
  Proceedings. pp. 1--12 (2023)

\bibitem{turkulainen2024dn}
Turkulainen, M., Ren, X., Melekhov, I., Seiskari, O., Rahtu, E., Kannala, J.:
  {DN-Splatter}: Depth and normal priors for {G}aussian splatting and meshing.
  arXiv preprint arXiv:2403.17822  (2024)

\bibitem{wang2003multiscale}
Wang, Z., Simoncelli, E.P., Bovik, A.C.: Multiscale structural similarity for
  image quality assessment. In: The Thrity-Seventh Asilomar Conference on
  Signals, Systems \& Computers. vol.~2, pp. 1398--1402. IEEE (2003)

\bibitem{yucer2016efficient}
Y{\"u}cer, K., Sorkine-Hornung, A., Wang, O., Sorkine-Hornung, O.: Efficient
  {3D} object segmentation from densely sampled light fields with applications
  to {3D} reconstruction. ACM Transactions on Graphics (TOG)  \textbf{35}(3),
  1--15 (2016)

\bibitem{zhang2018advances}
Zhang, C., B{\"u}tepage, J., Kjellstr{\"o}m, H., Mandt, S.: Advances in
  variational inference. IEEE Transactions on Pattern Analysis and Machine
  Intelligence  \textbf{41}(8),  2008--2026 (2018)

\bibitem{zhang2020nerf++}
Zhang, K., Riegler, G., Snavely, N., Koltun, V.: {NeRF++}: Analyzing and
  improving neural radiance fields. arXiv preprint arXiv:2010.07492  (2020)

\bibitem{zhang2018unreasonable}
Zhang, R., Isola, P., Efros, A.A., Shechtman, E., Wang, O.: The unreasonable
  effectiveness of deep features as a perceptual metric. In: Proceedings of the
  IEEE Conference on Computer Vision and Pattern Recognition (CVPR). pp.
  586--595 (2018)

\end{thebibliography}

\clearpage
\appendix

\renewcommand{\thetable}{A\arabic{table}}
\renewcommand{\thefigure}{A\arabic{figure}}

\section{Method Details}
\label{app:methods}
This section contains additional details about the uncertainty estimation techniques that we have used in our experiments in combination with NeRF- and GS-based methods.

\paragraph{Active-NeRF} We follow \cite{pan2022activenerf} and render the uncertainty using
\begin{equation}
    \beta = \sum_{i=1}^{N_s} T_i^2 \alpha_i^2 \beta_i
\end{equation} 
and use the Gaussian log-likelihood as the loss:
\begin{equation}\label{eq:loss-active-nerf}
    \gL_{\text{Active-NeRF}} = \frac{|| \vc_{\text{NeRF}} - \vc_{\text{GT}}||_2^2}{2\beta^2 } + \frac{\log{\beta^2}}{2} + \frac{\lambda}{N_s} \sum_{k=1}^{N_s} \sigma_k ,
\end{equation}
where the last term is an $L1$-regularizer with strength $\lambda$ to enforce sparse density values. We set $\lambda = 0.01$ in all experiments. For depth prediction in \cref{sec:reducible}, we estimate the depth variance as proposed in \cite{roessle2022dense} using 
\begin{equation}\label{eq:depth-variance-roessle}
    \text{Var}(d_{\text{NeRF}}) = \sum_{k=1}^{N_s} T_i \alpha_i (t_k - d_{\text{NeRF}})^2 , \quad \text{where} \quad d_{\text{NeRF}} = \sum_{k=1}^{N_s} T_i \alpha_i t_k, 
\end{equation}
where $d_{\text{NeRF}}$ is the depth estimate and 
$t_k \in [t_n, t_f ]$ is the sampling location along the ray within the near and far planes. 

\paragraph{Active-GS}
We render the uncertainty $\beta = \sum_{i=1}^{N_s} T_i \alpha_i \beta_i$, similarly as done in NeRF-W~\cite{martin2021nerf}. We leave it as future work to use the uncertainty rendering from Active-NeRF for GS since squaring the rendering weights requires modifying the rasterization code in CUDA. 
We replace the $\gL_1$-loss with the Gaussian log-likelihood as: %
\begin{equation}\label{eq:loss-active-gs}
    \gL_{\text{Active-GS}} = (1-\lambda) \left( \frac{|| \vc_{\text{GS}} - \vc_{\text{GT}}||_2^2}{2\beta^2 } + \frac{\log{\beta^2}}{2} \right) + \lambda \gL_{\text{SSIM}} + \frac{\lambda_r}{N_p} \sum_{k=1}^{N_p} o_k ,
\end{equation}
where the last term is an $L1$-regularizer with strength $\lambda_r$ to enforce sparse opacity values for the sorted $N_p$ Gaussians. We set $\lambda_r = 0.01$ in all experiments, except for the Mip-NeRF 360 scenes where it was increased to $\lambda_r = 0.02$ to avoid out-of-memory errors. The weighting factor for SSIM is set to the default $\lambda=0.2$. 
As depth uncertainty, we use the depth variance proposed in \cite{roessle2022dense} here as
\begin{equation}\label{eq:depth-variance-gs}
    \text{Var}(d_{\text{GS}}) = \sum_{k=1}^{N_p} T_i \alpha_i (d_k - d_{\text{GS}})^2 ,  \quad \text{where} \quad d_{\text{GS}} = \sum_{k=1}^{N_s} T_i \alpha_i d_k, 
\end{equation}
where $d_{\text{GS}}$ is the per-pixel $z$–depth estimates rendered using the discrete volume rendering and where $d_k$ is the $k$\textsuperscript{th} Gaussian $z$-depth coordinate in view space~\cite{turkulainen2024dn}.

\paragraph{MC-Dropout NeRF} We apply MC-Dropout~\cite{gal2016dropout} before the last layer of both the color and density MLPs. The dropout rate is set to $p=0.2$ during both training and testing, where $p$ is the probability of a weight to be masked. The training loss is the mean-squared-error which is the standard for NeRFs (see \cref{sec:background}). We use $M=5$ forward-passes to estimate the RGB variance in \cref{eq:mcdropout-nerf-color-and-unc} at test time. For depth prediction, we use the same NeRF depth estimate as Active-NeRF, \ie, $d_{\text{NeRF}} = \sum_{k=1}^{N_s} T_i \alpha_i t_k$, and average the depth maps obtained with the $M$ forward-passes to estimate its variance.

\paragraph{Laplace NeRF} 
We apply the Laplace approximation~\cite{mackay1992bayesian,daxberger2021laplace} on the last layer of both the color and density MLPs of NeRF. We use a diagonal approximation and the generalized Gauss-Newton matrix~(GGN, \cite{immer2021improving}) to approximate the two Hessian matrices. We compute the diagonal GGN using 100 batches of 4096 rays randomly sampled from the training images. The covariance matrices for the approximate posteriors of the two last layers are $\mSigma = (\Hess + \gamma \mathbf{I})^{-1}$ where $\gamma$ is the prior precision which we set to $\gamma = 1.0$ and $\mathbf{I}$ is the identity matrix. We use $M=100$ samples drawn from the approximate posterior on the weights $q(\vtheta)$ to estimate the mean and variance of the color values along the rays as, 
\begin{align}
    \hat{\vmu}_{\vc} & = \frac{1}{M} \sum_{m=1}^{M} f_{\vtheta^{(m)}}(\vx, \vd), \quad \text{where} \quad \vtheta^{(m)} \sim \mathcal{N}(\vtheta; \vtheta^*, \mSigma), \\
    \hat{\beta}_{\vc} & \textstyle\approx \frac{1}{M} \sum_{m=1}^{M} f_{\vtheta^{(m)}}(\vx, \vd)^2  - \hat{\vmu}_{\vc}^2 .
\end{align}
We then render the color and its uncertainty as for Active-NeRF~\cite{pan2022activenerf}.

The mean and variance for the density values along the ray are computed the same way as the color above. However, since the density is nested in both the transmittance $T$ and the blending weight $\alpha$, the rendering weight $w = T \alpha$ has the distributional form of the difference between two log-Normal random variables~\cite{lee2024bayesian}. This makes rendering an uncertainty using the density variance challenging as this has no closed-form solution. We opt to compute an empirical mean of the rendering weights using the density distribution as 
\begin{equation}
    \hat{w}_i = \frac{1}{L} \sum_{l=1}^{L} T_i^{(l)} \alpha_i^{(l)} \text{~with~} T_i^{(l)} = \textstyle\prod_{i=1}^{j-1} (1 - \alpha_i^{(l)}), \quad \alpha_j^{(l)} = 1 - \exp{(-\sigma_j^{(l)} \delta_j)} ,
\end{equation}
where $\sigma_j^{(l)} \sim \mathcal{N}(\sigma; \mu_{\sigma}, \beta_{\sigma})$ and with $L=100$ samples. We then use the rendering weights $\hat{w}$ to compute the depth estimate and its variance with the same approach as for Active-NeRF above\cite{roessle2022dense}.

\paragraph{Ensemble NeRF/GS} 
We train $M=5$ models for both NeRF and GS ensmbles, which is the default value in \cite{lakshminarayanan2017simple}. The models are trained with the standard losses for NeRF and GS described in \cref{sec:background}. We compute the depth estimates the same way as for Active-NeRF/GS, but average the $M$ depth maps to estimate the depth variance.

\section{Additional Experimental Details}
\label{app:additional_experimental_results}
In this section, we first describe details on the evaluation metrics (\cref{app:evaluation_metrics}) and the data sets employed in our experiments (\cref{app:datasets}).
Then, we provide additional results for each experiment section, \ie, aleatoric uncertainty~(\cref{app:aleatoric}), epistemic uncertainty~(\cref{app:epistemic}), sensitivity to cluttered inputs~(\cref{app:outliers}), and sensitivity in the input poses~(\cref{app:pose}).

\subsection{Details on Evaluation Metrics}
\label{app:evaluation_metrics}

\paragraph{Negative Log-likelihood (NLL)} We use the Gaussian log-likelihood to compute the NLL metric. We set the minimum standard deviation to $\beta_{\text{min}} = 0.03$ for the RGB and to $\beta_{\text{min}} = 2.0$ for the depth, since the NLL can produce diverging values for outliers. 

\paragraph{Area Under Sparsification Error (AUSE)}
The Area Under Sparsification Error~(AUSE) is used to assess the correlation between predicted uncertainty of the model and the errors in the prediction values. 
The prediction errors for every pixel are sorted based on their value in descending order. 
We then compute the mean of the pixel errors for different sets where a ratio of the sorted pixels have been removed. 
The same procudre is performed for the prediction errors that are sorted by their predicted uncertainty. 
The difference between the mean pixel error ratios sorted by their error and uncertainty is the sparsification error. We integrate over the sparsification error to obtain a scalar metric for when the prediction error coincides with the predictied uncertainty. 
We use the the mean-squared error (MSE) as the pixel error metric when reporting the AUSE for all experiments. 

\paragraph{Area Under Calibration Error (AUCE)} 
The main idea with the AUCE metric is to construct prediction intervals for each pixel and check the proportion of pixels for which their interval covers their corresponding true target value. The prediction intervals are constructed pixel-wise using $\mu \pm \mathbf{\Phi}^{-1}(\frac{p+1}{2})\beta$, where $p \in (0,1)$ is the confidence level, $\mathbf{\Phi}$ is the cumulative distribution function (CDF) of the standard Normal distribution, and $\mu$ and $\beta$ are the predicted mean and standard deviation respectively, \eg, rendered RGB and its uncertainty. The empirical {\it coverage} $\hat{p}$ of each confidence level $p$ is computed by constructing a prediction interval for every $(\mu_i, \beta_i)$-pair and then count how many targets $y_i$ that fall within their corresponding interval. The empirical coverage $\hat{p}$ should be equal to the corresponding $p$ for a perfectly calibrated model. As proposed in \cite{gustafsson2020evaluating}, the metric is computed using the absolute error with respect to perfect calibration, \ie, $|\hat{p} - p|$, for 100 values of $p \in (0, 1)$ and taking the area under the curve to obtain a scalar.

\subsection{Data Sets}
\label{app:datasets}
In our experiments, we use scenes from different data sets to evaluate the performances of the selected uncertainty estimations techniques. The taxonomy we introduced in \cref{sec:related_work} requires that our data presents a particular type of uncertainty source in order to quantify the robustness of the method to it.

We use the {\bf Mip-NeRF 360}~\cite{barron2022mip} data set, which consists of 9 real-world scenes that contain a complex object in the center with detailed background. Out of the 9 scenes, 5 are outdoors and 4 are indoors. The data set is captured such that there is minimum light variance and no presence of moving objects. We used this data set for experiments on aleatoric uncertainty~\num{1}, the epistemic uncertainty~\num{2}, and sensitivity to the camera pose uncertainty~\num{4}. 
For~\num{1}, we vary the proportion of training views available at training time by randomly sampling 10\%, 25\%, 50\%, and 100\% of the training views. 
We regularly subsample 10\% of the input images as the test set before the training set sampling to ensure the test set to be the same across the training view proportions. 
For~\num{2}, we created an out-of-distribution (OOD) setting by splitting all available views based on the camera pose. more specifically, the training views are selected based on where the $x$-translation is positive, except for the scene Room where the training views are selected based on the positive $z$-translation. We then evaluate on the remaining views that are from camera views that are from unobserved regions of the scene.  
Finally, for \num{4}, we compute the pose gradient over the scenes Garden, Bicycle, and Kitchen.

For the {\bf Light Field} (LF)~\cite{yucer2016efficient,zhang2020nerf++} data set, we follow the sparse setting introduced in \cite{shen2021stochastic, shen2022conditional}. In this setting, from the 4 real-world scenes in LF, \ie, Africa, Basket, Ship, and Torch, only 4-5 views are used for training on each scene. This data set is used for epistemic uncertainty experiments.

For the uncertainty coming from a cluttered scene \num{3}, we used two data sets {\bf RobustNeRF}~\cite{sabour2023robustnerf} data set and the {\bf On-the-go}~\cite{ren2024nerf} data set. Both data sets consist of real-world scenes that present different level of occlusions due to the presence of confounding objects, \eg, toys or people moving around the scene.\\
For RobustNeRF, the results reported in \cref{tab:outliers-robustnerf} are based on the scenes And-bot, Crab, Balloon Statue, and Yoda. In \cref{fig:outliers-yoda-clutter-visual}, we used the different levels of cluttered views, from only clean images to only cluttered ones, in order to assess the uncertainty predictions increasing the difficulty for the model to explain the presence of higher percentage of confounders.\\
In our experiments on On-the-go data set, we used 6 scenes with varying levels of occlusion. In particular, we have: {\em (i)}~Low occlusion, Fountain and Mountain scenes; {\em (ii)}~Medium occlusion, Corner and 
Patio; {\em (iii)}~High occlusion, Patio\_high and Spot. 

Finally, in \cref{app:aleatoric} we show some additional results for the aleatoric and epistemic uncertainty experiments, \num{1} and \num{2}, over all scenes from the Blender synthetic data set~\cite{mildenhall2020nerf}, which consists of scenes obtained from realistic Blender 3D models. 
Similarly as for the Mip-NeRF 360 data set, for~\num{1}, we vary the proportion of training views available at training time by randomly sampling 10\%, 25\%, 50\%, and 100\% of the training views, and then evaluate on the standard test set. For~\num{2}, we select the views that are in the left hemisphere in the scenes (negative $x$-translation) as the training views, and evaluate on the remaining views on the right hemisphere of the scene.

\begin{figure}[t]
  \centering
  \setlength{\figurewidth}{0.31\textwidth}
  \begin{subfigure}[c]{\linewidth}
  \begin{tikzpicture}[image/.style = {inner sep=0, outer sep=0, minimum width=\figurewidth, anchor=north west, text width=\figurewidth}, node distance = 1pt and 1pt, every node/.style={font= {\tiny}}, label/.style = {scale=0.75,font={\tiny},anchor=south,inner sep=0pt,outer sep=2pt,rotate=0}, spy using outlines={rectangle, magnification=2, size=0.90cm, connect spies}] 

    \node [image] (garden-noise-original) {\includegraphics[width=\figurewidth]{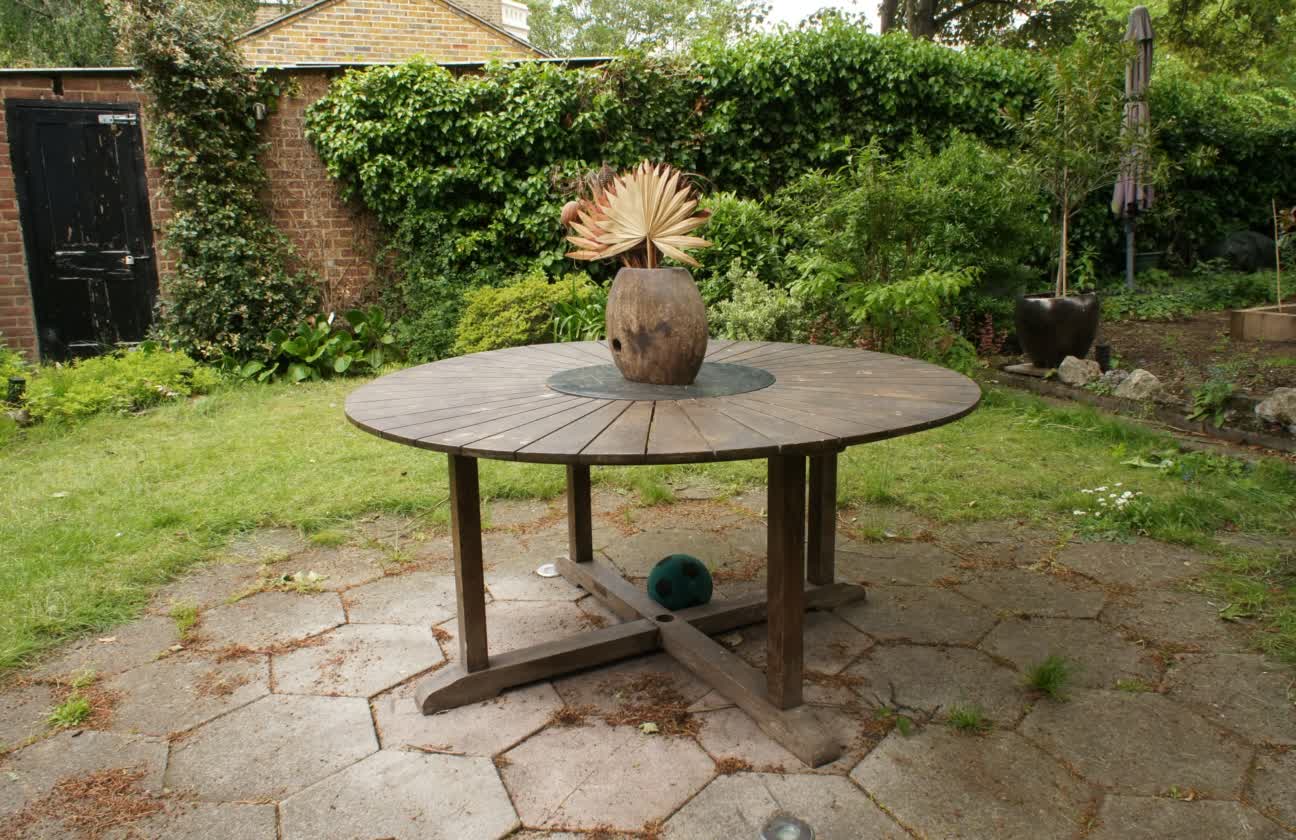}}; 
    \node [image,right= of garden-noise-original] (garden-noise-1) {\includegraphics[width=\figurewidth]{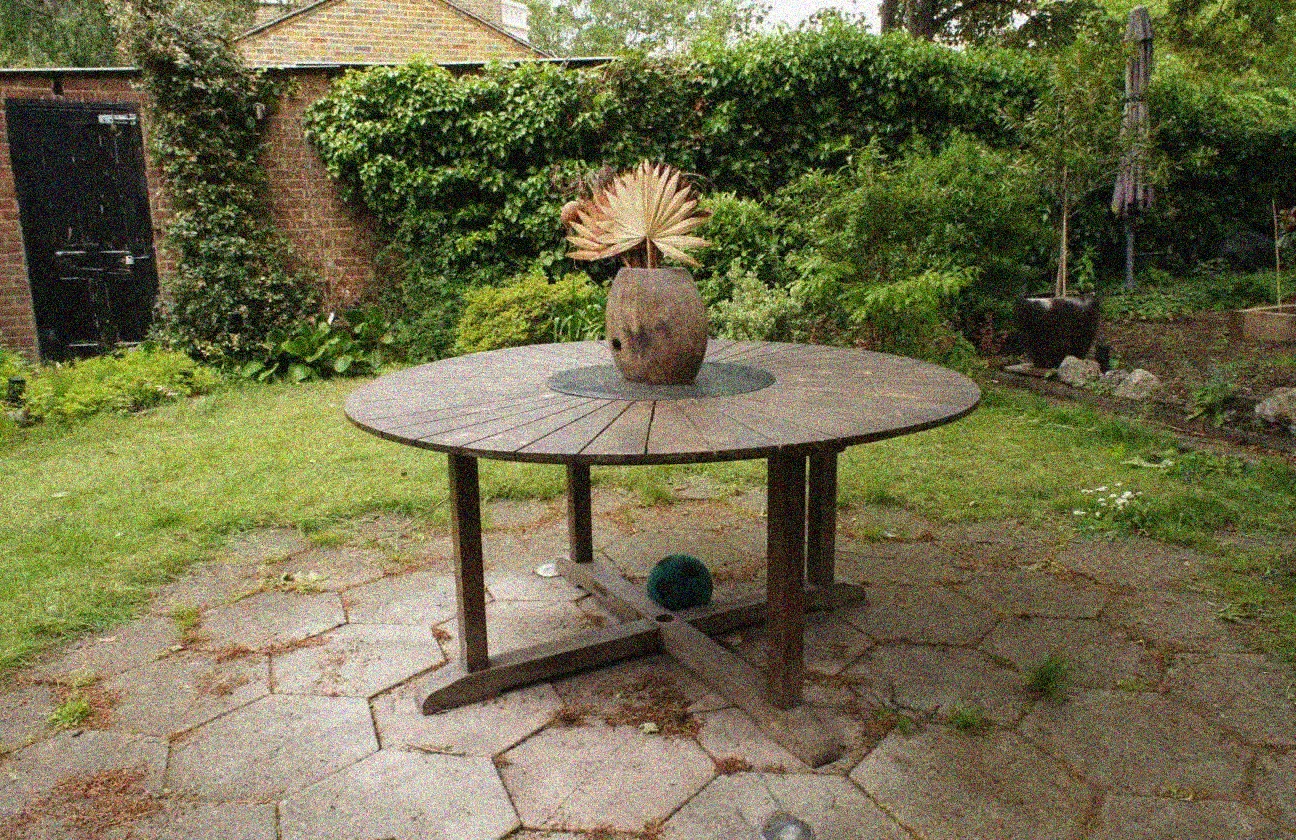}}; 
    \node [image,right=of garden-noise-1] (garden-noise-2) {\includegraphics[width=\figurewidth]{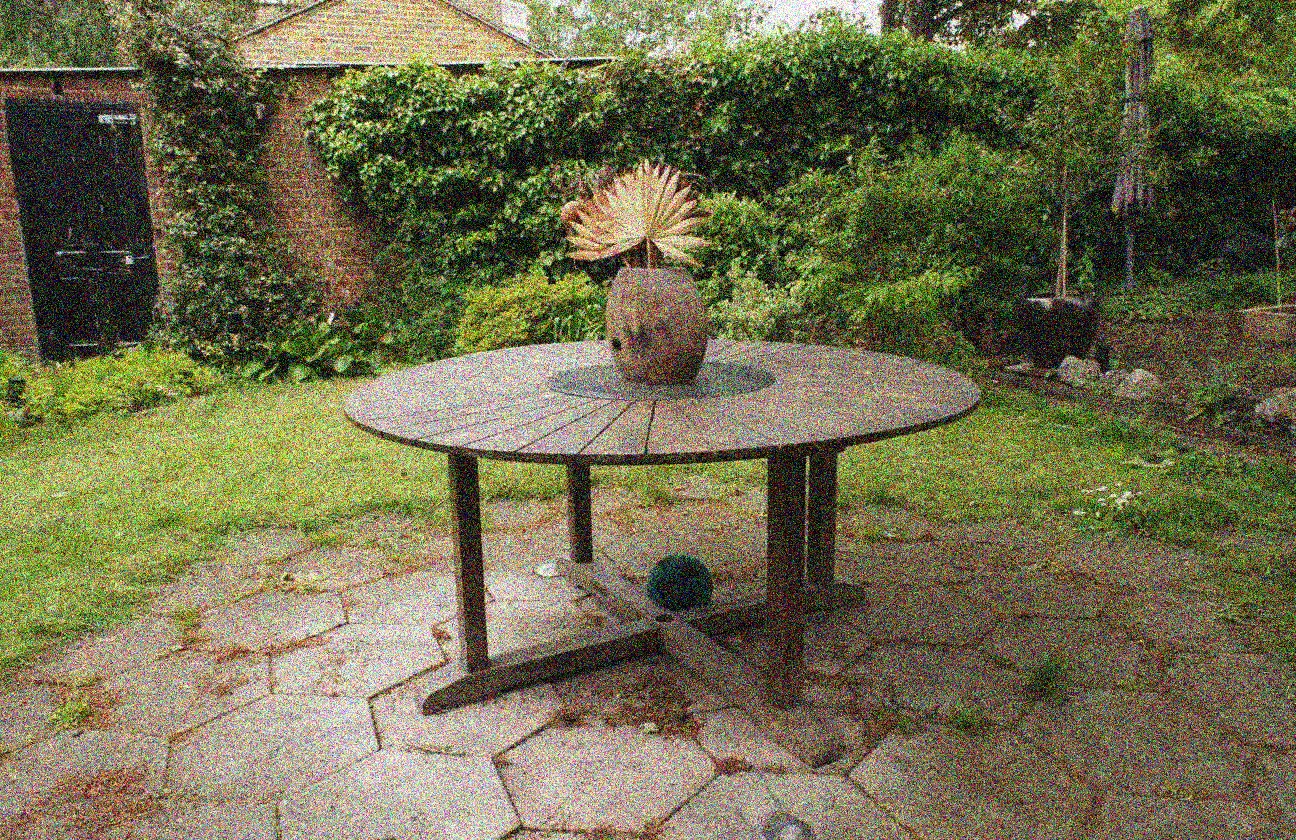}};

    \spy [red] on ($(garden-noise-original) + (-0.0,0.5)$) in node [right,below] at ($(garden-noise-original.north west) + (0.5,0.0)$);
    \spy [red] on ($(garden-noise-1) + (-0.0,0.5)$) in node [right,below] at ($(garden-noise-1.north west) + (0.5,0.0)$);
    \spy [red] on ($(garden-noise-2) + (-0.0,0.5)$) in node [right,below] at ($(garden-noise-2.north west) + (0.5,0.0)$);

    \spy [blue] on ($(garden-noise-original) + (1.0,-0.5)$) in node [right] at ($(garden-noise-original.south west) + (0,0.5)$);
    \spy [blue] on ($(garden-noise-1) + (1.0,-0.5)$) in node [right] at ($(garden-noise-1.south west) + (0,0.5)$);
    \spy [blue] on ($(garden-noise-2) + (1.0,-0.5)$) in node [right] at ($(garden-noise-2.south west) + (0,0.5)$);

    \node [image,below=of garden-noise-original] (kitchen-noise-original) {\includegraphics[width=\figurewidth]{figures-main/noisyview-mipnerf360-examples/kitchen/frame_00001.JPG}}; 
    \node [image,right=of kitchen-noise-original] (kitchen-noise-1) {\includegraphics[width=\figurewidth]{figures-main/noisyview-mipnerf360-examples/kitchen/frame_00001_noise_stddev_0.1.JPG}}; 
    \node [image,right=of kitchen-noise-1] (kitchen-noise-2) {\includegraphics[width=\figurewidth]{figures-main/noisyview-mipnerf360-examples/kitchen/frame_00001_noise_stddev_0.2.JPG}}; 

    \spy [red] on ($(kitchen-noise-original) + (-0.0,0.8)$) in node [right,below] at ($(kitchen-noise-original.north west) + (0.5,0.0)$);
    \spy [red] on ($(kitchen-noise-1) + (-0.0,0.8)$) in node [right,below] at ($(kitchen-noise-1.north west) + (0.5,0.0)$);
    \spy [red] on ($(kitchen-noise-2) + (-0.0,0.8)$) in node [right,below] at ($(kitchen-noise-2.north west) + (0.5,0.0)$);

    \spy [blue] on ($(kitchen-noise-original) + (-0.8,-0.4)$) in node [right] at ($(kitchen-noise-original.south west) + (0,0.5)$);
    \spy [blue] on ($(kitchen-noise-1) + (-0.8,-0.4)$) in node [right] at ($(kitchen-noise-1.south west) + (0,0.5)$);
    \spy [blue] on ($(kitchen-noise-2) + (-0.8,-0.4)$) in node [right] at ($(kitchen-noise-2.south west) + (0,0.5)$);

    \node[label] (label5) at (garden-noise-original.north) {Original\vphantom{p}};
    \node[label] (label1) at (garden-noise-1.north) {Noise Scale 0.1\vphantom{p}};
    \node[label] (label2) at (garden-noise-2.north) {Noise Scale 0.2\vphantom{p}};

    \node[anchor=south,inner sep=1pt,rotate=90] (scene2) at (garden-noise-original.west) {Garden\vphantom{p}};
    \node[anchor=south,inner sep=1pt,rotate=90] (scene3) at (kitchen-noise-original.west) {Kitchen\vphantom{p}};
    
  \end{tikzpicture}
  \caption{Examples with Gaussian blur for different noise scales (original means without noise).}
  \end{subfigure}
  \begin{subfigure}[c]{\linewidth}
    \begin{tikzpicture}[image/.style = {inner sep=0, outer sep=0, minimum width=\figurewidth, anchor=north west, text width=\figurewidth}, node distance = 1pt and 1pt, every node/.style={font= {\tiny}}, label/.style = {scale=0.75,font={\tiny},anchor=south,inner sep=0pt,outer sep=2pt,rotate=0}, spy using outlines={rectangle, red, magnification=2, size=0.90cm, connect spies}] 

      \node [image] (garden-blur-original) {\includegraphics[width=\figurewidth]{figures-appendix/noisyview-mipnerf360-examples/garden/frame_00001.JPG}}; 
      \node [image,right= of garden-blur-original] (garden-blur-1) {\includegraphics[width=\figurewidth]{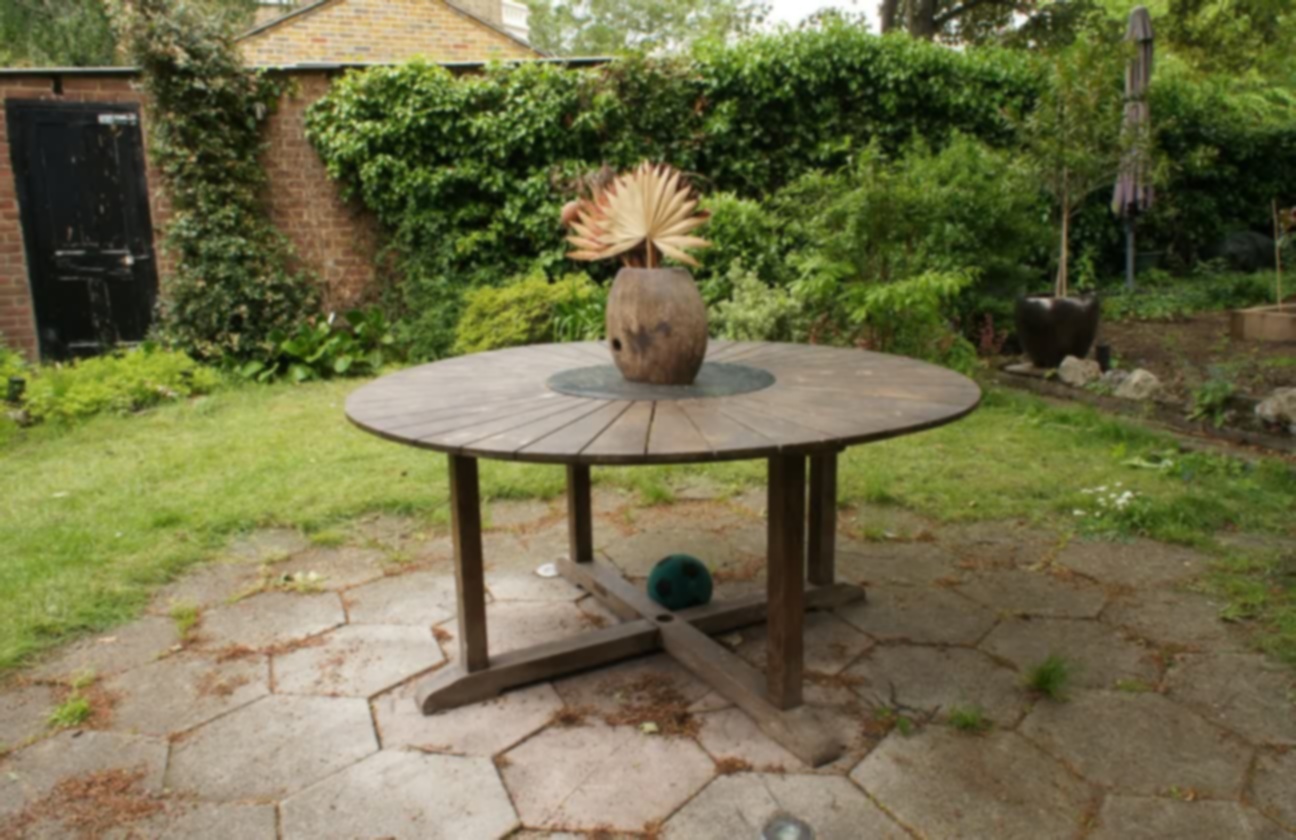}}; 
      \node [image,right=of garden-blur-1] (garden-blur-2) {\includegraphics[width=\figurewidth]{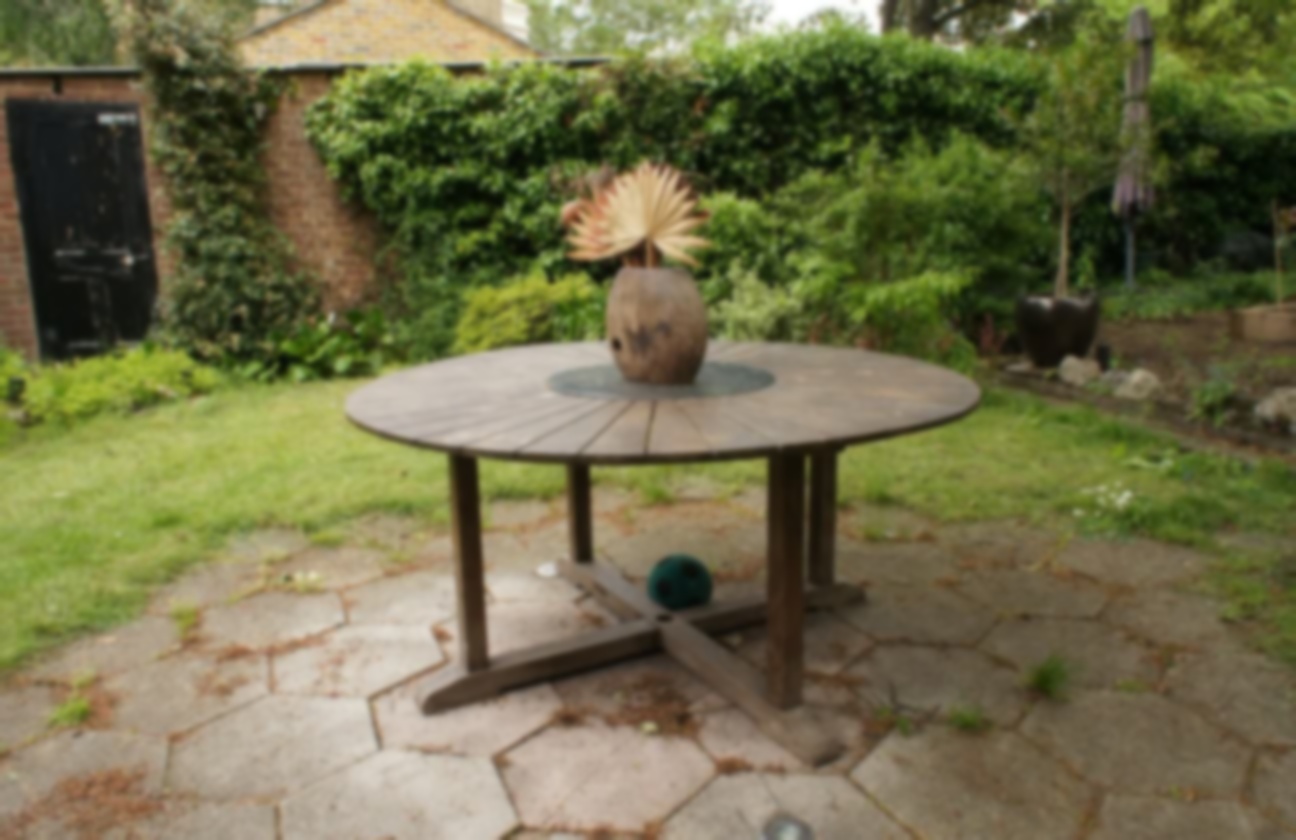}}; 
  
      \spy [red] on ($(garden-blur-original) + (-0.3,0.4)$) in node [right,below] at ($(garden-blur-original.north west) + (0.5,0.0)$);
      \spy [red] on ($(garden-blur-1) + (-0.3,0.4)$) in node [right,below] at ($(garden-blur-1.north west) + (0.5,0.0)$);
      \spy [red] on ($(garden-blur-2) + (-0.3,0.4)$) in node [right,below] at ($(garden-blur-2.north west) + (0.5,0.0)$);

      \spy [blue] on ($(garden-blur-original) + (-0.70,-0.75)$) in node [right] at ($(garden-blur-original.south west) + (0,0.5)$);
      \spy [blue] on ($(garden-blur-1) + (-0.70,-0.75)$) in node [right] at ($(garden-blur-1.south west) + (0,0.5)$);
      \spy [blue] on ($(garden-blur-2) + (-0.70,-0.75)$) in node [right] at ($(garden-blur-2.south west) + (0,0.5)$);
      \node [image,below=of garden-blur-original] (kitchen-blur-original) {\includegraphics[width=\figurewidth]{figures-main/noisyview-mipnerf360-examples/kitchen/frame_00001.JPG}}; 
      \node [image,right=of kitchen-blur-original] (kitchen-blur-1) {\includegraphics[width=\figurewidth]{figures-main/noisyview-mipnerf360-examples/kitchen/frame_00001_blur_kernelsize_7.JPG}}; 
      \node [image,right=of kitchen-blur-1] (kitchen-blur-2) {\includegraphics[width=\figurewidth]{figures-main/noisyview-mipnerf360-examples/kitchen/frame_00001_blur_kernelsize_15.JPG}}; 
  
      \spy [red] on ($(kitchen-blur-original) + (-0.0,0.8)$) in node [right,below] at ($(kitchen-blur-original.north west) + (0.5,0.0)$);
      \spy [red] on ($(kitchen-blur-1) + (-0.0,0.8)$) in node [right,below] at ($(kitchen-blur-1.north west) + (0.5,0.0)$);
      \spy [red] on ($(kitchen-blur-2) + (-0.0,0.8)$) in node [right,below] at ($(kitchen-blur-2.north west) + (0.5,0.0)$);
  
      \spy [blue] on ($(kitchen-blur-original) + (-0.8,-0.4)$) in node [right] at ($(kitchen-blur-original.south west) + (0,0.5)$);
      \spy [blue] on ($(kitchen-blur-1) + (-0.8,-0.4)$) in node [right] at ($(kitchen-blur-1.south west) + (0,0.5)$);
      \spy [blue] on ($(kitchen-blur-2) + (-0.8,-0.4)$) in node [right] at ($(kitchen-blur-2.south west) + (0,0.5)$);

      \node[label] (label5) at (garden-blur-original.north) {Original\vphantom{p}};
      \node[label] (label1) at (garden-blur-1.north) {Kernel Size 7\vphantom{p}};
      \node[label] (label2) at (garden-blur-2.north) {Kernel Size 15\vphantom{p}};

      \node[anchor=south,inner sep=1pt,rotate=90] (scene2) at (garden-noise-original.west) {Garden\vphantom{p}};
      \node[anchor=south,inner sep=1pt,rotate=90] (scene3) at (kitchen-noise-original.west) {Kitchen\vphantom{p}};
      
    \end{tikzpicture}
    \caption{Examples with Gaussian blur for different kernel sizes (original means without blur).}
    \end{subfigure}
  \caption{Aleatoric uncertainty \num{1}: Examples training views from Garden and Kitchen scenes from Mip-NeRF 360 data set with Gaussian noise or blur applied. }
  \label{fig:noisyview-mipnerf360-examples-appendix}
\end{figure}

\subsection{Experiments on Aleatoric Uncertainty (\num{1})}
\label{app:aleatoric}

\paragraph{Additional Figures} \cref{fig:noisyview-mipnerf360-examples-appendix} shows examples of a training view with applied Gaussian noise and blur from the Garden and Kitchen scenes from the Mip-NeRF 360 data set. We zoom  in on various parts of the scene to show how these effects affect the image when applying stronger noise or blurring. The Gaussian noise introduces random variations on a per-pixel basis, while fine details and sharp edges are lost when blurring the images.  

\begin{figure}[t]
  \centering
  \setlength{\figurewidth}{0.2\textwidth}
  \setlength{\figureheight}{0.18\textwidth}
  \pgfplotsset{every axis title/.append style={at={(0.5,0.80)}}} 
\pgfplotsset{every axis x label/.append style={at={(0.5,0.05)}}}
\pgfplotsset{every axis y label/.append style={at={(0.15,0.5)}}}

\begin{tikzpicture}
\tikzstyle{every node}=[font=\scriptsize]
\definecolor{crimson2143940}{RGB}{214,39,40}
\definecolor{darkgray176}{RGB}{176,176,176}
\definecolor{darkorange25512714}{RGB}{255,127,14}
\definecolor{forestgreen4416044}{RGB}{44,160,44}
\definecolor{mediumpurple148103189}{RGB}{148,103,189}
\definecolor{sienna1408675}{RGB}{140,86,75}
\definecolor{steelblue31119180}{RGB}{31,119,180}

\definecolor{lightgray204}{RGB}{204,204,204}

\begin{groupplot}[
  group style={group size= 2 by 1, horizontal sep=1.25cm, vertical sep=.5cm},
  tick align=outside,
  tick pos=left,
  grid=both,
  xlabel={Noise scale},
  xlabel={Avg. variance},
  xmin=-0.1, xmax=2.1,
  xtick={0,1,2},
  xticklabels={0,0.1,0.2},
  xtick style={color=black},
  x grid style={darkgray176,solid},
  y grid style={darkgray176,solid},
  ytick style={color=black},
  tick label style={font=\tiny}
]

\nextgroupplot[
height=\figureheight,
width=\figurewidth,
legend cell align={left},
legend columns=1,
legend style={
  nodes={scale=0.8},
  fill opacity=0.8, 
  draw opacity=1, 
  text opacity=1, 
  at={(2.65,0.0)}, 
  anchor=south west,
  draw=lightgray204},
xlabel={Noise scale},
xticklabels={0,0.1,0.2},
ymin=-0.000914076997923985, ymax=0.0194333730252058,
ytick style={color=black}
]
\addplot [thick, steelblue31119180, mark=*, mark size=1, mark options={solid}]
table {%
0 0.00296415873647978
1 0.00722684380080965
2 0.0185084889332453
};
\addlegendentry{Active-Nerfacto}
\addplot [thick, darkorange25512714, mark=*, mark size=1, mark options={solid}]
table {%
0 0.00345189025392756
1 0.00615475181904104
2 0.0169340694944064
};
\addlegendentry{Active-Splatfacto}
\addplot [thick, forestgreen4416044, mark=*, mark size=1, mark options={solid}]
table {%
0 0.000153489754464115
1 0.000145964218972949
2 0.00012634604354389
};
\addlegendentry{MC-Dropout-Nerfacto}
\addplot [thick, crimson2143940, mark=*, mark size=1, mark options={solid}]
table {%
0 2.28801544229403e-05
1 1.08070940364592e-05
2 1.35734823743405e-05
};
\addlegendentry{Laplace-NeRF}
\addplot [thick, mediumpurple148103189, mark=*, mark size=1, mark options={solid}]
table {%
0 0.000834113146993332
1 0.000809037213912234
2 0.000777921632915321
};
\addlegendentry{Ensemble-Nerfacto}
\addplot [thick, sienna1408675, mark=*, mark size=1, mark options={solid}]
table {%
0 0.00284104024467524
1 0.00481508229131578
2 0.00799068006583386
};
\addlegendentry{Ensemble-Splatfacto}

\nextgroupplot[
height=\figureheight,
width=\figurewidth,
xlabel={Kernel size},
xticklabels={0,7,15},
ymin=-0.000158942134392722, ymax=0.00362383465337138,
ytick style={color=black}
]
\addplot [thick, steelblue31119180, mark=*, mark size=1, mark options={solid}]
table {%
0 0.00296415873647978
1 0.00086590326585186
2 0.000394140645059653
};
\addplot [thick, darkorange25512714, mark=*, mark size=1, mark options={solid}]
table {%
0 0.00345189025392756
1 0.0024974697153084
2 0.00227640502594618
};
\addplot [thick, forestgreen4416044, mark=*, mark size=1, mark options={solid}]
table {%
0 0.000153489754464115
1 0.000113226341151555
2 9.27954731903608e-05
};
\addplot [thick, crimson2143940, mark=*, mark size=1, mark options={solid}]
table {%
0 2.28801544229403e-05
1 1.66323841565625e-05
2 1.30022650511011e-05
};
\addplot [thick, mediumpurple148103189, mark=*, mark size=1, mark options={solid}]
table {%
0 0.000834113146993332
1 0.000561221303845135
2 0.000435609103684934
};
\addplot [thick, sienna1408675, mark=*, mark size=1, mark options={solid}]
table {%
0 0.00284104024467524
1 0.0016524754804171
2 0.00109150490041227
};

\end{groupplot}

\end{tikzpicture} %
  \caption{Aleatoric uncertainty \num{1}: Average predicted variance over rendered pixels averaged across the Mip-NeRF 360 scenes for Gaussian noise (left) and Gaussian blur (right) applied to the training views. The variance increases when adding more noise to the input images, while it decreases when the images are blurred more.  
  }
  \label{fig:noisyview-mipnerf360-average-variance-appendix}
\end{figure}
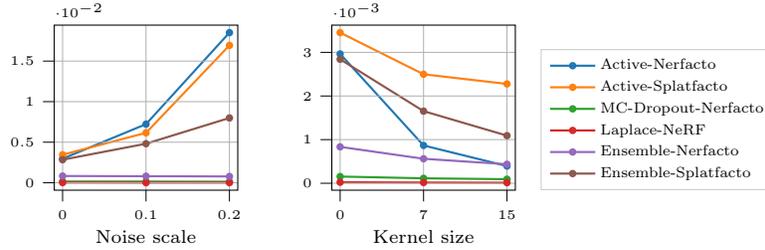

In \cref{fig:noisyview-mipnerf360-average-variance-appendix}, we show the average predicted variance across the rendered pixels for each method trained using noisy and blurred input images on the Mip-NeRF 360 scenes. The trends are that the average variance increases when adding more Gaussian noise, but decreases when adding more blur for the kernel sizes we selected. As there is less variability in the blurred images, the methods become more certain of their per-pixel predictions which reduces their average predicted variance.

\subsection{Experiments on Epistemic Uncertainty (\num{2})}
\label{app:epistemic}
In this section, we provide additional qualitative and quantitative results for the epistemic uncertainty experiments in \cref{sec:reducible}.

\begin{figure}[t]
    \centering
    \setlength{\figurewidth}{0.2\textwidth}
    \setlength{\figureheight}{0.2\textwidth}
    \input{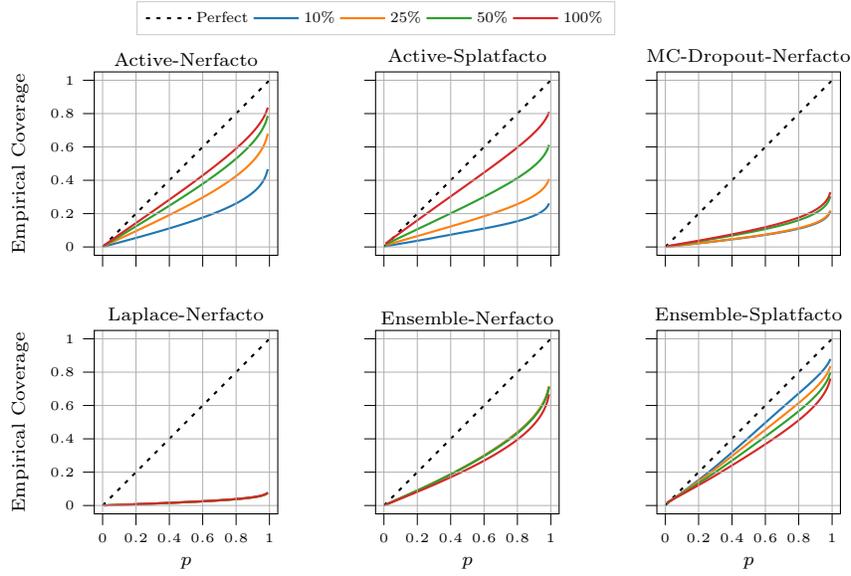} %
    \caption{Epistemic uncertainty \num{2}: Empirical coverage curves from the AUCE metric averaged across the Mip-Nerf 360 scenes for all models with different proportions of training views. %
    }
    \label{fig:varyingview-mipnerf360-auce-coverage}
\end{figure}

\begin{figure}[h!]
    \centering
    \setlength{\figurewidth}{0.2\textwidth}
    \setlength{\figureheight}{.08\textheight}
    \pgfplotsset{every axis title/.append style={at={(0.5,0.80)}}} 
\pgfplotsset{every axis x label/.append style={at={(0.5,0.05)}}}
\pgfplotsset{every axis y label/.append style={at={(0.15,0.5)}}}

\begin{tikzpicture}
\tikzstyle{every node}=[font=\scriptsize]
\definecolor{crimson2143940}{RGB}{214,39,40}
\definecolor{darkgray176}{RGB}{176,176,176}
\definecolor{darkorange25512714}{RGB}{255,127,14}
\definecolor{forestgreen4416044}{RGB}{44,160,44}
\definecolor{lightgray204}{RGB}{204,204,204}
\definecolor{mediumpurple148103189}{RGB}{148,103,189}
\definecolor{sienna1408675}{RGB}{140,86,75}
\definecolor{steelblue31119180}{RGB}{31,119,180}

\begin{groupplot}[
  group style={group size= 3 by 2, horizontal sep=1.25cm, vertical sep=.5cm},
  tick align=outside,
  tick pos=left,
  xlabel={\#Training views},
  grid=both,
  xmin=-0.1, xmax=3.1,
  xtick={0,1,2,3},
  xticklabels={10,25,50,100},
  xtick style={color=black},
  x grid style={darkgray176,solid},
  y grid style={darkgray176,solid},
  ytick style={color=black},
  tick label style={font=\tiny}
]

\nextgroupplot[
height=\figureheight,
width=\figurewidth,
legend cell align={left},
legend columns=3,
legend style={
  nodes={scale=0.8},
  fill opacity=0.8, 
  draw opacity=1, 
  text opacity=1, 
  at={(0.03,1.27)}, 
  anchor=south west,
  draw=lightgray204},
xlabel=\empty,
xticklabel=\empty,
ylabel={PSNR~$\rightarrow$},
ymin=9.72981764674187, ymax=33.5846515715122,
]
\addplot [thick, steelblue31119180, mark=*, mark size=1, mark options={solid}]
table {%
0 16.6636991202831
1 18.9260800778866
2 22.089431732893
3 25.8913583755493
};
\addlegendentry{Active-Nerfacto}
\addplot [thick, darkorange25512714, mark=*, mark size=1, mark options={solid}]
table {%
0 15.0339905023575
1 23.0020382404327
2 28.7999484539032
3 32.2375695705414
};
\addlegendentry{Active-Splatfacto}
\addplot [thick, forestgreen4416044, mark=*, mark size=1, mark options={solid}]
table {%
0 11.6420383453369
1 16.939617395401
2 25.5751316547394
3 24.1629854440689
};
\addlegendentry{MC-Dropout-Nerfacto}
\addplot [thick, crimson2143940, mark=*, mark size=1, mark options={solid}]
table {%
0 10.814128279686
1 14.834463596344
2 16.2295293807983
3 18.8653793334961
};
\addlegendentry{Laplace-Nerfacto}
\addplot [thick, mediumpurple148103189, mark=*, mark size=1, mark options={solid}]
table {%
0 14.4432364702225
1 19.2040857076645
2 22.7264596223831
3 26.2551016807556
};
\addlegendentry{Ensemble-Nerfacto}
\addplot [thick, sienna1408675, mark=*, mark size=1, mark options={solid}]
table {%
0 14.8397352695465
1 23.8710536956787
2 28.9771358966827
3 32.5003409385681
};
\addlegendentry{Ensemble-Splatfacto}

\nextgroupplot[
height=\figureheight,
width=\figurewidth,
ylabel={SSIM~$\rightarrow$},
xlabel=\empty,
xticklabel=\empty,
ymin=0.674020602554083, ymax=0.98147541359067,
]
\addplot [thick, steelblue31119180, mark=*, mark size=1, mark options={solid}]
table {%
0 0.738683849398512
1 0.774218402628321
2 0.802094682876486
3 0.926547989249229
};
\addplot [thick, darkorange25512714, mark=*, mark size=1, mark options={solid}]
table {%
0 0.79322224855423
1 0.897723346948624
2 0.942174151539803
3 0.962310820817947
};
\addplot [thick, forestgreen4416044, mark=*, mark size=1, mark options={solid}]
table {%
0 0.687995821237564
1 0.83484023809433
2 0.916148163378239
3 0.913577787578106
};
\addplot [thick, crimson2143940, mark=*, mark size=1, mark options={solid}]
table {%
0 0.720224343240261
1 0.810824871063232
2 0.841312795877457
3 0.863632023334503
};
\addplot [thick, mediumpurple148103189, mark=*, mark size=1, mark options={solid}]
table {%
0 0.801753900945187
1 0.884820304811001
2 0.915272228419781
3 0.932250469923019
};
\addplot [thick, sienna1408675, mark=*, mark size=1, mark options={solid}]
table {%
0 0.765757396817207
1 0.906390659511089
2 0.949835740029812
3 0.967500194907188
};

\nextgroupplot[
height=\figureheight,
width=\figurewidth,
xlabel=\empty,
xticklabel=\empty,
ylabel={$\leftarrow$~LPIPS},
ymin=-0.000818198791239413, ymax=0.5783532019821,
]
\addplot [thick, steelblue31119180, mark=*, mark size=1, mark options={solid}]
table {%
0 0.279985062777996
1 0.224961265455931
2 0.176664152182639
3 0.0700303260236979
};
\addplot [thick, darkorange25512714, mark=*, mark size=1, mark options={solid}]
table {%
0 0.279799371957779
1 0.0900458237156272
2 0.0389212146401405
3 0.0255077739711851
};
\addplot [thick, forestgreen4416044, mark=*, mark size=1, mark options={solid}]
table {%
0 0.461992550641298
1 0.234630200080574
2 0.0740542337298393
3 0.114078081445768
};
\addplot [thick, crimson2143940, mark=*, mark size=1, mark options={solid}]
table {%
0 0.552027229219675
1 0.428656270727515
2 0.387025682255626
3 0.34832024667412
};
\addplot [thick, mediumpurple148103189, mark=*, mark size=1, mark options={solid}]
table {%
0 0.307248320430517
1 0.137690205127001
2 0.0847697923891246
3 0.071297108894214
};
\addplot [thick, sienna1408675, mark=*, mark size=1, mark options={solid}]
table {%
0 0.368123140186071
1 0.119201406370848
2 0.0535014395136386
3 0.032853213720955
};

\nextgroupplot[
height=\figureheight,
width=\figurewidth,
ylabel={$\leftarrow$~NLL},
ymin=-5.23879613287281, ymax=59.8187038495904,
]
\addplot [thick, steelblue31119180, mark=*, mark size=1, mark options={solid}]
table {%
0 56.8615447594784
1 53.660405933857
2 53.0643546283245
3 -1.98511311411858
};
\addplot [thick, darkorange25512714, mark=*, mark size=1, mark options={solid}]
table {%
0 3.2326698154211
1 -0.116794333793223
2 -1.78987236320972
3 -2.28163704276085
};
\addplot [thick, forestgreen4416044, mark=*, mark size=1, mark options={solid}]
table {%
0 38.5643103122711
1 18.654163941741
2 -0.128537941724062
3 5.01765625923872
};
\addplot [thick, crimson2143940, mark=*, mark size=1, mark options={solid}]
table {%
0 31.9133546706289
1 11.6237800064264
2 3.92043653363362
3 4.54360410497975
};
\addplot [thick, mediumpurple148103189, mark=*, mark size=1, mark options={solid}]
table {%
0 11.4016357064247
1 3.1857741791755
2 -0.856360723730177
3 0.51405768096447
};
\addplot [thick, sienna1408675, mark=*, mark size=1, mark options={solid}]
table {%
0 8.49309593439102
1 0.0146828480064869
2 -1.71714656054974
3 -2.18524698913097
};

\nextgroupplot[
height=\figureheight,
width=\figurewidth,
ylabel={$\leftarrow$~AUSE},
ymin=0.0170564089668915, ymax=0.54449402524624,
]
\addplot [thick, steelblue31119180, mark=*, mark size=1, mark options={solid}]
table {%
0 0.122965187183581
1 0.0886776973493397
2 0.0917113258037716
3 0.0451249424368143
};
\addplot [thick, darkorange25512714, mark=*, mark size=1, mark options={solid}]
table {%
0 0.266199482604861
1 0.140032486990094
2 0.0793280289508402
3 0.0410308460704982
};
\addplot [thick, forestgreen4416044, mark=*, mark size=1, mark options={solid}]
table {%
0 0.520519588142633
1 0.400198087096214
2 0.0863040359690785
3 0.260876159183681
};
\addplot [thick, crimson2143940, mark=*, mark size=1, mark options={solid}]
table {%
0 0.45965345390141
1 0.406226929277182
2 0.363700481131673
3 0.342410831712186
};
\addplot [thick, mediumpurple148103189, mark=*, mark size=1, mark options={solid}]
table {%
0 0.354431057348847
1 0.171594938728958
2 0.0688917730003595
3 0.120604276889935
};
\addplot [thick, sienna1408675, mark=*, mark size=1, mark options={solid}]
table {%
0 0.267936123535037
1 0.127147163264453
2 0.079321458702907
3 0.0651468141004443
};

\nextgroupplot[
height=\figureheight,
width=\figurewidth,
ylabel={$\leftarrow$~AUCE},
ymin=0.136782439513428, ymax=0.370215757639893,
]
\addplot [thick, steelblue31119180, mark=*, mark size=1, mark options={solid}]
table {%
0 0.330646223636068
1 0.347224324169922
2 0.359605152270508
3 0.345673597783203
};
\addplot [thick, darkorange25512714, mark=*, mark size=1, mark options={solid}]
table {%
0 0.226187806575521
1 0.278296448789062
2 0.31322091735026
3 0.349556705966797
};
\addplot [thick, forestgreen4416044, mark=*, mark size=1, mark options={solid}]
table {%
0 0.260252238450521
1 0.271074628847656
2 0.293982694659831
3 0.292683799899089
};
\addplot [thick, crimson2143940, mark=*, mark size=1, mark options={solid}]
table {%
0 0.211864731246745
1 0.220710273990885
2 0.238496600764974
3 0.257283086082357
};
\addplot [thick, mediumpurple148103189, mark=*, mark size=1, mark options={solid}]
table {%
0 0.147393044882813
1 0.200832438077799
2 0.215983462286784
3 0.243144213616536
};
\addplot [thick, sienna1408675, mark=*, mark size=1, mark options={solid}]
table {%
0 0.277905577958984
1 0.264977445817057
2 0.289095819967448
3 0.296232796582031
};

\end{groupplot}

\end{tikzpicture} %
    \caption{Epistemic uncertainty \num{2}: Including more training views (sampled uniformly) improves test metrics. The metrics over varying number of training views averaged across scenes from the Blender data set. 
    }
    \label{fig:varyingview-blender}
\end{figure}
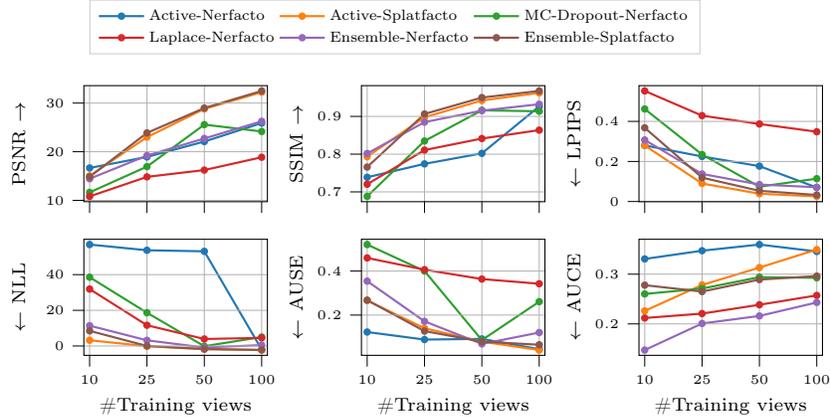

\paragraph{Mip-NeRF 360 AUCE Coverage} 
In \cref{fig:varyingview-mipnerf360-auce-coverage}, we show the empirical coverage urves used for computing the AUCE metric in \cref{fig:varyingview-mipnerf360}. All methods under-estimate their uncertainties as their curves are below the black dashed line for `perfect' calibration. We see that the coverage for Active-Nerfacto and Active-Splatfacto achieve better calibrated uncertainties when more views are added seen by the curves approaching the `perfect' line. MC-Dropout-Nerfacto also improves its calibration but slower than the Active methods, while the improvement is tiny for Laplace-Nerfacto. 
Surprisingly, the Ensemble methods achieve slightly worse calibration when more training viewss are added, which indicates that a fewer target pixels fall within their corresponding prediction intervals when the confidence level $p$ increases. However, as their uncertainties correlate well with the errors seen on the AUSE metric (\cref{fig:varyingview-mipnerf360}), the uncertainties may still be useful although they are under-estimated.

\paragraph{Blender Results} \cref{fig:varyingview-blender} shows the performance metrics over varying number of training views averaged over all scenes in Blender. We observe that Active-Splatfacto and Ensemble-Splatfacto achieves the highest image quality between the methods in general. Interestingly, the Splatfacto methods trained with 25 views performs similarly as Active-Nerfacto and Ensemble-Nerfacto trained with all views. However, the Splatfacto methods performs slightly worse than their Nerfacto counterparts when trained with only 10 views, which tells that GS-based methods need views with good coverage of the scene to perform well. 

For uncertainty estimation, we observe that Active-Nerfacto achieves the best AUSE score for low number of training views. The AUSE score for the Ensemble methods and Active-Splatfacto decreases performs on par with Active-Nerfacto when adding more views, which demonstrates that the quality of the uncertainty maps improve when adding more views as expected. Furthermore, we observe a trend that the AUCE scores increase for each method when the number of training views increases. This might be due to the white background in the Blender scenes which causes a large proportion of pixels to be covered in small prediction intervals. 

\begin{figure}[t]
    \centering
    \setlength{\figurewidth}{0.136\textwidth}
    \begin{tikzpicture}[image/.style = {inner sep=0, outer sep=0, minimum width=\figurewidth, anchor=north west, text width=\figurewidth}, node distance = 1pt and 1pt, every node/.style={font= {\tiny}}, label/.style = {scale=0.75,font={\tiny},anchor=south,inner sep=0pt,outer sep=2pt,rotate=0}] 

      \node [image] (kitchen-1) {\includegraphics[width=\figurewidth]{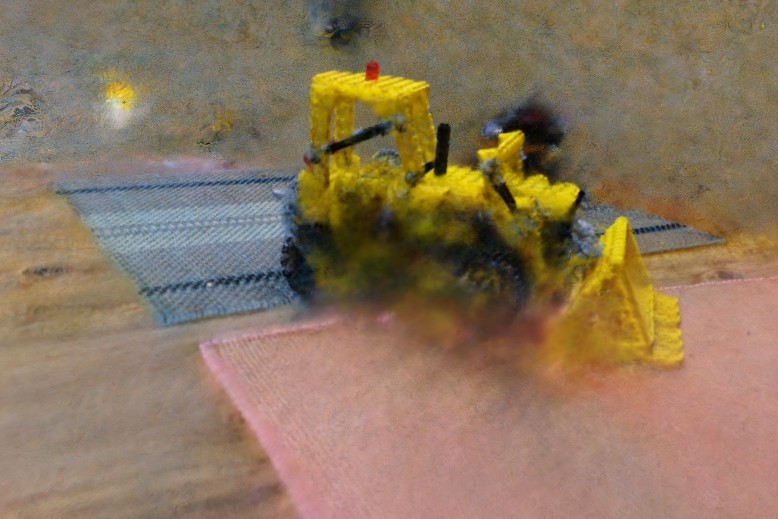}};%
      \node [image,right=of kitchen-1] (kitchen-2) {\includegraphics[width=\figurewidth]{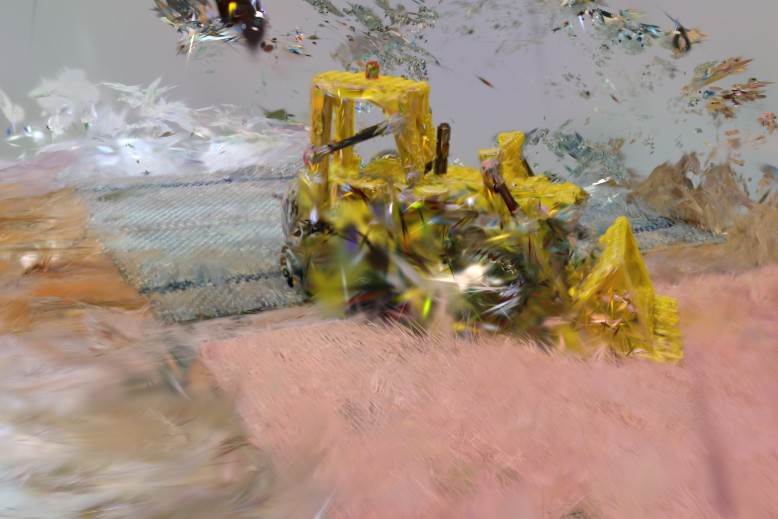}};
      \node [image,right=of kitchen-2] (kitchen-3) {\includegraphics[width=\figurewidth]{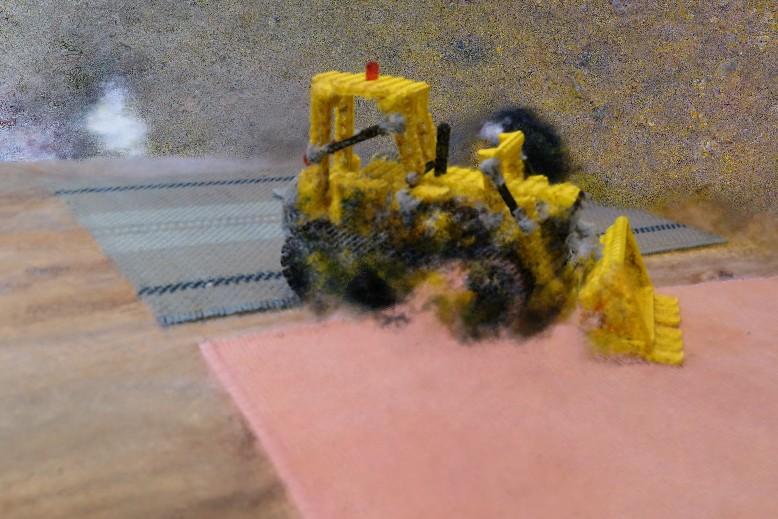}};
      \node [image,right=of kitchen-3] (kitchen-4) {\includegraphics[width=\figurewidth]{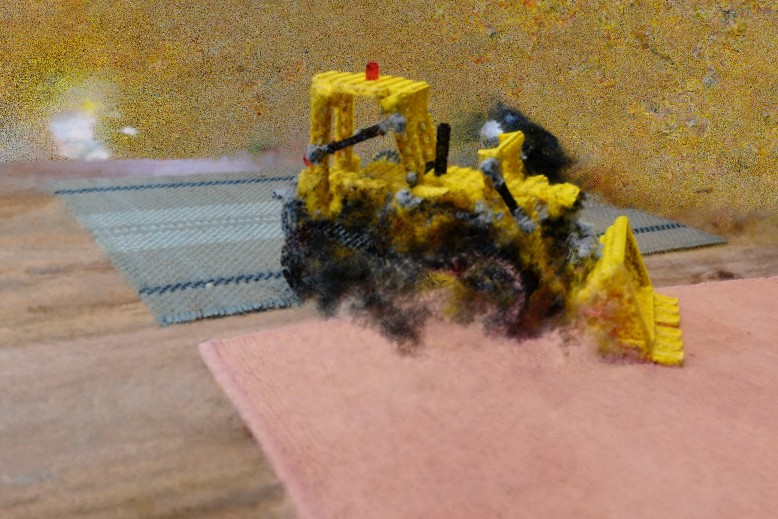}};
      \node [image,right=of kitchen-4] (kitchen-5) {\includegraphics[width=\figurewidth]{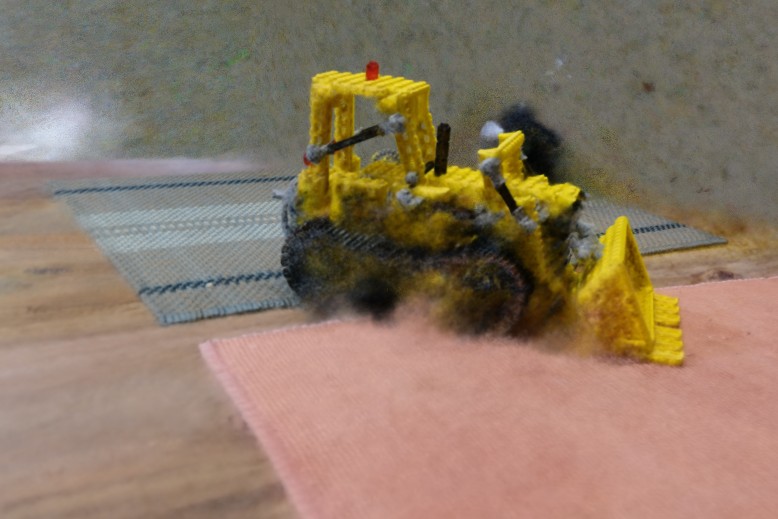}};
      \node [image,right=of kitchen-5] (kitchen-6) {\includegraphics[width=\figurewidth]{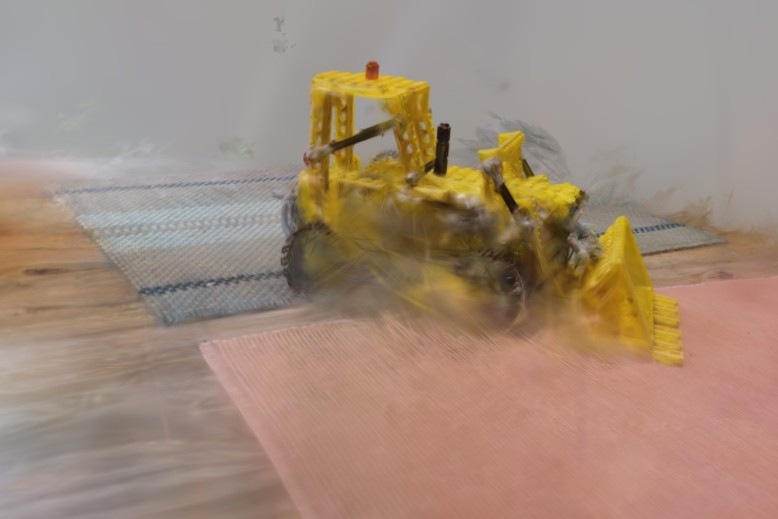}};
      \node [image,right=of kitchen-6] (kitchen-gt) {\includegraphics[width=\figurewidth]{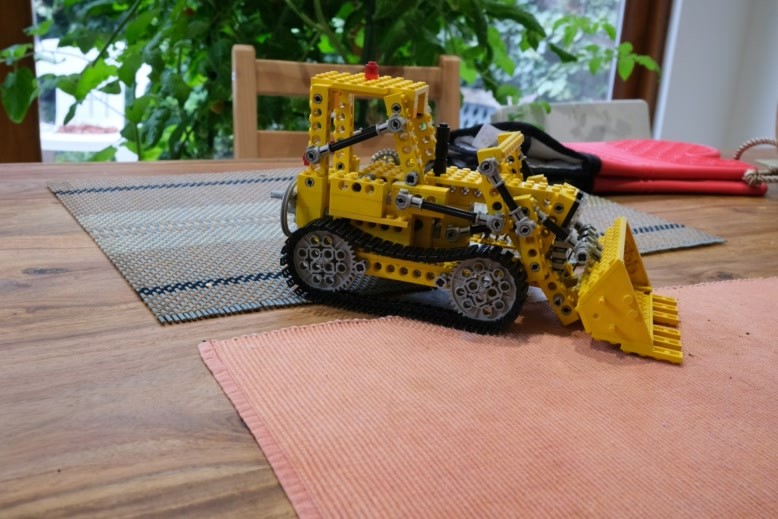}};
      
      \node[image,below=of kitchen-1] (kitchen-unc-1) {\includegraphics[width=\figurewidth]{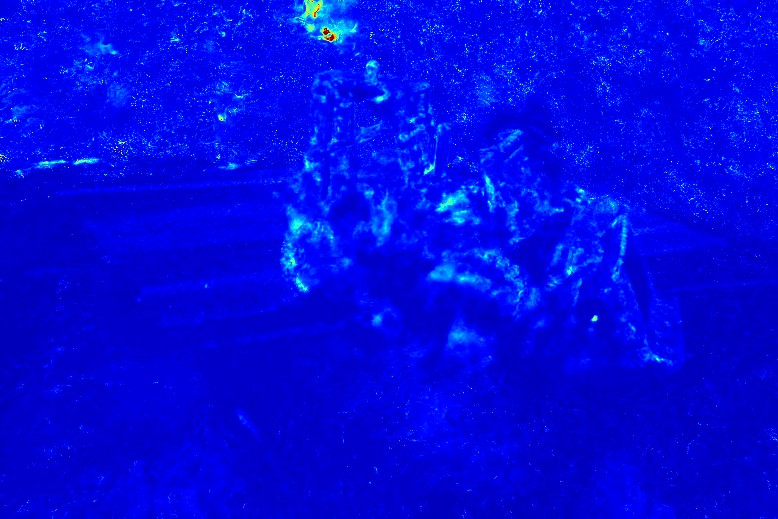}};
      \node[image,right=of kitchen-unc-1] (kitchen-unc-2) {\includegraphics[width=\figurewidth]{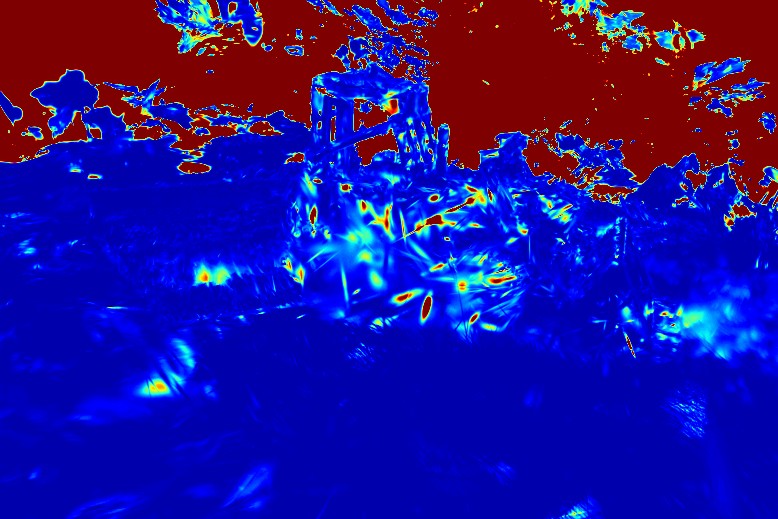}};
      \node[image,right=of kitchen-unc-2] (kitchen-unc-3) {\includegraphics[width=\figurewidth]{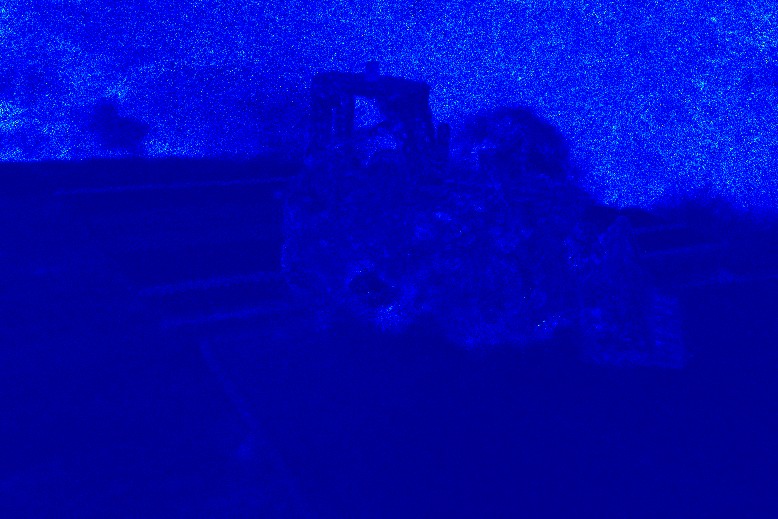}};
      \node[image,right=of kitchen-unc-3] (kitchen-unc-4) {\includegraphics[width=\figurewidth]{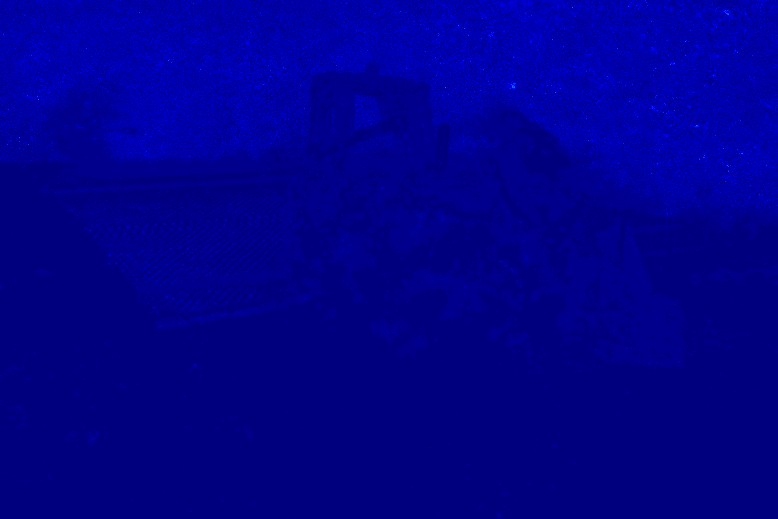}};
      \node[image,right=of kitchen-unc-4] (kitchen-unc-5) {\includegraphics[width=\figurewidth]{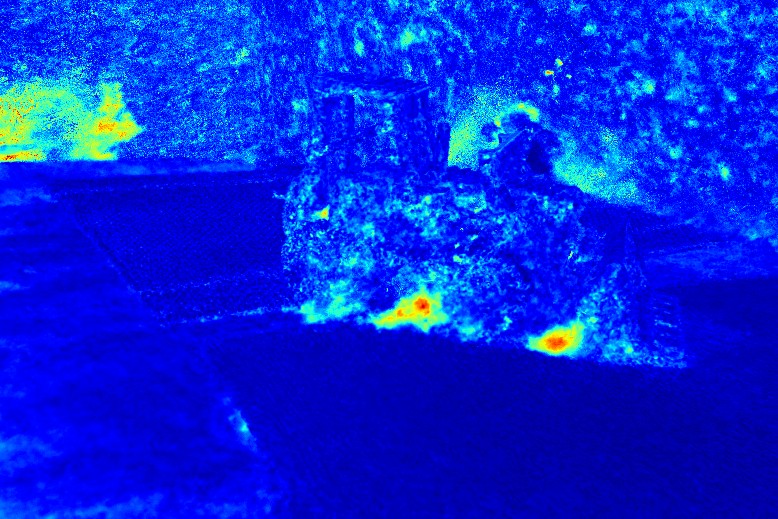}};
      \node[image,right=of kitchen-unc-5] (kitchen-unc-6) {\includegraphics[width=\figurewidth]{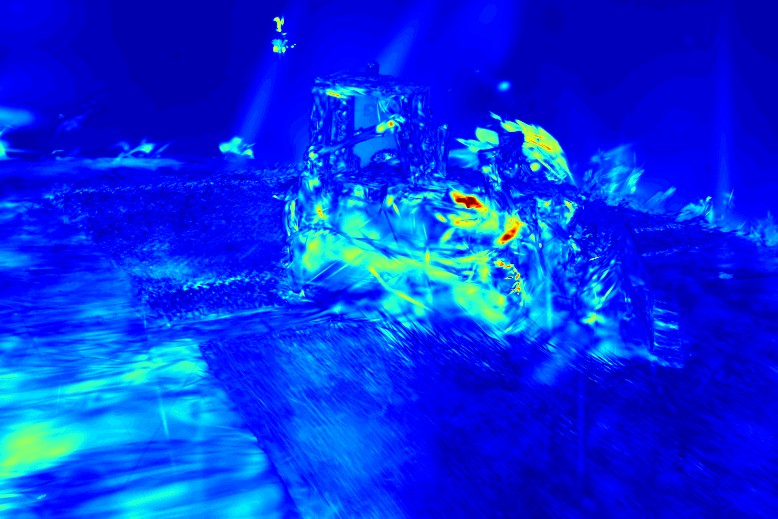}};

      \node [image,below=of kitchen-unc-1] (room-1) {\includegraphics[width=\figurewidth]{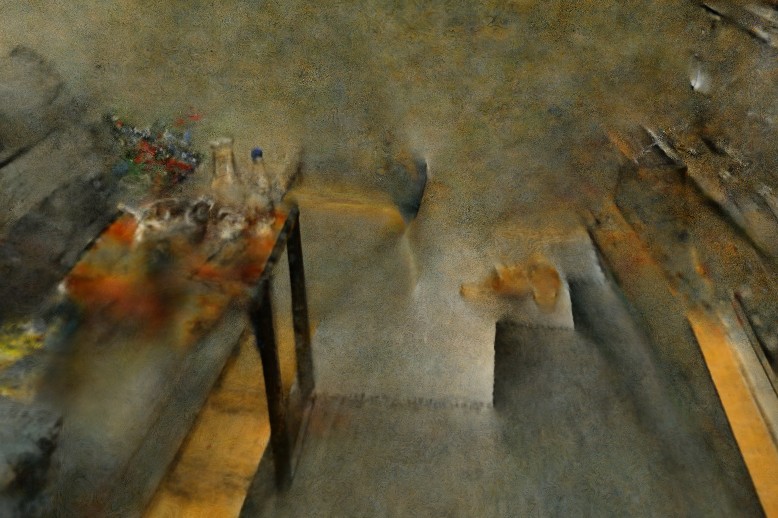}};
      \node [image,right=of room-1] (room-2) {\includegraphics[width=\figurewidth]{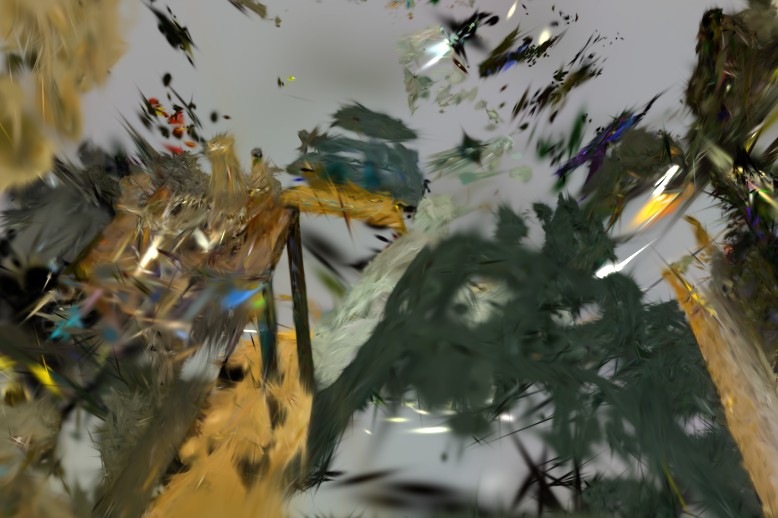}};
      \node [image,right=of room-2] (room-3) {\includegraphics[width=\figurewidth]{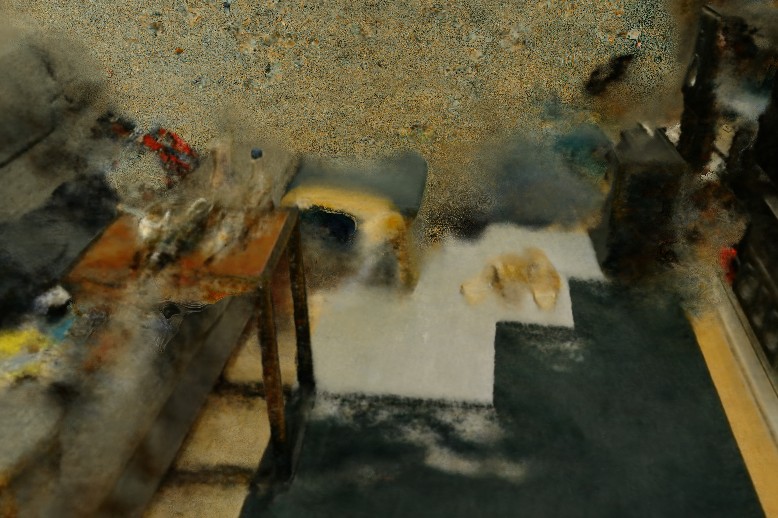}};
      \node [image,right=of room-3] (room-4) {\includegraphics[width=\figurewidth]{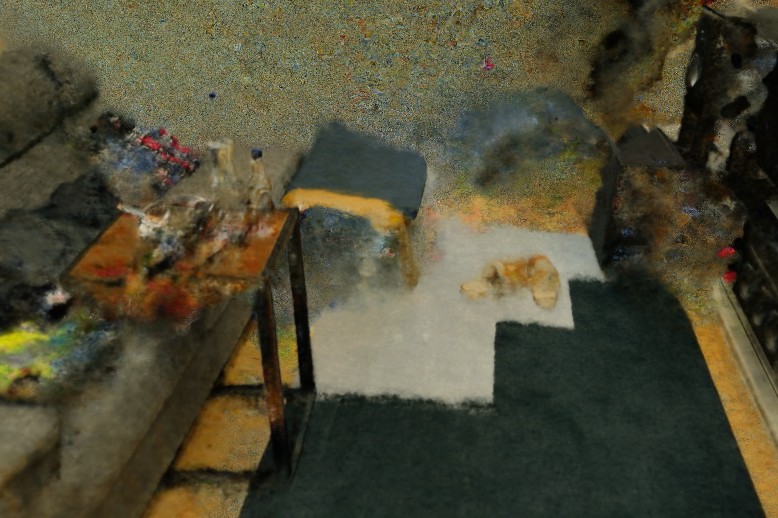}};
      \node [image,right=of room-4] (room-5) {\includegraphics[width=\figurewidth]{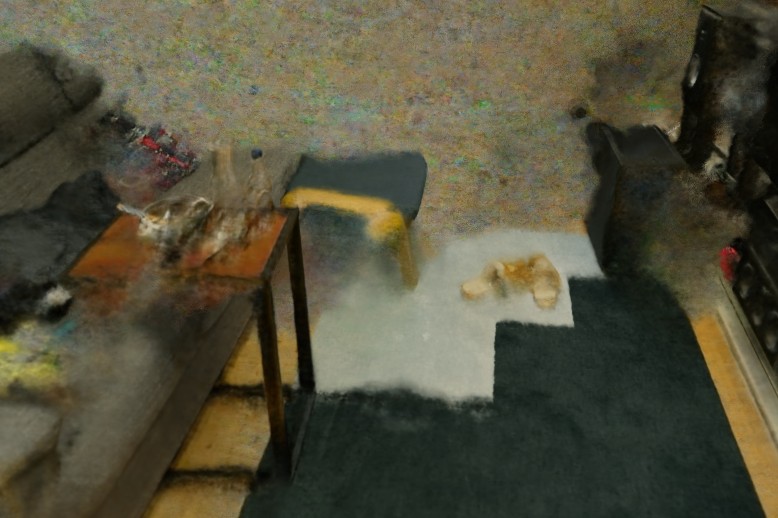}};
      \node [image,right=of room-5] (room-6) {\includegraphics[width=\figurewidth]{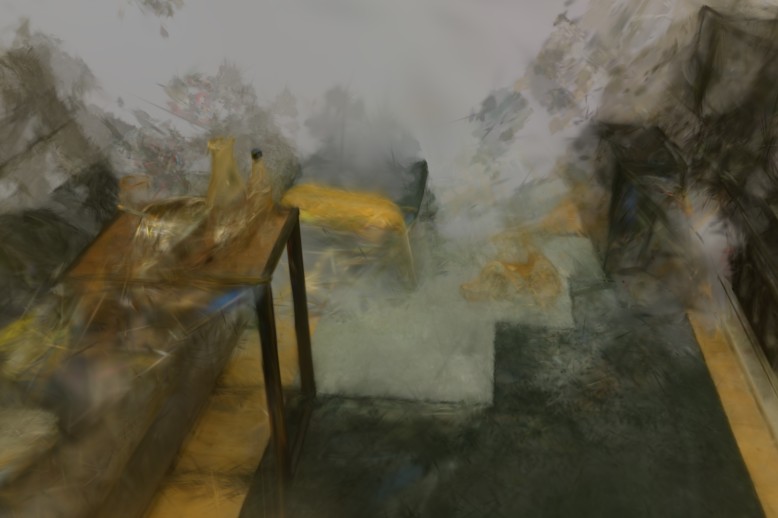}};
      \node [image,right=of room-6] (room-gt) {\includegraphics[width=\figurewidth]{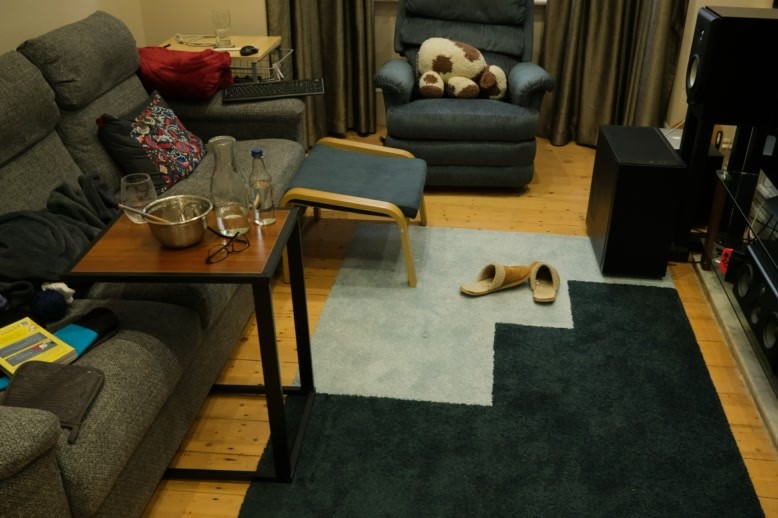}};
      
      \node[image,below=of room-1] (room-unc-1) {\includegraphics[width=\figurewidth]{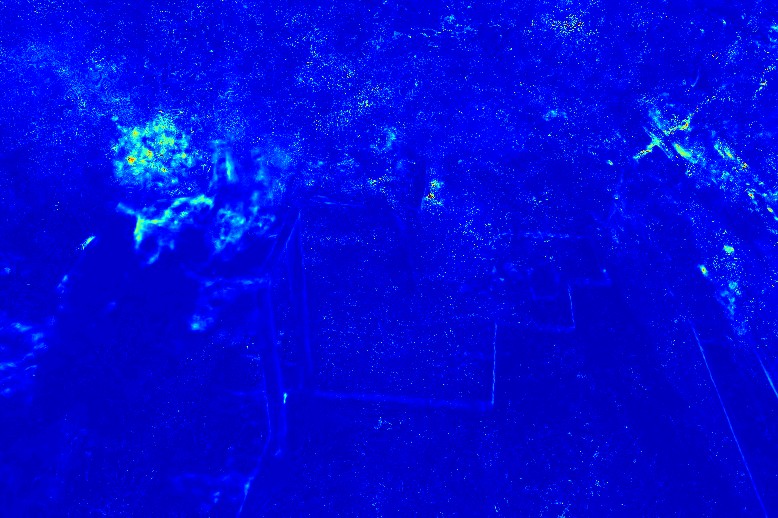}};
      \node[image,right=of room-unc-1] (room-unc-2) {\includegraphics[width=\figurewidth]{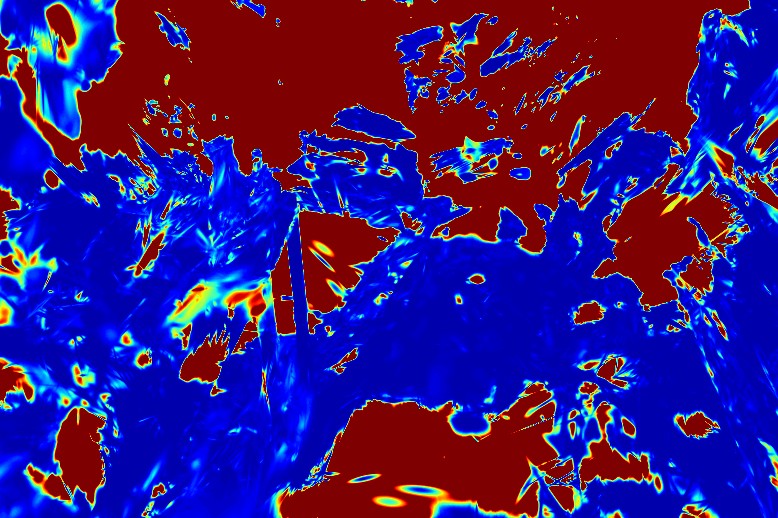}};
      \node[image,right=of room-unc-2] (room-unc-3) {\includegraphics[width=\figurewidth]{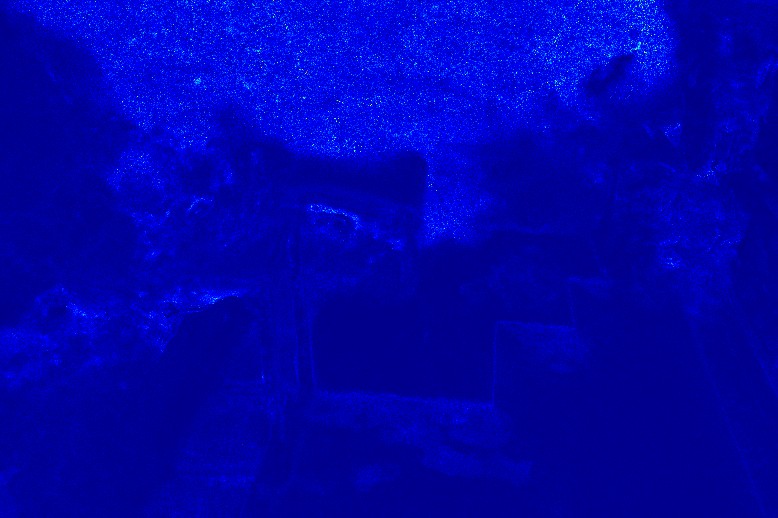}};
      \node[image,right=of room-unc-3] (room-unc-4) {\includegraphics[width=\figurewidth]{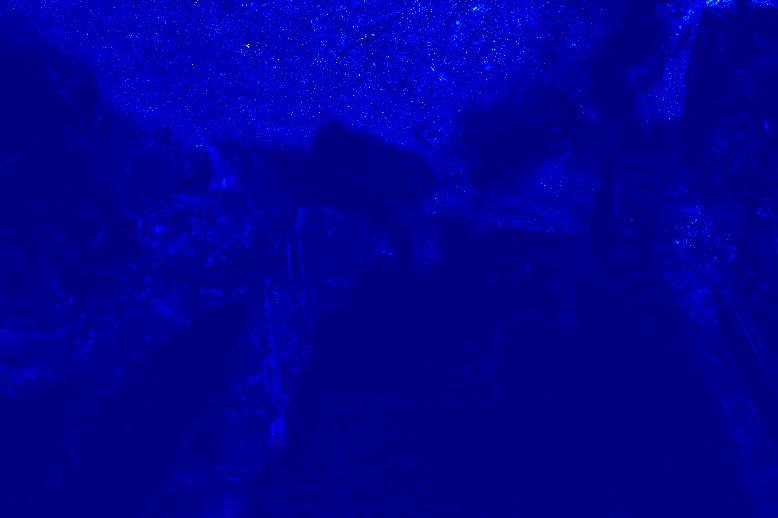}};
      \node[image,right=of room-unc-4] (room-unc-5) {\includegraphics[width=\figurewidth]{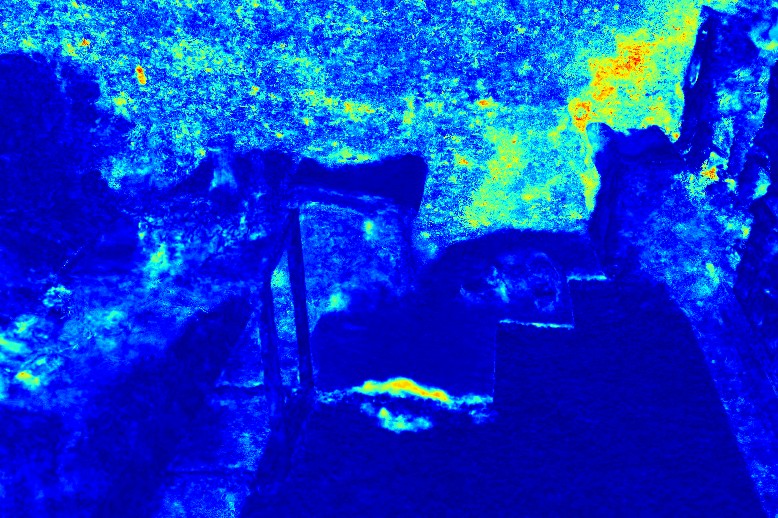}};
      \node[image,right=of room-unc-5] (room-unc-6) {\includegraphics[width=\figurewidth]{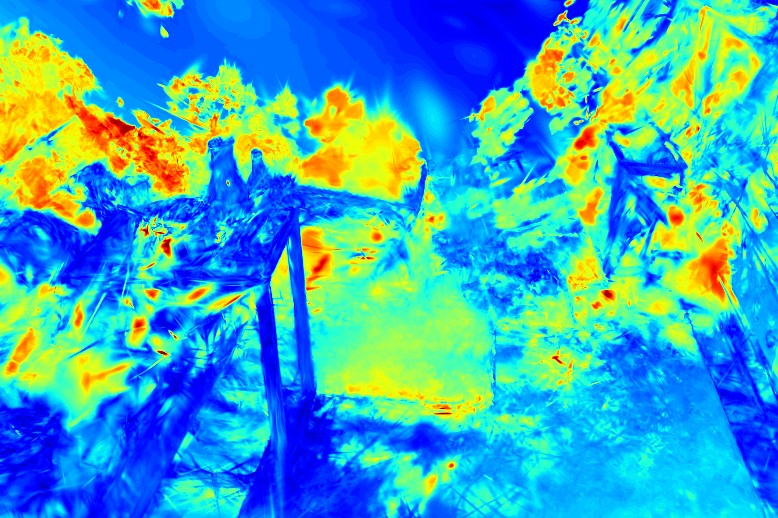}};

      \node[label] (label1) at (kitchen-1.north) {Active-Nerfacto\vphantom{p}};
      \node[label] (label2) at (kitchen-2.north) {Active-Splatfacto\vphantom{p}};
      \node[label] (label3) at (kitchen-3.north) {MC-Dropout-Nerfacto\vphantom{p}};
      \node[label] (label4) at (kitchen-4.north) {Laplace-Nerfacto\vphantom{p}};
      \node[label] (label5) at (kitchen-5.north) {Ensemble-Nerfacto\vphantom{p}};
      \node[label] (label6) at (kitchen-6.north) {Ensemble-Splatfacto\vphantom{p}};
      \node[label] (label-gt) at (kitchen-gt.north) {Ground Truth\vphantom{p}};
      \node[anchor=south,inner sep=1pt,rotate=90] (scene3) at (kitchen-1.south west) {Kitchen\vphantom{p}};
      \node[anchor=south,inner sep=1pt,rotate=90] (scene4) at (room-1.south west) {Room\vphantom{p}};
      
      \end{tikzpicture}
    \caption{Epistemic uncertainty \num{2}: Rendered RGB and uncertainty for test views from Mip-NeRF 360 scenes next to the ground truth RGB. The uncertainty is visualized by the standard deviation (0.0~\protect\includegraphics[width=3em,height=.7em]{figures-main/jet.png}~0.3). Extended version of \cref{fig:ood-mipnerf360-qualitative}.}
    \label{fig:ood-mipnerf360-qualitative-appendix}
\end{figure}

\paragraph{Mip-NeRF 360 Results} In
\cref{fig:ood-mipnerf360-qualitative-appendix}, we extend \cref{fig:ood-mipnerf360-qualitative} by providing additional qualitative results on scenes from Mip-NeRF 360 for the rendered RGB values and their predicted uncertainty using different models. The behaviour of the models on the additional scenes is consistent with the scenes Bicycle and Garden presented also in the main text. 

\begin{table}[t]
\centering
\caption{Epistemic uncertainty \num{2}: Performance metrics in the OOD setting on the Blender data set. The \first{first}, \second{second}, and \third{third} values are highlighted.}
\resizebox{0.8\textwidth}{!}{
\begin{tabular}{lcccccc}
\toprule
Method & PSNR ($\uparrow$) & SSIM ($\uparrow$) & LPIPS ($\downarrow$) & NLL ($\downarrow$) & AUSE ($\downarrow$) & AUCE ($\downarrow$) \\
\midrule
Active-Nerfacto & 14.96 & 0.74 & 0.31 & 53.33 & \cellcolor{red!25}0.09 & 0.34 \\ 
Active-Splatfacto & 13.53 & 0.82 & 0.24 & \cellcolor{orange!25}2.49 & \cellcolor{orange!25}0.11 & \cellcolor{red!25}0.22 \\ 
MC-Dropout-Nerfacto & \cellcolor{orange!25}18.68 & \cellcolor{yellow!25}0.83 & \cellcolor{orange!25}0.20 & 6.97 & \cellcolor{yellow!25}0.26 & \cellcolor{yellow!25}0.29 \\ 
Laplace-Nerfacto & \cellcolor{yellow!25}16.04 & 0.79 & 0.43 & \cellcolor{yellow!25}3.95 & 0.34 & \cellcolor{orange!25}0.26 \\ 
Ensemble-Nerfacto & \cellcolor{red!25}19.44 & \cellcolor{red!25}0.86 & \cellcolor{red!25}0.16 & \cellcolor{red!25}-0.36 & \cellcolor{red!25}0.09 & \cellcolor{red!25}0.22 \\ 
Ensemble-Splatfacto & 13.78 & \cellcolor{orange!25}0.85 & \cellcolor{yellow!25}0.23 & 31.42 & 0.37 & 0.32 \\
\bottomrule
\end{tabular}
}
\label{tab:ood-blender}
\end{table}

\begin{figure}[h]
    \centering
    \setlength{\figurewidth}{0.136\textwidth}
    \begin{tikzpicture}[image/.style = {inner sep=0, outer sep=0, minimum width=\figurewidth, anchor=north west, text width=\figurewidth}, node distance = 1pt and 1pt, every node/.style={font= {\tiny}}, label/.style = {scale=0.75,font={\tiny},anchor=south,inner sep=0pt,outer sep=2pt,rotate=0}] 
      \node [image] (frame1) {\includegraphics[width=\figurewidth]{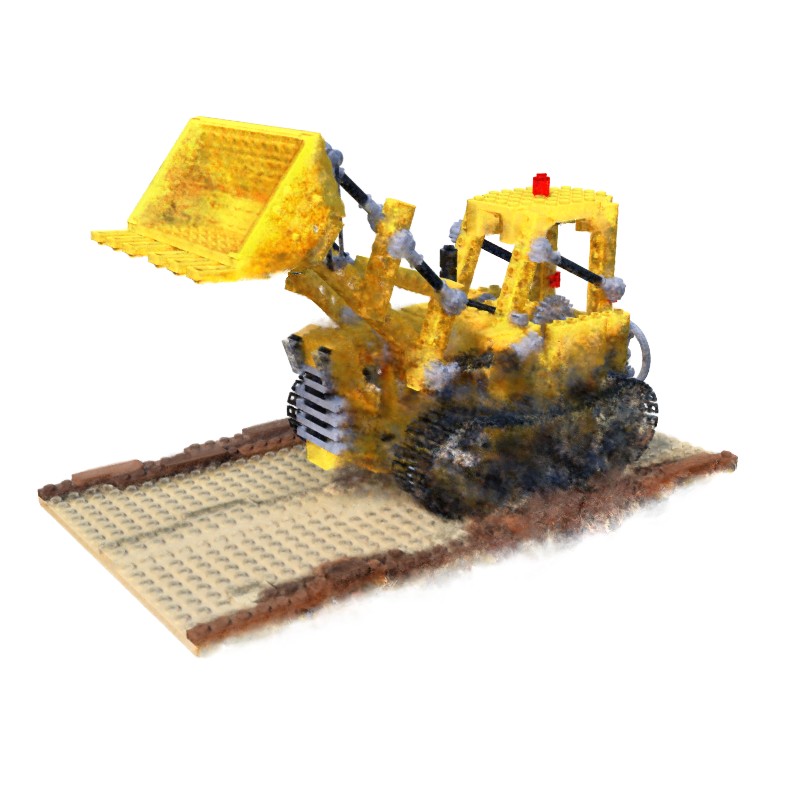}};
      \node [image,right=of frame1] (frame2) {\includegraphics[width=\figurewidth]{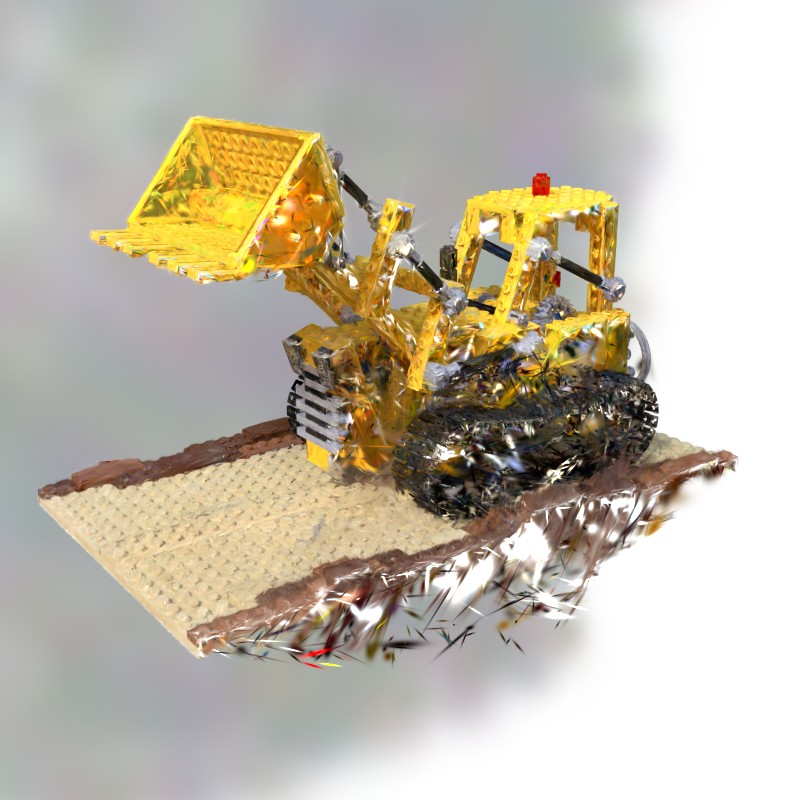}};
      \node [image,right=of frame2] (frame3) {\includegraphics[width=\figurewidth]{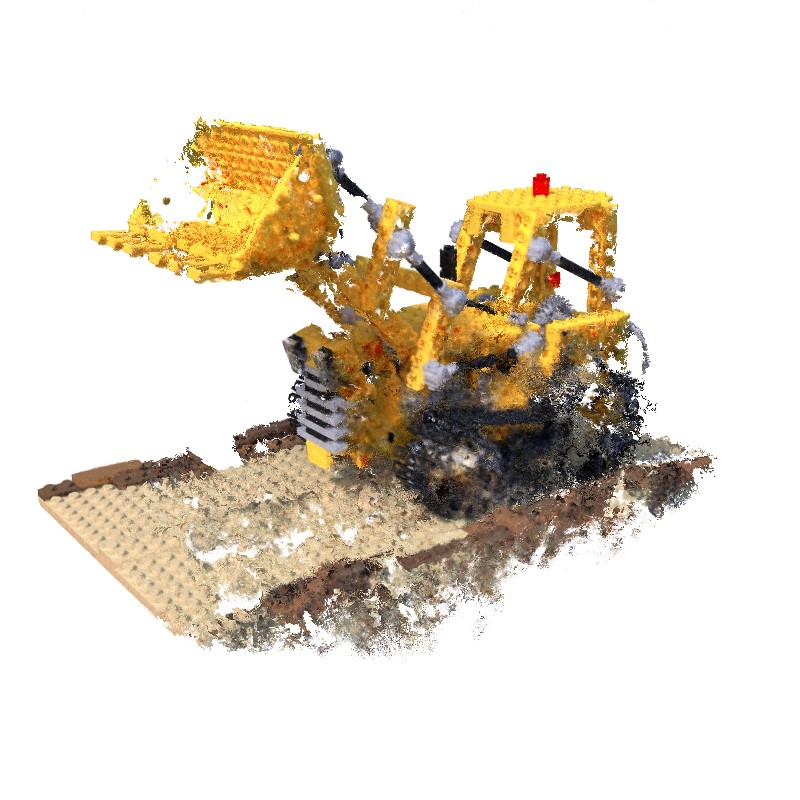}};
      \node [image,right=of frame3] (frame4) {\includegraphics[width=\figurewidth]{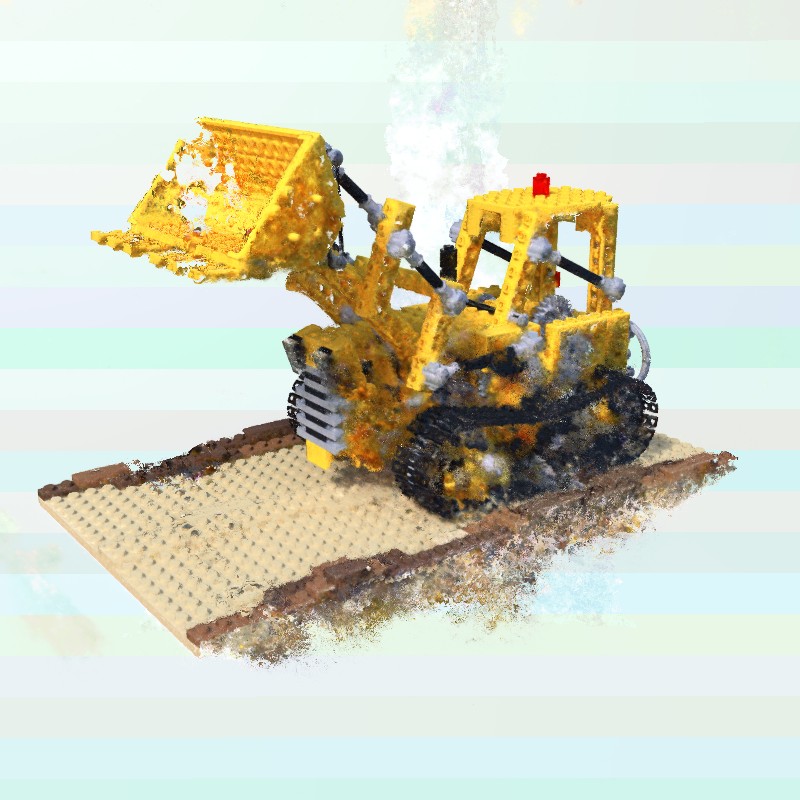}};
      \node [image,right=of frame4] (frame5) {\includegraphics[width=\figurewidth]{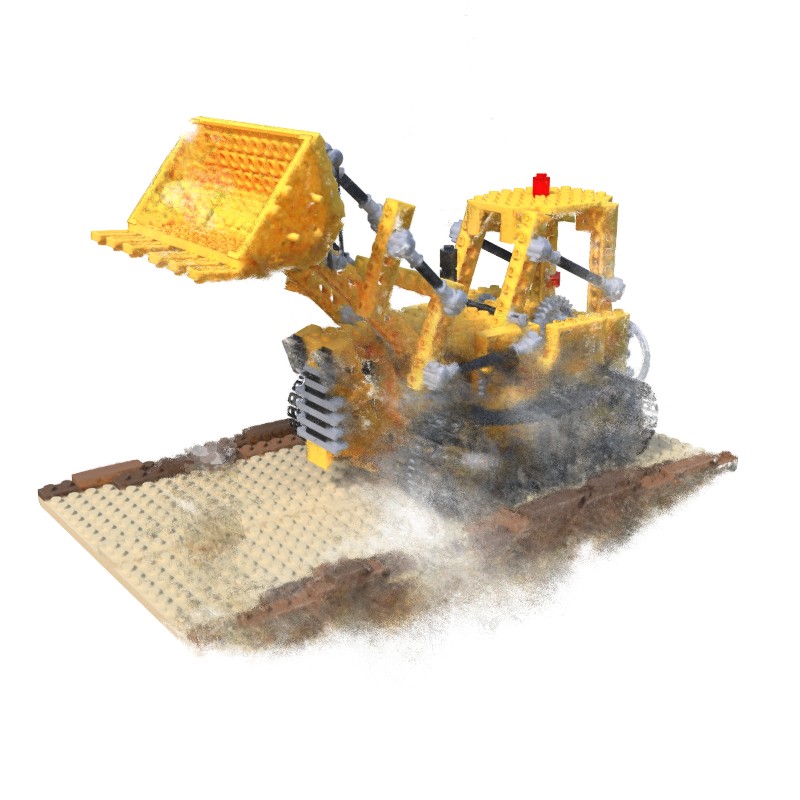}};
      \node [image,right=of frame5] (frame6) {\includegraphics[width=\figurewidth]{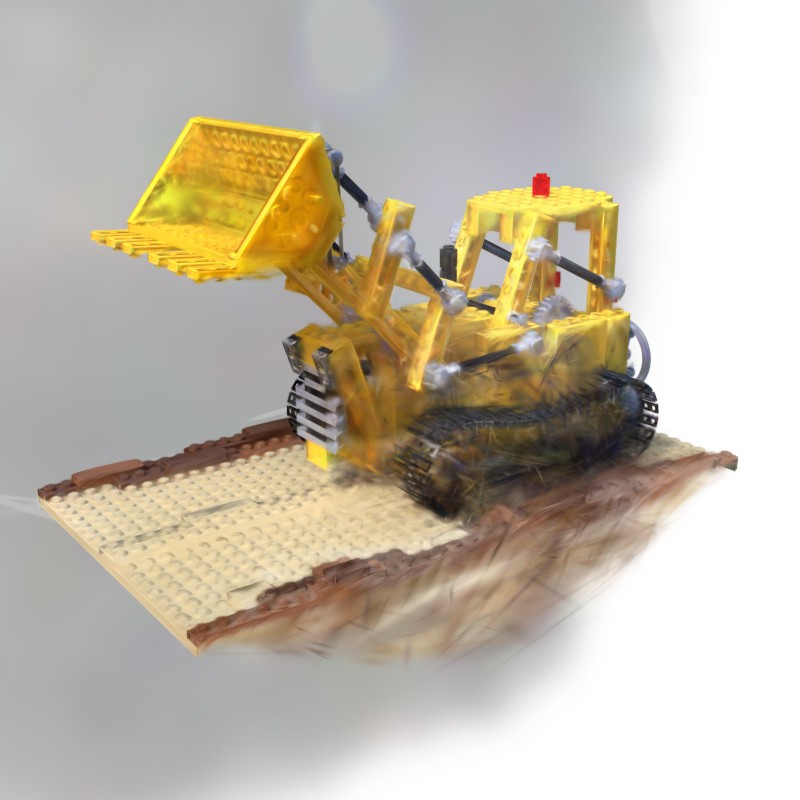}};
      \node [image,right=of frame6] (frame-gt) {\includegraphics[width=\figurewidth]{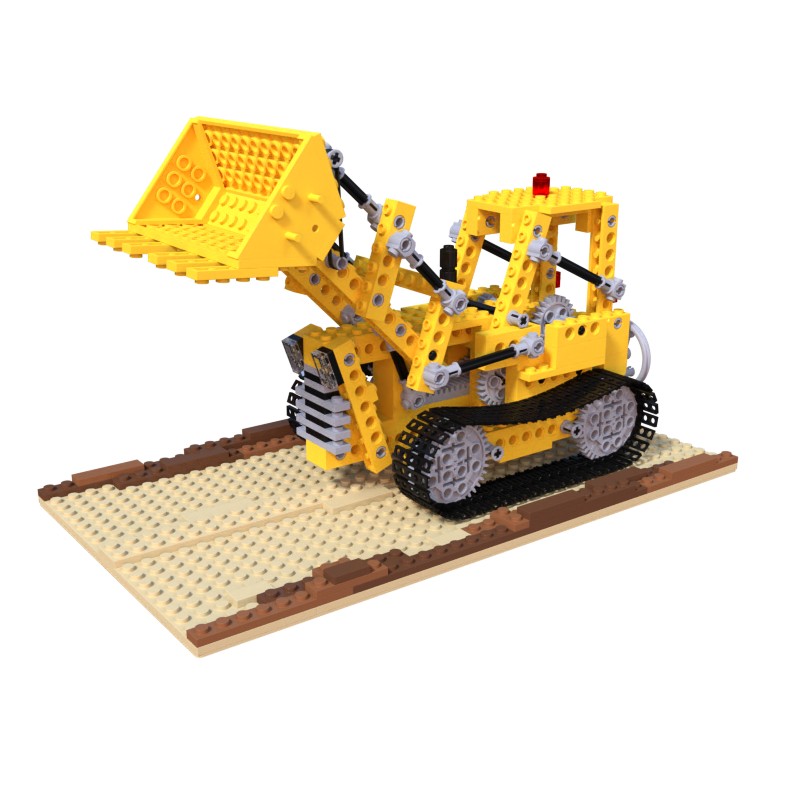}};
      
      \node[image,below=of frame1] (unc1) {\includegraphics[width=\figurewidth]{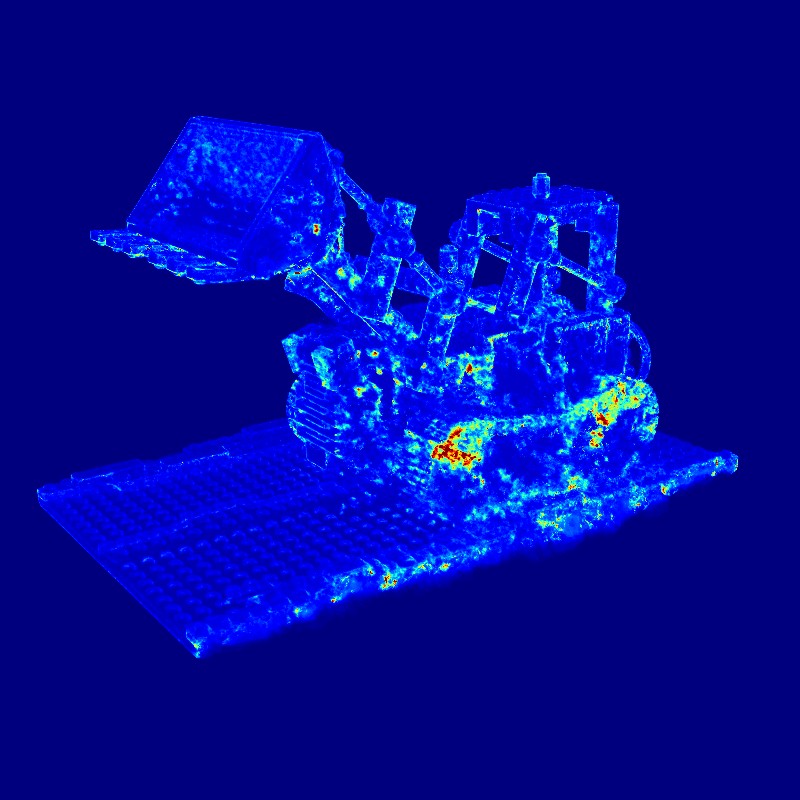}};
      \node[image,right=of unc1] (unc2) {\includegraphics[width=\figurewidth]{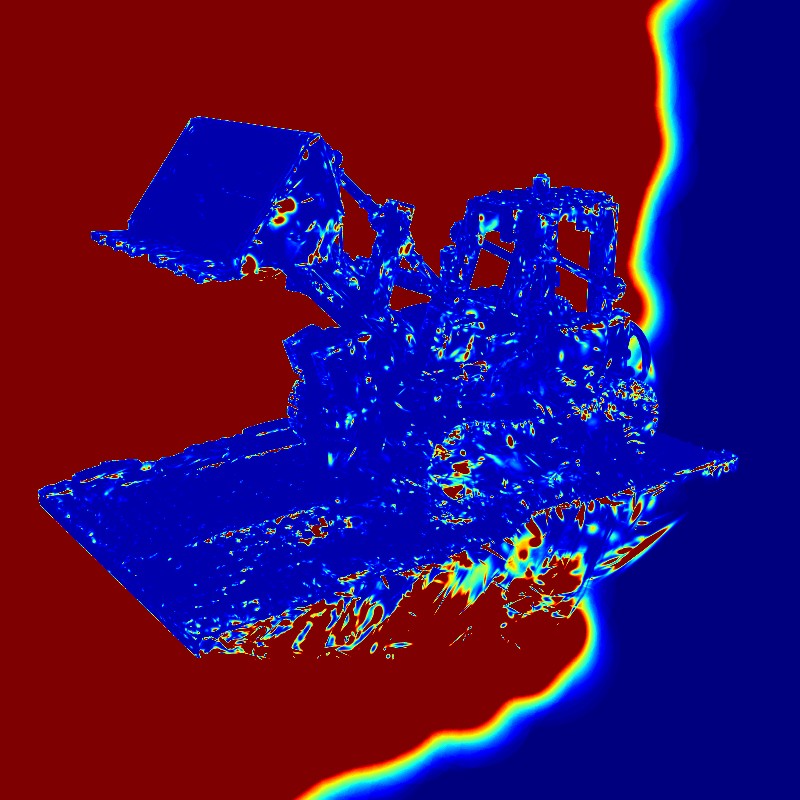}};
      \node[image,right=of unc2] (unc3) {\includegraphics[width=\figurewidth]{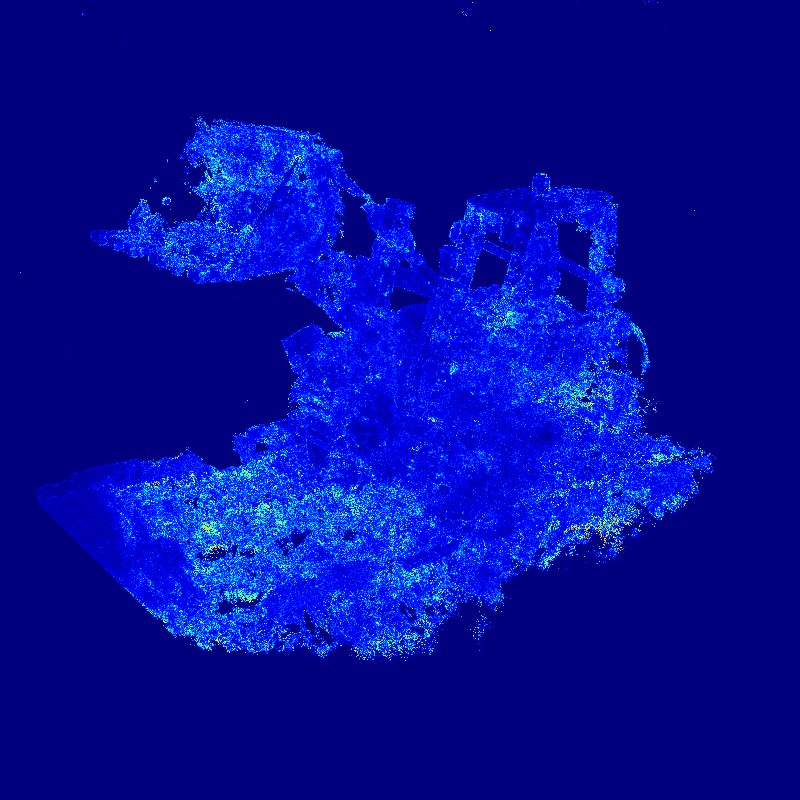}};
      \node[image,right=of unc3] (unc4) {\includegraphics[width=\figurewidth]{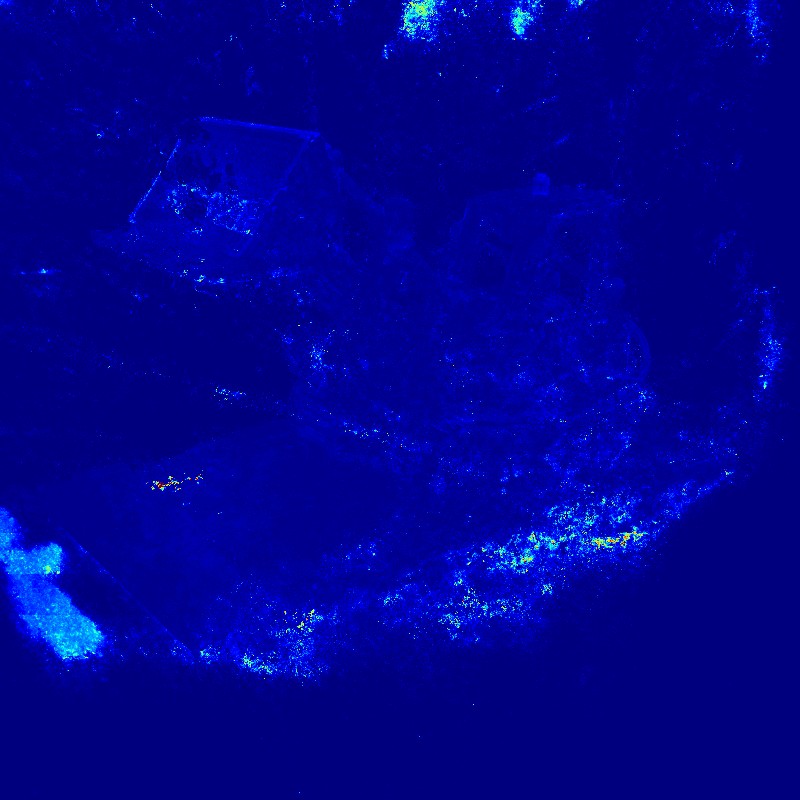}};
      \node[image,right=of unc4] (unc5) {\includegraphics[width=\figurewidth]{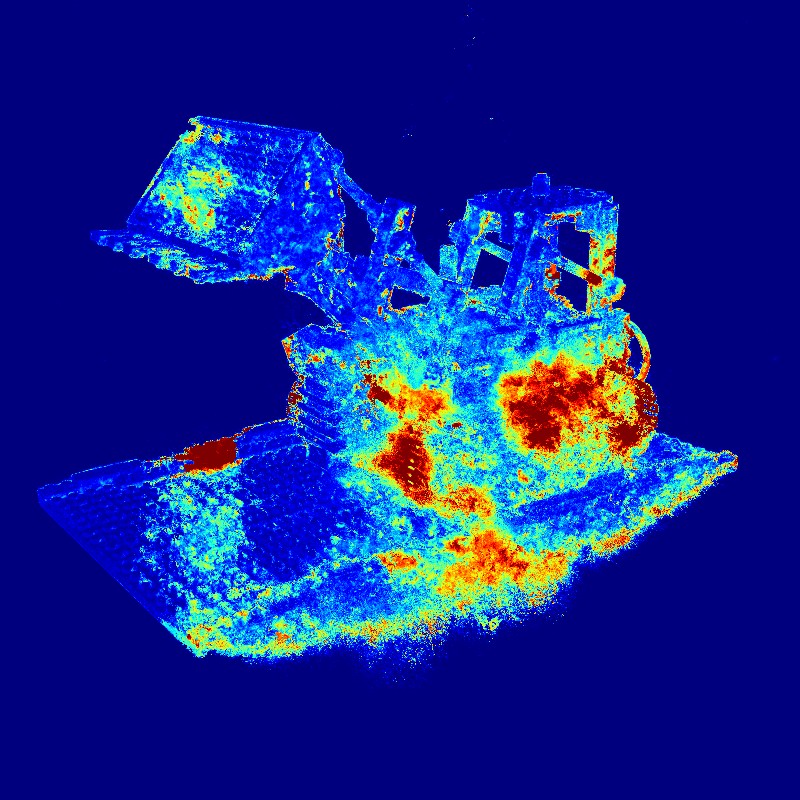}};
      \node[image,right=of unc5] (unc6) {\includegraphics[width=\figurewidth]{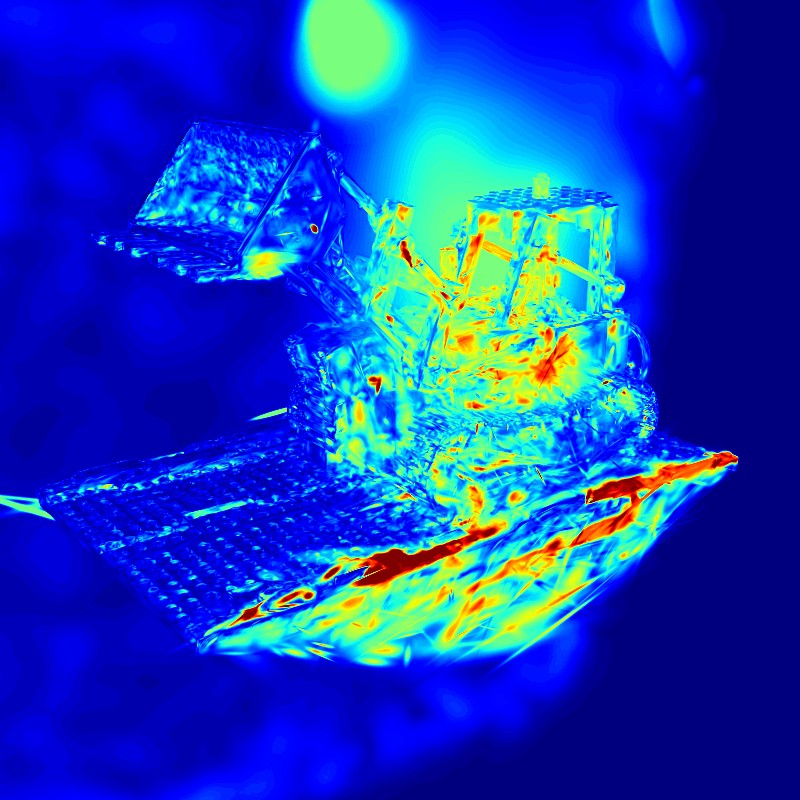}};

      \node[label] (label1) at (frame1.north) {Active-Nerfacto\vphantom{p}};
      \node[label] (label2) at (frame2.north) {Active-Splatfacto\vphantom{p}};
      \node[label] (label3) at (frame3.north) {MC-Dropout-Nerfacto\vphantom{p}};
      \node[label] (label4) at (frame4.north) {Laplace-Nerfacto\vphantom{p}};
      \node[label] (label5) at (frame5.north) {Ensemble-Nerfacto\vphantom{p}};
      \node[label] (label6) at (frame6.north) {Ensemble-Splatfacto\vphantom{p}};
      \node[label] (label-gt) at (frame-gt.north) {Ground Truth\vphantom{p}};

    \end{tikzpicture}
    \caption{Epistemic uncertainty \num{2}: Rendered RGB and uncertainty for one test view of the Lego scene from Blender next to the ground truth RGB. The uncertainty is visualized by the standard deviation (0.0~\protect\includegraphics[width=3em,height=.7em]{figures-main/jet.png}~0.3).}
    \label{fig:ood-blender-lego-qualitative}
\end{figure}

\paragraph{Blender Results} In \cref{tab:ood-blender}, we present the results for the epistemic uncertainty setting using Blender scenes. We can observe that the Splatfacto-based models tend to perform worse then their Nerfacto alternatives, both Active- and Ensemble-, in terms of quality metrics (PSNR, SSIM, LPIPS). This points out that Splatfacto-based model tends to have a poor generalization on OOD novel-view rendering, as also pointed out in \cite{he2024nerfs}.
In \cref{fig:ood-blender-lego-qualitative}, we show some qualitative results on the Lego scene. We can see that the source of the poor performance on quality metrics of the Splatfacto-based methods is related to the emergence in the reconstruction of unwanted floaters. On the positive note, we can see that both methods produce a large standard deviation value over those areas.

\begin{figure}[t]
  \centering
  \setlength{\figurewidth}{0.136\textwidth}
  \begin{tikzpicture}[image/.style = {inner sep=0, outer sep=0, minimum width=\figurewidth, anchor=north west, text width=\figurewidth}, node distance = 1pt and 1pt, every node/.style={font= {\tiny}}, label/.style = {scale=0.75,font={\tiny},anchor=south,inner sep=0pt,outer sep=2pt,rotate=0}] 

    \node [image] (africa-1) {\includegraphics[width=\figurewidth]{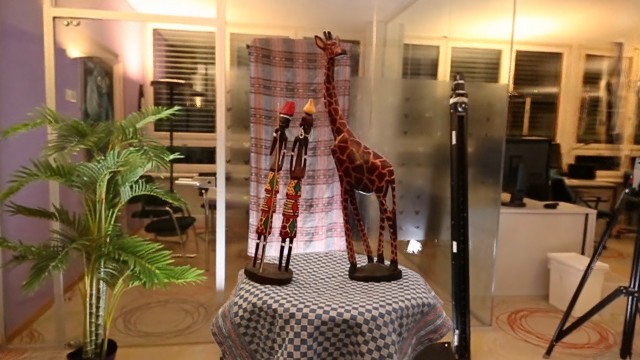}};
    \node [image,right=of africa-1] (africa-2) {\includegraphics[width=\figurewidth]{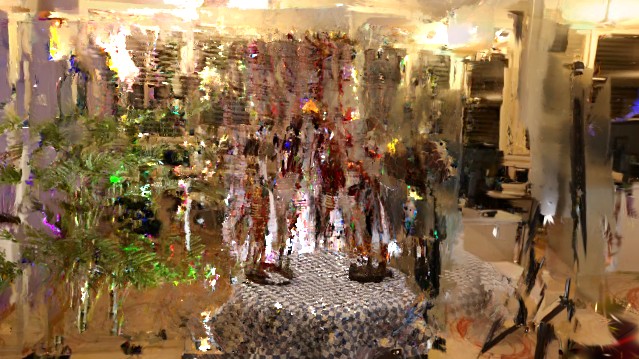}};
    \node [image,right=of africa-2] (africa-3) {\includegraphics[width=\figurewidth]{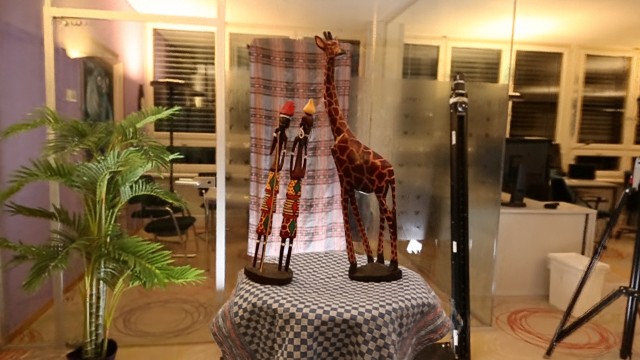}};
    \node [image,right=of africa-3] (africa-4) {\includegraphics[width=\figurewidth]{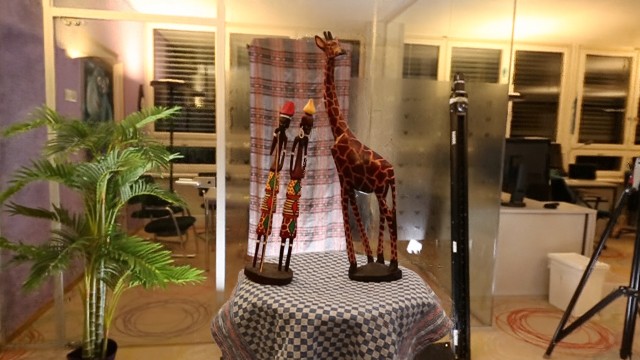}};
    \node [image,right=of africa-4] (africa-5) {\includegraphics[width=\figurewidth]{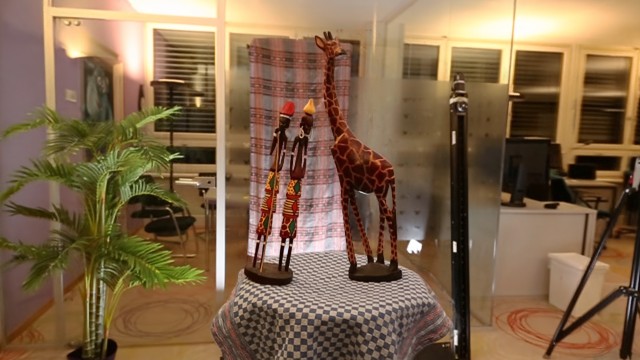}};
    \node [image,right=of africa-5] (africa-6) {\includegraphics[width=\figurewidth]{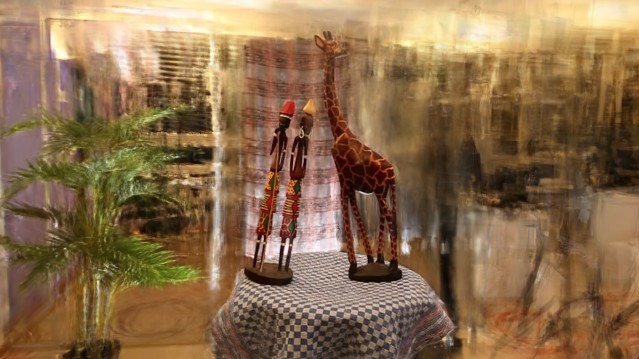}};
    \node [image,right=of africa-6] (africa-gt) {\includegraphics[width=\figurewidth]{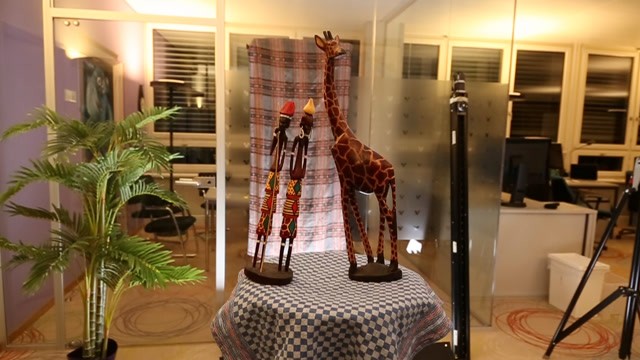}};
    
    \node[image,below=of africa-1] (africa-unc-1) {\includegraphics[width=\figurewidth]{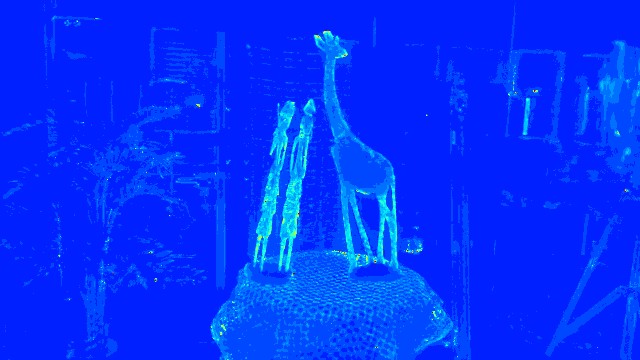}};
    \node[image,right=of africa-unc-1] (africa-unc-2) {\includegraphics[width=\figurewidth]{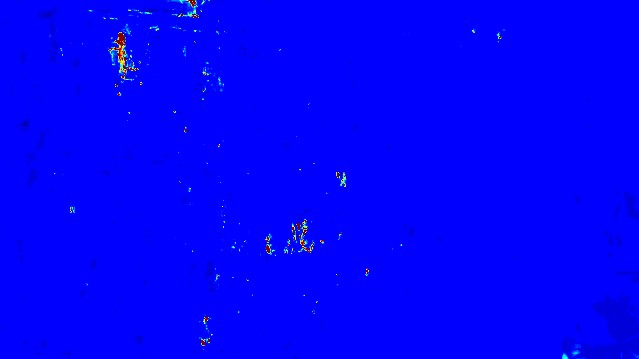}};
    \node[image,right=of africa-unc-2] (africa-unc-3) {\includegraphics[width=\figurewidth]{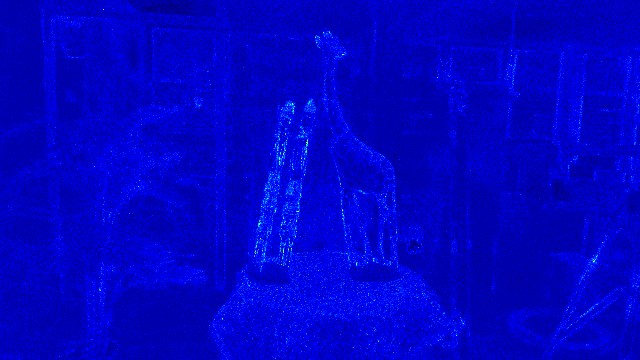}};
    \node[image,right=of africa-unc-3] (africa-unc-4) {\includegraphics[width=\figurewidth]{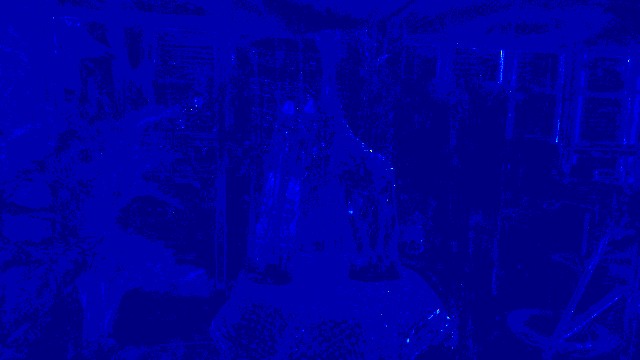}};
    \node[image,right=of africa-unc-4] (africa-unc-5) {\includegraphics[width=\figurewidth]{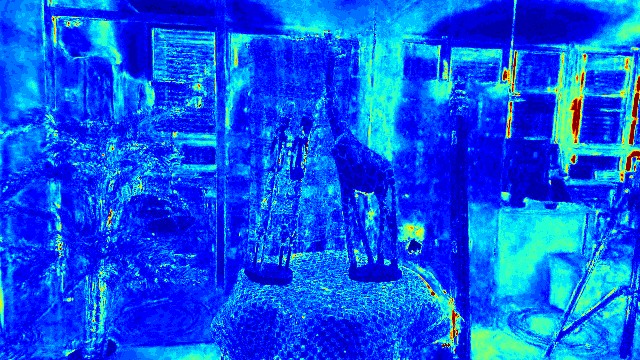}};
    \node[image,right=of africa-unc-5] (africa-unc-6) {\includegraphics[width=\figurewidth]{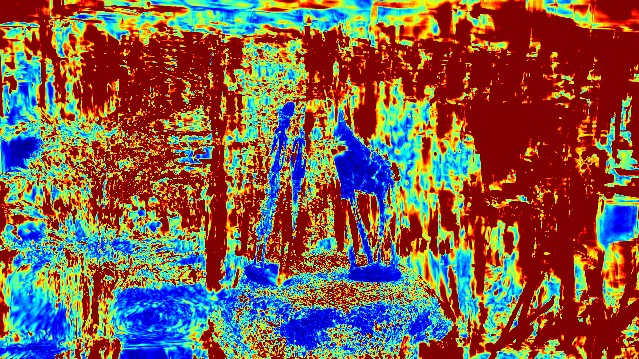}};

    \node [image,below=of africa-unc-1] (basket-1) {\includegraphics[width=\figurewidth]{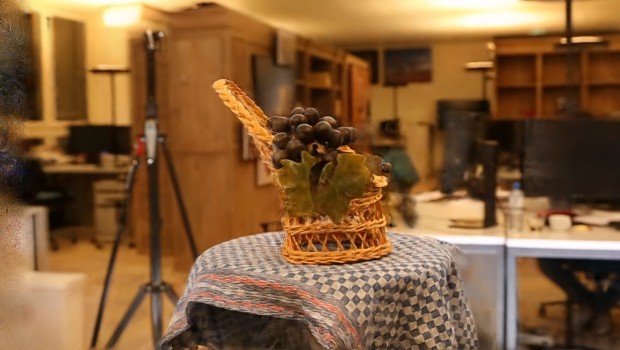}};
    \node [image,right=of basket-1] (basket-2) {\includegraphics[width=\figurewidth]{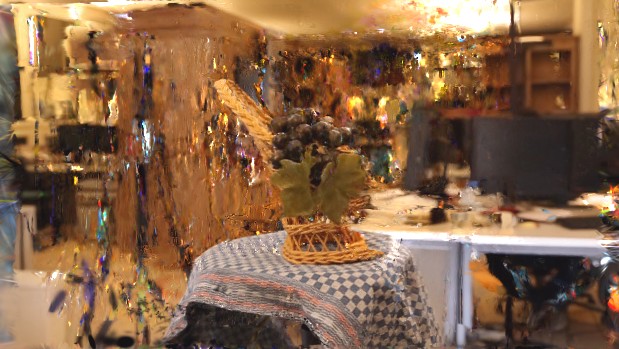}};
    \node [image,right=of basket-2] (basket-3) {\includegraphics[width=\figurewidth]{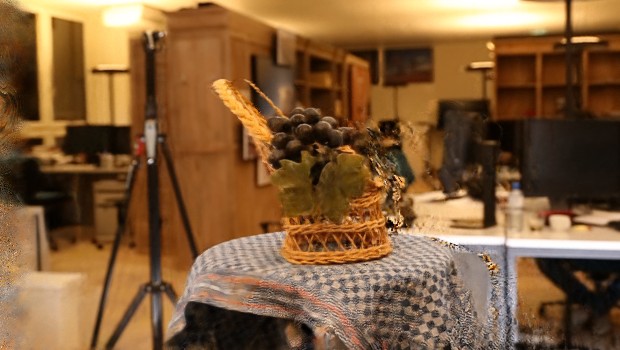}};
    \node [image,right=of basket-3] (basket-4) {\includegraphics[width=\figurewidth]{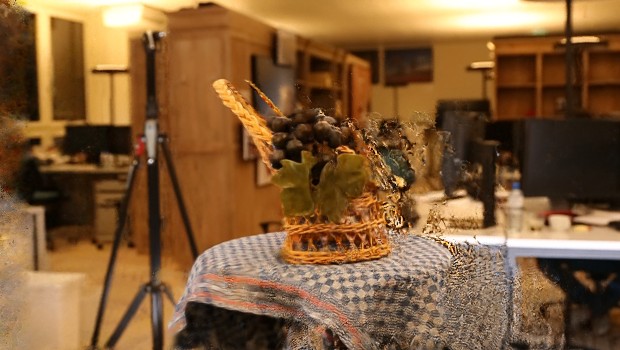}};
    \node [image,right=of basket-4] (basket-5) {\includegraphics[width=\figurewidth]{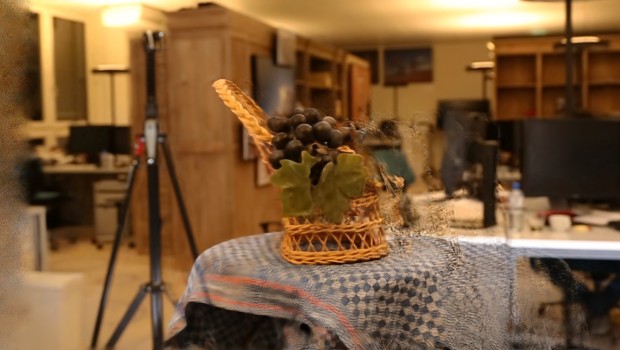}};
    \node [image,right=of basket-5] (basket-6) {\includegraphics[width=\figurewidth]{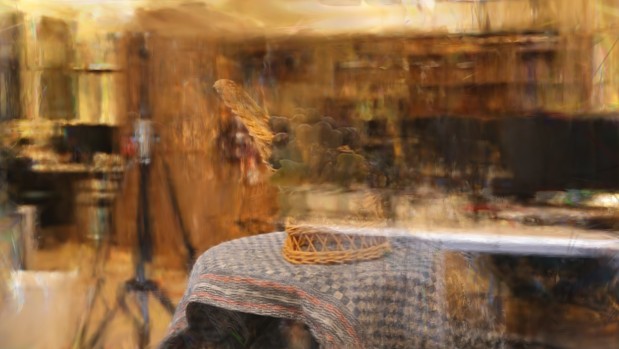}};
    \node [image,right=of basket-6] (basket-gt) {\includegraphics[width=\figurewidth]{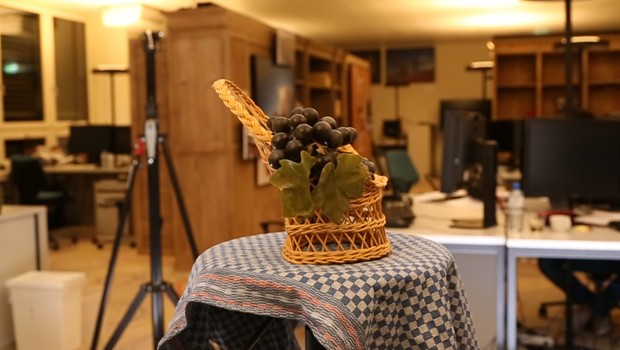}};
    
    \node[image,below=of basket-1] (basket-unc-1) {\includegraphics[width=\figurewidth]{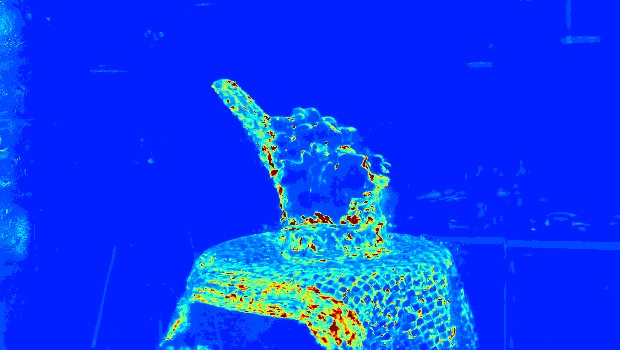}};
    \node[image,right=of basket-unc-1] (basket-unc-2) {\includegraphics[width=\figurewidth]{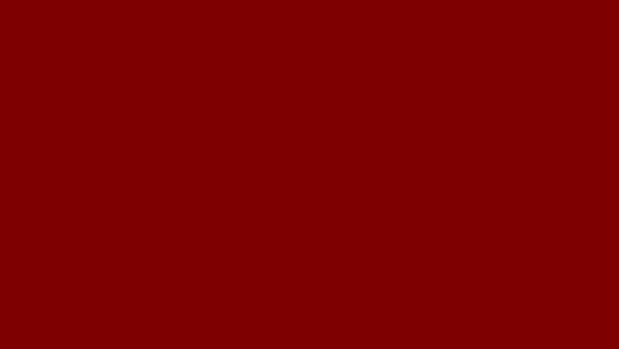}};
    \node[image,right=of basket-unc-2] (basket-unc-3) {\includegraphics[width=\figurewidth]{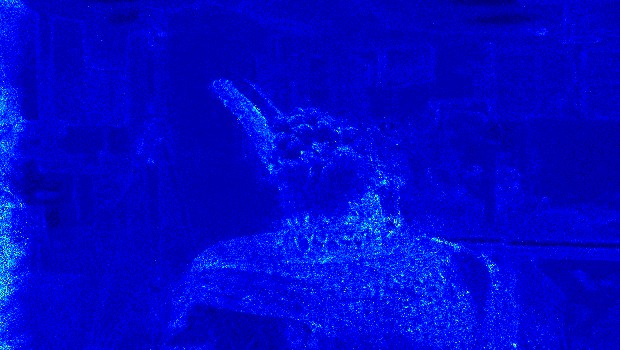}};
    \node[image,right=of basket-unc-3] (basket-unc-4) {\includegraphics[width=\figurewidth]{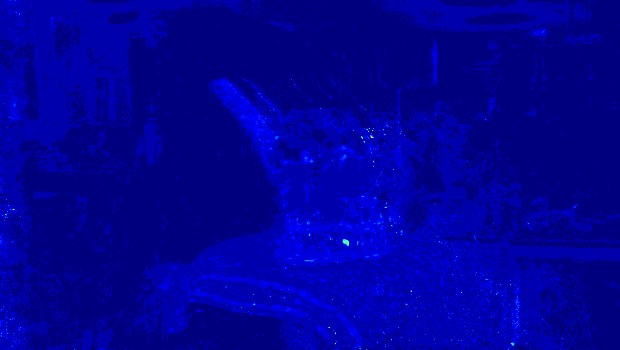}};
    \node[image,right=of basket-unc-4] (basket-unc-5) {\includegraphics[width=\figurewidth]{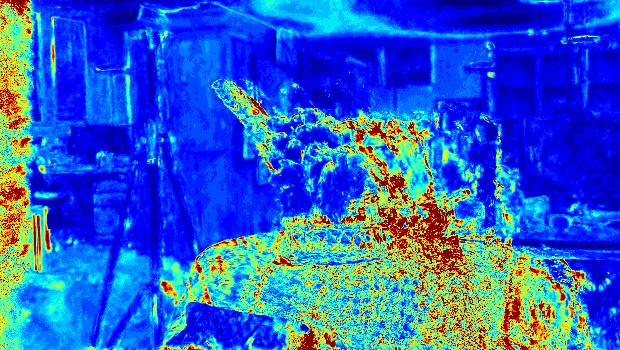}};
    \node[image,right=of basket-unc-5] (basket-unc-6) {\includegraphics[width=\figurewidth]{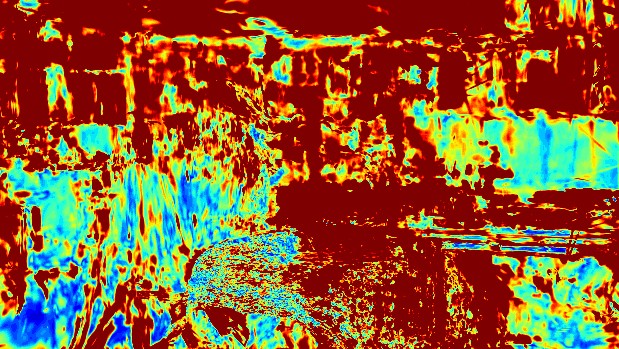}};

    \node [image,below=of basket-unc-1] (statue-1) {\includegraphics[width=\figurewidth]{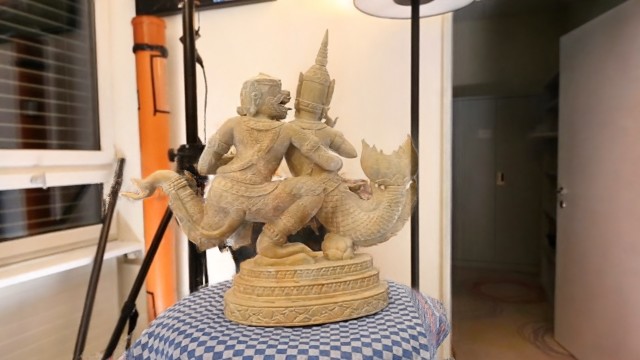}};
    \node [image,right=of statue-1] (statue-2) {\includegraphics[width=\figurewidth]{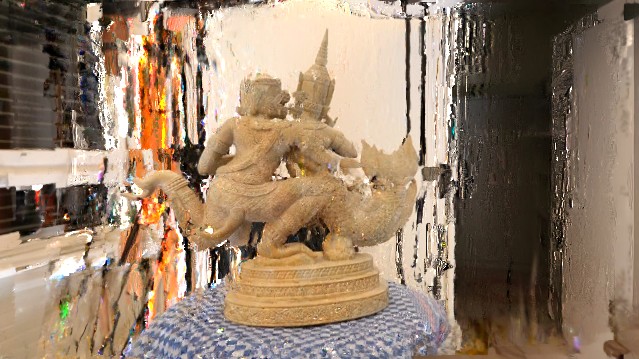}};
    \node [image,right=of statue-2] (statue-3) {\includegraphics[width=\figurewidth]{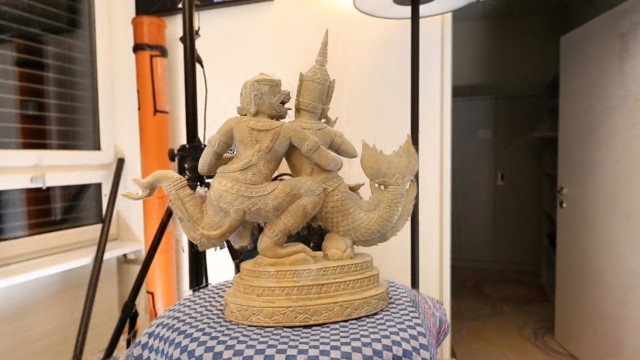}};
    \node [image,right=of statue-3] (statue-4) {\includegraphics[width=\figurewidth]{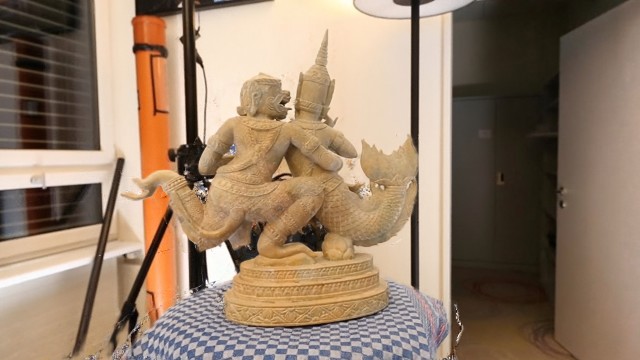}};
    \node [image,right=of statue-4] (statue-5) {\includegraphics[width=\figurewidth]{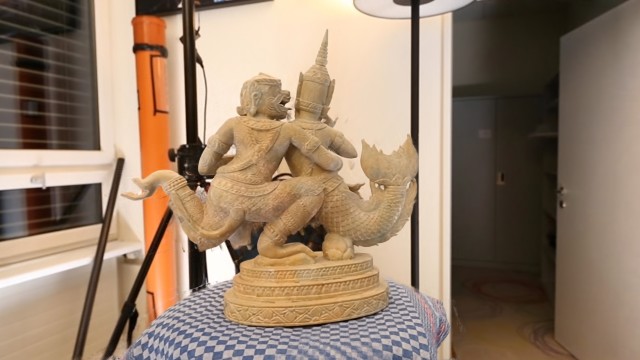}};
    \node [image,right=of statue-5] (statue-6) {\includegraphics[width=\figurewidth]{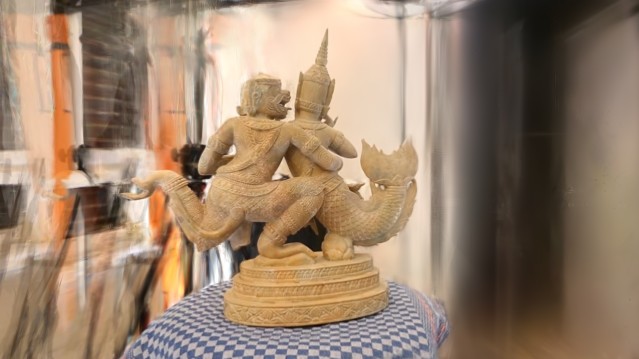}};
    \node [image,right=of statue-6] (statue-gt) {\includegraphics[width=\figurewidth]{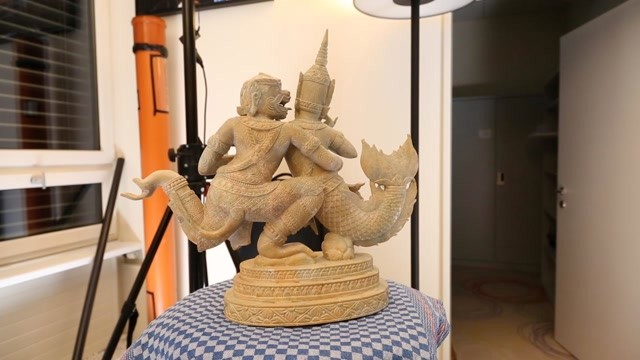}};
    
    \node[image,below=of statue-1] (statue-unc-1) {\includegraphics[width=\figurewidth]{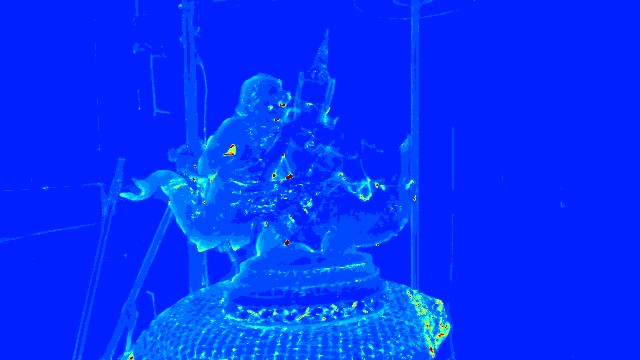}};
    \node[image,right=of statue-unc-1] (statue-unc-2) {\includegraphics[width=\figurewidth]{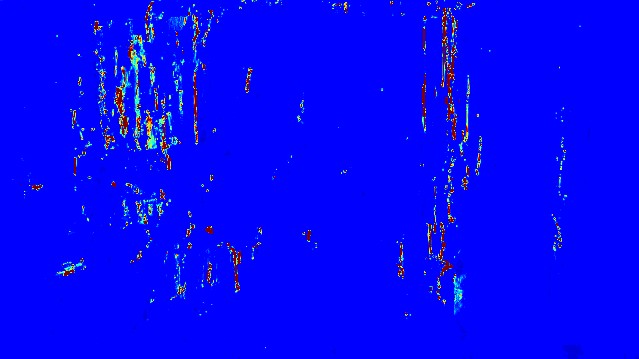}};
    \node[image,right=of statue-unc-2] (statue-unc-3) {\includegraphics[width=\figurewidth]{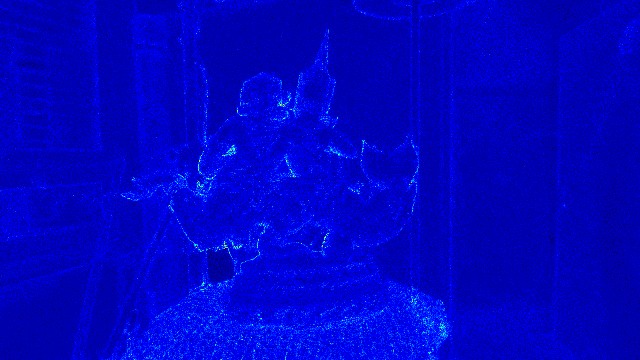}};
    \node[image,right=of statue-unc-3] (statue-unc-4) {\includegraphics[width=\figurewidth]{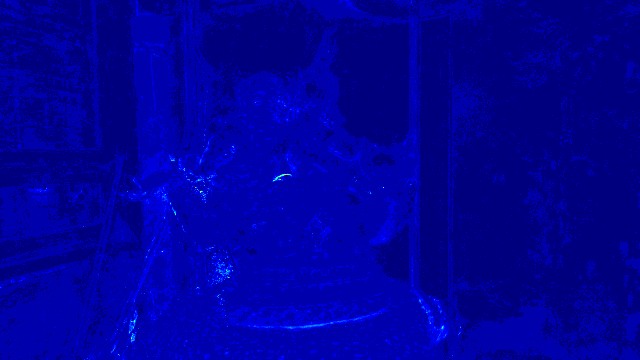}};
    \node[image,right=of statue-unc-4] (statue-unc-5) {\includegraphics[width=\figurewidth]{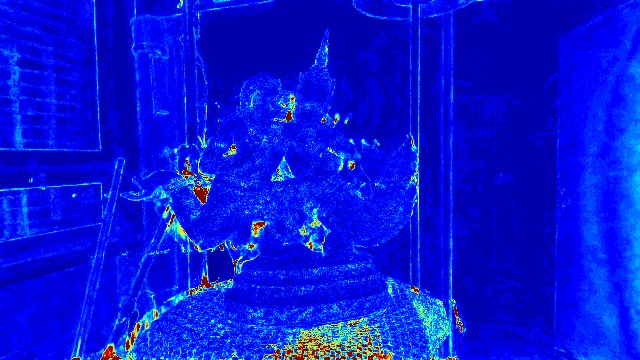}};
    \node[image,right=of statue-unc-5] (statue-unc-6) {\includegraphics[width=\figurewidth]{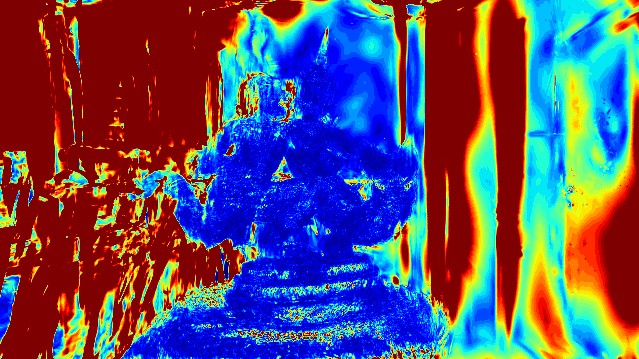}};

    \node [image,below=of statue-unc-1] (torch-1) {\includegraphics[width=\figurewidth]{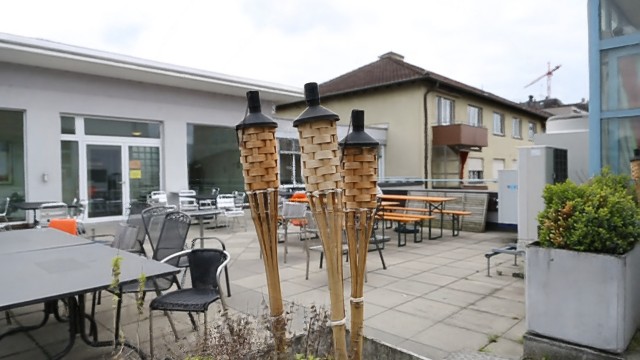}};
    \node [image,right=of torch-1] (torch-2) {\includegraphics[width=\figurewidth]{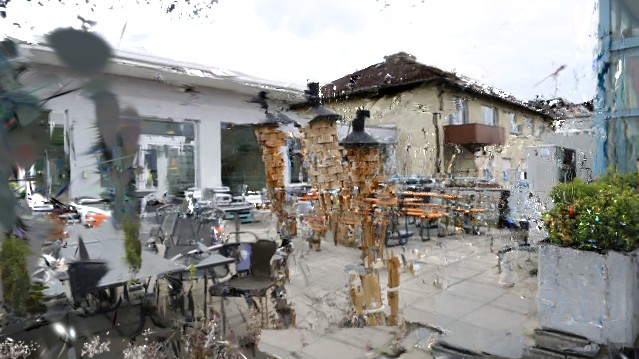}};
    \node [image,right=of torch-2] (torch-3) {\includegraphics[width=\figurewidth]{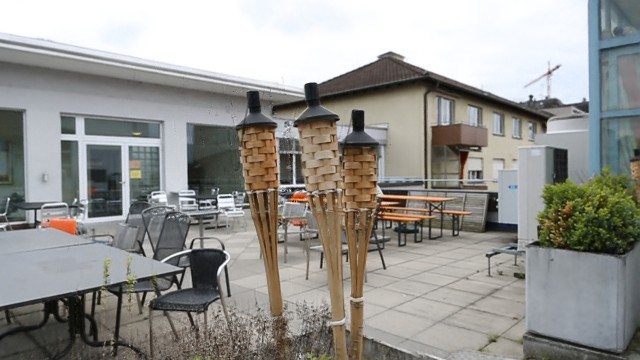}};
    \node [image,right=of torch-3] (torch-4) {\includegraphics[width=\figurewidth]{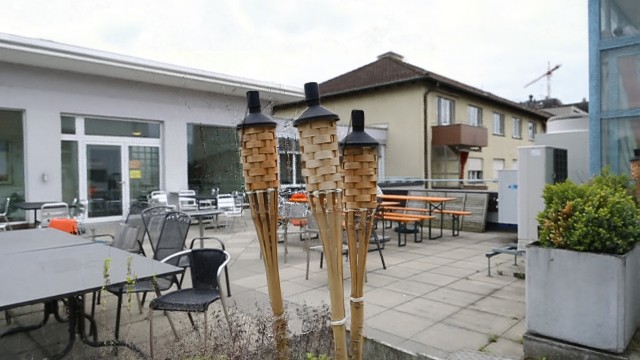}};
    \node [image,right=of torch-4] (torch-5) {\includegraphics[width=\figurewidth]{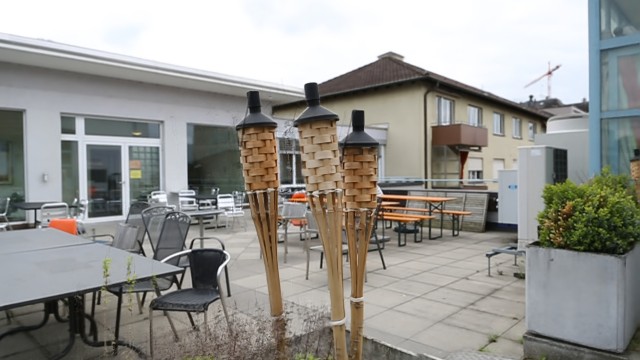}};
    \node [image,right=of torch-5] (torch-6) {\includegraphics[width=\figurewidth]{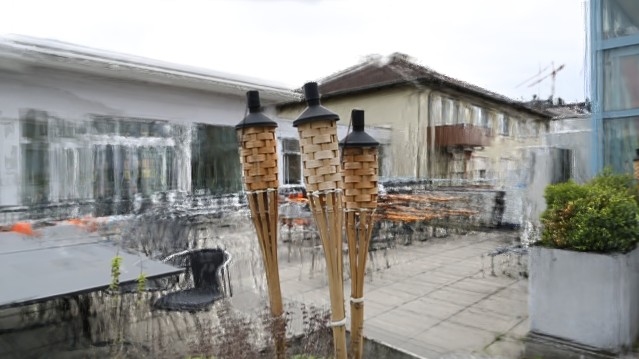}};
    \node [image,right=of torch-6] (torch-gt) {\includegraphics[width=\figurewidth]{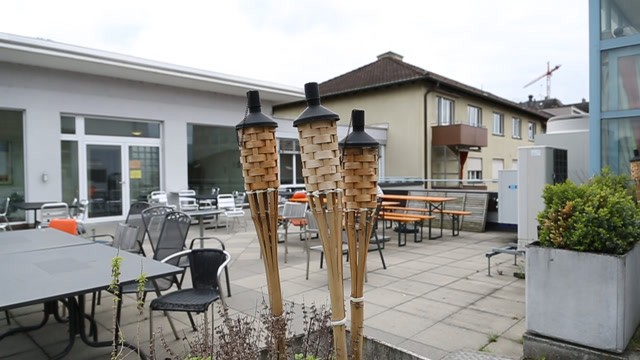}};
    
    \node[image,below=of torch-1] (torch-unc-1) {\includegraphics[width=\figurewidth]{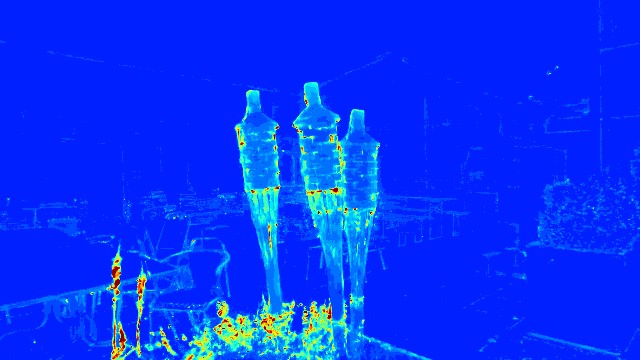}};
    \node[image,right=of torch-unc-1] (torch-unc-2) {\includegraphics[width=\figurewidth]{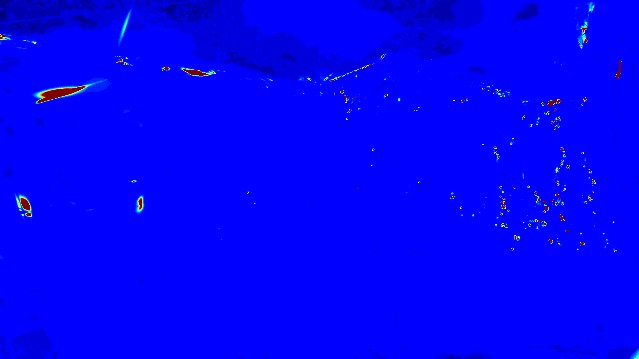}};
    \node[image,right=of torch-unc-2] (torch-unc-3) {\includegraphics[width=\figurewidth]{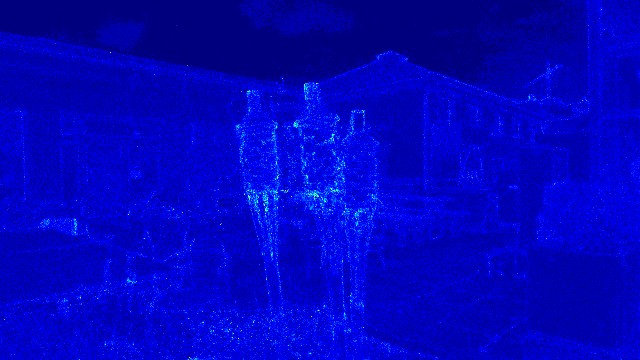}};
    \node[image,right=of torch-unc-3] (torch-unc-4) {\includegraphics[width=\figurewidth]{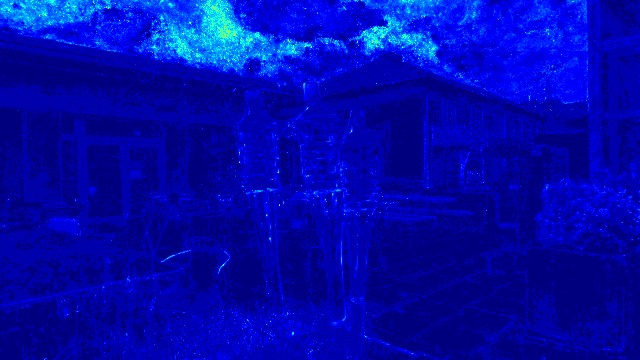}};
    \node[image,right=of torch-unc-4] (torch-unc-5) {\includegraphics[width=\figurewidth]{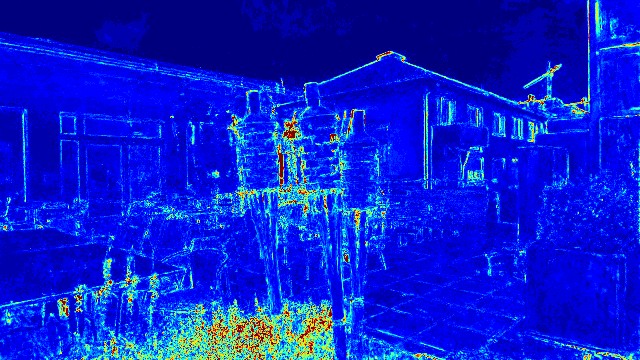}};
    \node[image,right=of torch-unc-5] (torch-unc-6) {\includegraphics[width=\figurewidth]{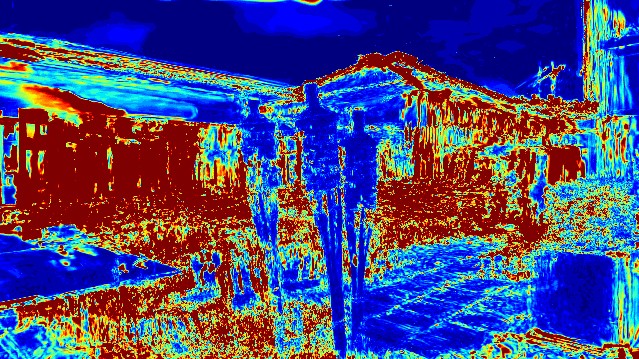}};

    \node[label] (label1) at (africa-1.north) {Active-Nerfacto\vphantom{p}};
    \node[label] (label2) at (africa-2.north) {Active-Splatfacto\vphantom{p}};
    \node[label] (label3) at (africa-3.north) {MC-Dropout-Nerfacto\vphantom{p}};
    \node[label] (label4) at (africa-4.north) {Laplace-Nerfacto\vphantom{p}};
    \node[label] (label5) at (africa-5.north) {Ensemble-Nerfacto\vphantom{p}};
    \node[label] (label6) at (africa-6.north) {Ensemble-Splatfacto\vphantom{p}};
    \node[label] (label-gt) at (africa-gt.north) {Ground Truth\vphantom{p}};

    \node[anchor=south,inner sep=1pt,rotate=90] (scene1) at (africa-1.south west) {Africa\vphantom{p}};
    \node[anchor=south,inner sep=1pt,rotate=90] (scene2) at (basket-1.south west) {Basket\vphantom{p}};
    \node[anchor=south,inner sep=1pt,rotate=90] (scene3) at (statue-1.south west) {Statue\vphantom{p}};
    \node[anchor=south,inner sep=1pt,rotate=90] (scene4) at (torch-1.south west) {Torch\vphantom{p}};
    
    \end{tikzpicture}
  \caption{Epistemic uncertainty \num{2}: Rendered RGB and uncertainty for test views from LF scenes~\cite{shen2022conditional} next to the ground truth RGB. The uncertainty is visualized by the standard deviation (0.0~\protect\includegraphics[width=3em,height=.7em]{figures-main/jet.png}~0.1). 
  }
  \label{fig:fewview-lf-qualitative-appendix}
\end{figure}

\paragraph{LF Results} We take the few-view setting from \cite{shen2021stochastic, shen2022conditional} which uses 4-5 forward-facing views from LF data set scenes~\cite{yucer2016efficient,zhang2020nerf++}. Here, we evaluate the uncertainty estimation on both RGB and depth (see details in \cref{app:methods}). In \cref{tab:ood-lfdataset-rgb-depth-appendix}, we note that the Nerfacto-based methods clearly outperform the Splatfacto methods on the rendered RGB quality and depth error. 
Although the test views are nearby the training views, Active-Splatfacto and Ensemble-Splatfacto need larger number of views of the scenes to achieve a lower reconstruction error. 
For the uncertainty metrics, Ensemble-Nerfacto yields the best AUSE scores on both RGB and depth predictions. Active-Nerfacto and Laplace-Nerfacto achieve the best AUCE scores for the depth, which indicates that the depth uncertainty proposed in \cite{roessle2022dense} is better calibrated than averaging depth maps from an ensemble. See \cref{fig:fewview-lf-qualitative-appendix} in \cref{app:epistemic} for qualitative results. 

In \cref{fig:fewview-lf-qualitative-appendix}, we provide the RGB reconstruction and uncertainties, for all the scenes from the LF data set. 
All methods, except Active-Splatfacto, yield plausible uncertainty maps, where Ensemble-Nerfacto captures the uncertainty around edges and fine-grained object details. %
Active-Nerfacto yields highest uncertainty on the foreground objects and estimates higher uncertainty on the background compared to the other NeRF methods. Ensemble-Splatfacto gives high uncertainty mainly on the background, while Active-Splatfacto is unable to produce reasonable uncertainty maps.\looseness-1

\begin{table}[t]
  \centering
  \caption{Epistemic uncertainty \num{2}: Performance metrics for RGB and depth predictions averaged across scenes from the LF data set. 
  The \first{first}, \second{second}, and \third{third} values are highlighted.
  }
  \setlength{\tabcolsep}{5pt}
  \resizebox{\textwidth}{!}{
  \begin{tabular}{l|cccccc|cccc}
  \toprule
   & \multicolumn{6}{|c}{\textbf{RGB}} & \multicolumn{4}{|c}{\textbf{Depth}} \\
  {Method} & PSNR $\uparrow$ & SSIM $\uparrow$ & LPIPS $\downarrow$ & NLL $\downarrow$ & AUSE $\downarrow$ & AUCE $\downarrow$ & RMSE $\downarrow$ & NLL $\downarrow$ & AUSE $\downarrow$ & AUCE $\downarrow$ \\
  \midrule
  Active-Nerfacto & \cellcolor{yellow!25}27.33 & \cellcolor{red!25}0.92 & \cellcolor{red!25}0.05 & \cellcolor{yellow!25}-1.52 & \cellcolor{orange!25}0.24 & \cellcolor{red!25}0.09 & \cellcolor{orange!25}60.20 & \cellcolor{red!25}5.08 & \cellcolor{yellow!25}0.42 & \cellcolor{orange!25}0.14 \\ 
  Active-Splatfacto & 13.39 & 0.32 & 0.54 & 21.62 & 0.55 & 0.35 & 107.28 & 1574.38 & 0.49 & 0.45 \\ 
  MC-Dropout-Nerfacto & \cellcolor{red!25}27.97 & \cellcolor{orange!25}0.91 & \cellcolor{red!25}0.05 & \cellcolor{orange!25}-1.54 & \cellcolor{yellow!25}0.29 & \cellcolor{yellow!25}0.29 & \cellcolor{yellow!25}61.83 & 1044.31 & 0.47 & 0.47 \\ 
  Laplace-Nerfacto & 27.05 & \cellcolor{yellow!25}0.90 & \cellcolor{orange!25}0.06 & -1.26 & 0.34 & 0.40 & 64.66 & \cellcolor{orange!25}72.98 & \cellcolor{orange!25}0.38 & \cellcolor{red!25}0.13 \\ 
  Ensemble-Nerfacto & \cellcolor{orange!25}27.74 & \cellcolor{red!25}0.92 & \cellcolor{red!25}0.05 & \cellcolor{red!25}-1.88 & \cellcolor{red!25}0.14 & \cellcolor{orange!25}0.15 & \cellcolor{red!25}58.73 & \cellcolor{yellow!25}61.77 & \cellcolor{red!25}0.31 & \cellcolor{yellow!25}0.32 \\ 
  Ensemble-Splatfacto & 15.91 & 0.52 & \cellcolor{yellow!25}0.33 & 0.75 & 0.33 & \cellcolor{red!25}0.09 & 104.55 & 960.21 & 0.45 & 0.44 \\
  \bottomrule
  \end{tabular}
  }
  \label{tab:ood-lfdataset-rgb-depth-appendix}
  \end{table}

\subsection{Experiments on Sensitivity to Cluttered Inputs}
\label{app:outliers}

\begin{figure}[t]
    \centering
    \setlength{\figurewidth}{0.2\textwidth}
    \setlength{\figureheight}{.08\textheight}
    \input{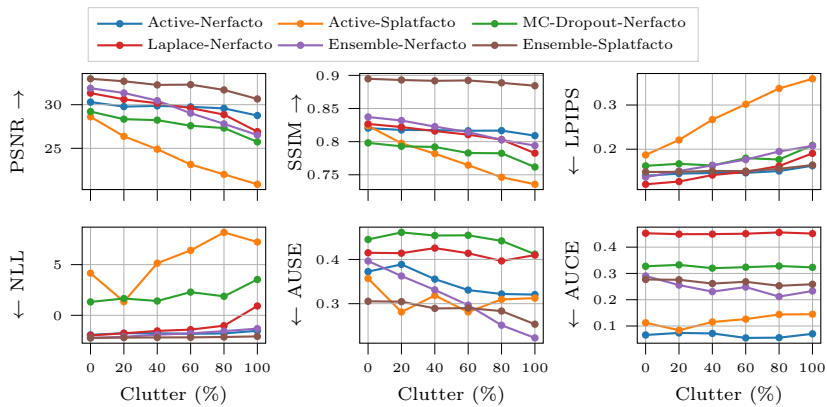} %
    \caption{Increasing clutter \num{3} affects performance metrics. Increasing number of cluttered training views (in \%) in the Yoda scene from the RobustNeRF data set. 
    }
    \label{fig:outliers-robustnerf-yoda}
\end{figure}

\paragraph{RobustNeRF Results} In \cref{fig:outliers-robustnerf-yoda}, we evaluate on the Yoda scene with varying amounts of cluttered training views. On rendered RGB, Ensemble-Splatfacto is most robust to an increasing number of cluttered views followed by Active-Nerfacto. On the AUSE scores, we see that most methods obtain a better correlation between their estimated uncertainty and the error, which could be due to the floaters from dynamic objects being easier to detect with the uncertainty rather than details about the static objects. The calibration errors are stable when using more cluttered training views for each method, where the Active-Nerfacto followed by the Ensemble methods achieve the best calibration. 

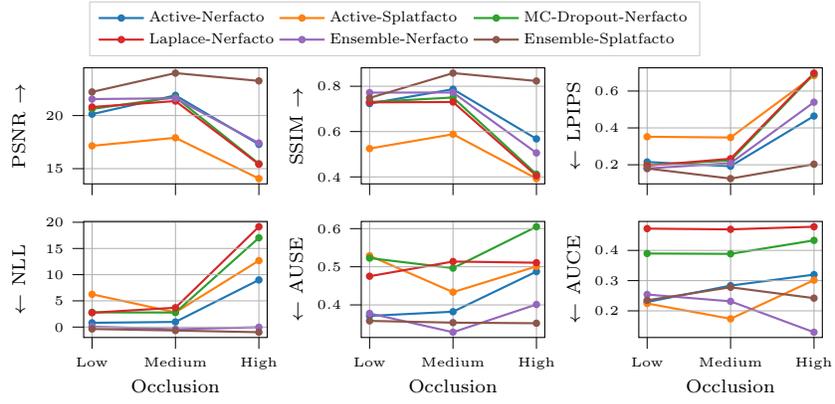
\begin{figure}[t]
  \centering
  \setlength{\figurewidth}{0.2\textwidth}
  \setlength{\figureheight}{.08\textheight}
  \pgfplotsset{every axis title/.append style={at={(0.5,0.80)}}} 
\pgfplotsset{every axis x label/.append style={at={(0.5,0.05)}}}
\pgfplotsset{every axis y label/.append style={at={(0.15,0.5)}}}

\begin{tikzpicture}
\tikzstyle{every node}=[font=\scriptsize]
\definecolor{crimson2143940}{RGB}{214,39,40}
\definecolor{darkgray176}{RGB}{176,176,176}
\definecolor{darkorange25512714}{RGB}{255,127,14}
\definecolor{forestgreen4416044}{RGB}{44,160,44}
\definecolor{lightgray204}{RGB}{204,204,204}
\definecolor{mediumpurple148103189}{RGB}{148,103,189}
\definecolor{sienna1408675}{RGB}{140,86,75}
\definecolor{steelblue31119180}{RGB}{31,119,180}

\begin{groupplot}[
  group style={group size= 3 by 2, horizontal sep=1.25cm, vertical sep=.5cm},
  tick align=outside,
  tick pos=left,
  grid=both,
  xlabel={Occlusion},
  xmin=-0.1, xmax=2.1,
  xtick={0,1,2},
  xticklabels={Low,Medium,High},
  xtick style={color=black},
  x grid style={darkgray176,solid},
  y grid style={darkgray176,solid},
  ytick style={color=black},
  tick label style={font=\tiny}
]

\nextgroupplot[
height=\figureheight,
width=\figurewidth,
legend cell align={left},
legend columns=3,
legend style={
  nodes={scale=0.8},
  fill opacity=0.8, 
  draw opacity=1, 
  text opacity=1, 
  at={(0.03,1.1)}, 
  anchor=south west,
  draw=lightgray204},
xlabel=\empty,
xticklabel=\empty,
ylabel={PSNR~$\rightarrow$},
ymin=13.5683740615845, ymax=24.4979330062866,
ytick style={color=black}
]
\addplot [thick, steelblue31119180, mark=*, mark size=1, mark options={solid}]
table {%
0 20.1330404281616
1 21.9035453796387
2 17.2878751754761
};
\addlegendentry{Active-Nerfacto}
\addplot [thick, darkorange25512714, mark=*, mark size=1, mark options={solid}]
table {%
0 17.1491088867188
1 17.9064741134644
2 14.0651721954346
};
\addlegendentry{Active-Splatfacto}
\addplot [thick, forestgreen4416044, mark=*, mark size=1, mark options={solid}]
table {%
0 20.623607635498
1 21.7319622039795
2 15.46622133255
};
\addlegendentry{MC-Dropout-Nerfacto}
\addplot [thick, crimson2143940, mark=*, mark size=1, mark options={solid}]
table {%
0 20.8078327178955
1 21.3666868209839
2 15.4145460128784
};
\addlegendentry{Laplace-Nerfacto}
\addplot [thick, mediumpurple148103189, mark=*, mark size=1, mark options={solid}]
table {%
0 21.5546407699585
1 21.6325874328613
2 17.4112930297852
};
\addlegendentry{Ensemble-Nerfacto}
\addplot [thick, sienna1408675, mark=*, mark size=1, mark options={solid}]
table {%
0 22.2263355255127
1 24.0011348724365
2 23.2667007446289
};
\addlegendentry{Ensemble-Splatfacto}

\nextgroupplot[
height=\figureheight,
width=\figurewidth,
xlabel=\empty,
xticklabel=\empty,
ylabel={SSIM~$\rightarrow$},
ymin=0.369186376780272, ymax=0.881393680721521,
ytick style={color=black}
]
\addplot [thick, steelblue31119180, mark=*, mark size=1, mark options={solid}]
table {%
0 0.723465234041214
1 0.786900162696838
2 0.567728370428085
};
\addplot [thick, darkorange25512714, mark=*, mark size=1, mark options={solid}]
table {%
0 0.525000095367432
1 0.588202580809593
2 0.392468526959419
};
\addplot [thick, forestgreen4416044, mark=*, mark size=1, mark options={solid}]
table {%
0 0.7302266061306
1 0.750393569469452
2 0.411962285637856
};
\addplot [thick, crimson2143940, mark=*, mark size=1, mark options={solid}]
table {%
0 0.729029953479767
1 0.730449855327606
2 0.405045494437218
};
\addplot [thick, mediumpurple148103189, mark=*, mark size=1, mark options={solid}]
table {%
0 0.772009760141373
1 0.772371917963028
2 0.50573955476284
};
\addplot [thick, sienna1408675, mark=*, mark size=1, mark options={solid}]
table {%
0 0.748315244913101
1 0.858111530542374
2 0.823262214660645
};

\nextgroupplot[
height=\figureheight,
width=\figurewidth,
xlabel=\empty,
xticklabel=\empty,
ylabel={$\leftarrow$~LPIPS},
ymin=0.0964400926604867, ymax=0.726028132252395,
ytick style={color=black}
]
\addplot [thick, steelblue31119180, mark=*, mark size=1, mark options={solid}]
table {%
0 0.215141721069813
1 0.192094556987286
2 0.464275881648064
};
\addplot [thick, darkorange25512714, mark=*, mark size=1, mark options={solid}]
table {%
0 0.351991474628448
1 0.347907111048698
2 0.6836878657341
};
\addplot [thick, forestgreen4416044, mark=*, mark size=1, mark options={solid}]
table {%
0 0.198496550321579
1 0.223164774477482
2 0.692182391881943
};
\addplot [thick, crimson2143940, mark=*, mark size=1, mark options={solid}]
table {%
0 0.195029236376286
1 0.232793241739273
2 0.697410494089127
};
\addplot [thick, mediumpurple148103189, mark=*, mark size=1, mark options={solid}]
table {%
0 0.179915256798267
1 0.208812981843948
2 0.539479732513428
};
\addplot [thick, sienna1408675, mark=*, mark size=1, mark options={solid}]
table {%
0 0.180103600025177
1 0.125057730823755
2 0.203079625964165
};

\nextgroupplot[
height=\figureheight,
width=\figurewidth,
ylabel={$\leftarrow$~NLL},
ymin=-1.99267206043005, ymax=20.1557464644313,
ytick style={color=black}
]
\addplot [thick, steelblue31119180, mark=*, mark size=1, mark options={solid}]
table {%
0 0.810395002365112
1 0.977057844400406
2 8.99483752250671
};
\addplot [thick, darkorange25512714, mark=*, mark size=1, mark options={solid}]
table {%
0 6.25316321849823
1 2.80561584234238
2 12.6826195716858
};
\addplot [thick, forestgreen4416044, mark=*, mark size=1, mark options={solid}]
table {%
0 2.80239868164062
1 2.76995050907135
2 17.0389976501465
};
\addplot [thick, crimson2143940, mark=*, mark size=1, mark options={solid}]
table {%
0 2.73574614524841
1 3.69063830375671
2 19.1490001678467
};
\addplot [thick, mediumpurple148103189, mark=*, mark size=1, mark options={solid}]
table {%
0 0.0939229000359774
1 -0.423177175223827
2 -0.0353035777807236
};
\addplot [thick, sienna1408675, mark=*, mark size=1, mark options={solid}]
table {%
0 -0.372241258621216
1 -0.660018116235733
2 -0.985925763845444
};

\nextgroupplot[
height=\figureheight,
width=\figurewidth,
ylabel={$\leftarrow$~AUSE},
ymin=0.314597538858652, ymax=0.618801140040159,
ytick style={color=black}
]
\addplot [thick, steelblue31119180, mark=*, mark size=1, mark options={solid}]
table {%
0 0.371211677789688
1 0.382167533040047
2 0.487401783466339
};
\addplot [thick, darkorange25512714, mark=*, mark size=1, mark options={solid}]
table {%
0 0.529008969664574
1 0.433491975069046
2 0.500921934843063
};
\addplot [thick, forestgreen4416044, mark=*, mark size=1, mark options={solid}]
table {%
0 0.522689074277878
1 0.496458038687706
2 0.604973703622818
};
\addplot [thick, crimson2143940, mark=*, mark size=1, mark options={solid}]
table {%
0 0.475179240107536
1 0.513704940676689
2 0.510635197162628
};
\addplot [thick, mediumpurple148103189, mark=*, mark size=1, mark options={solid}]
table {%
0 0.377806484699249
1 0.328424975275993
2 0.401163682341576
};
\addplot [thick, sienna1408675, mark=*, mark size=1, mark options={solid}]
table {%
0 0.358079716563225
1 0.353514716029167
2 0.351965233683586
};

\nextgroupplot[
height=\figureheight,
width=\figurewidth,
ylabel={$\leftarrow$~AUCE},
ymin=0.111784739004289, ymax=0.495794014638915,
ytick style={color=black}
]
\addplot [thick, steelblue31119180, mark=*, mark size=1, mark options={solid}]
table {%
0 0.229265464055116
1 0.283444595452834
2 0.319819997486306
};
\addplot [thick, darkorange25512714, mark=*, mark size=1, mark options={solid}]
table {%
0 0.224433300454487
1 0.173645399689291
2 0.30157093039562
};
\addplot [thick, forestgreen4416044, mark=*, mark size=1, mark options={solid}]
table {%
0 0.389730845515816
1 0.388515340473203
2 0.433132076787619
};
\addplot [thick, crimson2143940, mark=*, mark size=1, mark options={solid}]
table {%
0 0.472032289836471
1 0.469605414552257
2 0.478339047564613
};
\addplot [thick, mediumpurple148103189, mark=*, mark size=1, mark options={solid}]
table {%
0 0.25406270536267
1 0.231455467930683
2 0.129239706078591
};
\addplot [thick, sienna1408675, mark=*, mark size=1, mark options={solid}]
table {%
0 0.23483982477912
1 0.277944951839921
2 0.24198258395496
};

\end{groupplot}

\end{tikzpicture}%
  \caption{Performance metrics over increasing occlusion level in the training views averaged across the corresponding scenes from the On-the-go data set. 
  }
  \label{fig:outliers-nerfonthego-appendix}
\end{figure}

\begin{figure}[h]
  \centering
  \setlength{\figurewidth}{0.136\textwidth}
  \begin{tikzpicture}[image/.style = {inner sep=0, outer sep=0, minimum width=\figurewidth, anchor=north west, text width=\figurewidth}, node distance = 1pt and 1pt, every node/.style={font= {\tiny}}, label/.style = {scale=0.75,font={\tiny},anchor=south,inner sep=0pt,outer sep=2pt,rotate=0}] 

    \node [image] (corner-1) {\includegraphics[width=\figurewidth]{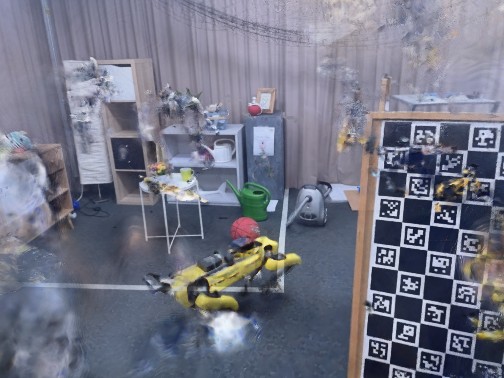}};
    \node [image,right=of corner-1] (corner-2) {\includegraphics[width=\figurewidth]{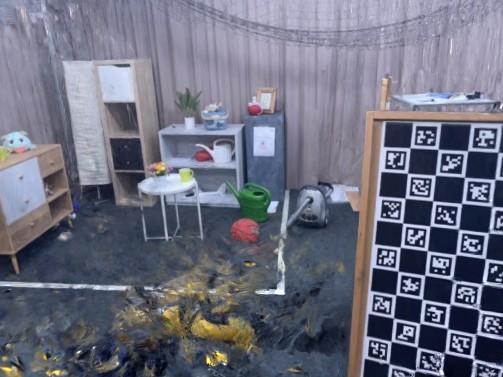}};
    \node [image,right=of corner-2] (corner-3) {\includegraphics[width=\figurewidth]{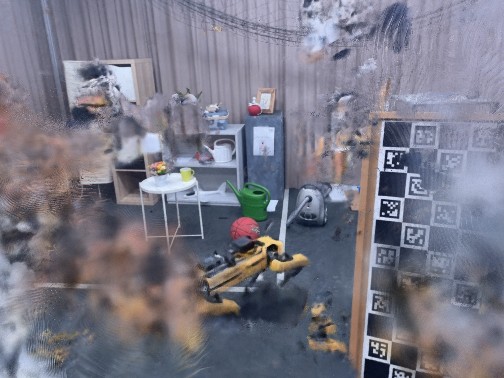}};
    \node [image,right=of corner-3] (corner-4) {\includegraphics[width=\figurewidth]{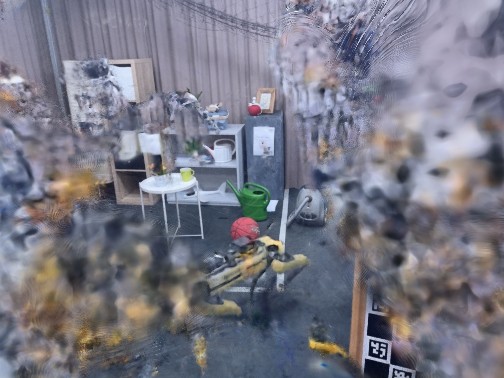}};
    \node [image,right=of corner-4] (corner-5) {\includegraphics[width=\figurewidth]{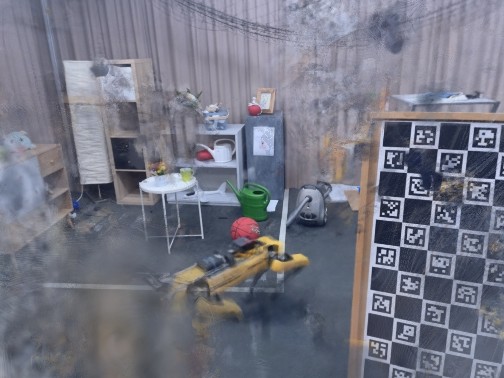}};
    \node [image,right=of corner-5] (corner-6) {\includegraphics[width=\figurewidth]{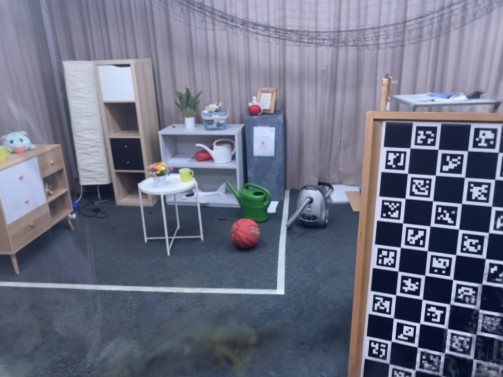}};
    \node [image,right=of corner-6] (corner-gt) {\includegraphics[width=\figurewidth]{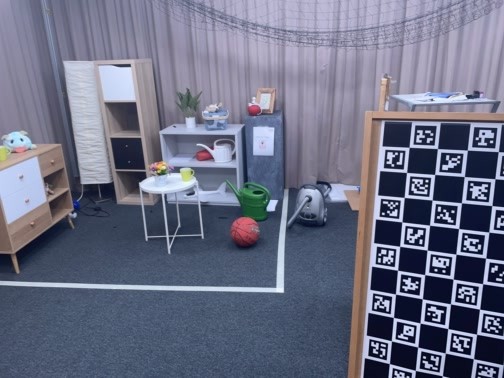}};
    
    \node[image,below=of corner-1] (corner-unc-1) {\includegraphics[width=\figurewidth]{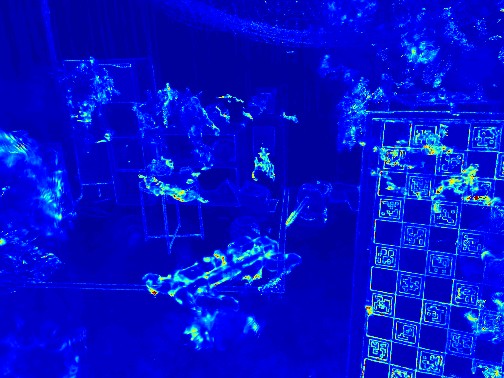}};
    \node[image,right=of corner-unc-1] (corner-unc-2) {\includegraphics[width=\figurewidth]{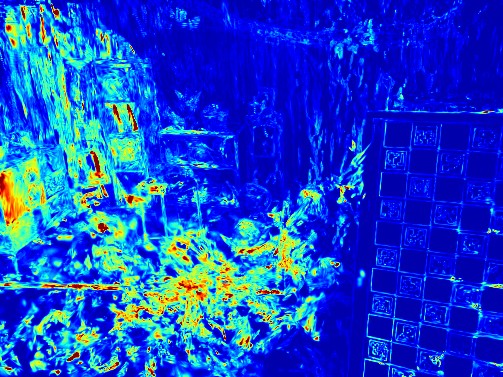}};
    \node[image,right=of corner-unc-2] (corner-unc-3) {\includegraphics[width=\figurewidth]{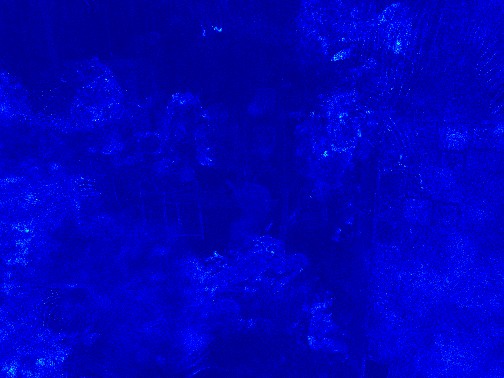}};
    \node[image,right=of corner-unc-3] (corner-unc-4) {\includegraphics[width=\figurewidth]{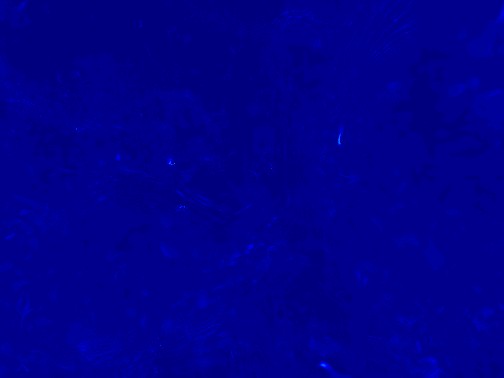}};
    \node[image,right=of corner-unc-4] (corner-unc-5) {\includegraphics[width=\figurewidth]{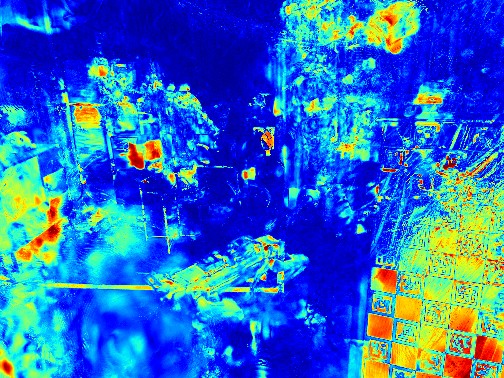}};
    \node[image,right=of corner-unc-5] (corner-unc-6) {\includegraphics[width=\figurewidth]{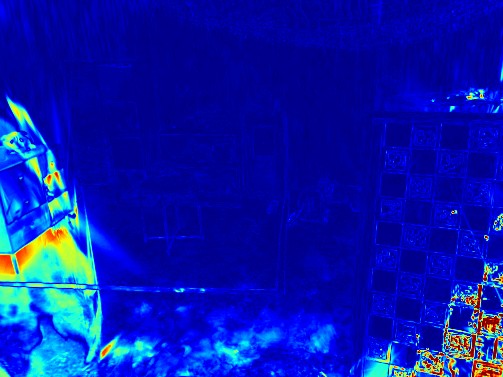}};
    \node [image,below=of corner-unc-1] (patio_high-1) {\includegraphics[width=\figurewidth]{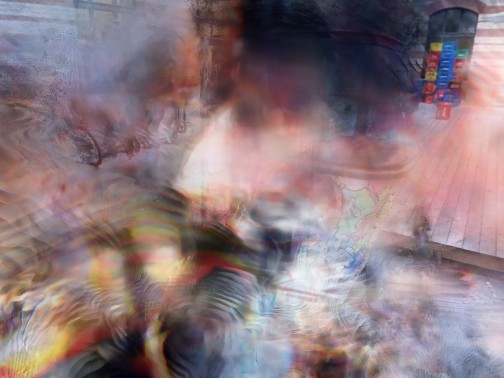}}; %
    \node [image,right=of patio_high-1] (patio_high-2) {\includegraphics[width=\figurewidth]{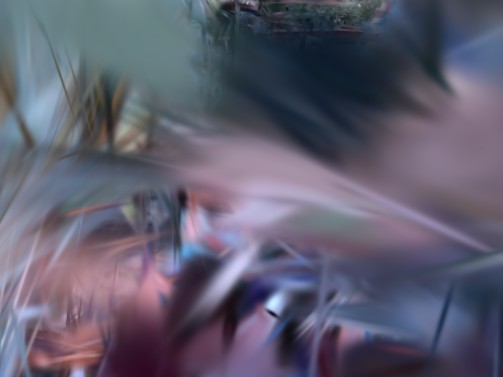}};
    \node [image,right=of patio_high-2] (patio_high-3) {\includegraphics[width=\figurewidth]{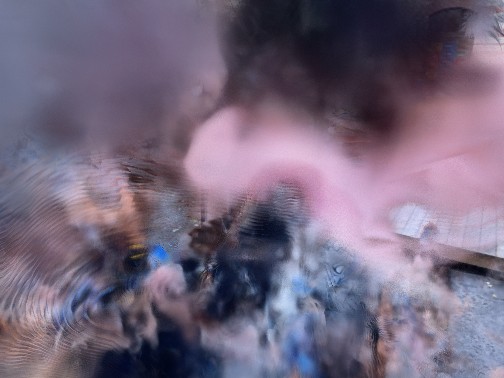}};
    \node [image,right=of patio_high-3] (patio_high-4) {\includegraphics[width=\figurewidth]{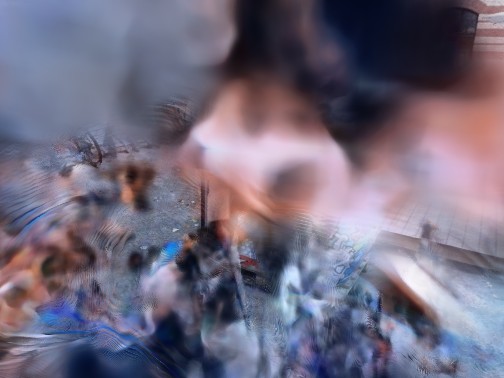}};
    \node [image,right=of patio_high-4] (patio_high-5) {\includegraphics[width=\figurewidth]{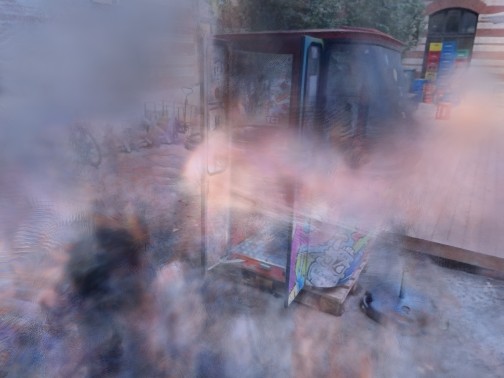}};
    \node [image,right=of patio_high-5] (patio_high-6) {\includegraphics[width=\figurewidth]{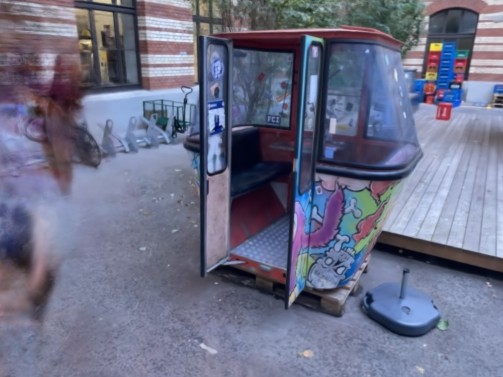}};
    \node [image,right=of patio_high-6] (patio_high-gt) {\includegraphics[width=\figurewidth]{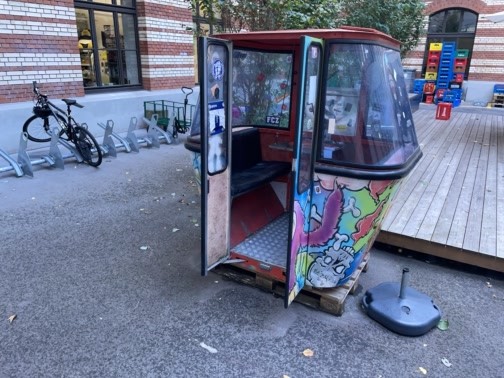}};
    
    \node[image,below=of patio_high-1] (patio_high-unc-1) {\includegraphics[width=\figurewidth]{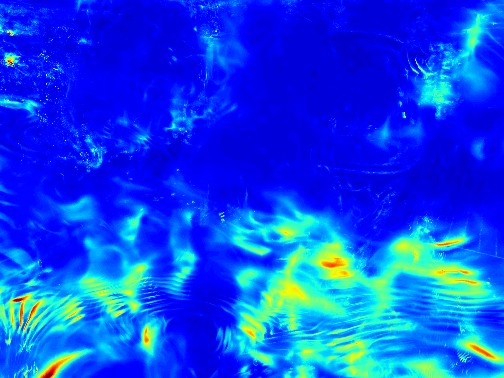}};
    \node[image,right=of patio_high-unc-1] (patio_high-unc-2) {\includegraphics[width=\figurewidth]{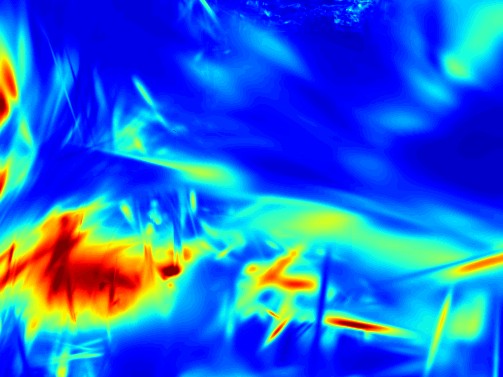}};
    \node[image,right=of patio_high-unc-2] (patio_high-unc-3) {\includegraphics[width=\figurewidth]{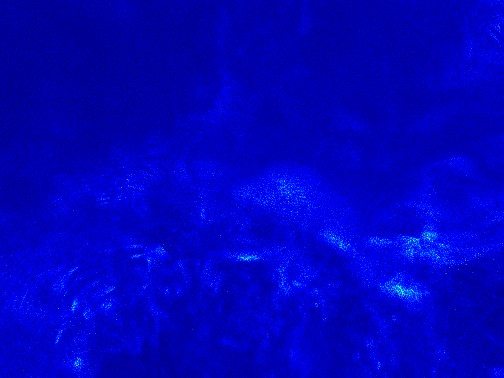}};
    \node[image,right=of patio_high-unc-3] (patio_high-unc-4) {\includegraphics[width=\figurewidth]{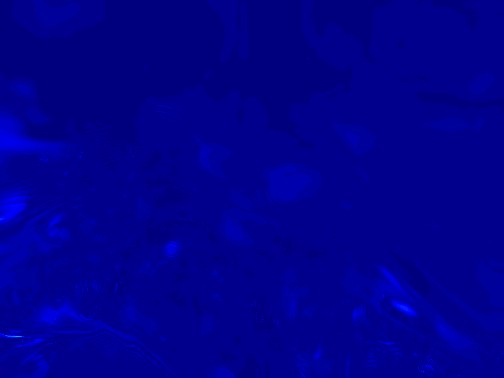}};
    \node[image,right=of patio_high-unc-4] (patio_high-unc-5) {\includegraphics[width=\figurewidth]{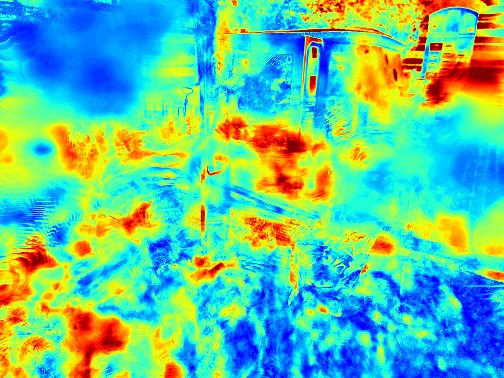}};
    \node[image,right=of patio_high-unc-5] (patio_high-unc-6) {\includegraphics[width=\figurewidth]{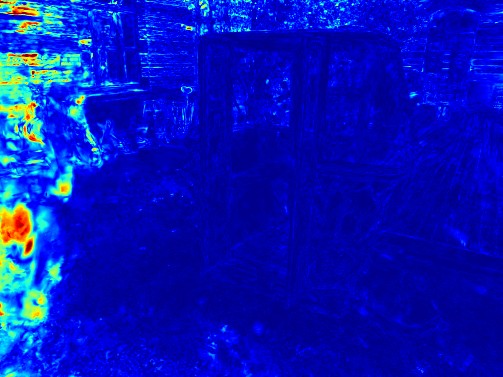}};

    \node[label] (label1) at (corner-1.north) {Active-Nerfacto\vphantom{p}};
    \node[label] (label2) at (corner-2.north) {Active-Splatfacto\vphantom{p}};
    \node[label] (label3) at (corner-3.north) {MC-Dropout-Nerfacto\vphantom{p}};
    \node[label] (label4) at (corner-4.north) {Laplace-Nerfacto\vphantom{p}};
    \node[label] (label5) at (corner-5.north) {Ensemble-Nerfacto\vphantom{p}};
    \node[label] (label6) at (corner-6.north) {Ensemble-Splatfacto\vphantom{p}};
    \node[label] (label-gt) at (corner-gt.north) {Ground Truth\vphantom{p}};

    \node[anchor=south,inner sep=1pt,rotate=90] (scene1) at (corner-1.south west) {Corner\vphantom{p}};
    \node[anchor=south,inner sep=1pt,rotate=90] (scene4) at (patio_high-1.south west) {Patio\_high\vphantom{p}};
    
    \end{tikzpicture}
  \caption{Rendered RGB and uncertainty for test views from On-the-go scenes~\cite{ren2024nerf} next to the ground truth RGB. The uncertainty is visualized by the standard deviation (0.0~\protect\includegraphics[width=3em,height=.7em]{figures-main/jet.png}~0.3). Extended version of \cref{fig:outliers-nerfonthego-qualitative}.}
  \label{fig:outliers-nerfonthego-qualitative-appendix}
\end{figure}

\paragraph{On-the-go Results} \cref{fig:outliers-nerfonthego-appendix} shows all performance metrics on the On-the-go scenes across different occlusion levels. In \cref{fig:outliers-nerfonthego-qualitative-appendix}, we extend \cref{fig:outliers-nerfonthego-qualitative} by providing additional qualitative results on scenes Patio\_high and Spot (high occlusion) for the rendered RGB values and their predicted uncertainty using different models. The behaviour of the models on the additional scenes is consistent with the scenes Mountain and Patio presented also in the main text.

\subsection{Expriments on Input Pose Uncertainty}
\label{app:pose}

\begin{figure}[t]
  \centering
  \setlength{\figurewidth}{0.205\textwidth}
  \begin{subfigure}[c]{\linewidth}
    \centering
    \begin{tikzpicture}[image/.style = {inner sep=0, outer sep=0, minimum width=\figurewidth, anchor=north west, text width=\figurewidth}, node distance = 1pt and 1pt, every node/.style={font= {\tiny}}, label/.style = {scale=0.75,font={\tiny},anchor=south,inner sep=0pt,outer sep=2pt,rotate=0}, spy using outlines={rectangle, red, magnification=2, size=0.75cm, connect spies}] 

      \node [image] (rgb-0) {\includegraphics[width=\figurewidth]{figures-main/posegrad-visuals/garden-downsample-8/rendered_rgb_0.jpg}};
      \node [image,right=of rgb-0] (rgb-2) {\includegraphics[width=\figurewidth]{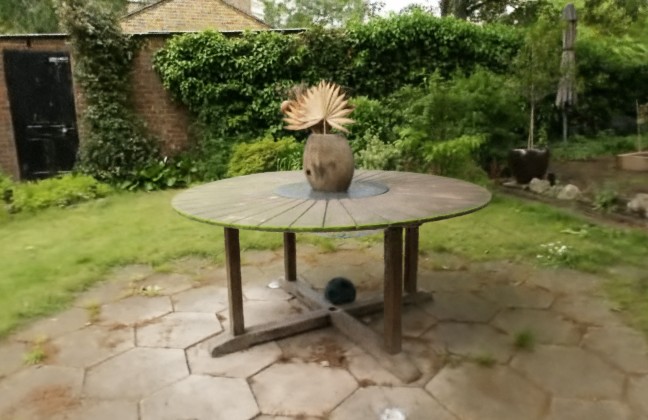}};
      \node [image,right=of rgb-2] (rgb-3) {\includegraphics[width=\figurewidth]{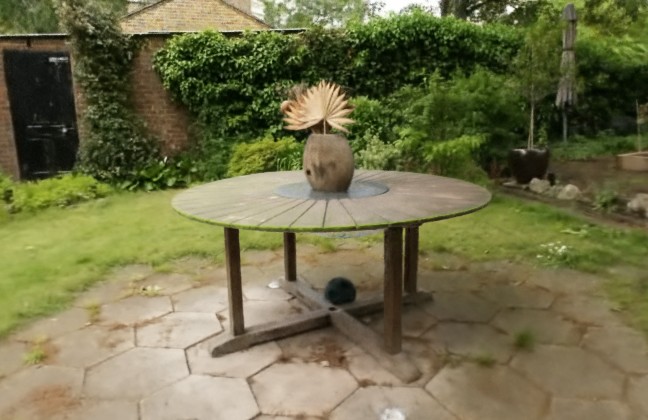}};

      \spy on ($(rgb-0) + (-0.30,-0.60)$) in node [left] at ($(rgb-0.south east) + (0,0.38)$);
      \spy on ($(rgb-2) + (-0.30,-0.60)$) in node [left] at ($(rgb-2.south east) + (0,0.38)$);
      \spy on ($(rgb-3) + (-0.30,-0.60)$) in node [left] at ($(rgb-3.south east) + (0,0.38)$);

      \node [image,below=of rgb-0] (gradnorm-0) {\includegraphics[width=\figurewidth]{figures-main/posegrad-visuals/garden-downsample-8/gradnorm_0.jpg}};
      \node [image,right=of gradnorm-0] (gradnorm-2) {\includegraphics[width=\figurewidth]{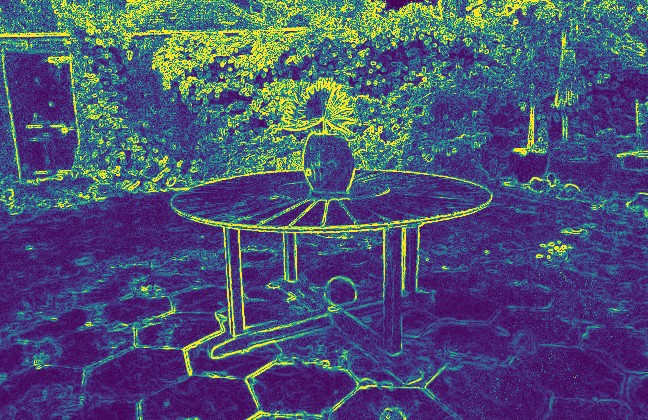}};
      \node [image,right=of gradnorm-2] (gradnorm-3) {\includegraphics[width=\figurewidth]{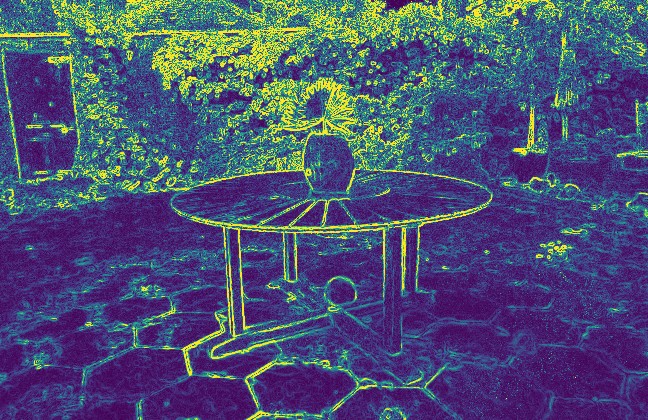}};

      \spy on ($(gradnorm-0) + (-0.30,-0.60)$) in node [left] at ($(gradnorm-0.south east) + (0,0.38)$);
      \spy on ($(gradnorm-2) + (-0.30,-0.60)$) in node [left] at ($(gradnorm-2.south east) + (0,0.38)$);
      \spy on ($(gradnorm-3) + (-0.30,-0.60)$) in node [left] at ($(gradnorm-3.south east) + (0,0.38)$);

      \node [image,draw,below=of gradnorm-0,width=\figurewidth,height=0.645\figurewidth] (err-0) {};
      \node [image,below=of gradnorm-2] (err-2) {\includegraphics[width=\figurewidth]{figures-main/posegrad-visuals/garden-downsample-8/gradnorm_error_1.jpg}};
      \node [image,right=of err-2] (err-3) {\includegraphics[width=\figurewidth]{figures-main/posegrad-visuals/garden-downsample-8/gradnorm_error_3.jpg}};
      \spy on ($(err-2) + (-0.30,-0.60)$) in node [left] at ($(err-2.south east) + (0,0.38)$);
      \spy on ($(err-3) + (-0.30,-0.60)$) in node [left] at ($(err-3.south east) + (0,0.38)$);

      \node[label] (label1) at (rgb-0.north) {Zero Shift};
      \node[label] (label2) at (rgb-2.north) {+1e-6 Shift};
      \node[label] (label3) at (rgb-3.north) {+1e-4 Shift};

      \node[anchor=south,inner sep=1pt,rotate=90] (scene1) at (rgb-0.west) {RGB\vphantom{p}};
      \node[anchor=south,inner sep=1pt,rotate=90] (scene2) at (gradnorm-0.west) {Gradient Norm\vphantom{p}};
      \node[anchor=south,inner sep=1pt,rotate=90] (scene3) at (err-0.west) {Difference \vphantom{p}};

      \end{tikzpicture}
      \caption{Garden scene.}
      \label{fig:posegrad-mipnerf360-qualitative-garden}
  \end{subfigure}
  \begin{subfigure}[c]{\linewidth}
    \centering
    \begin{tikzpicture}[image/.style = {inner sep=0, outer sep=0, minimum width=\figurewidth, anchor=north west, text width=\figurewidth}, node distance = 1pt and 1pt, every node/.style={font= {\tiny}}, label/.style = {scale=0.75,font={\tiny},anchor=south,inner sep=0pt,outer sep=2pt,rotate=0}, spy using outlines={rectangle, red, magnification=2, size=0.75cm, connect spies}] 

      \node [image] (rgb-0) {\includegraphics[width=\figurewidth]{figures-main/posegrad-visuals/bicycle-downsample-8/rendered_rgb_0.jpg}};
      \node [image,right=of rgb-0] (rgb-2) {\includegraphics[width=\figurewidth]{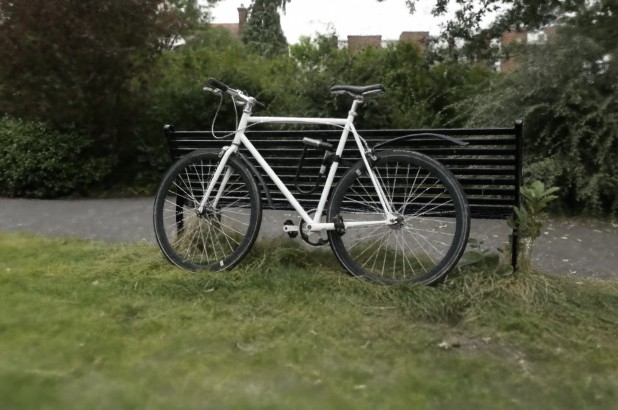}};
      \node [image,right=of rgb-2] (rgb-3) {\includegraphics[width=\figurewidth]{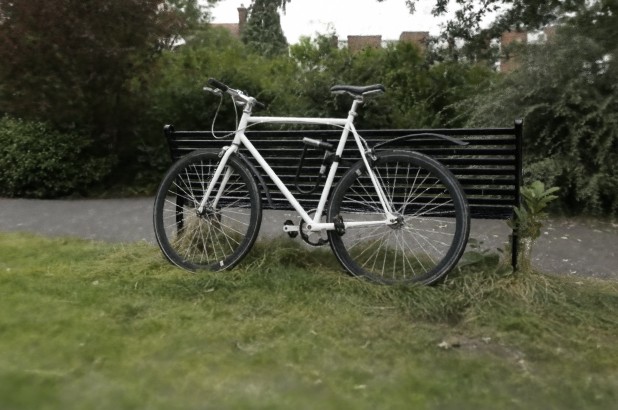}};

      \spy on ($(rgb-0) + (-0.40,0.35)$) in node [right] at ($(rgb-0.south west) + (0,0.38)$);
      \spy on ($(rgb-2) + (-0.40,0.35)$) in node [right] at ($(rgb-2.south west) + (0,0.38)$);
      \spy on ($(rgb-3) + (-0.40,0.35)$) in node [right] at ($(rgb-3.south west) + (0,0.38)$);

      \node [image,below=of rgb-0] (gradnorm-0) {\includegraphics[width=\figurewidth]{figures-main/posegrad-visuals/bicycle-downsample-8/gradnorm_0.jpg}};
      \node [image,right=of gradnorm-0] (gradnorm-2) {\includegraphics[width=\figurewidth]{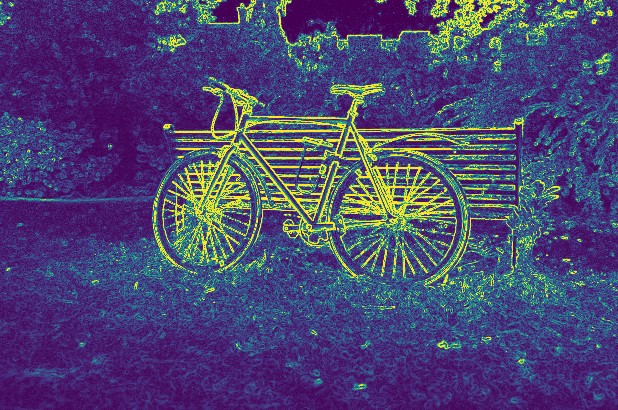}};
      \node [image,right=of gradnorm-2] (gradnorm-3) {\includegraphics[width=\figurewidth]{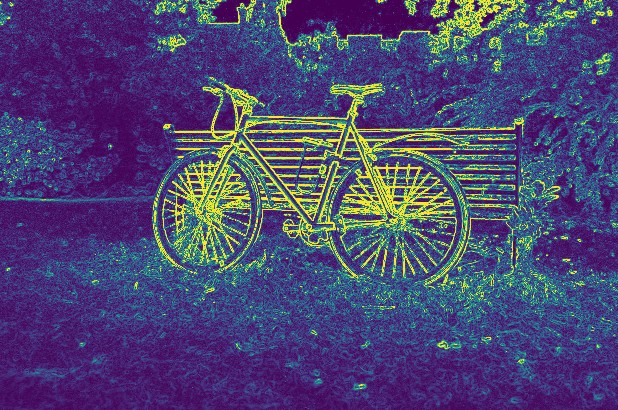}};

      \spy on ($(gradnorm-0) + (-0.40,0.35)$) in node [right] at ($(gradnorm-0.south west) + (0,0.38)$);
      \spy on ($(gradnorm-2) + (-0.40,0.35)$) in node [right] at ($(gradnorm-2.south west) + (0,0.38)$);
      \spy on ($(gradnorm-3) + (-0.40,0.35)$) in node [right] at ($(gradnorm-3.south west) + (0,0.38)$);

      \node [image,draw,below=of gradnorm-0,width=\figurewidth,height=0.645\figurewidth] (err-0) {};
      \node [image,below=of gradnorm-2] (err-2) {\includegraphics[width=\figurewidth]{figures-main/posegrad-visuals/bicycle-downsample-8/gradnorm_error_1.jpg}};
      \node [image,right=of err-2] (err-3) {\includegraphics[width=\figurewidth]{figures-main/posegrad-visuals/bicycle-downsample-8/gradnorm_error_3.jpg}};

      \spy on ($(err-2) + (-0.40,0.35)$) in node [right] at ($(err-2.south west) + (0,0.38)$);
      \spy on ($(err-3) + (-0.40,0.35)$) in node [right] at ($(err-3.south west) + (0,0.38)$);

      \node[label] (label1) at (rgb-0.north) {Zero Shift};
      \node[label] (label2) at (rgb-2.north) {+1e-6 Shift};
      \node[label] (label3) at (rgb-3.north) {+1e-4 Shift};

      \node[anchor=south,inner sep=1pt,rotate=90] (scene1) at (rgb-0.west) {RGB\vphantom{p}};
      \node[anchor=south,inner sep=1pt,rotate=90] (scene2) at (gradnorm-0.west) {Gradient Norm\vphantom{p}};
      \node[anchor=south,inner sep=1pt,rotate=90] (scene3) at (err-0.west) {Difference \vphantom{p}};
      
      \end{tikzpicture}
      \caption{Bicycle scene.}
      \label{fig:posegrad-mipnerf360-qualitative-bicycle}
  \end{subfigure}
  \begin{subfigure}[c]{\linewidth}
    \centering
    \begin{tikzpicture}[image/.style = {inner sep=0, outer sep=0, minimum width=\figurewidth, anchor=north west, text width=\figurewidth}, node distance = 1pt and 1pt, every node/.style={font= {\tiny}}, label/.style = {scale=0.75,font={\tiny},anchor=south,inner sep=0pt,outer sep=2pt,rotate=0}, spy using outlines={rectangle, red, magnification=2, size=0.75cm, connect spies}] 

      \node [image] (rgb-0) {\includegraphics[width=\figurewidth]{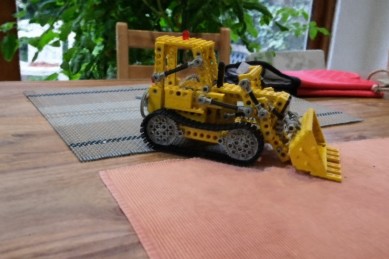}};
      \node [image,right=of rgb-0] (rgb-2) {\includegraphics[width=\figurewidth]{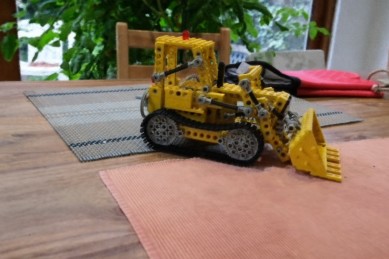}};
      \node [image,right=of rgb-2] (rgb-3) {\includegraphics[width=\figurewidth]{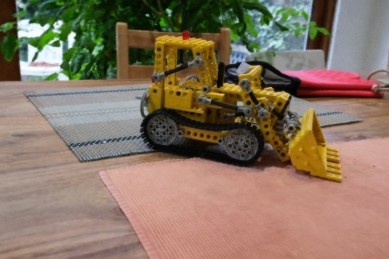}};

      \spy on ($(rgb-0) + (-0.65,-0.30)$) in node [left] at ($(rgb-0.south east) + (0,0.38)$);
      \spy on ($(rgb-2) + (-0.65,-0.30)$) in node [left] at ($(rgb-2.south east) + (0,0.38)$);
      \spy on ($(rgb-3) + (-0.65,-0.30)$) in node [left] at ($(rgb-3.south east) + (0,0.38)$);

      \node [image,below=of rgb-0] (gradnorm-0) {\includegraphics[width=\figurewidth]{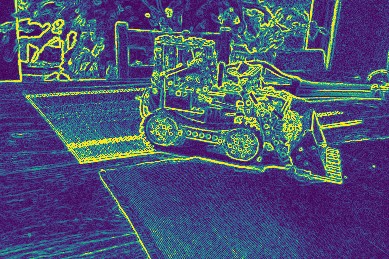}};
      \node [image,right=of gradnorm-0] (gradnorm-2) {\includegraphics[width=\figurewidth]{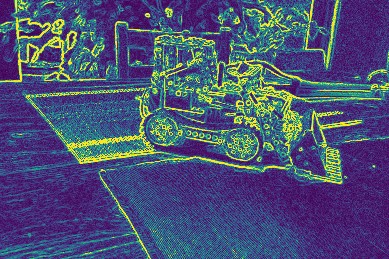}};
      \node [image,right=of gradnorm-2] (gradnorm-3) {\includegraphics[width=\figurewidth]{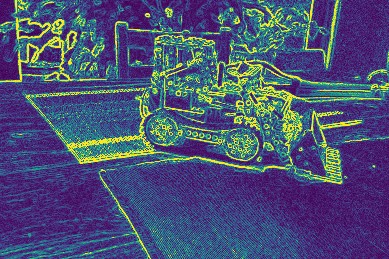}};

      \spy on ($(gradnorm-0) + (-0.65,-0.30)$) in node [left] at ($(gradnorm-0.south east) + (0,0.38)$);
      \spy on ($(gradnorm-2) + (-0.65,-0.30)$) in node [left] at ($(gradnorm-2.south east) + (0,0.38)$);
      \spy on ($(gradnorm-3) + (-0.65,-0.30)$) in node [left] at ($(gradnorm-3.south east) + (0,0.38)$);

      \node [image,draw,below=of gradnorm-0,width=\figurewidth,height=0.645\figurewidth] (err-0) {};
      \node [image,below=of gradnorm-2] (err-2) {\includegraphics[width=\figurewidth]{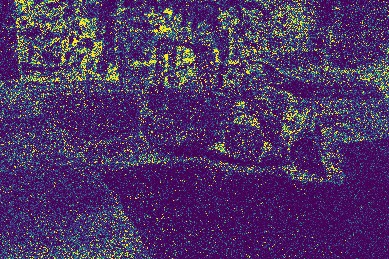}};
      \node [image,right=of err-2] (err-3) {\includegraphics[width=\figurewidth]{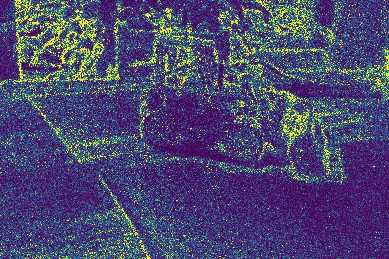}};

      \spy on ($(err-2) + (-0.65,-0.30)$) in node [left] at ($(err-2.south east) + (0,0.38)$);
      \spy on ($(err-3) + (-0.65,-0.30)$) in node [left] at ($(err-3.south east) + (0,0.38)$);

      \node[label] (label1) at (rgb-0.north) {Zero Shift};
      \node[label] (label2) at (rgb-2.north) {+1e-6 Shift};
      \node[label] (label3) at (rgb-3.north) {+1e-4 Shift};

      \node[anchor=south,inner sep=1pt,rotate=90] (scene1) at (rgb-0.west) {RGB\vphantom{p}};
      \node[anchor=south,inner sep=1pt,rotate=90] (scene2) at (gradnorm-0.west) {Gradient Norm\vphantom{p}};
      \node[anchor=south,inner sep=1pt,rotate=90] (scene3) at (err-0.west) {Difference \vphantom{p}};
      
      \end{tikzpicture}
      \caption{Kitchen scene.}
      \label{fig:posegrad-mipnerf360-qualitative-kitchen}
    \end{subfigure}
  \caption{Pose sensitivity \num{4}: Rendered RGB (row 1), gradient norm maps (row 2), and difference between gradient norms from perturbed camera poses vs.\ non-shifted pose (row 3) in the scenes Garden, Bicycle, and Kitchen. We threshold the difference values by the 95\textsuperscript{th} percentile of the gradient norm maps individually. A larger shift results in larger gradient differences around edges with high-contrast. }
  \label{fig:posegrad-mipnerf360-qualitative-appendix}
\end{figure}

Here, we provide additional details in the experiment on assessing sensitivity to imprecise camera poses. The gradient maps are obtained by computing a pixel-wise gradient of the rendered RGB w.r.t.\ the camera pose parameters using automatic differentiation, and then compute the gradient magnitude using the 2-norm (Euclidean) to obtain a scalar value for each pixel. 
In the additional examples in \cref{fig:posegrad-mipnerf360-qualitative-appendix}, 
we show that RGB and gradient norms from the perturbed camera poses are visually similar to the RGB and gradient norm from the non-perturbed (zero-shift) camera pose. This is becuase we keep the perturbation shifts small enough to make pixel-wise comparisons by computing the mean absolute error of the gradient norm maps between the erturbed vs. non-perturbed poses to show their differences. Moreover, we threshold each the gradient norm difference map by its own 95\textsuperscript{th} percentile value, such that we have 95\% of the gradient norms below the threshold value.

\paragraph{Mip-NeRF 360 Results} In \cref{fig:posegrad-mipnerf360-qualitative-garden,fig:posegrad-mipnerf360-qualitative-bicycle}, we extend \cref{fig:posegrad-mipnerf360-qualitative} by providing the rendered RGB, gradient norm maps and the differences between the perturbed vs. non-perturbed gradient maps for the Bicycle and Garden view respectively. We also show a similar visualization for a view from the Kitchen scene in \cref{fig:posegrad-mipnerf360-qualitative-kitchen}. The behaviour of the gradient norm differences on the Kitchen scenes is consistent with the Garden and Bicycle. For instance, the table cloths, the plant in the background, and the Lego tractor are more highlighted in the difference map with larger shift.

\end{document}